\documentclass[twoside]{article}

\usepackage[accepted]{aistats2024}
\usepackage[round]{natbib}

\usepackage[utf8]{inputenc} 
\usepackage[T1]{fontenc}    
\usepackage[hyphens]{url}
\usepackage{hyperref}       
\usepackage{booktabs}       
\usepackage{amsfonts}       
\usepackage{nicefrac}       
\usepackage{microtype}      
\usepackage{xcolor}         
\usepackage{multirow}
\usepackage{cite}
\usepackage{amsmath,amssymb,amsfonts}
\usepackage[linguistics]{forest}
\usepackage{tabularx}
\usepackage{textcomp}
\usepackage{xcolor,colortbl}
\usepackage{amsthm}
\usepackage{booktabs}
\usepackage{makecell}
\usepackage{subcaption}
\usepackage{enumitem}

\usepackage{newfloat}
\usepackage{listings}
\usepackage{algorithm}
\usepackage{algorithmic}

\def\x{\mathbf{x}}
\def\c{\mathbf{c}}
\def\y{\mathbf{y}}

\def\X{\mathbf{X}}

\def\S{\mathbf{S}}
\newcommand{\capt}{{\textrm{cap}}}
\newcommand{\fix}{{\textrm{\rm fix}}}
\newcommand{\splita}{{\textrm{\rm split}}}
\newcommand{\lf}{{\textrm{left}}}
\newcommand{\ri}{{\textrm{right}}}

\newtheorem{theorem}{Theorem}[section]
\newtheorem{lemma}[theorem]{Lemma}

\newtheorem{corollary}{Corollary}[theorem]

\newcommand{\ourmethod}{{\textrm{OSST}}}
\DeclareMathOperator*{\argmin}{\arg\!\min}

\begin{document}
%

%
\twocolumn[

\aistatstitle{Optimal Sparse Survival Trees}

\aistatsauthor{Rui Zhang \And Rui Xin \And  Margo Seltzer \And Cynthia Rudin }

\aistatsaddress{Duke University \And  Duke University \And University of British Columbia \And Duke University } ]

\begin{abstract}
Interpretability is crucial for doctors, hospitals, pharmaceutical companies and biotechnology corporations to analyze and make decisions for high stakes problems that involve human health. Tree-based methods have been widely adopted for \textit{survival analysis} due to their appealing interpretablility and their ability to capture complex relationships. However, most existing methods to produce survival trees rely on heuristic (or greedy) algorithms, which risk producing sub-optimal models. We present a dynamic-programming-with-bounds approach that finds provably-optimal sparse survival tree models, frequently in only a few seconds.
\end{abstract}

\section{INTRODUCTION}
Interpretability is essential for high stakes decisions \citep{RudinEtAlSurvey2022}, particularly in healthcare. Thus, when a machine learning model estimates the answers to critical questions such as ``how long is this patient expected to survive?'', the reasoning process of the model must be understandable to humans. There has been little overlap between the fields of \textit{interpretable machine learning} and \textit{survival analysis}, despite the importance of survival analysis problems in healthcare and beyond. Survival analysis is used in reliability modeling for mechanical failure of equipment (where it is called reliability analysis), customer churn prediction, and questions in the social sciences such as predicting economic events. Early work in survival analysis either did not use covariates, such as Kaplan-Meier curves \citep{kaplan1958nonparametric}, or used a (non-sparse) linear combination of covariates, such as Cox proportional hazard models \citep{cox1972regression}. Interpretable machine learning tools introduce the ability to use  \textit{sparse} and \textit{nonlinear} combinations of variables to form accurate survival models. This makes survival analyses more powerful, while maintaining the interpretability necessary for use in practice.

In a classical setup for survival analysis, we might wish to predict the time at which an event (``death'') will occur. However, estimating time-to-event is problematic, because some of the data are  \textit{censored}, meaning that the time of the event is not known; all we know is that the individual survived beyond the last observation time. Because we cannot distinguish between samples who have different death times when neither time is observed, we instead estimate the probability that a sample with variables $\x$ survives past time $y$, which is defined as the survival function $S_{\x}(y)$. Since $S_{\x}(y)$ is a probability estimate, it suggests that techniques for classification or regression might work, but the loss function is completely different, so these techniques do not apply directly. 
Some machine learning techniques can adapt to many loss functions to achieve \textit{high performing} models but not \textit{interpretable} models. Specifically, it is easy to adapt standard machine learning techniques to produce black box survival  models, simply by changing the loss function to a survival loss \citep{che2018recurrent,ching2018cox, giunchiglia2018rnn,hothorn2006unbiased,ishwaran2007random,katzman2018deepsurv,  ripley2001neural}. Since the survival curves created by these methods are black box, they are unlikely to be useful in practice. Another easy way to get survival models is to use greedy methods for minimizing the loss function; greedy tree-based models 
for the survival function $S_{\x}(y)$ are popular
\citep{ciampi1988recpam,
davis1989exponential,
gordon1985tree,
hothorn2006unbiased, jin2004alternative, 
kelecs2002residual,
leblanc1992relative,
leblanc1993survival,  molinaro2004tree,
segal1988regression, therneau1990martingale,
zhang1995splitting}. These methods generally choose and fix the top split first according to heuristic splitting rules, and continue to split according to heuristics until the tree is formed, perhaps using additional heuristics to prune the tree afterwards to prevent overfitting. However, their performance is often limited as a bad split cannot be fixed once it has been made. 

To get interpretable models that achieve good performance requires optimization. As soon as the requirement of sparsity is added to any machine learning problem, the problem becomes computationally extremely hard, and techniques must be tailored to the specific loss functions in order to maintain performance.
Specialized techniques have been developed for classification and regression for sparse decision trees \citep{grubinger2014evtree, Dunn2018,
lin2020generalized,nijssen2020,zhang2023optimal}. 
We have learned from these modern methods that their single -- very sparse -- trees are often as accurate as black box models for tabular datasets. We have also learned that each problem requires whole new algorithms. However, fully optimizing a survival tree is much harder than a classification or regression tree.

The only previous attempt to construct optimized sparse survival trees that we know of is that of \citet{bertsimas2022optimal} who attempted to solve the problem using Mixed Integer Programming (MIP) and local search. However, their method assumes that the ratio of the hazard functions for any two individuals is constant over time. Their search method also can get stuck at local optima. Their code is proprietary and tends to crash very often (about 90\% of runs).

The method presented in this paper is first algorithm for optimal sparse survival trees with public code. We chose a dynamic-programming-with-bounds framework, where theorems based on the survival loss -- in this case, the Integrated Brier Score -- are used to narrow the search space. The tighter these bounds are, the better they prune the search space and reduce computation. Our bounds are tight enough that optimal sparse survival trees can be found in seconds or minutes for all public survival analysis datasets we know of. We call our algorithm \textbf{Optimal Sparse Survival Trees} (OSST). In Section \ref{sec:notation}, we present notation and the survival objective, and Sections \ref{sec:dp} and \ref{sec:bounds} present our theorems that make training optimal sparse survival trees possible. Section \ref{sec:exp} contains the experimental results. 
An extended related work section appears in the appendix.

\section{METHODOLOGY}\label{sec:method}
\subsection{Notation and Objective}\label{sec:notation}
We denote the training dataset $(\X, \c, \y)$ as $\{(\x_i, c_i, y_i)\}^N_{i=1} $, where $c_i \in \{0,1\}$ is a binary variable indicating whether the
sample has an observed death $(c_i = 1)$ or is censored $(c_i = 0)$, and $y_i \in \mathbb{R} $ is the time of last observation for sample $i$. $\x_i \in \{0,1\}^M$ is a binary feature vector, where real-valued variables are converted to a set of binary variables using either all possible splits or a subset of splits chosen by a black box model \citep{McTavishZhongEtAl2022}. The union of $\{\x_i\}$ is the set of possible splits for any decision tree.  

The Integrated Brier Score (IBS) for censored observations proposed by \citet{graf1999assessment} is used as our performance metric. We denote the loss of tree $t$ on the training dataset as
\begin{equation}\label{eq:tree_loss}
    \mathcal{L}(t,\X, \c, \y) := \frac{1}{y_{\max}} \int_0^{y_{\max}}BS(y) dy
\end{equation} 
where $y_{\max} = \max \{y_i\}^N_{i=1}$ is the latest time point of all observed samples, and $BS(y)$ is the Brier Score of tree $t$ at given time $y$, which can be interpreted as the mean square error between the data and the predicted survival function $\hat{S}_{\x_i}(y)$ by tree $t$, weighted by Inverse Probability of Censoring Weights (IPCW):
\begin{eqnarray}\label{eq:brierscore}\nonumber
    BS(y) & = & \frac{1}{N} \sum_{i=1}^N \frac{(\hat{S}_{\x_i}(y) - 0)^2}{\hat{G}(y_i)}\cdot \mathbf{1}_{y_i \leq y, c_i = 1} \\
    &  &  + \frac{(\hat{S}_{\x_i}(y) - 1)^2}{\hat{G}(y)}\cdot \mathbf{1}_{y_i > y}
\end{eqnarray}
where $\hat{S}_{\x_i}(\cdot)$ is estimated by a non-parametric Kaplan–Meier estimator at the leaf node to which $\x_i$ gets assigned, and $\hat{G}(\cdot)$ is the Kaplan–Meier estimate of the censoring distribution $\c$,  which is assumed to be independent of any covariates \citep{molinaro2004tree}. The Kaplan–Meier estimator, known as the product limit estimator, is given by
\begin{equation}\label{eq:KM_estimator}
    \hat{S}(y) = \prod_{i: y_i \leq y} \left( 1 - \frac{d_i}{n_i}\right)
\end{equation}
where $d_i$ is the number of deaths at time $y_i$, and $n_i$ is the number of samples known to have survived up to time $y_i$.
The first term in Equation \ref{eq:brierscore} applies to non-censored data and encourages the survival function to be close to 0 after the death event. The second term applies to censored and non-censored data, encouraging the survival function to be 1 up until the censoring (for censored data) or up until a death event is observed (for non-censored data).

We define the objective function  $R(t,\X,\y)$ of tree $t$, as a combination of the tree loss defined above and a complexity penalty: $\mathcal{L}(t,\X, \c, \y) + \lambda \cdot \text{complexity}(t),$ where the complexity penalty is  $H_t$, the number of leaves in tree $t$:
\begin{equation}\label{eq:obj1}
     R(t,\X, \c, \y) := \mathcal{L}(t,\X, \c, \y) + \lambda H_t.
\end{equation}
In addition to the soft constraint on complexity, we also add an optional hard constraint:
\begin{equation}\label{eq:obj2}
     \mathcal{L}(t,\X, \c, \y) + \lambda H_t, \text{ s.t. } \textrm{depth}(t) \leq d.
\end{equation}
The hard constraint on depth makes computation substantially easier, because each increase in depth creates an exponentially larger search space. The soft complexity constraint is always used, because it excludes unnecessary leaves. That is, the optimal tree with a depth limit of 6 might need only 8 leaves, but the hard constraint would permit trees having up to 64 leaves. 
Our algorithm finds optimal trees that minimize Equation \ref{eq:obj2} in seconds. It can also minimize Equation \ref{eq:obj1} without a depth constraint but takes more time to find the globally optimal solution.

\subsection{Dynamic Programming}\label{sec:dp}
 We use dynamic-programming-with-bounds (Algorithm \ref{alg:opt}) for optimization \citep[see ][]{HuRuSe2019, lin2020generalized,McTavishZhongEtAl2022, zhang2023optimal}; the question here is how to derive the bounds for survival analysis so that the search space is pruned effectively (Line 35-38). 
 If the bounds determine that a partially constructed tree cannot be part of an optimal solution, then the part of the search space it extends to can be eliminated. 

We start with a single leaf node to which all sample points are assigned (\text{Line 1-5}), and then try splitting this single leaf by all possible features and all possible ways of splitting those features (\text{Line 12-17}). This process produces more leaf nodes, each containing only a subset of the data; recursive splitting creates a large number of nodes/subsets. Each node/subset represents a \textit{sub-problem} for which we want to find the corresponding optimal sub-tree. Each \textit{sub-problem} is identified by a support set $s = \{s_1, s_2, \ldots, s_{N}\}$, where $s_i$ is a Boolean value indicating whether sample $i$ is in the support set $s$. With each sub-problem, the algorithm records and updates the lower bound and upper bound (current best score) of its corresponding sub-tree objective (Line 19). The sub-problem is considered \textit{solved} when these two bounds converge (Line 9, 22). If the bounds determine that a sub-problem cannot benefit from splitting (Line 37), then we can mark it as solved without further exploration of that part of the search space. We store all sub-problems and the parent/child relationships between them in a \textit{dependency graph}. Since a sub-problem can have more than one parent problem (i.e., a sub-problem can arise by multiple different sequences of splits), the dependency graph lets us avoid duplicate computations. A priority queue is maintained to store the copies of all unsolved sub-problems, and there can be multiple copies of a specific sub-problem in the queue at the same time (i.e., the subproblem 
can be pushed by its child problems, Line 21; or it can be pushed by its parent problem, Line 29-30). We use the difference between the fraction of captured points and the objective lower bound as the priority value for ordering the queue. Once all sub-problems in the graph are solved, the optimization is complete, and we can extract all optimal survival trees from the graph.
 
\definecolor{comment}{rgb}{0.2,0.4,0.2}
\def\comment#1{\textcolor{comment}{\textit{// #1 }}}
\setlength{\textfloatsep}{0pt}
\begin{algorithm}
\begin{minipage}{\columnwidth}
\caption{\textbf{dynamic-programming-with-bounds} $(\X, \c, \y, \lambda, T, R)$ \label{alg:opt}\\
\comment{Optional input: reference model $T$ and initial risk score $R$}}

\begin{tabbing}
xxx \= xx \= xx \= xx \= xx \= xx \kill
1: $Q \leftarrow  \emptyset$ \comment{priority queue} \\
2: $G \leftarrow  \emptyset$ \comment{dependency graph} \\
3: $s_0 \leftarrow \{1, \dots, 1\}$\comment{support set for root problem}\\
4: $p_0 \leftarrow $ \text{find\_or\_create\_node}$(G, s_0)$ \comment{root problem}\\
5: $Q.push(s_0)$ \comment{add to priority queue}\\
6: $\textbf{while  } p_0.lb \neq p_0.ub \textbf{  do}$ \\
7: \>$s \leftarrow Q.{\rm pop}()$  \\
8: \> $p \leftarrow G.{\rm find}(s)$  \\
9: \> \textbf{if} $p.lb = p.ub$ \textbf{then}\\
10:\>\> \textbf{break} \comment{problem already solved}\\
11:\> $lb', ub' \leftarrow (\inf, \inf)$ \comment{initialize starting bounds}\\
12:\> \textbf{for} each feature $j \in \{1, \dots, M\}$ \textbf{do}\\
\>\>\comment{support sets for child problems}\\
13:\>\>$s_l, s_r \leftarrow \text{split}(s,j,\X)$\\
14:\>\> $p_l^j\leftarrow$\text{find\_or\_create\_node}$(G,s_l)$  \\ 
15:\>\> $p_r^j\leftarrow$\text{find\_or\_create\_node}$(G,s_r)$\\
\>\> \comment{create bounds as if $j$ were chosen}\\
\>\> \comment{for splitting}\\
16:\>\>  $lb' \leftarrow \min(lb', p_l^j.lb + p_r^j.lb)$\\
17:\>\> $ub' \leftarrow \min(ub', p_l^j.ub + p_r^j.ub)$\\ 
\> \comment{signal the parents if an update occurred}\\
18:\> \textbf{if} $p.lb \neq lb'$ \textbf{or} $p.ub \neq ub'$  \textbf{then}\\
19:\>\> $(p.lb, p.ub) \leftarrow (lb', ub')$\\
20:\>\>\textbf{for} $p_{\pi} \in G.{\rm parent}(p)$ \textbf{do} \\
\>\>\> \comment{propagate information upwards}\\
21: \>\>\>$Q.{\rm push}(p_{\pi}.{\rm id}, {\rm priority}=count(s)/N - p.lb)$\\
22:\>\textbf{if} $p.lb \geq p.ub$ \textbf{then} \\
23:\>\>\textbf{continue} \comment{problem solved just now} \\
\comment{loop, enqueue all children that are dependencies} \\
24:\> \textbf{for} each feature $j \in [1,M]$ \textbf{do} \\
25:\>\> repeat line 14-16\\
26:\>\> $lb' \leftarrow p_l^j.lb + p_r^j.lb$ \\
27:\>\> $ub' \leftarrow p_l^j.ub + p_r^j.ub$ \\
28:\>\> \textbf{if} $lb' < ub'$ \textbf{and} $lb' \le p.ub$ \textbf{then} \\
29:\>\>\>$Q.{\rm push}(s_{l}, {\rm priority}=count(s_{l})/N - p_l.lb)$ \\
30:\>\>\>$Q.{\rm push}(s_{r}, {\rm priority}=count(s_{r})/N - p_r.lb)$ \\
31: \textbf{return}\\
-----------------------------------------------\\
32: \textbf{subroutine}: \textbf{find\_or\_create\_node}$(G,s)$\\
\comment{$p$ not yet in dependency graph}\\
33: if $G.{\rm find}(s) = {\rm NULL}$\\
34: \> $p.id \leftarrow s$ \comment{identify $p$ by $s$}\\
\> \comment{compute initial lower and upper bounds}\\
35: \> $p.lb \leftarrow {\rm get\_lower\_bound}(s,\X, \c, \y)$\\
36: \> $p.ub \leftarrow {\rm get\_upper\_bound}(s,\X, \c, \y)$\\
37: \> \textbf{if} fails\_bounds$(p)$ \textbf{then}\\
38: \>\> $p.lb=p.ub$ \comment{no more splitting allowed}\\
39: \> G.insert(p) \comment{put $p$ in dependency graph}\\
40: \textbf{return} G.find(s)
\end{tabbing}
\end{minipage}
\end{algorithm}
\setlength{\textfloatsep}{0pt}
\subsection{Bounds}\label{sec:bounds}
Using notation similar to \citet{lin2020generalized}, we represent a tree $t$ as a set of $H_t$ distinct leaves:
$t = \{l_1, l_2, \dots, l_{H_t}\}.$
It can also be written as:
\[t = (t_{\textrm{fix}}, \delta_{\textrm{fix}}, t_{\textrm{split}}, \delta_{\textrm{split}}, K, H_t),\]
where $t_{\textrm{fix}}=\{l_1, l_2, \dots, l_K\}$ are a set of $K$ \textbf{fixed leaves} that are not allowed to be further split in this part of the search space, $\delta_{\fix} = \{\hat{S}_{l_1},\hat{S}_{l_2}, \dots, \hat{S}_{l_K}\}$ are predicted survival functions for the fixed leaves, $t_{\splita} = \{l_{K+1}, l_{K+2}, \dots, l_{H_t}\}$ are $H_t - K$ \textbf{splitting leaves} that can be further split in this part of the search space, and their predicted survival functions are $\delta_{\textrm{split}} = \{\hat{S}_{l_{K+1}},\hat{S}_{l_{K+2}}, \dots, \hat{S}_{H_t}\}$. 

A child tree $t' = (t'_{\fix}, \delta'_{\fix}, t'_{\splita}, \delta'_{\splita}, K', H_{t'})$ can be created by splitting a subset of splitting leaves $t_{\textrm{split}}$ in tree $t$. $t'_{\fix}$ is a superset of $t_{\fix}$. We denote $\sigma(t)$ as the set of all child trees of $t$.
\textit{Note that proofs for all theorems are in Appendix \ref{sec:thms_and_prfs}.}

\subsubsection{Lower Bounds}
A large portion of the search space is reduced through leveraging lower bounds of the survival tree objective. Specifically, as we will show, if the objective lower bound of tree $t$ exceeds the current best objective so far, $R^c$, then neither tree $t$ nor any of its children $t' \in \sigma(t)$ can be an optimal tree. This is typically called the hierarchical objective lower bound, but we cannot prove this without some work. Let us start by deriving some principles of the objective. 
\begin{theorem}\label{thm:additive}
The loss of a survival tree is an additive function of the observations and leaves. 
\end{theorem}
Because of Theorem \ref{thm:additive} and that fixed leaves are not allowed to be further split in this part of the search space, the loss of observations in fixed leaves provides a lower bound for tree $t$:
\begin{equation}\label{eq:lb}
    R(t,\X, \c, \y) \geq \mathcal{L}(t_{\fix},\X, \c, \y) + \lambda H_t,
\end{equation}
where $\mathcal{L}(t_{\fix},\X, \c,\y)$ is the sum of losses for the fixed leaves:
\begin{align*}
    \mathcal{L}(t_{\fix},\X, \c, \y) & = \frac{1}{y_{\max}}\frac{1}{N} \sum_{i=1}^N \biggl\{\int_0^{y_i} \frac{(\hat{S}_{\x_i}(y) - 1)^2}{\hat{G}(y)}dy \\
    + & c_i \int_{y_i}^{y_{\max}} \frac{(\hat{S}_{\x_i}(y) - 0)^2}{\hat{G}(y_i)}dy \biggr\} \cdot \mathbf{1}_{\capt(t_{\fix}, \x_i)}, 
\end{align*}
where $\mathbf{1}_{\capt(t_{\fix}, \x_i)}$ is 1 if a leaf in $t_{\fix}$ captures $\x_i$ (when $\x_i$ falls into one of the fixed leaves of $t$), 0 otherwise. 

\begin{theorem}\label{thm:hierarchical_lb}
(Hierarchical Objective Lower Bound). Any tree $t' = (t'_{\fix}, \delta'_{\fix}, t'_{\splita}, \delta'_{\splita}, K', H_{t'}) \in \sigma(t)$ in the child tree set of tree $t = (t_{\fix}, \delta_{\fix}, t_{\splita}, \delta_{\splita}, K, H_t)$ obeys: \[R(t',\X, \c, \y) \geq \mathcal{L}(t_{\fix},\X, \c,\y) + \lambda H_t. \]
\end{theorem}
Theorem \ref{thm:hierarchical_lb} indicates that the objective lower bound of the parent tree also holds for all its child trees, which means that if the parent tree can be pruned via the lower bound then so can its child trees.

Sometimes even if the parent tree $t$ satisfies $\mathcal{L}(t_{\fix},\X, \c,\y) + \lambda H_t \leq R_c$, we still can prune all of its child trees, which is shown in Theorem \ref{thm:look_ahead}. 

\begin{theorem}\label{thm:look_ahead}
(Objective Lower Bound with One-step Lookahead). Let $t = (t_{\fix}, \delta_{\fix}, t_{\splita}, \delta_{\splita}, K, H_t)$ be a tree with $H_t$ leaves. If $\mathcal{L}(t_{\fix},\X, \c, \y) + \lambda H_t + \lambda > R^c$, even if its objective lower bound obeys  $\mathcal{L}(t_{\fix},\X, \c,\y) + \lambda H_t \leq R^c$, then for any child tree $t' \in \sigma(t)$, $R(t', \X, \c, \y) > R^c$.
\end{theorem}

This bound shows that although the parent tree cannot be pruned via the lower bound, we should not explore any of its child trees if $\mathcal{L}(t_{\fix},\X, \c, \y) + \lambda H_t + \lambda > R^c$, because all of them are sub-optimal.

To make the lower bound tighter, we leverage the property that some points cannot be partitioned into different leaves in any survival tree, by introducing the concept of equivalent points. 
\subsubsection*{Equivalent Points}
We denote the loss from sample $\{\x_i, c_i, y_i\}$ with prediction $\hat{S}_{\x_i}(\cdot)$,  as $\mathcal{L}(\hat{S}_{\x_i}(\cdot), \x_i, c_i, y_i)$, where
\begin{eqnarray}\label{eq:sample_loss}\nonumber
    \mathcal{L}(\hat{S}_{\x_i}(\cdot), \x_i, c_i, y_i) & = &\frac{1}{y_{\max}}\frac{1}{N} \int_0^{y_i} \frac{(\hat{S}_{\x_i}(y) - 1)^2}{\hat{G}(y)}dy \\
    & + & c_i \int_{y_i}^{y_{\max}} \frac{(\hat{S}_{\x_i}(y) - 0)^2}{\hat{G}(y_i)}dy.
\end{eqnarray}
In the simplest case, $\mathcal{L}(\hat{S}_{\x_i}(\cdot),\x_i, c_i, y_i) = 0$ if the leaf node contains only $\{\x_i, c_i, y_i\}$. More formally, $\mathcal{L}(\hat{S}_{\x_i}(\cdot),\x_i, c_i, y_i)$ can be zero only if two conditions are satisfied: (1) there is no sample in the same leaf that died prior to sample $\{\x_i, c_i, y_i\}$, i.e., $\{\x_j, c_j = 1, y_j < y_i \}$ (here, the first term in \ref{eq:sample_loss} is nonzero because the survival function will be $<1$ at $y_i$), and (2) if $c_i = 1$, then all the points in $i$'s leaf that were observed to survive up to time $y_i$ must also die at $y_i$. To explain, if $c_i = 1$ and there are (censored or uncensored) points in $i$'s leaf node whose observation time is later than $y_i$, i.e., $\{\x_j, c_j , y_j > y_i \}$, the second term in \ref{eq:sample_loss} will be nonzero (those points will force the survival curve to be nonzero even after $i$ dies at time $y_i$). Also if $c_i=1$, we cannot have censored points with the same observation time as $i$, i.e., $\{\x_j, c_j = 0, y_j = y_i \}$, since it would mean that when $i$ dies at $y_i$, another point does not, so the survival function cannot go to 0 at $y_i$.

If each sample point can be placed in a leaf by itself, the tree $t$ can achieve zero loss, but this tree is clearly overfitted. Further, such a tree may not be possible if there exist two samples $\{\x_i,c_i, y_i\}$ and $\{\x_j, c_j, y_j\}$ such that $\x_i = \x_j$, but $c_i \ne c_j$ or $y_i \ne y_j$; we call such points equivalent points.
There is no tree that can partition these samples into different leaves, so the loss contributed from these samples must be non-zero if the two conditions above cannot be satisfied. 

Let us generalize this argument. Let $u$ be a set of equivalent points where samples have exactly the same feature vector $\x$, such that $\forall j_1,j_2,...j_{|u|}\in u$:
\begin{equation}\label{eq:equiv_set}
    \x_{j_1} = \x_{j_2} = \dots = \x_{j_{|u|}}.
\end{equation}
Assume set $u$ does not satisfy the two conditions, which means loss is non-zero:
\begin{equation}\label{eq:sum_of_sample_loss}
    \sum_{k=1}^{|u|}\mathcal{L}(\hat{S}_{\x_{j_k}}(\cdot),\x_{j_k}, c_{j_k}, y_{j_k}) > 0.
\end{equation}
Let us derive the lower bound for Equation \ref{eq:sum_of_sample_loss}. Recall that  samples in a leaf share the same predicted survival function.
\begin{lemma}\label{lm:equiv_loss}
(Equivalent Loss). Let $u$ be a set of equivalent points defined as in \eqref{eq:equiv_set}. We denote $^{\ast}S $ as the optimal step function that minimizes  IBS loss only for set $u$ (leaf contains set $u$ only), such that:
\[^{\ast}S = \argmin_{S} \sum_{k=1}^{|u|}\mathcal{L}(S,\x_{j_k}, c_{j_k}, y_{j_k}).\]
We define \textbf{Equivalent Loss} for set $u$ as 
$\mathcal{E}_u = \sum_{k=1}^{|u|}\mathcal{L}(^{\ast} S,\x_{j_k}, c_{j_k}, y_{j_k}).$
Then, any leaf $l$ that captures set $u$ in a survival tree has loss  $\mathcal{L}(l,\X, \c, \y) \geq \mathcal{E}_u$.
\end{lemma}

\begin{lemma}\label{lm:equiv_loss_2}
Let $l$ be a leaf node that captures $n$ equivalent sets: $\{u_i\}_{i=1}^n$ and corresponding $\mathcal{E}_{u_i}$. The loss of $l$: $\mathcal{L}(l,\X, \c, \y) \geq \sum_{i=1}^n\mathcal{E}_{u_i}$. 
\end{lemma}
That is, the lower bound of a leaf is the sum of equivalent losses of the equivalent sets it captures.

\begin{theorem}\label{thm:equiv_lb}
(Equivalent Points Lower Bound). Let $t = (t_{\fix}, \delta_{\fix}, t_{\splita}, \delta_{\splita}, K, H_t)$ be a tree with $K$ fixed leaves and $H_t - K$ splitting leaves. For any child tree $t'= (t'_{\fix}, \delta'_{\fix}, t'_{\splita}, \delta'_{\splita}, K', H_{t'}) \in \sigma(t)$:
\begin{eqnarray}\label{eq:equiv_lb}\nonumber
    R(t', \X,\c, \y) &\geq & \mathcal{L}(t_{\fix},\X,\c, \y) + \lambda H_{t}\\ & + &\sum_{u\in U}  \mathcal{E}_u \cdot \mathbf{1}_{\capt(t_{\splita}, u)},
\end{eqnarray}
where $U$ is the set of equivalent points sets in the training dataset $(\X, \c, \y)$ and $\mathbf{1}_{\capt(t_{\splita}, u)}$ is 1 when $t_{\splita}$ captures set $u$, 0 otherwise. Combining with Theorem \ref{thm:look_ahead}, we have a tighter bound: for any child tree $t'$:
\begin{eqnarray}\label{eq:equiv_lb+lookahead}\nonumber
    R(t', \X, \c, \y) &\geq  &\mathcal{L}(t_{\fix},\X,\c, \y) + \lambda H_{t} + \lambda \\  &+ &\sum_{u\in U}  \mathcal{E}_u \cdot \mathbf{1}_{\capt(t_{\splita}, u)}.
\end{eqnarray}
\end{theorem}
\subsubsection*{Lower Bound from Reference Models}
\label{sec:guessing}
Given that the hardness of the optimization problem to solve varies, the Equivalent Points Lower Bound may not be tight enough in some cases.  When it is not tight enough to sufficiently prune the search space, calculating the bound adds overhead and slows the optimization procedure. Moreover, some datasets do not have many equivalent points, resulting in a looser lower bound. 

To efficiently prune the search space, we adopt the \textit{guessing} technique of \citet{McTavishZhongEtAl2022}. Specifically, we use a reference survival model that we believe will make errors that will also be made by an optimal survival tree and use the errors made by the reference model as a lower bound in our branch-and-bound method. 
We denote the reference model as $T$ and let $\hat{S}^{T}_{\x_i}(\cdot)$ be its predicted survival function for sample $i$. Define $s_a$ as the subset of training samples that satisfy a boolean assertion $a$, such that:
$
s_a := \{ i: a(\x_i) = \text{True}, i \in \{1, 2, \cdots, N\}\}$, 
$\X(s_a):= \{ \x_i: i \in s_a\}$,
$\c(s_a):= \{ c_i: i \in s_a\}$, and
$\y(s_a):= \{ y_i: i \in s_a\}$,
which corresponds to a subproblem defined in Section \ref{sec:dp} (e.g., the boolean value at index $i$ in $s$ is True for $i\in s_a$). 
We define our guessed lower bound of subproblem $s_a$ as the error made by the reference model $T$ on these samples plus a complexity penalty $\lambda$:
\begin{equation}\label{eq:lb_guess}
    lb_\text{guess}(s_a) = \mathcal{L}(\hat{\S}^{T},\X(s_a), \c(s_a), \y(s_a)) + \lambda
\end{equation}
where $\hat{\S}^{T} = \{ \hat{S}^{T}_{\x_i}(\cdot): i \in s_a\}$ and 
\begin{align*}
    &\mathcal{L}(\hat{\S}^{T},\X(s_a), \c(s_a), \y(s_a)) = \\
    &\sum_{i \in s_a} \mathcal{L}(\hat{S}^{T}_{\x_i}(\cdot),\X(s_a), \c(s_a), \y(s_a)),
\end{align*} which is the sum of sample losses defined in Equation \ref{eq:sample_loss}. 
We set the guessed lower bound $lb_\text{guess}$ as the initial lower bound for each subproblem. If the initial upper bound is less than or equal to $lb_\text{guess}$, then we consider this subproblem solved without further exploration of that portion of search space. Even when the initial lower bound is less than the initial upper bound, $lb_\text{guess}$ still improves runtime. Recall that during the optimization process, the lower bound of a subproblem never decreases and the upper bound of it never increases. Since the updates of a subproblem's lower bound rely on the lower bounds of its children, tighter initial lower bounds of children help the parent's lower bound converge to its upper bound. 

It can happen that we miss the true optimal solution of the subproblem $s_a$ if $lb_\text{guess} > R^{\ast}(s_a)$, where $R^{\ast}(s_a)$ is the optimal objective of subproblem $s_a$. This could impact the optimality of the optimization problem. We next quantify how much performance we might sacrifice by using $lb_\text{guess}$. 
Theorem \ref{thm:guess_obj_opt} shows that the distance to the true optimal solution depends on the performance of the reference model, and, under certain circumstances, we do not lose optimality at all. 
\begin{theorem}\label{thm:guess_obj_opt}
(Guarantee on guessed survival tree performance). Given dataset $\{\X, \c, \y\}$, depth constraint $d$, leaf penalty $\lambda$ and a reference model $T$, let $t_\text{guess}$ be the tree returned using $lb_\text{guess}$ defined in Equation \ref{eq:lb_guess} and $R(t_\text{guess}, \X, \c, \y)$ be its objective. Let $t^\ast$ be the true optimal tree on the training set according to the regularized objective defined in Equation \ref{eq:obj1}. We have:
\begin{eqnarray}\label{eq:guarantee}\nonumber
R(t_\text{guess}, \X, \c, \y) \leq \lambda H_{t^\ast} + \\
   \sum_{i=1}^N \max \left\{\mathcal{L}(\hat{S}^{T}_{\x_i}(\cdot),\x_i, c_i, y_i), \mathcal{L}(\hat{S}^{t^\ast}_{\x_i}(\cdot),\x_i, c_i, y_i) \right \}
\end{eqnarray}
where $\hat{S}^{T}_{\x_i}(\cdot)$ is the predicted survival function of reference model $T$ on sample $i$, and $\hat{S}^{t^\ast}_{\x_i}(\cdot)$ is the predicted survival function of optimal tree $t^\ast$.
That is, the objective of the guessed tree is no worse than the union of errors made by the reference model and the optimal tree.\end{theorem}
Under certain circumstances, the tree returned does not lose optimality.
\begin{corollary}\label{corollary}
Let $T, t_\text{guess}, t^\ast$ defined as in Theorem \ref{thm:guess_obj_opt}. If the reference model $T$ performs no worse than $t^\ast$ on each sample, the tree returned using $lb_\text{guess}$ is still optimal. 
\end{corollary}
In other words, if the reference model has good training performance (even if it is overfitted), the tree returned after using this optional guessing  technique is still optimal. 

\section{EXPERIMENTS}\label{sec:exp}
We ran experiments on 17 datasets (11 real-world survival datasets and 6 synthetic datasets from regression tasks), whose details are described in Appendix \ref{sec:setup}. Our experiments answer the following questions. 1) How far from optimal are existing survival tree methods (Section \ref{sec:optimality})?
2) How well do optimal sparse survival trees generalize (Section \ref{sec:generalization})?
3) How long does \ourmethod{} take to find the optimal survival trees (Section \ref{sec:run_time})?
4) How does \ourmethod{} scale on large datasets (Section \ref{sec:scalability})?
5) What do optimal survival trees look like (Section \ref{sec:opt_trees})?

We used Conditional Inference Trees (CTree) \citep{hothorn2015ctree},
Recursive Partitioning and Regression Trees (RPART)\citep{rpart}, and
the SurvivalTree model in Scikit-survival (SkSurv) \citep{sksurv} as baselines. Interpretable AI (IAI) implements the OST algorithm proposed by \citet{bertsimas2022optimal}. We were given a license to the (proprietary) software, but source code was unavailable. When we tried to run the experiments, it frequently crashed. As a result, we were unable to include results from it below. We discuss Interpretable AI further in Appendix \ref{exp:iai}. 

We use several metrics to evaluate the quality of our survival trees, discussed in Appendix \ref{sec:metrics}.
This includes the Integrated Brier Score Ratio (IBS Ratio -- higher is better). 
The IBS Ratio of tree $t$ is:
\begin{equation}
    \text{IBS Ratio}(t) = 1 - \frac{\mathcal{L}(t,\X, \c, \y)}{\mathcal{L}(t_0,\X, \c, \y)}
\end{equation}
where $\mathcal{L}$ is the IBS loss defined in Equation \ref{eq:tree_loss} and $t_0$ is a single node containing all samples (equivalently a KM estimator ignoring all features). \textit{All IBS scores in our experiments below refer to IBS Ratios}.

We also use other metrics, namely
Harrell's C-index and Uno's C-index (which are concordance metrics, higher is better), and Cumulative-Dynamic-AUC (higher is better); see Appendix \ref{sec:metrics}. 
Note that these metrics are not additive, which means that if we want to optimize them, then we must construct the entire survival tree. Therefore we can use these metrics only in the evaluation stage, not for training. We found that even though we optimize the IBS, our optimal survival trees often perform better on the other metrics as well, which we show in Section \ref{sec:quality} and Appendix \ref{exp:quality}.
\subsection{Optimality}\label{sec:optimality}
\begin{figure*}[htbp]
    \centering
    \includegraphics[width=0.32\textwidth]{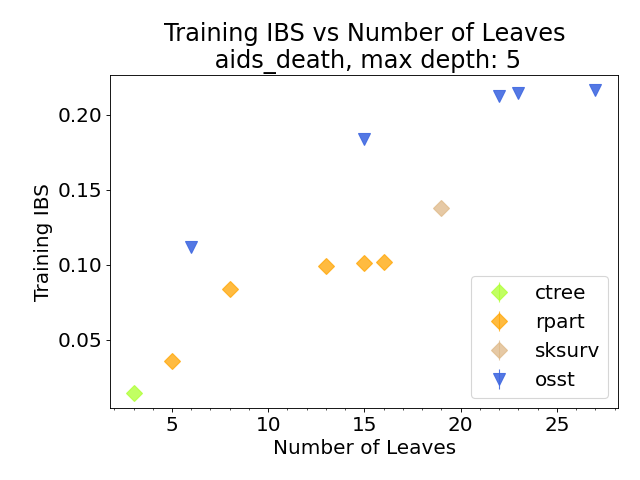}
    \includegraphics[width=0.32\textwidth]{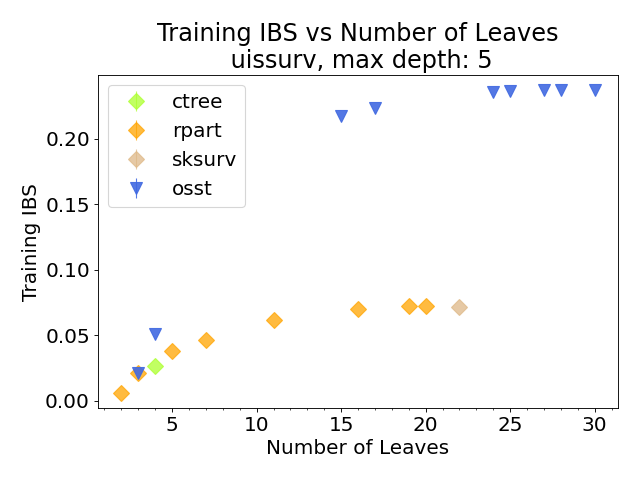}
    \includegraphics[width=0.32\textwidth]{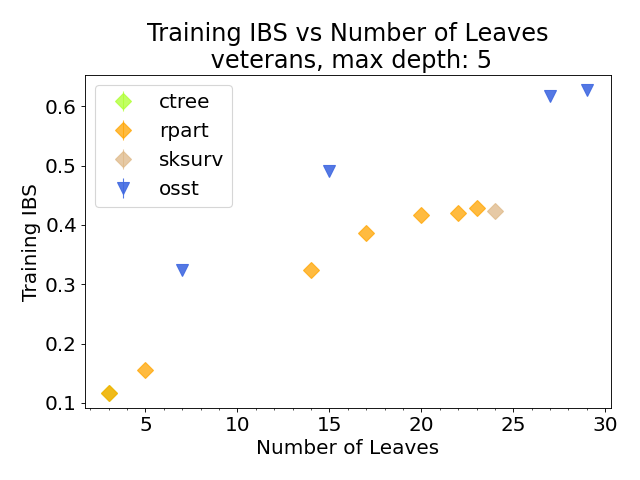}
    \caption{Training Score (IBS Ratio) of CTree, RPART, SkSurv and OSST on datasets: aids\_death, uissurv, veterans, max depth 5.}
    \label{fig:optimality-main}
\end{figure*}

For our method and baselines, we used different hyperparameters to generate trees of various sizes and show the relationship between loss and sparsity. 
Figure \ref{fig:optimality-main} demonstrates that OSST \textit{produces better performance (IBS Ratio) than all other methods}. Compared with other methods, the trees returned by OSST have higher performance and fewer leaves, which means that the trees found by OSST have \textit{higher quality in both test performance and sparsity}. Importantly, \textit{since OSST is the only method to produce optimal trees, other methods cannot quantify closeness to optimality without OSST}. More extensive results appear in Appendix \ref{exp:lvs}.
\begin{figure*}[htbp]
    \centering
    \includegraphics[width=0.24\textwidth]{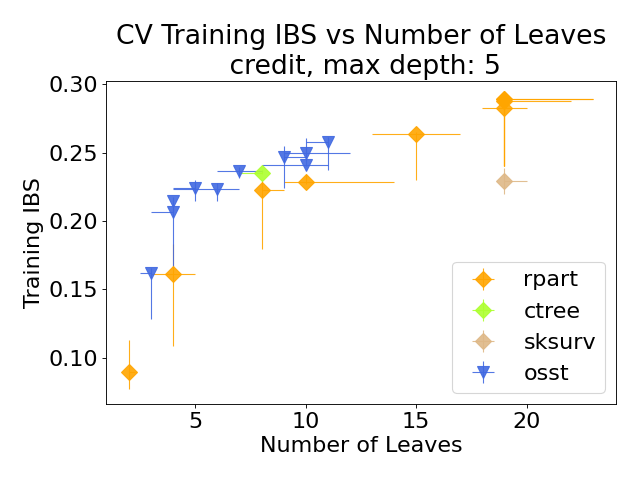}
    \includegraphics[width=0.24\textwidth]{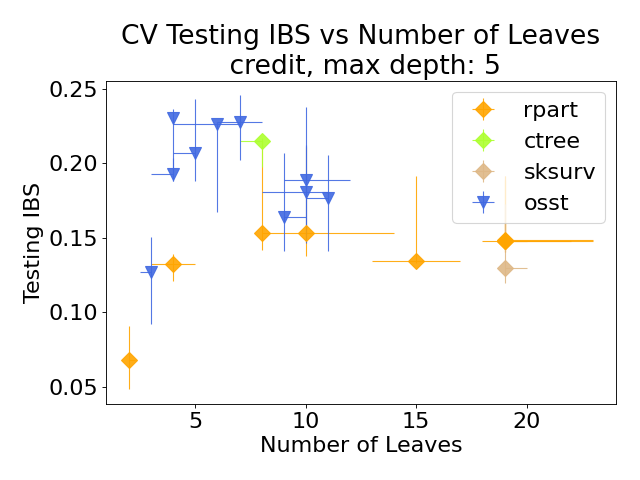}
    \includegraphics[width=0.24\textwidth]{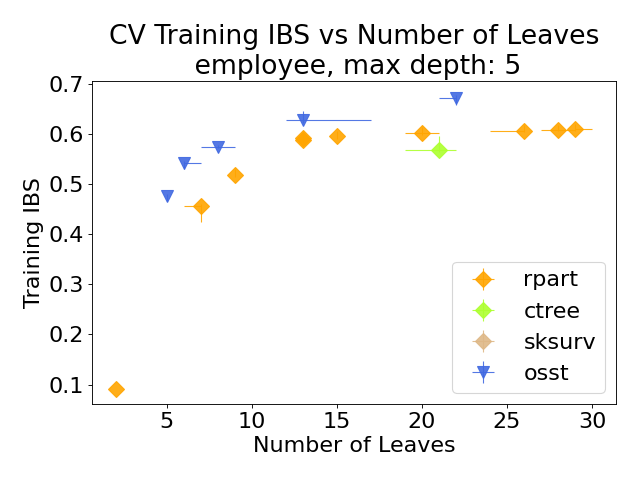}
    \includegraphics[width=0.24\textwidth]{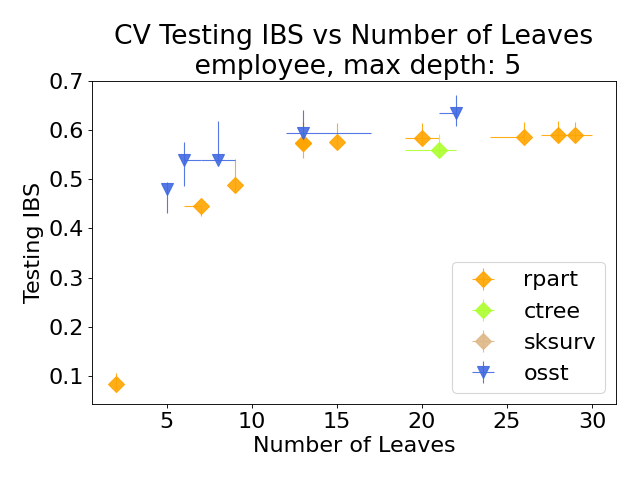}
    \caption{5-fold cross validation of CTree, RPART, SkSurv and OSST on datasets: credit, employee, max depth 5.}
    \label{fig:cv-main}
\end{figure*}
\begin{figure*}[htbp]
    \centering
    \includegraphics[width=0.31\textwidth]{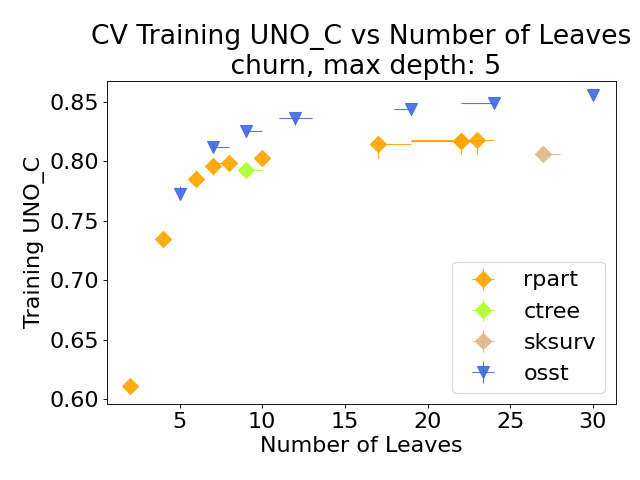}
    \includegraphics[width=0.31\textwidth]{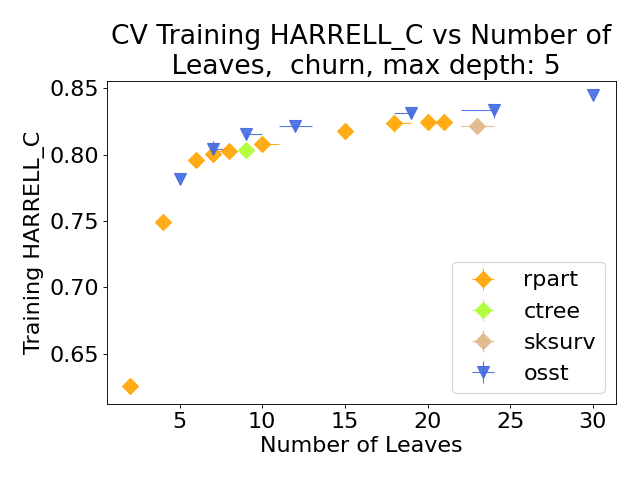}
    \includegraphics[width=0.31\textwidth]{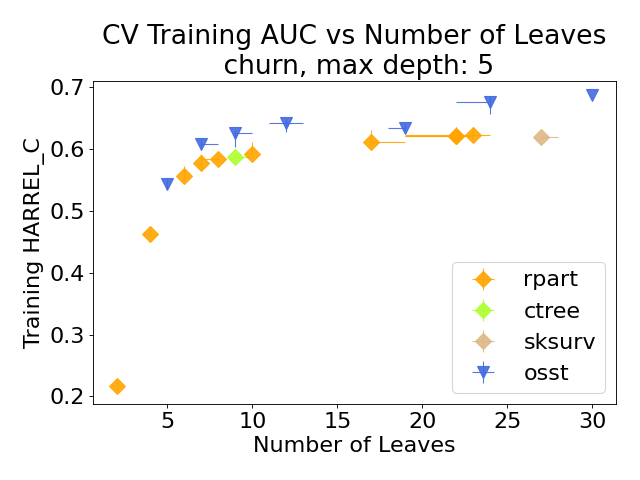}
    \includegraphics[width=0.31\textwidth]{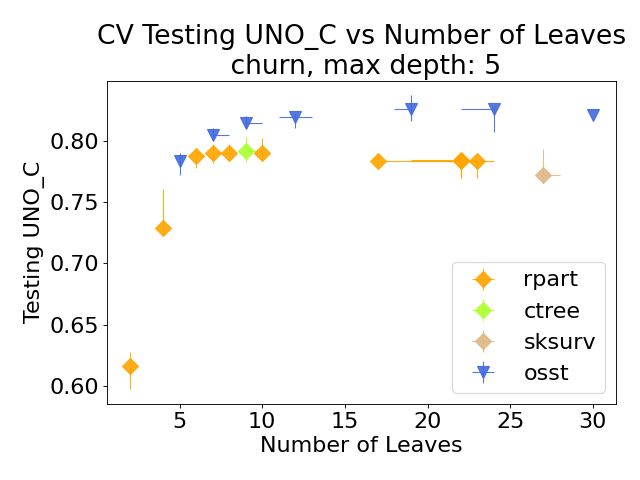}
    \includegraphics[width=0.31\textwidth]{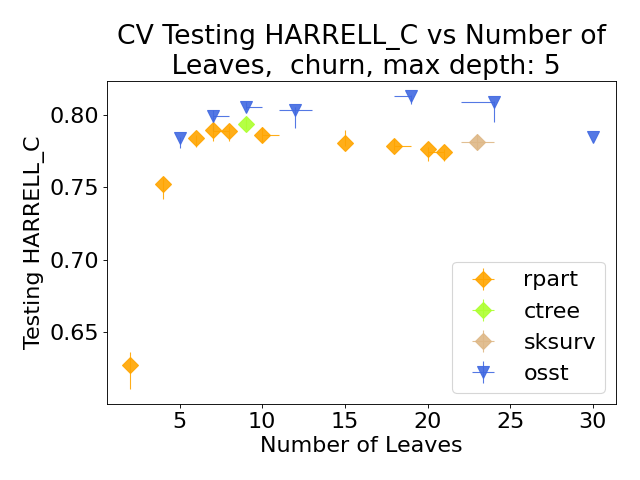}
    \includegraphics[width=0.31\textwidth]{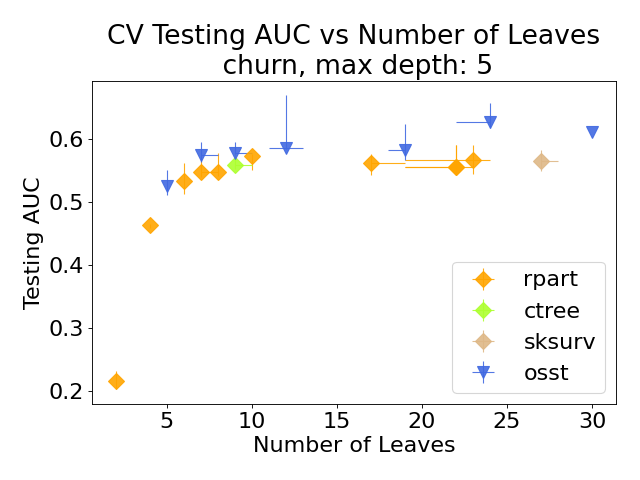}
    \caption{Testing performance of CTree, RPART, SkSurv and OSST on churn dataset, max depth 5, using different metrics. Cross-validation was used for confidence intervals.}
    \label{fig:metrics-main}
\end{figure*}
\subsection{Generalization}\label{sec:generalization}
We ran 5-fold cross-validation experiments on various datasets; the results show that OSST \textit{generalizes well}. Figure \ref{fig:cv-main} shows that optimal sparse survival trees \textit{also obtain higher testing performance.} More cross-validation results appear in Appendix \ref{exp:cv}.
\subsection{Comprehesive Quality}\label{sec:quality}
We evaluated trees returned by all methods again, using metrics defined in Appendix \ref{sec:metrics}. Figure \ref{fig:metrics-main} shows that the trees found by OSST also have higher Uno's C-index, Harrell's C-index and Cumulative-Dynamic-AUC, which indicates \textit{better overall quality}.
\subsection{Running Time}\label{sec:run_time}
Table \ref{tab:run_time-main} shows the 60-trial average run time and standard deviation of OSST on each dataset, using different configurations. The trees vary in complexity between 4 and 64 leaves. \textit{OSST is often able to find the optimal survival trees within a few seconds.} More details appear in Appendix \ref{exp:time}.
\begin{table}[htbp]
    \centering
    \begin{tabular}{|c|c|}
    \hline
        Dataset &  running time(s) \\\hline
         aids & $2.28 (\pm 3.66)$ \\\hline 
         aids\_death & $2.05 (\pm 3.82)$ \\\hline 
         maintenance & $0.78 (\pm 1.12)$ \\\hline 
         uissurv & $3.41 (\pm 2.86)$ \\\hline 
         veterans & $0.35 (\pm 0.53)$ \\\hline 
         whas500 & $8.76 (\pm 14.7)$\\\hline
         gbsg2 & $0.20 (\pm 0.25)$ \\\hline 
         insurance & $0.08 (\pm 0.02)$ \\\hline 
         sync &  $2.65 (\pm 1.93)$\\\hline
    \end{tabular}
    \setlength{\abovecaptionskip}{10pt}
    \caption{Summary of OSST average running time with different configurations on various datasets.}
    \label{tab:run_time-main}
\end{table}
\subsection{Scalability}\label{sec:scalability}
\begin{figure*}[htbp]
    \centering
    \includegraphics[width=0.32\textwidth]{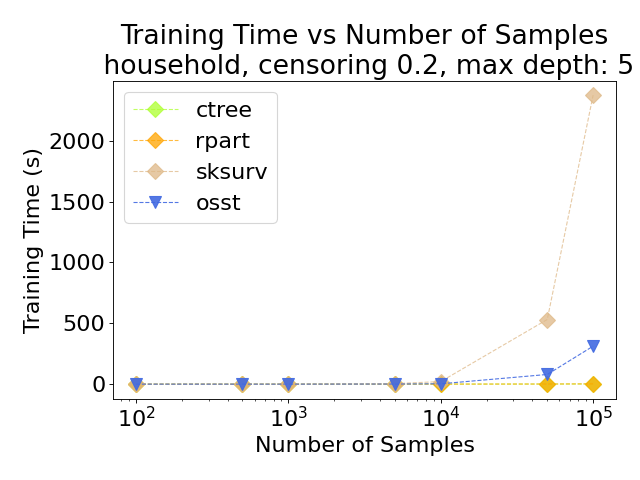}
    \includegraphics[width=0.32\textwidth]{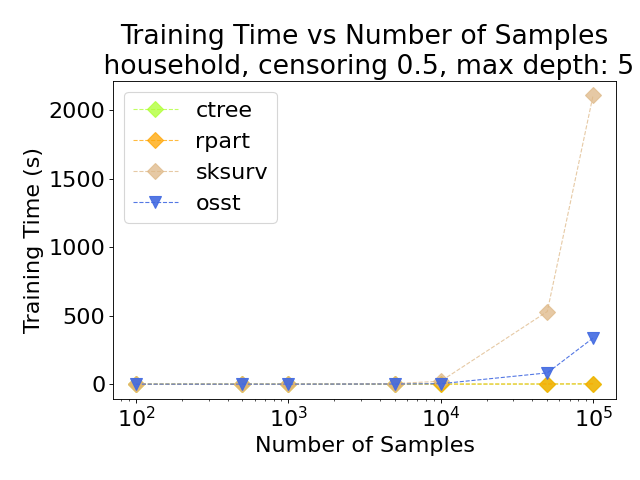}
    \includegraphics[width=0.32\textwidth]{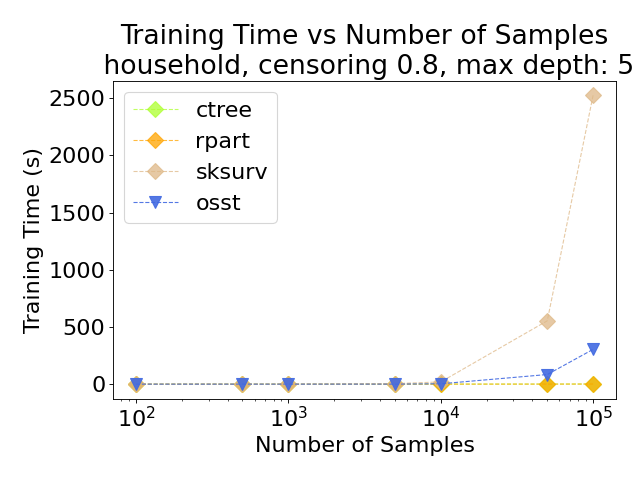}
    \caption{Training time of CTree, RPART, SkSurv and OSST as a function of sample size on household dataset, $d=5,\lambda=0.01$ (60-minutes time limit).}
    \label{fig:scalability_main}
\end{figure*}

Figure \ref{fig:scalability_main} shows that OSST achieved similar scalability performance to greedy methods with datasets of fewer than 10K samples. It is slower than CTree and RPART due to their greedy nature, but interestingly, it scales better than SkSurv when the number of samples exceeds 10K. The details of this experiment can be found in Appendix \ref{exp:scalability}. Again, we note that greedy methods do not have any performance guarantees (unlike OSST).
\subsection{Optimal Survival Trees}\label{sec:opt_trees}
\begin{figure*}[htbp]
    \centering
    \includegraphics[height = 3in]{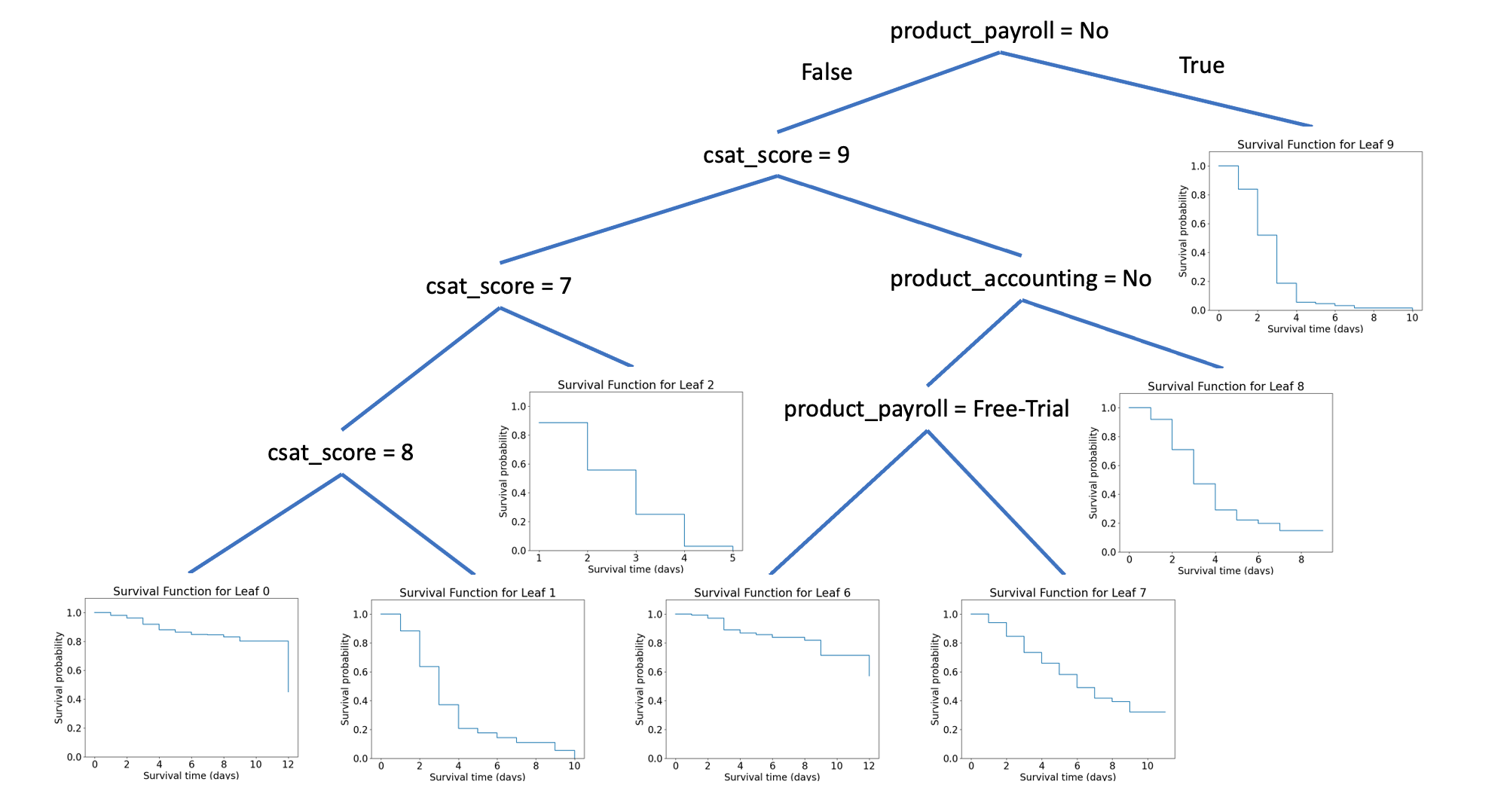}
    \caption{Optimal survival tree produced by OSST for churn dataset, 7 leaves. IBS ratio: $48.68\%$}
    \label{fig:churn_osst_main}
\end{figure*}

\begin{figure*}[htbp]
    \centering
    \includegraphics[height = 3in]{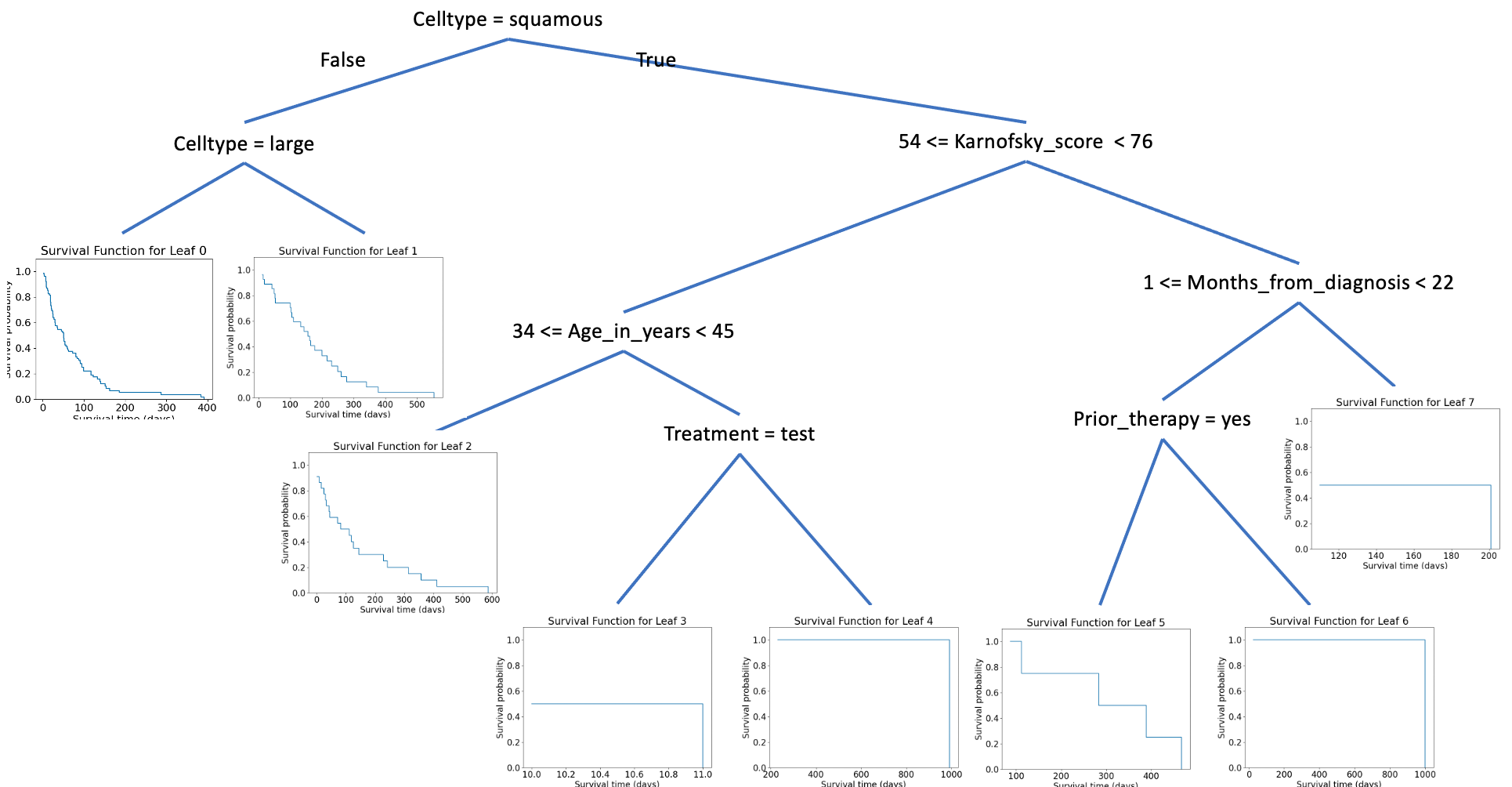}
    \caption{Optimal survival tree produced by OSST for veterans dataset, 8 leaves. IBS ratio: $32.83\%$}
    \label{fig:veterans_osst_main}
\end{figure*}

Figure \ref{fig:churn_osst_main} and \ref{fig:veterans_osst_main} show two example optimal sparse survival trees trained on the veterans and churn dataset. More trees can be found in Appendix \ref{exp:opt_trees}.
\section{LIMITATIONS}
We limited our baselines to other interpretable models. In high-stakes decision making, practitioners are ethically unable to use uninterpretable models.
Particularly in medical domains where data are messy and incomplete, interpretability is essential \citep{Council2019,Ellis2022}.
At the same time, sparsity comes with a cost; it is not clear whether sparse survival techniques achieve the performance of black box models (see Appendix \ref{sec:black_box} for more detail).

\section{CONCLUSION}
We provide a practical algorithm that is able to find provably-optimal sparse survival trees within a reasonable time, despite the hardness of fully optimizing a survival tree. Our method quickly finds optimal sparse survival models that generalize well and scales nicely to large datasets. 
There are many possible directions for future work. One is extending the optimized objective to other metrics mentioned before and possibly new metrics defined by users. The other one is to find systematic ways to create reference models that substantially speed up the search without impacting the performance of the returned trees.

\section*{Code Availability}
The implementation of \textbf{Optimal Sparse Survival Trees} can be found at \url{https://github.com/ruizhang1996/optimal-sparse-survival-trees-public}.

\section*{Acknowledgements}
We acknowledge the support of the Natural Sciences and Engineering Research Council of Canada (NSERC). Nous remercions le Conseil de recherches en sciences naturelles et en génie du Canada (CRSNG) de son soutien.

We acknowledge the following grants: NIH 1R01HL166233-01 and NIH/NIDA R01DA054994.

\clearpage

\bibliography{references}

\begin{thebibliography}{55}
\providecommand{\natexlab}[1]{#1}
\providecommand{\url}[1]{\texttt{#1}}
\expandafter\ifx\csname urlstyle\endcsname\relax
  \providecommand{\doi}[1]{doi: #1}\else
  \providecommand{\doi}{doi: \begingroup \urlstyle{rm}\Url}\fi

\bibitem[Bender et~al.(2018)Bender, Groll, and Scheipl]{bender2018generalized}
A.~Bender, A.~Groll, and F.~Scheipl.
\newblock A generalized additive model approach to time-to-event analysis.
\newblock \emph{Statistical Modelling}, 18\penalty0 (3-4):\penalty0 299--321,
  2018.

\bibitem[Bertsimas et~al.(2022)Bertsimas, Dunn, Gibson, and
  Orfanoudaki]{bertsimas2022optimal}
D.~Bertsimas, J.~Dunn, E.~Gibson, and A.~Orfanoudaki.
\newblock Optimal survival trees.
\newblock \emph{Machine Learning}, 111\penalty0 (8):\penalty0 2951--3023, 2022.

\bibitem[Che et~al.(2018)Che, Purushotham, Cho, Sontag, and
  Liu]{che2018recurrent}
Z.~Che, S.~Purushotham, K.~Cho, D.~Sontag, and Y.~Liu.
\newblock Recurrent neural networks for multivariate time series with missing
  values.
\newblock \emph{Scientific Reports}, 8\penalty0 (1):\penalty0 6085, 2018.

\bibitem[Ching et~al.(2018)Ching, Zhu, and Garmire]{ching2018cox}
T.~Ching, X.~Zhu, and L.~X. Garmire.
\newblock Cox-nnet: an artificial neural network method for prognosis
  prediction of high-throughput omics data.
\newblock \emph{PLoS Computational Biology}, 14\penalty0 (4):\penalty0
  e1006076, 2018.

\bibitem[Choi(2018)]{kaggle:insurance}
M.~Choi.
\newblock Kaggle insurance data, 2018.
\newblock URL \url{https://www.kaggle.com/datasets/mirichoi0218/insurance}.

\bibitem[Ciampi et~al.(1988)Ciampi, Hogg, McKinney, and
  Thiffault]{ciampi1988recpam}
A.~Ciampi, S.~A. Hogg, S.~McKinney, and J.~Thiffault.
\newblock Recpam: a computer program for recursive partition and amalgamation
  for censored survival data and other situations frequently occurring in
  biostatistics. i. methods and program features.
\newblock \emph{Computer Methods and Programs in Biomedicine}, 26\penalty0
  (3):\penalty0 239--256, 1988.

\bibitem[Council(2019)]{Council2019}
J.~Council.
\newblock Data challenges are halting {AI} projects, {IBM} executive says.
\newblock \emph{The Wall Street Journal}, 2019.
\newblock URL
  \url{https://www.wsj.com/articles/data-challenges-are-halting-ai-projects-ibm-executive-says-11559035800}.

\bibitem[Cox(1972)]{cox1972regression}
D.~R. Cox.
\newblock Regression models and life-tables.
\newblock \emph{Journal of the Royal Statistical Society: Series B
  (Methodological)}, 34\penalty0 (2):\penalty0 187--202, 1972.

\bibitem[Cox(2018)]{cox2018analysis}
D.~R. Cox.
\newblock \emph{Analysis of binary data}.
\newblock Routledge, 2018.

\bibitem[Davis and Anderson(1989)]{davis1989exponential}
R.~B. Davis and J.~R. Anderson.
\newblock Exponential survival trees.
\newblock \emph{Statistics in Medicine}, 8\penalty0 (8):\penalty0 947--961,
  1989.

\bibitem[Dispenzieri et~al.(2012)Dispenzieri, Katzmann, Kyle, Larson, Therneau,
  Colby, Clark, Mead, Kumar, Melton~III, et~al.]{dispenzieri2012use}
A.~Dispenzieri, J.~A. Katzmann, R.~A. Kyle, D.~R. Larson, T.~M. Therneau, C.~L.
  Colby, R.~J. Clark, G.~P. Mead, S.~Kumar, L.~J. Melton~III, et~al.
\newblock Use of nonclonal serum immunoglobulin free light chains to predict
  overall survival in the general population.
\newblock In \emph{Mayo Clinic Proceedings}, volume~87, pages 517--523.
  Elsevier, 2012.

\bibitem[Dua and Graff(2017)]{Dua:2019}
D.~Dua and C.~Graff.
\newblock {UCI} machine learning repository.
\newblock http://archive.ics.uci.edu/ml, 2017.
\newblock Accessed: 2022-04-01.

\bibitem[Dunn(2018)]{Dunn2018}
J.~Dunn.
\newblock \emph{Optimal Trees for Prediction and Prescription}.
\newblock PhD thesis, Massachusetts Institute of Technology, 2018.

\bibitem[Ellis et~al.(2022)Ellis, Sander, and Limon]{Ellis2022}
R.~J. Ellis, R.~M. Sander, and A.~Limon.
\newblock Twelve key challenges in medical machine learning and solutions.
\newblock \emph{Intelligence-Based Medicine}, 6:\penalty0 100068, 2022.
\newblock ISSN 2666-5212.
\newblock \doi{https://doi.org/10.1016/j.ibmed.2022.100068}.
\newblock URL
  \url{https://www.sciencedirect.com/science/article/pii/S2666521222000217}.

\bibitem[Fotso et~al.(2019)]{pysurvival_cite}
S.~Fotso et~al.
\newblock {PySurvival}: Open source package for survival analysis modeling,
  2019.
\newblock URL \url{https://www.pysurvival.io/}.

\bibitem[Giunchiglia et~al.(2018)Giunchiglia, Nemchenko, and van~der
  Schaar]{giunchiglia2018rnn}
E.~Giunchiglia, A.~Nemchenko, and M.~van~der Schaar.
\newblock Rnn-surv: A deep recurrent model for survival analysis.
\newblock In \emph{Artificial Neural Networks and Machine Learning--ICANN 2018:
  27th International Conference on Artificial Neural Networks, Rhodes, Greece,
  October 4-7, 2018, Proceedings, Part III 27}, pages 23--32. Springer, 2018.

\bibitem[Gordon and Olshen(1985)]{gordon1985tree}
L.~Gordon and R.~A. Olshen.
\newblock Tree-structured survival analysis.
\newblock \emph{Cancer Treatment Reports}, 69\penalty0 (10):\penalty0
  1065--1069, 1985.

\bibitem[Graf et~al.(1999)Graf, Schmoor, Sauerbrei, and
  Schumacher]{graf1999assessment}
E.~Graf, C.~Schmoor, W.~Sauerbrei, and M.~Schumacher.
\newblock Assessment and comparison of prognostic classification schemes for
  survival data.
\newblock \emph{Statistics in Medicine}, 18\penalty0 (17-18):\penalty0
  2529--2545, 1999.

\bibitem[Grubinger et~al.(2014)Grubinger, Zeileis, and
  Pfeiffer]{grubinger2014evtree}
T.~Grubinger, A.~Zeileis, and K.-P. Pfeiffer.
\newblock evtree: Evolutionary learning of globally optimal classification and
  regression trees in r.
\newblock \emph{Journal of Statistical Software}, 61:\penalty0 1--29, 2014.

\bibitem[Harrell~Jr et~al.(1996)Harrell~Jr, Lee, and
  Mark]{harrell1996multivariable}
F.~E. Harrell~Jr, K.~L. Lee, and D.~B. Mark.
\newblock Multivariable prognostic models: issues in developing models,
  evaluating assumptions and adequacy, and measuring and reducing errors.
\newblock \emph{Statistics in Medicine}, 15\penalty0 (4):\penalty0 361--387,
  1996.

\bibitem[Hothorn et~al.(2006)Hothorn, Hornik, and Zeileis]{hothorn2006unbiased}
T.~Hothorn, K.~Hornik, and A.~Zeileis.
\newblock Unbiased recursive partitioning: A conditional inference framework.
\newblock \emph{Journal of Computational and Graphical Statistics}, 15\penalty0
  (3):\penalty0 651--674, 2006.

\bibitem[Hothorn et~al.(2015)Hothorn, Hornik, and Zeileis]{hothorn2015ctree}
T.~Hothorn, K.~Hornik, and A.~Zeileis.
\newblock ctree: Conditional inference trees.
\newblock \emph{The Comprehensive R Archive Network}, 8, 2015.

\bibitem[Hu et~al.(2019)Hu, Rudin, and Seltzer]{HuRuSe2019}
X.~Hu, C.~Rudin, and M.~Seltzer.
\newblock Optimal sparse decision trees.
\newblock In \emph{Proceedings of Conference on Neural Information Processing
  Systems {(NeurIPS)}}, 2019.

\bibitem[Hung and Chiang(2010)]{hung2010estimation}
H.~Hung and C.-T. Chiang.
\newblock Estimation methods for time-dependent auc models with survival data.
\newblock \emph{Canadian Journal of Statistics}, 38\penalty0 (1):\penalty0
  8--26, 2010.

\bibitem[Ishwaran and Kogalur(2007)]{ishwaran2007random}
H.~Ishwaran and U.~B. Kogalur.
\newblock Random survival forests for r.
\newblock \emph{R News}, 7\penalty0 (2):\penalty0 25--31, 2007.

\bibitem[Jin et~al.(2004)Jin, Lu, Stone, and Black]{jin2004alternative}
H.~Jin, Y.~Lu, K.~Stone, and D.~M. Black.
\newblock Alternative tree-structured survival analysis based on variance of
  survival time.
\newblock \emph{Medical Decision Making}, 24\penalty0 (6):\penalty0 670--680,
  2004.

\bibitem[Kalbfleisch and Prentice(2011)]{kalbfleisch2011statistical}
J.~D. Kalbfleisch and R.~L. Prentice.
\newblock \emph{The statistical analysis of failure time data}.
\newblock John Wiley \& Sons, 2011.

\bibitem[Kaplan and Meier(1958)]{kaplan1958nonparametric}
E.~L. Kaplan and P.~Meier.
\newblock Nonparametric estimation from incomplete observations.
\newblock \emph{Journal of the American Statistical Association}, 53\penalty0
  (282):\penalty0 457--481, 1958.

\bibitem[Katzman et~al.(2018)Katzman, Shaham, Cloninger, Bates, Jiang, and
  Kluger]{katzman2018deepsurv}
J.~L. Katzman, U.~Shaham, A.~Cloninger, J.~Bates, T.~Jiang, and Y.~Kluger.
\newblock Deepsurv: personalized treatment recommender system using a cox
  proportional hazards deep neural network.
\newblock \emph{BMC Medical Research Methodology}, 18\penalty0 (1):\penalty0
  1--12, 2018.

\bibitem[Keles and Segal(2002)]{kelecs2002residual}
S.~Keles and M.~R. Segal.
\newblock Residual-based tree-structured survival analysis.
\newblock \emph{Statistics in Medicine}, 21\penalty0 (2):\penalty0 313--326,
  2002.

\bibitem[Kleinbaum and Klein(1996)]{kleinbaum1996survival}
D.~G. Kleinbaum and M.~Klein.
\newblock \emph{Survival analysis, a self-learning text}.
\newblock Springer, 1996.

\bibitem[Kvamme et~al.(2019)Kvamme, Borgan, and Scheel]{kvamme2019time}
H.~Kvamme, {\O}.~Borgan, and I.~Scheel.
\newblock Time-to-event prediction with neural networks and cox regression.
\newblock \emph{Journal of Machine Learning Research}, 20:\penalty0 1--30,
  2019.

\bibitem[Lambert and Chevret(2016)]{lambert2016summary}
J.~Lambert and S.~Chevret.
\newblock Summary measure of discrimination in survival models based on
  cumulative/dynamic time-dependent roc curves.
\newblock \emph{Statistical Methods in Medical Research}, 25\penalty0
  (5):\penalty0 2088--2102, 2016.

\bibitem[Lawless(2011)]{lawless2011statistical}
J.~F. Lawless.
\newblock \emph{Statistical models and methods for lifetime data}.
\newblock John Wiley \& Sons, 2011.

\bibitem[LeBlanc and Crowley(1992)]{leblanc1992relative}
M.~LeBlanc and J.~Crowley.
\newblock Relative risk trees for censored survival data.
\newblock \emph{Biometrics}, pages 411--425, 1992.

\bibitem[LeBlanc and Crowley(1993)]{leblanc1993survival}
M.~LeBlanc and J.~Crowley.
\newblock Survival trees by goodness of split.
\newblock \emph{Journal of the American Statistical Association}, 88\penalty0
  (422):\penalty0 457--467, 1993.

\bibitem[Lee et~al.(2018)Lee, Zame, Yoon, and Van Der~Schaar]{lee2018deephit}
C.~Lee, W.~Zame, J.~Yoon, and M.~Van Der~Schaar.
\newblock Deephit: A deep learning approach to survival analysis with competing
  risks.
\newblock In \emph{Proceedings of the {AAAI} Conference on Artificial
  Intelligence}, volume~32, 2018.

\bibitem[Lemeshow et~al.(2011)Lemeshow, May, and
  Hosmer~Jr]{lemeshow2011applied}
S.~Lemeshow, S.~May, and D.~W. Hosmer~Jr.
\newblock \emph{Applied survival analysis: regression modeling of time-to-event
  data}.
\newblock John Wiley \& Sons, 2011.

\bibitem[Lin et~al.(2020)Lin, Zhong, Hu, Rudin, and
  Seltzer]{lin2020generalized}
J.~Lin, C.~Zhong, D.~Hu, C.~Rudin, and M.~Seltzer.
\newblock Generalized and scalable optimal sparse decision trees.
\newblock In \emph{Proceedings of International Conference on Machine Learning
  ({ICML})}, pages 6150--6160, 2020.

\bibitem[McTavish et~al.(2022)McTavish, Zhong, Achermann, Karimalis, Chen,
  Rudin, and Seltzer]{McTavishZhongEtAl2022}
H.~McTavish, C.~Zhong, R.~Achermann, I.~Karimalis, J.~Chen, C.~Rudin, and
  M.~Seltzer.
\newblock Fast sparse decision tree optimization via reference ensembles.
\newblock In \emph{Proceedings of {AAAI} Conference on Artificial
  Intelligence}, 2022.

\bibitem[Molinaro et~al.(2004)Molinaro, Dudoit, and Van~der
  Laan]{molinaro2004tree}
A.~M. Molinaro, S.~Dudoit, and M.~J. Van~der Laan.
\newblock Tree-based multivariate regression and density estimation with
  right-censored data.
\newblock \emph{Journal of Multivariate Analysis}, 90\penalty0 (1):\penalty0
  154--177, 2004.

\bibitem[Nijssen et~al.(2020)Nijssen, Schaus, et~al.]{nijssen2020}
S.~Nijssen, P.~Schaus, et~al.
\newblock Learning optimal decision trees using caching branch-and-bound
  search.
\newblock In \emph{Proceedings of AAAI Conference on Artificial Intelligence
  (AAAI)}, 2020.

\bibitem[Norton(1988)]{norton1988gompertzian}
L.~Norton.
\newblock A gompertzian model of human breast cancer growth.
\newblock \emph{Cancer Research}, 48\penalty0 (24\_Part\_1):\penalty0
  7067--7071, 1988.

\bibitem[P{\"o}lsterl(2020)]{sksurv}
S.~P{\"o}lsterl.
\newblock scikit-survival: A library for time-to-event analysis built on top of
  scikit-learn.
\newblock \emph{Journal of Machine Learning Research}, 21\penalty0
  (212):\penalty0 1--6, 2020.
\newblock URL \url{http://jmlr.org/papers/v21/20-729.html}.

\bibitem[Ripley and Ripley(2001)]{ripley2001neural}
B.~D. Ripley and R.~M. Ripley.
\newblock Neural networks as statistical methods in survival analysis.
\newblock \emph{Clinical Applications of Artificial Neural Networks},
  237:\penalty0 255, 2001.

\bibitem[Rudin et~al.(2022)Rudin, Chen, Chen, Huang, Semenova, and
  Zhong]{RudinEtAlSurvey2022}
C.~Rudin, C.~Chen, Z.~Chen, H.~Huang, L.~Semenova, and C.~Zhong.
\newblock {Interpretable machine learning: Fundamental principles and 10 grand
  challenges}.
\newblock \emph{Statistics Surveys}, 16\penalty0 (none):\penalty0 1 -- 85,
  2022.
\newblock \doi{10.1214/21-SS133}.
\newblock URL \url{https://doi.org/10.1214/21-SS133}.

\bibitem[Schumacher et~al.(1994)Schumacher, Bastert, Bojar, H{\"u}bner,
  Olschewski, Sauerbrei, Schmoor, Beyerle, Neumann, and
  Rauschecker]{schumacher1994randomized}
M.~Schumacher, G.~Bastert, H.~Bojar, K.~H{\"u}bner, M.~Olschewski,
  W.~Sauerbrei, C.~Schmoor, C.~Beyerle, R.~Neumann, and H.~Rauschecker.
\newblock Randomized 2 x 2 trial evaluating hormonal treatment and the duration
  of chemotherapy in node-positive breast cancer patients. german breast cancer
  study group.
\newblock \emph{Journal of Clinical Oncology}, 12\penalty0 (10):\penalty0
  2086--2093, 1994.

\bibitem[Segal(1988)]{segal1988regression}
M.~R. Segal.
\newblock Regression trees for censored data.
\newblock \emph{Biometrics}, pages 35--47, 1988.

\bibitem[Therneau and Atkinson(2019)]{rpart}
T.~Therneau and B.~Atkinson.
\newblock \emph{rpart: Recursive Partitioning and Regression Trees}, 2019.
\newblock URL \url{https://CRAN.R-project.org/package=rpart}.
\newblock R package version 4.1-15.

\bibitem[Therneau et~al.(1990)Therneau, Grambsch, and
  Fleming]{therneau1990martingale}
T.~M. Therneau, P.~M. Grambsch, and T.~R. Fleming.
\newblock Martingale-based residuals for survival models.
\newblock \emph{Biometrika}, 77\penalty0 (1):\penalty0 147--160, 1990.

\bibitem[Uno et~al.(2007)Uno, Cai, Tian, and Wei]{uno2007evaluating}
H.~Uno, T.~Cai, L.~Tian, and L.-J. Wei.
\newblock Evaluating prediction rules for t-year survivors with censored
  regression models.
\newblock \emph{Journal of the American Statistical Association}, 102\penalty0
  (478):\penalty0 527--537, 2007.

\bibitem[Uno et~al.(2011)Uno, Cai, Pencina, D'Agostino, and Wei]{uno2011c}
H.~Uno, T.~Cai, M.~J. Pencina, R.~B. D'Agostino, and L.-J. Wei.
\newblock On the c-statistics for evaluating overall adequacy of risk
  prediction procedures with censored survival data.
\newblock \emph{Statistics in Medicine}, 30\penalty0 (10):\penalty0 1105--1117,
  2011.

\bibitem[Zhang(1995)]{zhang1995splitting}
H.~Zhang.
\newblock Splitting criteria in survival trees.
\newblock In \emph{Statistical Modelling: Proceedings of the 10th International
  Workshop on Statistical Modelling Innsbruck, Austria, 10--14 July, 1995},
  pages 305--313. Springer, 1995.

\bibitem[Zhang et~al.(2023)Zhang, Xin, Seltzer, and Rudin]{zhang2023optimal}
R.~Zhang, R.~Xin, M.~Seltzer, and C.~Rudin.
\newblock Optimal sparse regression trees.
\newblock In \emph{Proceedings of the AAAI Conference on Artificial
  Intelligence}, pages 11270--11279, 2023.

\bibitem[Zhao and Feng(2019)]{zhao2019dnnsurv}
L.~Zhao and D.~Feng.
\newblock Dnnsurv: Deep neural networks for survival analysis using pseudo
  values.
\newblock \emph{arXiv preprint arXiv:1908.02337}, 2019.

\end{thebibliography}
\bibliographystyle{abbrvnat}

\clearpage
\appendix
\onecolumn
\section{Extended Related Work}\label{sec:related_work}

\textbf{Early Survival Analysis:} 
In 1958, \citet{kaplan1958nonparametric} proposed the first survival models, Kaplan-Meier (KM) curves, which are simple non-parametric models that do not use  covariates. Parametric models \citep{cox2018analysis, kalbfleisch2011statistical, lawless2011statistical, norton1988gompertzian, kleinbaum1996survival} usually rely on strong assumptions either for the hazard rate or the underlying distribution of the survival time. They assume the hazard function follows a particular distribution, such as the exponential, Weibull, or log-normal distribution. Cox proportional hazards regression \citep{cox1972regression} is a practical semi-parametric approach proposed in 1972, which makes the strong assumption of a constant hazard ratio between individuals and fails to capture non-linear relationships between covariates and  predicted outcomes. Even though the coefficient of a Cox model is easy to interpret as a hazard ratio, it loses interpretability if the model is not sparse in the number of features. 

\textbf{Decision Tree Learning for Survival Analysis:} 
Attempts have been made to adapt traditional decision trees for censored survival data by proposing various splitting criteria to heuristically construct survival trees. \citet{gordon1985tree} chose splits by minimizing the Wasserstein distance between the two child node's Kaplan–Meier curves.  \citet{davis1989exponential} used exponential log-likelihood loss for splitting. \citet{leblanc1992relative} measured node deviance for splitting while \citet{therneau1990martingale} and \citet{kelecs2002residual} used martingale residuals as the splitting criteria. \citet{zhang1995splitting} combined impurity measurements for observed time and portion of censored samples. Various authors proposed other techniques to maximize the distance between the two child nodes using various statistics \citep{leblanc1993survival, ciampi1988recpam, segal1988regression, hothorn2006unbiased, jin2004alternative}. \citet{molinaro2004tree} used Inverse Probability of Censoring Weight (IPCW) to reduce the bias caused by a high degree of censoring. These splitting criteria either minimize the impurity within nodes or maximize the dissimilarity between different nodes, but all of them use greedy approaches, which means if a bad split is chosen at the top, there is no way to correct it. Importantly, there is no way to determine whether the trees are optimal without a method like ours that provably optimizes the tree structure.

\textbf{Black Box Survival Models:}
The use of decision trees for survival analysis was further extended to more sophisticated forest models such as random survival forests \citep{ishwaran2007random} and conditional inference forests \citep{hothorn2006unbiased}. Neural networks have been applied to survival analysis tasks as well \citep{katzman2018deepsurv, ching2018cox,che2018recurrent, ripley2001neural, giunchiglia2018rnn}. These black box models are useful for performance comparisons, but are not generally useful in practice, particularly for high-stakes decisions.

\textbf{Modern Decision Tree Methods:}
\citet{HuRuSe2019, lin2020generalized, zhang2023optimal, grubinger2014evtree, nijssen2020, Dunn2018} have shown the success of optimal sparse trees in both classification and regression. A single sparse optimal tree can achieve the performance of complex black box models while providing interpretability \citep{McTavishZhongEtAl2022}. \citet{bertsimas2022optimal} proposed a survival tree model using local search techniques, but it holds an assumption similar to that of the Cox model; that is the ratio of the hazard functions for any two individuals is assumed to be constant over time and independent of the values of the covariates. As we show in Appendix \ref{exp:iai}, in cases where we can get their code to run, OSST produces better IBS ratios than their OST algorithm. 

\section{Theorems and Proofs}\label{sec:thms_and_prfs}
\subsection{Proof of Theorem \ref{thm:additive}}
\textbf{Theorem \ref{thm:additive}}
\textit{The loss of a survival tree is an additive function of the observations and leaves.}

\begin{proof}
From Equation \ref{eq:tree_loss} and \ref{eq:brierscore}, we know:
\begin{equation}
    \mathcal{L}(t,\X, \c, \y) = \frac{1}{y_{\max}} \int_0^{y_{\max}}BS(y) dy
\end{equation}
\begin{equation}
    = \frac{1}{y_{\max}} \int_0^{y_{\max}} \frac{1}{N} \sum_{i=1}^N \left\{\frac{(\hat{S}_{\x_i}(y) - 0)^2}{\hat{G}(y_i)}\cdot \mathbf{1}_{y_i \leq y, c_i = 1} + \frac{(\hat{S}_{\x_i}(y) - 1)^2}{\hat{G}(y)}\cdot \mathbf{1}_{y_i > y}\right\}dy.
\end{equation}
Extracting the constant term $\frac{1}{N}$ and putting the integral into the summation yields:
\begin{eqnarray*}
    &=& \frac{1}{y_{\max}}\frac{1}{N} \sum_{i=1}^N \left\{ \int_0^{y_{\max}} \frac{(\hat{S}_{\x_i}(y) - 1)^2}{\hat{G}(y)}\cdot \mathbf{1}_{y_i > y}dy
    + \int_0^{y_{\max}} \frac{(\hat{S}_{\x_i}(y) - 0)^2}{\hat{G}(y_i)}\cdot \mathbf{1}_{y_i \leq y, c_i = 1} dy \right\}\\
    &=& \frac{1}{y_{\max}}\frac{1}{N} \sum_{i=1}^N \left\{ \int_0^{y_i} \frac{(\hat{S}_{\x_i}(y) - 1)^2}{\hat{G}(y)}dy
    + c_i \int_{y_i}^{y_{\max}} \frac{(\hat{S}_{\x_i}(y) - 0)^2}{\hat{G}(y_i)}dy\right\}.
\end{eqnarray*}
Now we have proved that the loss of a survival tree is additive in observations. And each sample can only fall into one leave at a time:
\begin{eqnarray*}
&=& \frac{1}{y_{\max}}\frac{1}{N} \sum_{i=1}^N \left\{ \int_0^{y_i} \frac{(\hat{S}_{\x_i}(y) - 1)^2}{\hat{G}(y)}dy
    + c_i \int_{y_i}^{y_{\max}} \frac{(\hat{S}_{\x_i}(y) - 0)^2}{\hat{G}(y_i)}dy\right\}\cdot \mathbf{1}_{\capt(t_{\fix}, \x_i),}\\
&&+ \frac{1}{y_{\max}}\frac{1}{N} \sum_{i=1}^N \left\{ \int_0^{y_i} \frac{(\hat{S}_{\x_i}(y) - 1)^2}{\hat{G}(y)}dy
    + c_i \int_{y_i}^{y_{\max}} \frac{(\hat{S}_{\x_i}(y) - 0)^2}{\hat{G}(y_i)}dy\right\}\cdot \mathbf{1}_{\capt(t_{\splita}, \x_i)}.
\end{eqnarray*}
Therefore, the objective is additive in the observations, and observations can be grouped by leaves, so the loss is additive in the leaves.
\end{proof}

\subsection{Proof of Theorem \ref{thm:hierarchical_lb}}
\textbf{Theorem \ref{thm:hierarchical_lb}}
(Hierarchical Objective Lower Bound). \textit{Any tree $t' = (t'_{\fix}, \delta'_{\fix}, t'_{\splita}, \delta'_{\splita}, K', H_{t'}) \in \sigma(t)$ in the child tree set of $t = (t_{\fix}, \delta_{\fix}, t_{\splita}, \delta_{\splita}, K, H_t)$ obeys: \[R(t',\X,\c,\y) \geq \mathcal{L}(t_{\fix},\X,\c,\y) + \lambda H_t. \]}
That is, the objective lower bound of the parent tree holds for all its child trees. This bound ensures that we do not further explore child trees if the parent tree can be pruned via the lower bound.
\begin{proof}
As we know, $K' \geq K, H_{t'} > H_t$, since $t'$ is a child tree of $t$. The objective lower bound (which holds for all trees) of $t'$ is:
\begin{equation}\label{eq:child_lb}
 R(t',\X, \c,\y) \geq \mathcal{L}(t'_{\fix},\X,\y) + \lambda H_{t'}.   
\end{equation}
Since  $\mathcal{L}(t'_{\fix},\X, \c, \y) = \mathcal{L}(t_{\fix},\X, \c,\y) + \mathcal{L}(t'_{\fix} \setminus t_{\fix},\X, \c,\y)$ and the loss of $K' - K$ fixed leaves in $t$ is
\begin{equation}\nonumber
    \mathcal{L}(t'_{\fix} \setminus t_{\fix},\X,\c,\y) = \frac{1}{y_{\max}}\frac{1}{N} \sum_{i=1}^N \left\{ \int_0^{y_i} \frac{(\hat{S}_{\x_i}(y) - 1)^2}{\hat{G}(y)}dy
    + c_i \int_{y_i}^{y_{\max}} \frac{(\hat{S}_{\x_i}(y) - 0)^2}{\hat{G}(y_i)}dy\right\}\cdot \mathbf{1}_{\capt(t'_{\fix} \setminus t_{\fix}, \x_i)}
\end{equation}
and $0 \leq \hat{G}(\cdot) \leq 1$, so
\begin{equation}
    \mathcal{L}(t'_{\fix} \setminus t_{\fix},\X,\c, \y) \geq 0,
\end{equation} and thus we have:
\begin{equation}\label{eq:fix_loss}
 \mathcal{L}(t'_{\fix},\X,\c,\y) \geq \mathcal{L}(t_{\fix},\X,\c,\y)   
\end{equation}
therefore:
\begin{equation}\label{eq:hierachical_lb}
    R(t',\X,\c,\y) \geq \mathcal{L}(t_{\fix},\X,\c,\y) + \lambda H_{t}. 
\end{equation}
\end{proof}

\subsection{Proof of Theorem \ref{thm:look_ahead}}
\textbf{Theorem \ref{thm:look_ahead}}
\textit{(Objective Lower Bound with One-step Lookahead). Let $t = (t_{\fix}, \delta_{\fix}, t_{\splita}, \delta_{\splita}, K, H_t)$ be a tree with $H_t$ leaves. If $\mathcal{L}(t_{\fix},\X,\c,\y) + \lambda H_t + \lambda > R^c$, even if its objective lower bound $\mathcal{L}(t_{\fix},\X,\c,\y) + \lambda H_t \leq R^c$, then for any child tree $t' \in \sigma(t)$, $R(t', \X, \c,\y) > R^c$.
That is, even if a parent tree cannot be pruned via its objective lower bound, if $\mathcal{L}(t_{\fix},\X,\y) + \lambda H_t + \lambda > R^c$, all of its child trees are sub-optimal and can be pruned (and never explored). }

\begin{proof}
This bound adapts directly from OSRT \citep{zhang2023optimal}, where the proof can be found.
\end{proof}

\subsection{Proof of Lemma \ref{lm:equiv_loss}}
\textbf{Lemma \ref{lm:equiv_loss}}
\textit{
(Equivalent Loss). Let $u$ be a set of equivalent points defined as Equation \ref{eq:equiv_set}. We denote $^{\ast}S $ as the optimal step function that minimizes the IBS loss only for set $u$ (leaf contains set $u$ only), such that:
\[^{\ast}S = \argmin_{S} \sum_{k=1}^{|u|}\mathcal{L}(S,\x_{j_k}, \c_{j_k}, \y_{j_k}).\]
We define \textbf{Equivalent Loss} for set $u$ as 
\[\mathcal{E}_u = \sum_{k=1}^{|u|}\mathcal{L}(^{\ast} S,\x_{j_k}, \c_{j_k}, \y_{j_k})\]
then any leaf that captures set $u$ in a survival tree trained on dataset $\{\X, \c, \y\}$ has loss $\geq \mathcal{E}_u$.
}
\begin{proof}
In any survival tree, the points in set $u$ always get assigned to the same leaf. Let $l_u$ be a leaf node that contains set $u$ only and let $\hat{S}$ be its KM estimator. Since the KM estimator is not the minimizer of the IBS loss, the loss of $l_u$ obeys:
\begin{equation}\label{eqn:epsi}
\mathcal{L}(l_u, \X, \c, \y) = \sum_{k=1}^{|u|}\mathcal{L}(\hat{S},\x_{j_k}, \c_{j_k}, \y_{j_k}) \geq \sum_{k=1}^{|u|}\mathcal{L}(^{\ast} S,\x_{j_k}, \c_{j_k}, \y_{j_k}) = \mathcal{E}_u.\end{equation}
Therefore the loss of $l_u$ is at least $\mathcal{E}_u$.

Let $l_u'$ be a leaf node that contains both set $u$ and $N'$ other sample points ($1 \leq N' \leq N - |u|$), and denote $\hat{S}'$ as its KM estimator. 
We need to prove that for leaf node $l_u'$, its loss is at least $\mathcal{E}_u$. The loss of $l_u$ is given by:


\[\mathcal{L}(l_u', \X, \c, \y) = \sum_{k=1}^{|u|}\mathcal{L}(\hat{S}',\x_{j_k}, \c_{j_k}, \y_{j_k}) + \sum_{k=1}^{N'} \mathcal{L}(\hat{S}',\x_{i_k}, \c_{i_k}, \y_{i_k})\]
Let $^{\ast} S'$ be the optimal step function that minimizes the IBS loss for leaf node $l_u'$, such that:
\[^{\ast} S' = \argmin_{S} \sum_{k=1}^{|u|}\mathcal{L}(S,\x_{j_k}, \c_{j_k}, \y_{j_k}) + \sum_{k=1}^{N'} \mathcal{L}(S,\x_{i_k}, \c_{i_k}, \y_{i_k}).\]
Then, the loss of $l_u'$ obeys:
\begin{equation}\label{eq:loss_of_l_u'}
  \sum_{k=1}^{|u|}\mathcal{L}(\hat{S}',\x_{j_k}, \c_{j_k}, \y_{j_k}) + \sum_{k=1}^{N'} \mathcal{L}(\hat{S}',\x_{i_k}, \c_{i_k}, \y_{i_k}) \geq \sum_{k=1}^{|u|}\mathcal{L}(^{\ast} S',\x_{j_k}, \c_{j_k}, \y_{j_k}) + \sum_{k=1}^{N'} \mathcal{L}(^{\ast} S',\x_{i_k}, \c_{i_k}, \y_{i_k}).  
\end{equation}
$^{\ast}S $ and $^{\ast} S'$ obey:
\begin{equation}\label{eq:minimizer_property}
  \sum_{k=1}^{|u|}\mathcal{L}(^{\ast} S',\x_{j_k}, \c_{j_k}, \y_{j_k}) + \sum_{k=1}^{N'} \mathcal{L}(^{\ast} S',\x_{i_k}, \c_{i_k}, \y_{i_k})  \geq  \sum_{k=1}^{|u|}\mathcal{L}(^{\ast} S',\x_{j_k}, \c_{j_k}, \y_{j_k}) \geq \sum_{k=1}^{|u|}\mathcal{L}(^{\ast} S,\x_{j_k}, \c_{j_k}, \y_{j_k}).  
\end{equation}
Substituting Equation \ref{eq:minimizer_property} into Equations \ref{eq:loss_of_l_u'} and \ref{eqn:epsi}, we have:
\[\sum_{k=1}^{|u|}\mathcal{L}(\hat{S}',\x_{j_k}, \c_{j_k}, \y_{j_k}) + \sum_{k=1}^{N'} \mathcal{L}(\hat{S}',\x_{i_k}, \c_{i_k}, \y_{i_k}) \geq \mathcal{E}_u.\]
Therefore, for any leaf that captures the equivalent set $u$, its loss is greater than or equal to the equivalent loss of the set $\mathcal{E}_u$. 
\end{proof}

\textbf{Lemma \ref{lm:equiv_loss_2}}
\textit{
Let $l$ be a leaf node that captures $n$ equivalent sets $\{u_i\}_{i=1}^n$ with corresponding $\mathcal{E}_{u_i}$. The loss of $l$ obeys: $\mathcal{L}(l,\X, \c, \y) \geq \sum_{i=1}^n\mathcal{E}_{u_i}$. 
That is, the lower bound of a leaf is the sum of equivalent losses of the equivalent sets it captures.}
\begin{proof}
    \begin{eqnarray*}
    \mathcal{L}(l,\X, \c, \y) &\geq& \sum_{i=1}^n \sum_{k=1}^{|u_i|} \mathcal{L}(\hat{S},\x_{j_k}, \c_{j_k}, \y_{j_k}) \\
    &\geq& \sum_{i=1}^n \sum_{k=1}^{|u_i|} \mathcal{L}(^{\ast}S,\x_{j_k}, \c_{j_k}, \y_{j_k}), \textrm{where $^{\ast}S = \argmin_S \sum_{i=1}^n \sum_{k=1}^{|u_i|} \mathcal{L}(S,\x_{j_k}, \c_{j_k}, \y_{j_k})$,}\\
    &\geq& \sum_{i=1}^n \sum_{k=1}^{|u_i|} \mathcal{L}(^{\ast}S_i,\x_{j_k}, \c_{j_k}, \y_{j_k}), \textrm{where $^{\ast}S_i = \argmin_S \sum_{k=1}^{|u_i|} \mathcal{L}(S,\x_{j_k}, \c_{j_k}, \y_{j_k})$}\\
    &\geq& \sum_{i=1}^n \mathcal{E}_{u_i}.
    \end{eqnarray*}
\end{proof}
\subsection{Proof of Theorem \ref{thm:equiv_lb}}
\textbf{Theorem \ref{thm:equiv_lb}}
\textit{(Equivalent Points Lower Bound). Let $t = (t_{\fix}, \delta_{\fix}, t_{\splita}, \delta_{\splita}, K, H_t)$ be a tree with $K$ fixed leaves and $H_t - K$ splitting leaves. For any child tree $t'= (t'_{\fix}, \delta'_{\fix}, t'_{\splita}, \delta'_{\splita}, K', H_{t'}) \in \sigma(t)$:
\begin{equation}
    R(t', \X, \c, \y) \geq  \mathcal{L}(t_{\fix},\X,\y) + \lambda H_{t} + \sum_{u=1}^U  \mathcal{E}_u \cdot \mathbf{1}_{\capt(t_{\splita}, u)},
\end{equation}
where $\mathbf{1}_{\capt(t_{\splita}, u)}$ is 1 when $t_{\splita}$ captures set $u$, 0 otherwise.
\noindent\textit{Combining with the idea of Theorem \ref{thm:look_ahead}, we have}:
\begin{equation}
    R(t', \X, \y) \geq  \mathcal{L}(t_{\fix},\X,\c,\y) + \lambda H_{t} + \lambda + \sum_{u=1}^U  \mathcal{E}_u \cdot \mathbf{1}_{\capt(t_{\splita}, u)}.
\end{equation}}

\begin{proof}
\begin{eqnarray}\nonumber
   \lefteqn{R(t', \X, \c,\y) = \mathcal{L}(t',\X,\c,\y) + \lambda H_{t'}}\\\nonumber
   &=& \mathcal{L}(t'_{\fix},\X,\c,\y) +  \mathcal{L}(t'_{\splita},\X,\c,\y) + \lambda H_{t'}\\
   \label{eq:child_tree_obj}
       &=& \mathcal{L}(t_{\fix},\X,\c,\y) + \mathcal{L}(t'_{\fix} \setminus t_{\fix},\X,\c,\y) + \mathcal{L}(t'_{\splita},\X,\c,\y) + \lambda H_{t'}.
   \end{eqnarray}
Since samples captured by $t_{\splita}$ are captured either by $t'_{\fix} \setminus t_{\fix}$ or $t'_{\splita}$, and equivalence loss cannot be eliminated in any tree, from Lemma \ref{lm:equiv_loss_2}, we have:
\begin{equation}\label{eq:split_leaf_equiv_lb}
\mathcal{L}(t'_{\fix} \setminus t_{\fix},\X,\c,\y)+\mathcal{L}(t'_{\splita},\X,\c,\y) \geq \sum_{u=1}^U  \mathcal{E}_u \cdot \mathbf{1}_{\capt(t_{\splita}, u)}.
\end{equation}
Substituting Equation \ref{eq:split_leaf_equiv_lb} into Equation \ref{eq:child_tree_obj}, we have:
\[R(t', \X, \c,\y) \geq \mathcal{L}(t_{\fix},\X,\c,\y) + \sum_{u=1}^U  \mathcal{E}_u \cdot \mathbf{1}_{\capt(t_{\splita}, u)} + \lambda H_{t'}.\]
Because $H_{t'} \geq H_t + 1$:
\[R(t', \X, \c,\y) \geq  \mathcal{L}(t_{\fix},\X,\c,\y) + \lambda H_{t} + \lambda + \sum_{u=1}^U  \mathcal{E}_u \cdot \mathbf{1}_{\capt(t_{\splita}, u)}.\]
\end{proof}
\subsection{Proof of Theorem \ref{thm:guess_obj_opt}}
\textbf{Theorem \ref{thm:guess_obj_opt}}
\textit{
(Guarantee on guessed survival tree performance). Given dataset $\{\X, \c, \y\}$, depth constraint $d$, leaf penalty $\lambda$ and  reference model $T$, let $t_\text{guess}$ be the tree returned using $lb_\text{guess}$ defined in Equation \eqref{eq:lb_guess} and $R(t_\text{guess}, \X, \c, \y)$ be its objective. Let $t^\ast$ be the true optimal tree. We have:
\begin{equation}
    R(t_\text{guess}, \X, \c, \y) \leq \sum_{i=1}^N \max \left\{\mathcal{L}(\hat{S}^{T}_{\x_i}(\cdot),\x_i, c_i, y_i),  \mathcal{L}(\hat{S}^{t^\ast}_{\x_i}(\cdot),\x_i, c_i, y_i) \right \} + \lambda H_{t^\ast}
\end{equation}
where $\hat{S}^{T}_{\x_i}(\cdot)$ is the predicted survival function of reference model $T$ on sample $i$, and $\hat{S}^{t^\ast}_{\x_i}(\cdot)$ is the predicted survival function of optimal tree $t^\ast$.
That is, the objective of the guessed tree is no worse than the union of errors made by the reference model and the optimal tree.}
\begin{proof}
Define $R_\text{guess}(s_a, d, \lambda)$ as the objective of the solution using the guessed lower bound for subproblem $s_a$. Then, $R(t_\text{guess}, \X, \c, \y)$ can be rewritten as $R_\text{guess}(s_a, d, \lambda)$ where $s_a = \{1, 2, 3,\cdots, N\}$. We want to prove that for any $s_a$, 
\begin{equation}
    R_\text{guess}(s_a, d, \lambda) \leq \sum_{i \in s_a} \max \left\{\mathcal{L}(\hat{S}^{T}_{\x_i}(\cdot),\x_i, c_i, y_i),  \mathcal{L}(\hat{S}^{t_a^\ast}_{\x_i}(\cdot),\x_i, c_i, y_i) \right \} + \lambda H_{t^\ast(s_a, d, \lambda)}
\end{equation}
where $t_a^\ast$ is the optimal solution to subproblem $s_a$. Since the lower bound of $s_a$, which we denote by $lb_a$, is non-decreasing during the optimization process, we denote the highest value $lb_a$ gets updated to as $lb_\text{max}(s_a, d, \lambda)$. The subproblem is solved when the upper bound of $s_a$, $ub_a \leq lb_a$, which means 
\begin{equation}
    R_\text{guess}(s_a, d, \lambda) \leq lb_\text{max}(s_a, d, \lambda).
\end{equation}

Now we will prove the following inequality holds using induction:
\begin{equation}\label{eq:lb_max}
    lb_\text{max}(s_a, d, \lambda) \leq \sum_{i \in s_a} \max \left\{\mathcal{L}(\hat{S}^{T}_{\x_i}(\cdot),\x_i, c_i, y_i),  \mathcal{L}(\hat{S}^{t_a^\ast}_{\x_i}(\cdot),\x_i, c_i, y_i) \right \} + \lambda H_{t_a^\ast}.
\end{equation}

Either the initial lower bound $lb_\text{guess}(s_a)$ is greater than the initial upper bound $ub$, or $lb_\text{guess}(s_a) \leq ub$. If $lb_\text{guess}(s_a) > ub$, the subproblem $s_a$ is solved without further exploration, and if $lb_\text{guess}(s_a) \leq ub$, then the lower bound $lb_a$ will get updated until it reaches $ub$. Since $t^\ast_a$ is a leaf node, the upper bound $ub_a$ is fixed at $ub$. Also, $lb_\text{max}(s_a, d, \lambda)$ will never be higher than the initial upper bound $ub$.

\textbf{Base case:} The optimal solution $t^\ast_a$ is a leaf node.
Combining these two possibilities, we have 
\begin{equation}\label{eq:lb_max_ub}
    lb_\text{max}(s_a, d, \lambda) \leq  \max\{lb_\text{guess}(s_a), ub\}.
\end{equation}
Substituting the definition of $lb_\text{guess}(s_a)$ and $ub$, we have:
\begin{equation*}
    \leq  \max \left \{\sum_{i \in s_a} \mathcal{L}(\hat{S}^{T}_{\x_i}(\cdot),\X(s_a), \c(s_a), \y(s_a))+ \lambda, \sum_{i \in s_a} \mathcal{L}(\hat{S}^{t^\ast}(\cdot),\X(s_a), \c(s_a), \y(s_a)) + \lambda \right \} 
\end{equation*}
and taking the sum out:
\begin{equation}
    \leq \sum_{i \in s_a} \max \left \{\mathcal{L}(\hat{S}^{T}_{\x_i}(\cdot),\X(s_a), \c(s_a), \y(s_a)), \mathcal{L}(\hat{S}^{t^\ast},\X(s_a), \c(s_a), \y(s_a)) \right \} + \lambda.  
\end{equation}
Because $t^\ast(s_a, d, \lambda)$ is a leaf node, $H_{t^\ast} = 1$. Equation \ref{eq:lb_max} holds.

\textbf{Inductive case:} The optimal solution $t_a^\ast$ is a tree. In this case $d > 0$. Assume Equation \ref{eq:lb_max} holds for any left and right children pair of $t^\ast(s_a, d, \lambda)$, $s_{j_l}, s_{j_r}$, created by splitting feature $j$, such that:
\begin{equation}\label{eq:hy_assumption1}
    lb_\text{max}(s_{j_l}, d - 1, \lambda) \leq \sum_{i \in s_{j_l}} \max \left\{\mathcal{L}(\hat{S}^{T}_{\x_i}(\cdot),\x_i, c_i, y_i),  \mathcal{L}(\hat{S}^{t^\ast_{j_l}}_{\x_i}(\cdot),\x_i, c_i, y_i) \right \} + \lambda H_{t^\ast_{j_l}}
\end{equation}
and
\begin{equation}\label{eq:hy_assumption2}
    lb_\text{max}(s_{j_r}, d - 1, \lambda) \leq \sum_{i \in s_{j_r}} \max \left\{\mathcal{L}(\hat{S}^{T}_{\x_i}(\cdot),\x_i, c_i, y_i),  \mathcal{L}(\hat{S}^{t^\ast_{j_r}}_{\x_i}(\cdot),\x_i, c_i, y_i) \right \} + \lambda H_{t^\ast_{j_r}}
\end{equation}
where $s_{j_l} \cap s_{j_r} = \emptyset, s_{j_l} \cup s_{j_r} = s_a$, $t^\ast_{j_l}, t^\ast_{j_r}$ are the optimal solutions to $s_{j_l}$ and $s_{j_r}$ respectively. 

If initial upper bound $ub \leq lb_\text{guess}(s_a) + \lambda$, then we consider $s_a$ solved without further exploration and its solution is a leaf node in $t_\text{guess}$. 

From Equation \ref{eq:lb_max_ub} we know:
\begin{eqnarray}\nonumber
    lb_\text{max}(s_a, d, \lambda) &\leq&  \max\{lb_\text{guess}(s_a), ub\}\\\nonumber
    &\leq&  \max\{lb_\text{guess}(s_a), lb_\text{guess}(s_a) + \lambda\}\\\nonumber
    &\leq & lb_\text{guess}(s_a) + \lambda\\\nonumber
    &\leq & \sum_{i \in s_a} \mathcal{L}(\hat{S}^{T}_{\x_i}(\cdot),\X(s_a), \c(s_a), \y(s_a))+ 2\lambda\\\nonumber
\end{eqnarray}
\begin{equation}
    \leq  \sum_{i \in s_a} \max \left\{\mathcal{L}(\hat{S}^{T}_{\x_i}(\cdot),\X(s_a), \c(s_a), \y(s_a)), \hat{S}^{t_a^\ast}_{\x_i}(\cdot),\X(s_a), \c(s_a), \y(s_a)) \right\}+ 2\lambda.
\end{equation}
Since $t_a^\ast$ is a tree, $H_{t_a^\ast} \geq 2$, therefore:
\begin{equation}
     lb_\text{max}(s_a, d, \lambda) \leq  \sum_{i \in s_a} \max \left\{\mathcal{L}(\hat{S}^{T}_{\x_i}(\cdot),\X(s_a), \c(s_a), \y(s_a)), \hat{S}^{t_a^\ast}_{\x_i}(\cdot),\X(s_a), \c(s_a), \y(s_a)) \right\}+ \lambda H_{t_a^\ast}.
\end{equation}

If initial upper bound $ub > lb_\text{guess}(s_a) + \lambda$, we explore this subproblem $s_a$ by splitting features and creating pairs of children subproblems $s_{j_l}, s_{j_r}$ split on feature $j$. The lower bound of $s_a$ will get updated until converged with the upper bound, and $lb_a$ is updated as: $lb_a \leftarrow \max (lb_a, \min(ub_a, lb_\text{split}))$, where $lb_\text{split} \leftarrow \min_{j \in \text{features}} lb_{j_l} + lb_{j_r}$. From the way of updating $lb_a$, we know for any feature $j$:
\begin{equation}
    lb_\text{max}(s_a, d, \lambda) \leq lb_\text{max}(s_{j_l}, d - 1, \lambda) + lb_\text{max}(s_{j_r}, d - 1, \lambda).
\end{equation}
Using the hypothesis assumption in Equation \ref{eq:hy_assumption1} and \ref{eq:hy_assumption2}:
\begin{eqnarray}\label{eq:intermediate_induction}
    &\leq &\sum_{i \in s_{j_l}} \max \left\{\mathcal{L}(\hat{S}^{T}_{\x_i}(\cdot),\x_i, c_i, y_i),  \mathcal{L}(\hat{S}^{t^\ast_{j_l}}_{\x_i}(\cdot),\x_i, c_i, y_i) \right \} + \lambda H_{t^\ast_{j_l}}\\\nonumber
    &+& \sum_{i \in s_{j_r}} \max \left\{\mathcal{L}(\hat{S}^{T}_{\x_i}(\cdot),\x_i, c_i, y_i),  \mathcal{L}(\hat{S}^{t^\ast_{j_r}}_{\x_i}(\cdot),\x_i, c_i, y_i) \right \} + \lambda H_{t^\ast_{j_r}}.
\end{eqnarray}
Equation \ref{eq:intermediate_induction} holds for $t^\ast_{j_l}, t^\ast_{j_r}$ that are subtrees of $t^\ast_a$,
\begin{equation}
    \leq \sum_{i \in s_a} \max \left\{\mathcal{L}(\hat{S}^{T}_{\x_i}(\cdot),\x_i, c_i, y_i),  \mathcal{L}(\hat{S}^{t^\ast}_{\x_i}(\cdot),\x_i, c_i, y_i) \right \} + \lambda H_{t^\ast}.
\end{equation}
Here we proved that Inequality \ref{eq:lb_max} holds for any subproblem $s_a$, including the subproblem that is the whole dataset, $s_a = \{1, 2, \cdots, N\}$.
\end{proof}
\subsection{Proof of Corollary \ref{corollary}}
\textit{
Let $T, t_\text{guess}, t^\ast$ be defined as in Theorem \ref{thm:guess_obj_opt}. If the reference model $T$ performs no worse than $t^\ast$ on each sample, the tree returned using $lb_\text{guess}$ is still optimal.}
In other words, if the reference model has good performance, the tree returned after using guessing is still optimal or very close to optimal.
\begin{proof}
Subtracting $R(t^\ast, \X, \c, \y) = \sum_{i=1}^N \mathcal{L}(\hat{S}^{t^\ast}_{\x_i}(\cdot),\x_i, c_i, y_i) + \lambda H_{t^\ast}$ from the two sides of Inequality \ref{eq:guarantee}, we have:
\[R(t_\text{guess}, \X, \c, \y) - R(t^\ast, \X, \c, \y) \leq \sum_{i=1}^N \max \left\{\mathcal{L}(\hat{S}^{T}_{\x_i}(\cdot),\x_i, c_i, y_i) - \hat{S}^{t^\ast}_{\x_i}(\cdot),\x_i, c_i, y_i), 0\right \}.\]
If $\mathcal{L}(\hat{S}^{T}_{\x_i}(\cdot),\x_i, c_i, y_i) \leq \hat{S}^{t^\ast}_{\x_i}(\cdot),\x_i, c_i, y_i)$ then $R(t_\text{guess}, \X, \c, \y) - R(t^\ast, \X, \c, \y) \leq 0$, which means the tree returned is still optimal. 
\end{proof}
\subsection{Splitting Bounds}\label{sec:split_bound}
When constructing a new child tree $t' =(t'_{\textrm{fix}}, \delta'_{\textrm{fix}}, t'_{\textrm{split}}, \delta'_{\textrm{split}}, K', H_{t'})$, $t'_{\textrm{split}}$ needs to be determined. Splitting bounds help determine which leaves in $t'$ \textit{cannot be further split} and which leaves \textit{must be further split}.

\begin{theorem}\label{thm:incre_pro_split}
(Incremental Progress Bound to Determine Splitting). For any optimal tree $t^{\ast}$, any parent node of its leaves must have loss at least $\geq \lambda$ when considered as a leaf.
\end{theorem}
\begin{proof}
This bound adapts directly from OSRT \citep{zhang2023optimal}, where the proof can be found.
\end{proof}

\begin{theorem}\label{thm:lb_incre_pro}
(Lower Bound on Incremental Progress). Consider any optimal tree $t^{\ast} = \{l_1, l_2, \dots, l_i, l_{i+1}, \dots, l_{H_{t^{\ast}}} \}$ with $H_{t^{\ast}}$ leaves. Let $t' = \{l_1, l_2, \dots, l_{i-1}, l_{i+2}, \dots, l_{H_{t^{\ast}}, l_j}\}$ be a tree created by deleting a pair of leaves $l_i$ and  $l_{i+1}$ in $t^{\ast}$ and adding their parent node $l_j$. The reduction in loss obeys:
\[\mathcal{L}(l_j,\X, \c,\y) -  \mathcal{L}(l_i,\X,\c, \y) - \mathcal{L}(l_{i+1},\X, \c,\y) \geq \lambda.\]
\end{theorem}
\begin{proof}
This bound adapts directly from OSRT \citep{zhang2023optimal}, where the proof can be found.
\end{proof}

When constructing new trees, if a leaf has loss less than $\lambda$ (it fails to meet Theorem \ref{thm:incre_pro_split}), then it cannot be further split. If a pair of leaves in that tree reduce loss from their parent node by less than $\lambda$ (the leaves fail to meet Theorem \ref{thm:lb_incre_pro}), then at least one of this pair of leaves must be further split to search for optimal trees.

\subsection{Leaf Bounds}
The following bounds on the number of leaves allow us to prune trees whose number of leaves exceed these upper bounds.
\begin{theorem}
(Upper Bound on the Number of Leaves). Let $H_t$ be the number of leaves of tree $t$ and let $R^c$ be the current best objective. For any optimal tree $t^{\ast}$ with $H_{t^{\ast}}$ leaves, it is true that:
\begin{equation}\label{eq:leaf_ub}
    H_{t^{\ast}} \leq \min \{ \lfloor R_c/\lambda \rfloor , 2^M\},
\end{equation}
 where $M$ is the number of features.
\end{theorem}
\begin{proof}
This bound adapts directly from OSDT \citep{HuRuSe2019}, where the proof can be found.
\end{proof}
\begin{theorem}\label{thm:parent_num_leaves_ub}
(Parent-specific upper bound on the number of leaves). Let $t = (t_{\textrm{fix}}, \delta_{\textrm{fix}}, t_{\textrm{split}}, \delta_{\textrm{split}}, K, H_t)$ be a tree with child tree $t' = (t'_{\textrm{fix}}, \delta'_{\textrm{fix}}, t'_{\textrm{split}}, \delta'_{\textrm{split}}, K', H_{t'}) \in \sigma(t)$ with $H_{t'}$ leaves. Then:
\begin{equation}\nonumber
     H_{t'} \leq \min \left\{ H_t + \left\lfloor \frac{R_c- \mathcal{L}(t_{\fix},\X, \c, \y) - \lambda H_t}{\lambda} \right \rfloor, 2^M\right\}.
\end{equation}
\end{theorem}
\begin{proof}
This bound adapts directly from OSDT \citep{HuRuSe2019}, where the proof can be found.
\end{proof}

\subsection{Permutation Bound}\label{sec:permutation}
\begin{theorem}\label{thm:permutation}
(Leaf Permutation Bound). Let $\pi$ be any permutation of $\{1 \dots H_t\}$. Let $t = \{l_1, l_2, \dots,l_{H_{t}}\}$, $T =\{l_{\pi(1)}, l_{\pi(2)}, \dots,l_{\pi(H_t)}\}$, that is, the leaves in $T$ are a permutation of the leaves in $t$. The objective lower bounds of $t$ and $T$ are the same and their child trees correspond to permutations of each other.
\end{theorem}
\begin{proof}
This bound adapts directly from OSDT \citep{HuRuSe2019}, where the proof can be found.
\end{proof}
This bound avoids duplicate computation of trees with leaf permutation.

\subsection{Hierarchical Objective Lower Bound for Sub-trees}
\begin{theorem}\label{thm:subtree_lb}
(Hierarchical Objective Lower Bound for Sub-trees). Let $R^c$ be the current best objective so far. Let $t$ be a tree such that the root node is split by a feature, where two sub-trees $t_{\lf}, t_{\ri}$ are generated with $H_{\lf}$ leaves for $t_{\lf}$ and $H_{\ri}$ leaves for $t_{\ri}$. The data captured by the left tree is $(\X_{\lf}, \y_{\lf})$ and the data captured by the right tree is $(\X_{\ri}, \y_{\ri})$. Then, the objective
lower bounds of the left sub-tree and right sub-tree are $b(t_{\lf}, \X_{\lf}, \c_{\lf}, \y_{\lf}) $ and $b(t_{\ri}, \X_{\ri}, \c_{\ri},\y_{\ri})$, which obey $R(t_{\lf}, \X_{\lf}, \c_{\lf},\y_{\lf})\geq b(t_{\lf}, \X_{\lf}, \c_{\lf},\y_{\lf})$, and $R(t_{\ri}, \X_{\ri},\c_{\ri}, \y_{\ri}) \geq b(t_{\ri}, \X_{\ri}, \c_{\ri},\y_{\ri})$. If $b(t_{\lf}, \X_{\lf}, \c_{\lf},\y_{\lf}) > R^c$ or $b(t_{\ri}, \X_{\ri}, \c_{\ri},\y_{\ri}) > R^c$ or $b(t_{\lf}, \X_{\lf},\c_{\lf}, \y_{\lf}) + b(t_{\ri}, \X_{\ri}, \c_{\ri},\y_{\ri}) > R^c$, then $t$ is not an optimal tree, and none of  its child trees are optimal.
\end{theorem}
\begin{proof}
This bound adapts directly from GOSDT \citep{lin2020generalized}, where the proof can be found.
\end{proof}
This bound can be applied to any tree, even if the tree is partially constructed. In a partially constructed tree $t$, if one of its subtrees has objective worse than current best objective $R^c$, we can prune tree $t$ and all of its child trees without constructing the other subtree.

\subsection{Subset Bound}\label{sec:subset}
\begin{theorem}\label{thm:subset}
Let $t$ and $T$ to be two trees
with the same root node, where $t$ uses feature $f_1$ to split the root node and $T$ uses feature $f_2$ to split the root node. Let $t_1, t_2$ be subtrees of $t$ under its root node, and $(\X_{t_1}, \c_{t_1},\y_{t_1}), (\X_{t_2}, \c_{t_2}, \y_{t_2})$ be samples captured by $t_1$ and $t_2$. Similarly, let $T_1, T_2$ be subtrees of $T$ under its root node, and $(\X_{T_1}, \c_{T_1},\y_{T_1}), (\X_{T_2}, \c_{T_2},\y_{T_2})$ be samples captured by $T_1$ and $T_2$. Suppose $t_1, t_2 $ are optimal trees for $(\X_{t_1}, \c_{t_1},\y_{t_1}), (\X_{t_2}, \c_{t_2},\y_{t_2})$ respectively, and $T_1, T_2$ are optimal trees for $(\X_{T_1}, \c_{T_1},\y_{T_1}), (\X_{T_2}, \c_{T_2},\y_{T_2})$ respectively. If $R(t_1,\X_{t_1}, \c_{t_1},\y_{t_1}) \leq R(T_1,\X_{T_1}, \c_{T_1},\y_{T_1}) $ and $(\X_{t_2}, \c_{t_2},\y_{t_2}) \subset (\X_{T_2}, \c_{T_2},\y_{T_2})$, then $R(t, \X, \c, \y ) \leq R(T, \X, \c, \y )$.
\end{theorem}
\begin{proof}
This bound adapts directly from GOSDT \citep{lin2020generalized}, where the proof can be found.
\end{proof}
Similar to Theorem  \ref{thm:subtree_lb}, this bound ensures that we can safely prune a partially constructed tree without harming optimality. It checks whether subtree $t_1$ has a better objective than $T_1$, despite handling more data.
\section{Metrics}\label{sec:metrics}
We use several metrics to evaluate the quality of our survival trees. Note that some of the following metrics are not additive, which means that if we want to optimize them then we must construct the entire survival tree. Therefore we use these metrics only in the evaluation stage. We found that even though we optimize the IBS, the optimal survival trees often perform better on other metrics as well, which we show in Section \ref{sec:quality} and Appendix \ref{exp:quality}.
\subsection{Integrated Brier Score Ratio}
To be consistent with other metrics (the higher the better), we used Integrated Brier Score Ratio (IBS Ratio) to report IBS loss in all experimental sections. 
The IBS Ratio of tree $t$ is defined as
\begin{equation}
    \text{IBS Ratio}(t) = 1 - \frac{\mathcal{L}(t,\X, \c, \y)}{\mathcal{L}(t_0,\X, \c, \y)}
\end{equation}
where $\mathcal{L}$ is the IBS loss defined in Equation \ref{eq:tree_loss} and $t_0$ is a single node containing all samples (equivalently a KM estimator ignoring all features). 
This ratio is very similar to the $R^2$ in regression.
\subsection{Concordance Indices}
\citet{harrell1996multivariable} adapted the Concordance Statistic from logistic regression into survival analysis. Given a pair of samples, $i$ and $j$, a good survival model should predict a lower survival probability for the sample that is closer to the actual death time. A \textit{concordant} pair is defined as a pair of samples that satisfy this expectation (e.g., $\hat{S}_{\x_i}(y_i) < \hat{S}_{\x_j}(y_i), y_i < y_j$). When the predicted survival probabilities are tied, we consider them have 0.5 expectation to be concordant due to random arrangement. A pair is \textit{comparable} either when both samples are observed (not censored) or when one sample died before the other was censored. The concordance index is defined as the number of concordant pairs divided by the number of comparable pairs. Formally, Harrell's C-index is given by:
\begin{equation}\label{eq:harrel_c}
    C_H = \frac{\sum_i \sum_j c_i \cdot \mathbf{1}_{y_i < y_j} \cdot \left(\mathbf{1}_{\hat{S}_{\x_i}(y_i) < \hat{S}_{\x_j}(y_i)} + 0.5 \cdot \mathbf{1}_{\hat{S}_{\x_i}(y_i) = \hat{S}_{\x_j}(y_i)}\right)}{\sum_i \sum_j c_i \cdot \mathbf{1}_{y_i < y_j}}.
\end{equation}
Since the predicted survival function is time-dependent, we use the earlier observation time ($\min\{y_i, y_j\}$) when evaluating whether a comparable pair is concordant. \\
However, Harrell's C-index is biased when there are a large number of censored samples. \citet{uno2011c} applied the inverse probability of censoring weights to Harrell's C-index to make it robust to a high degree of censoring. Uno's C-index is given by:
\begin{equation}\label{eq:uno_c}
C_U =  \frac{\sum_i \sum_j c_i \cdot \hat{G}^{-2}(y_i)\cdot \mathbf{1}_{y_i < y_j} \cdot \left( \mathbf{1}_{\hat{S}_{\x_i}(y_i) < \hat{S}_{\x_j}(y_i)} +  0.5 \cdot \mathbf{1}_{\hat{S}_{\x_i}(y_i) = \hat{S}_{\x_j}(y_i)}\right)}{\sum_i \sum_j c_i \cdot \hat{G}^{-2}(y_i) \cdot \mathbf{1}_{y_i < y_j} }.
\end{equation}
Remember, $\hat{G}(\cdot)$ is the Kaplan–Meier estimate of the censoring distribution $\c$ under the assumption that it is independent of the covariates. \citet{uno2011c} claimed that Uno's C-index is `quite robust even when the censoring is dependent on the covariates.'  
\subsection{Cumulative-Dynamic-AUC}
The receiver operating characteristic curve (ROC) plots the the true positive rate versus the false positive rate in binary classification tasks. The area under the ROC curve (AUC) can be extended to survival analysis: given a time $y$, the true positives are samples that died before or at time $y$, e.g., $c_i=1, y_i < y$ (cumulative cases), and true negatives are samples that are still alive at time $y$, e.g., $y_j > y$ (dynamic controls). 

As we change the threshold of predicted survival function, $\tau$,  to classify samples into cumulative cases and dynamic control cases, the true positive and false positive rates also change, which gives a time-dependent ROC curve. The time-dependent ROC curve measures how well a survival model separates cumulative cases from dynamic control cases at time $y$.

The specificity at time $t$ is given by the proportion of the samples with predicted survival function greater than or equal to $\tau$:
\begin{equation}\label{eq:specificity}
    \widehat{Sp}(\tau, y) = \frac{\sum_i \mathbf{1}_{y_i > y} \cdot \mathbf{1}_{\hat{S}(y|\x_i) \geq \tau}}{\sum_{i} \mathbf{1}_{y_i > y} }.
\end{equation}
\citet{uno2007evaluating} and \citet{hung2010estimation} proposed to perform IPCW when calculating the sensitivity:
\begin{equation}\label{eq:sensitivity}
    \widehat{Se}(\tau, y) = \frac{\sum_i c_i \cdot \hat{G}^{-1}(y_i)\cdot \mathbf{1}_{y_i \leq y} \cdot \mathbf{1}_{\hat{S}(y|\x_i) < \tau}}{\sum_{i} c_i \cdot \hat{G}^{-1}(y_i)\cdot \mathbf{1}_{y_i \leq y} }
\end{equation}
which gives the AUC at time $y$ as:
\begin{equation}\label{eq:auc}
    \widehat{AUC}(y) = \frac{\sum_i \sum_j c_i \cdot \hat{G}^{-1}(y_i) \cdot \mathbf{1}_{y_i \leq y} \cdot \mathbf{1}_{y_j > y} \cdot \mathbf{1}_{\hat{S}(y|\x_i) < \hat{S}(y|\x_j)}}{\left(\sum_i c_i \cdot \hat{G}^{-1}(y_i) \cdot \mathbf{1}_{y_i \leq y} \right) \left(\sum_j \mathbf{1}_{y_j > y}\right)}.
\end{equation}
\citet{lambert2016summary} proposed to restrict a time-dependent weighted AUC estimator to a fixed time interval, which summarizes the mean AUC over the time range. Here we compute the time range from $y_{\min} = \min\{y_i\}^N_{i=1}$ to $y_{\max} = \max\{y_i\}^N_{i=1}$:
\begin{equation}
    \overline{AUC} = \frac{1}{\hat{S}(y_\text{min}) - \hat{S}(y_\text{max})} \int_{y_\text{min}}^{y_\text{max}} \widehat{AUC}(y) d \hat{S}(y).
\end{equation}

\section{Experiment Setup Details}\label{sec:setup}
\subsection{Datasets}
We used 17 datasets of which 11 are real-world survival data and 6 are synthetic data (manually censoring on real-world regression datasets). 

\textbf{Real-world survival data:}
\begin{itemize}[leftmargin=*, topsep=0pt, noitemsep]
    \item \textbf{Aids}:AIDS Clinical Trial \citep{lemeshow2011applied}. The event is  AIDS-defining event.
    \item \textbf{Aids\_death}: AIDS Clinical Trial \citep{lemeshow2011applied}. The event is death.
    \item \textbf{Churn}: The task of predicting when a software as a service (SaaS) company's customers are likely to stop their monthly subscription. The event is end of subscription. This dataset is from pysurvival \citep{pysurvival_cite}. 
    \item \textbf{Credit}: Lenders need to predict when a borrower will repay a loan. The event is loan repayment. This dataset is from pysurvival \citep{pysurvival_cite}, adapted from UCI Machine Learning Repository \citep{Dua:2019}.
    \item \textbf{Employee}: The task of predicting when an employee will quit. The event is an employee's leaving. This dataset is from pysurvival \citep{pysurvival_cite}.
    \item \textbf{Maintenance}: The task of predicting when equipment failure will occur. The event is machine failure. This dataset is from pysurvival \citep{pysurvival_cite}.
    \item \textbf{Veterans}: Veterans’ Administration Lung Cancer Trial \citep{kalbfleisch2011statistical}. The event is death. 
    \item \textbf{Whas500}: Worcester Heart Attack Study \citep{lemeshow2011applied}. The event is death. 
    \item \textbf{Flchain}: Use of nonclonal serum immunoglobulin free light chains to predict overall survival in the general population \citep{dispenzieri2012use}. The event is death.
    \item \textbf{Gbsg2}: German Breast Cancer Study Group 2 \citep{schumacher1994randomized}. The event is recurrence free survival. 
    \item \textbf{Uissurv}: UMASS Aids Research Unit IMPACT Study \citep{lemeshow2011applied}. The event is returning to drugs.
\end{itemize}

\textbf{Synthetic data:}
We randomly censored samples with a censoring rate of 20\%. 
\begin{itemize}[leftmargin=*, topsep=0pt, noitemsep]
    \item \textbf{Airfoil}: Airfoil self-noise. This dataset is from UCI Machine Learning Repository \citep{Dua:2019}.
    \item \textbf{Insurance}: Medical cost personal. This dataset is from a Kaggle competition \citep{kaggle:insurance}.
    \item \textbf{Real-estate}: Real estate valuation. This dataset is from the UCI Machine Learning Repository \citep{Dua:2019}.
    \item \textbf{Sync}: Synchronous machine. This dataset is from the UCI Machine Learning Repository \citep{Dua:2019}.
    \item \textbf{Servo}: The rise time of a servomechanism. This dataset is from the UCI Machine Learning Repository \citep{Dua:2019}.
    \item \textbf{Household}: Individual household electric power consumption. This dataset is from the UCI Machine Learning Repository \citep{Dua:2019}. 
\end{itemize}

\subsection{Data Preprocessing}
First, we removed all observations with missing values. Because we are performing hundreds of experiments consuming months of compute time, for each dataset, we categorized some continuous features into 4 equal-width partitions and used one-hot encoding to convert all the features into binary features:\\\\
\textbf{Aids}: each of \textit{age, cd4, priorzdv} were discretized into 4 categories.\\
\textbf{Aids\_death}: each of \textit{age, cd4, priorzdv} were discretized into 4 categories.\\
\textbf{Churn}: each of \textit{articles\_viewed, minutes\_customer\_support, smartphone\_notifications\_viewed, social\_media\_ads\_viewed, and marketing\_emails\_clicked} were discretized into 4 categories.\\
\textbf{Credit}: each of \textit{amount, installment\_rate, present\_residence, age, number\_of\_credits, andpeople\_liable} were discretized into 4 categories.\\
\textbf{Employee}: feature \textit{average\_montly\_hours} was discretized into 4 categories. \\
\textbf{Maintenance}: each of \textit{pressureInd, moistureInd, and temperatureInd} were discretized into 4 categories.\\
\textbf{Veterans}: each of \textit{Age\_in\_years, Karnofsky\_score, Months\_from\_Diagnosis} were discretized into 4 categories.\\
\textbf{Whas500}: each of \textit{age, bmi, diasbp, hr, los, sysbp} were discretized into 4 categories.\\
\textbf{Flchain}: each of \textit{age, creatinine, kappa, lambda} were discretized into 4 categories.\\
\textbf{Gbsg2}: each of \textit{age, estrec, pnodes, progrec, tsize} were discretized into 4 categories.\\
\textbf{Uissurv}: each of \textit{age, beck, ndrugtx, los} were discretized into 4 categories.\\
\textbf{Airfoil}: We discretized each of the features \textit{frequency, angle of attack, suction side displacement thickness} into 4 categories.\\
\textbf{Insurance}: We discretized each of \textit{age, bmi} into 4 categories.\\
\textbf{Real-estate:} We discretized each continuous feature into 4 categories.\\
\textbf{Servo:} We directly use this dataset that only contains categorical features.\\
\textbf{Sync:} We discretized each feature into 4 categories.\\
\textbf{Household}: We transformed the \textit{Date} feature into \textit{Month}, \textit{Time} into \textit{Hour}. Then we discretized each of \textit{Month, Hour, Global\_reactive\_power, Voltage, Global\_intensity} into 4 categories.\\
\textbf{Table \ref{tab:data} summarizes all the datasets after preprocessing.}
\begin{table}[ht]
    \small
    \centering
    \begin{tabularx}{\linewidth}{|X|X|X|X|X|X|}\hline 
    Dataset & Samples & Orig. Features & Encoded Binary Features & Event (Censoring Rate) & Time to Event\\\hline
    churn & 2000 & 12 & 42 & churned ($53.4\%$) & months active\\\hline
    credit & 1000 & 19 & 56 & fully\_repaid ($30\%$)& duration \\ \hline
    employee & 15000 & 7 & 27 &left ($77.2\%$) & time spend company \\\hline
    maintenance & 1000 & 5 & 14 & broken ($60.3\%$)& lifetime \\\hline
    aids & 1151 & 11 & 25 & aids ($91.7\%$)& time \\\hline
    aids\_death & 1151 & 11 & 25 & death ($97.7\%$) & time \\\hline
    flchain & 7478 & 9 & 47 & death ($72.5\%$) & time \\\hline
    gbsg2 & 686 & 8 & 19 & recurrence free survival ($56.4\%$) & time \\\hline
    whas500 & 500 & 14 & 26 & death ($57\%$) & time \\\hline
    veterans & 137 & 6 & 14 & death ($6.6\%$) & time \\\hline
    uissurv & 628 & 12 & 26 & censoring ($19.1\%$) & time \\\hline
    airfoil & 1503 & 5 & 17 & censoring ($20\%$) & scaled sound pressure level \\\hline
    insurance & 1338 & 6 & 16 & censoring ($20\%$) & charges \\\hline
    real-estate & 414 & 6 & 18&  censoring ($20\%$)& house price of unit area \\\hline
    servo & 167 & 4 & 15 & censoring ($20\%$)&class\\\hline
    sync & 557& 4 & 12 & censoring ($20\%$) &``If'' \\\hline
    household & 2,049,280 & 5 & 15 & censoring ($20\%,
    50\%, 80\%$) & global active power \\\hline
    \end{tabularx}
    \setlength{\abovecaptionskip}{10pt}
    \caption{Datasets Summary.}
    \label{tab:data}
\end{table}

\subsection{Experiment Platform}
We ran all experiments on a 48-core TensorEX TS2-673917-DPN Intel Xeon Gold 6226 Processor, 2.7Ghz with 768GB RAM. We set a time limit of 60 minutes and a memory limit of 200GB. All algorithms ran single-threaded. 

\subsection{Software Packages}
\textbf{Recursive Partitioning and Regression Trees (RPART):}
CRAN package of Rpart, version 4.1.19 (https://cran.r-project.org/web/packages/rpart/index.html).\\ 
\textbf{Conditional Inference Trees (CTree):} 
CRAN package of partykit, version 1.2-16 (https://cran.r-project.org/web/packages/partykit/index.html).\\
\textbf{Scikit-survival(SkSurv):} scikit-survival version-0.20.0 (https://scikit-survival.readthedocs.io/en/stable/index.html).
\newpage 
\section{Experiments: Optimality}\label{exp:lvs}
\noindent\textbf{Collection and Setup:} We ran this experiment on 8 datasets: \textit{aids, aids\_death,  uissurv, gbsg2,  churn, maintenance, veterans, whas500}. We trained models on the entire dataset to measure time to convergence/optimality. For each dataset, we ran algorithms with different configurations:
\begin{itemize}[leftmargin=10pt]
    \item \textbf{CTree}: We ran this algorithm with 8 different configurations: depth limit, $d$, ranging from 2 to 9, and a corresponding maximum leaf limit $2^d$. All other parameters were set to the default.
    \item \textbf{SkSurv}: We ran this algorithm with 8 different configurations: depth limit, $d$, ranging from 2 to 9, and a corresponding maximum leaf limit $2^d$. The random state was set to 2023 and all other parameters were set to the default.
    \item \textbf{RPART}: We ran this algorithm with $8 \times 12$ different configurations: depth limits ranging from 2 to 9, and 12 different regularization coefficients (0.1, 0.05, 0.025, 0.01, 0.005, 0.0025, 0.001, 0.0005, 0.00025, 0.0001, 0.00005, 0.00001). All other parameters were set to the default.
    \item \textbf{OSST} (our method): We ran this algorithm with $8 \times 10$ different configurations: depth limits ranging from 2 to 9, and 12 different regularization coefficients (0.1, 0.05, 0.01, 0.005, 0.0025, 0.001, 0.0005, 0.00025, 0.0001, 0.00005, 0.000025, 0.00001). The minimum sample required in each leaf was set to 7, which is consistent with other three methods' default value. 
\end{itemize}

\noindent\textbf{Calculations:} For each combination of dataset and depth limit, we drew one plot (training loss against number of leaves). Under the same depth limit, runs of one algorithm with different regularization coefficients may result in trees with same number of leaves. In this case, we plotted the median loss of those trees and showed the best and worst loss among these trees as lower and upper error values respectively. These plots do not display trees where the number of leaves exceeds 30, as these trees are more likely to be overfitted and hard to interpret. A single root node is excluded from the plots as well, as it is not meaningful. 

\begin{figure*}[htbp]
    \centering
    \includegraphics[width=0.42\textwidth]{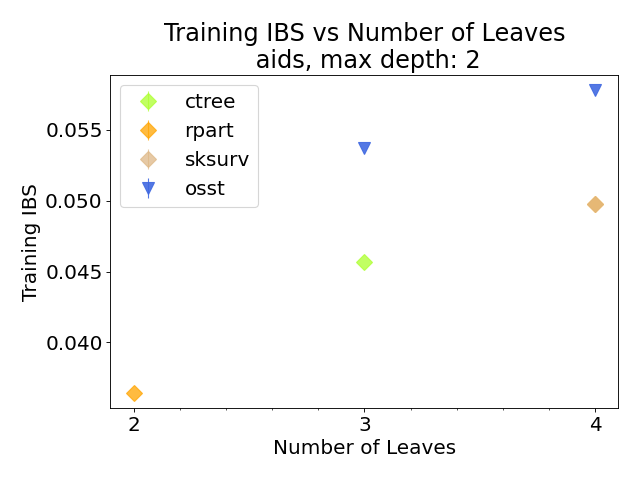}
    \includegraphics[width=0.42\textwidth]{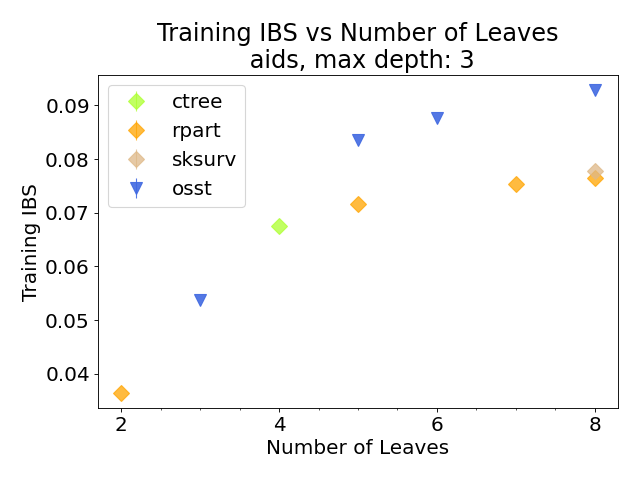}
    \includegraphics[width=0.42\textwidth]{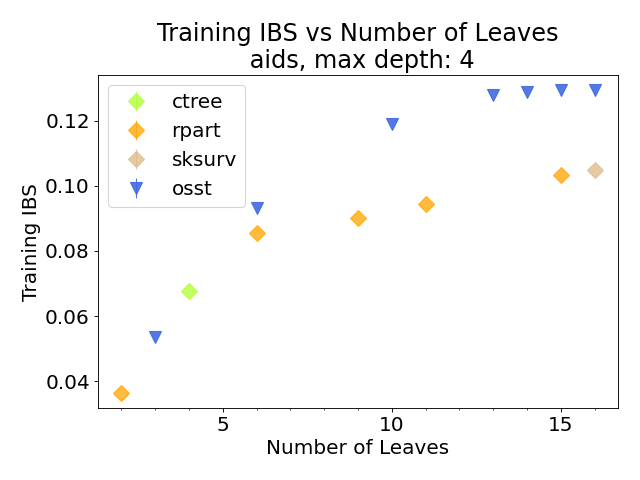}
    \includegraphics[width=0.42\textwidth]{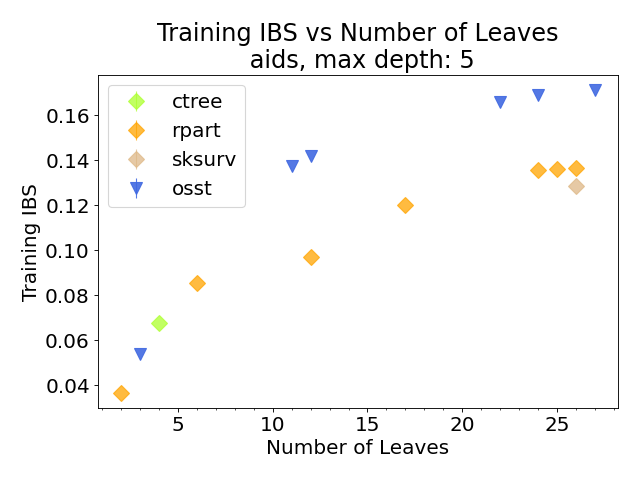}
    \includegraphics[width=0.42\textwidth]{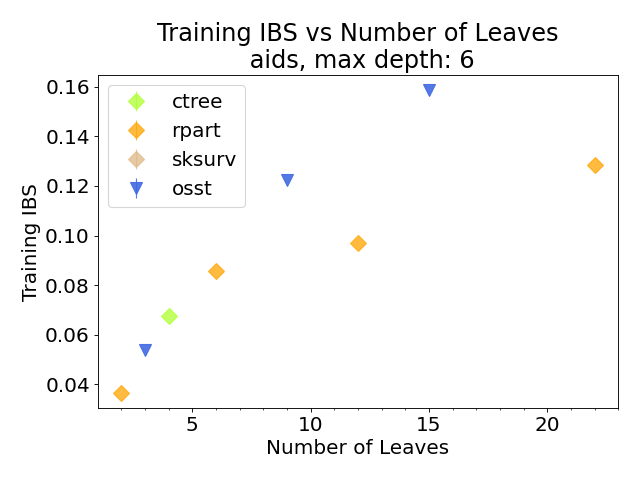}
    \includegraphics[width=0.42\textwidth]{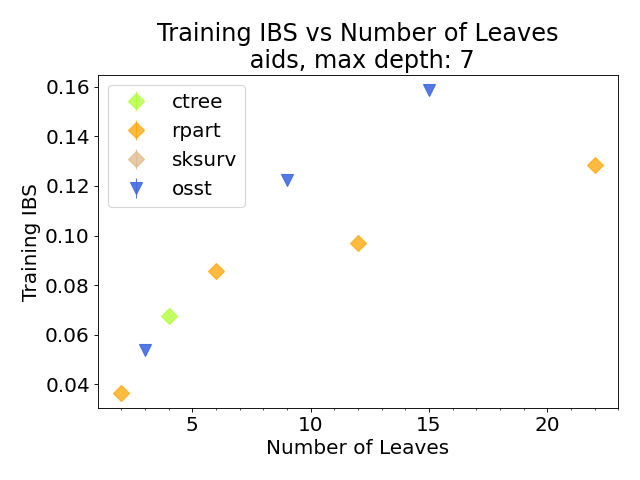}
    \includegraphics[width=0.42\textwidth]{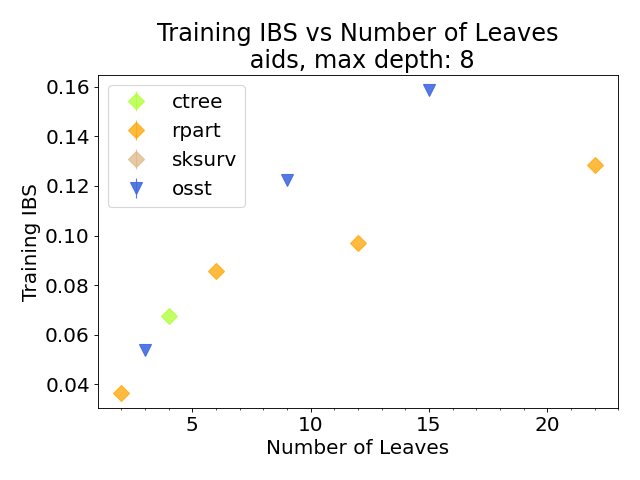}
    \includegraphics[width=0.42\textwidth]{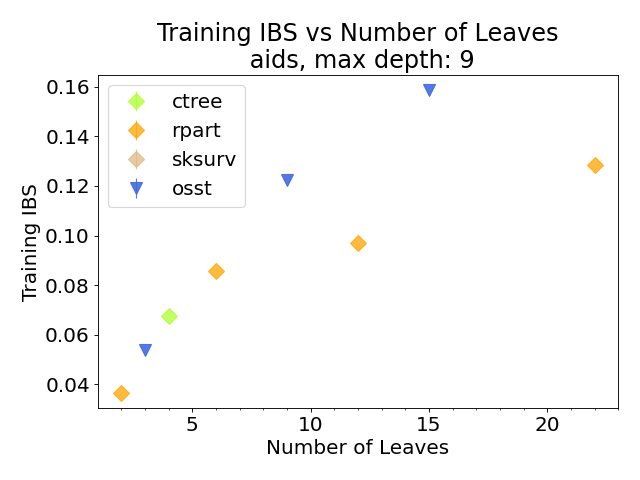}
    \caption{Training IBS achieved by CTree, RPART, SkSurv and OSST as a function of number of leaves on dataset: aids.}
    \label{fig:lvs:aids-ibs}
\end{figure*}

\begin{figure*}[htbp]
    \centering
    \includegraphics[width=0.42\textwidth]{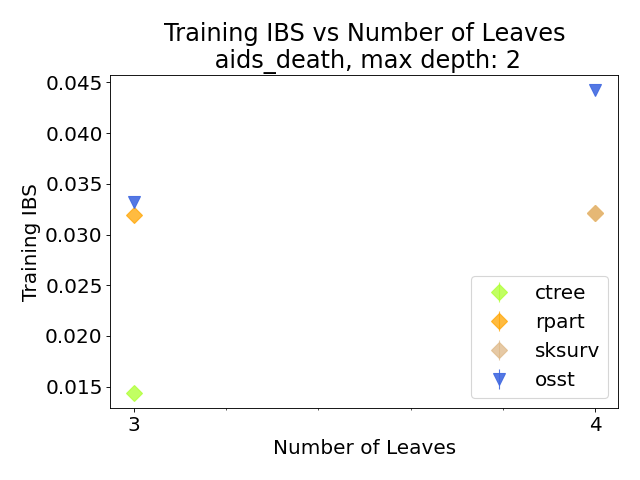}
    \includegraphics[width=0.42\textwidth]{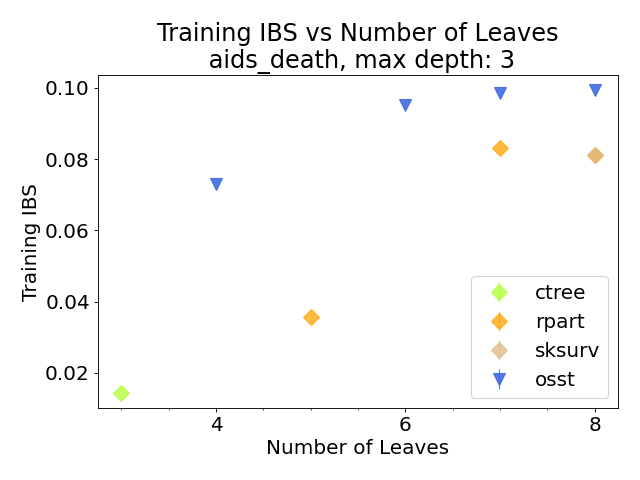}
    \includegraphics[width=0.42\textwidth]{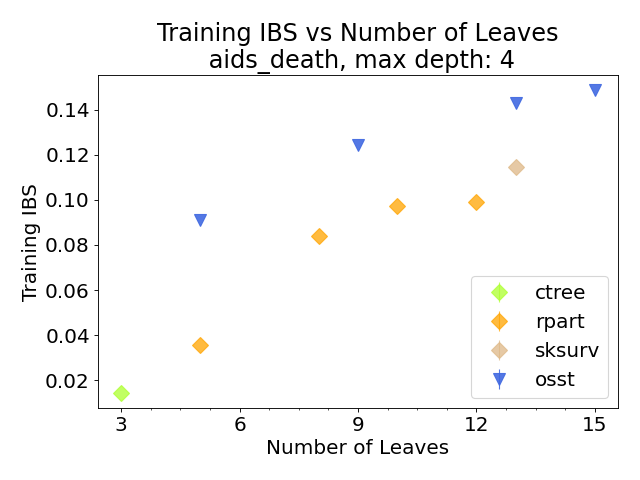}
    \includegraphics[width=0.42\textwidth]{figures/loss_vs_sparsity/aids_death/ibs/aids_death_depth_5.png}
    \includegraphics[width=0.42\textwidth]{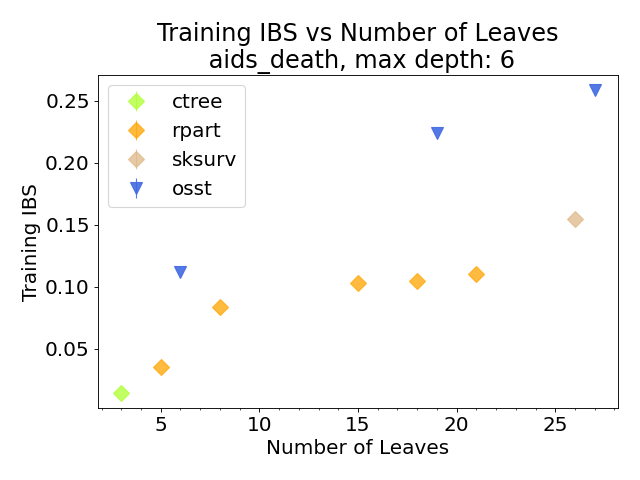}
    \includegraphics[width=0.42\textwidth]{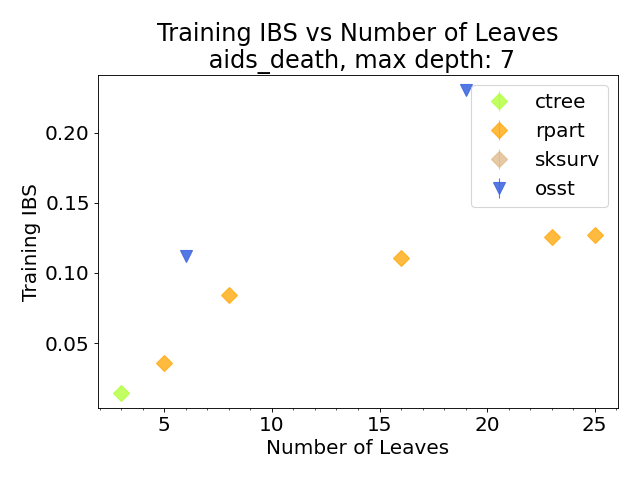}
    \includegraphics[width=0.42\textwidth]{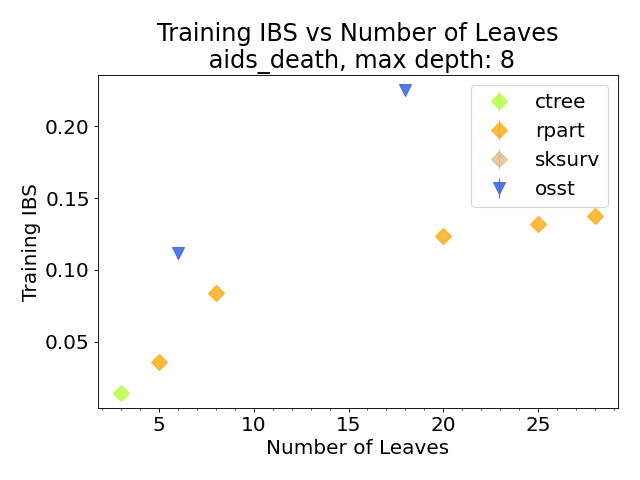}
    \includegraphics[width=0.42\textwidth]{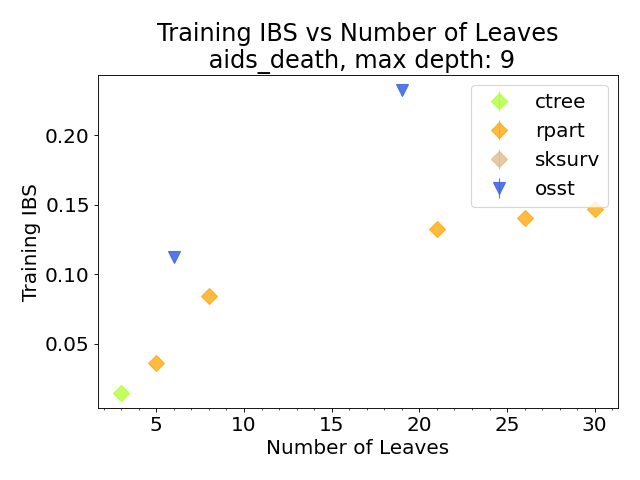}
    
    \caption{Training IBS achieved by CTree, RPART, SkSurv and OSST as a function of number of leaves on dataset: aids\_death.}
    \label{fig:lvs:aids_death-ibs}
\end{figure*}

\begin{figure*}[htbp]
    \centering
    \includegraphics[width=0.42\textwidth]{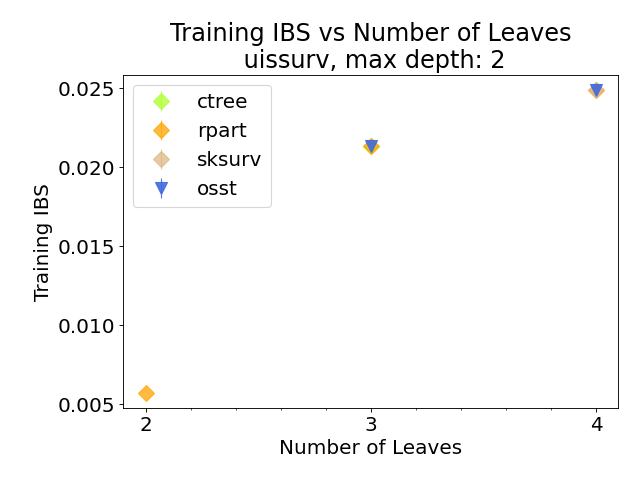}
    \includegraphics[width=0.42\textwidth]{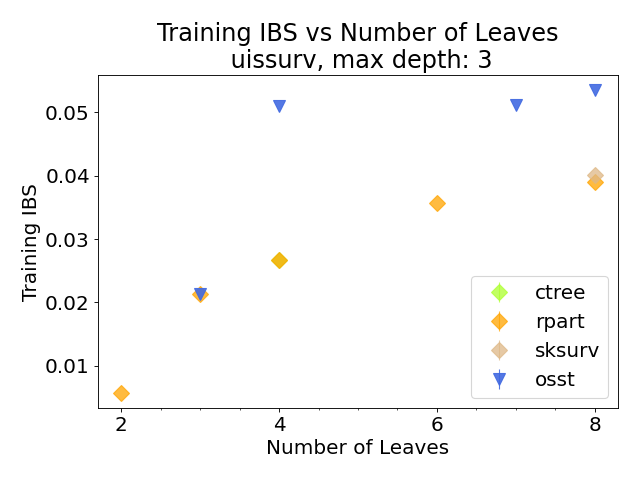}
    \includegraphics[width=0.42\textwidth]{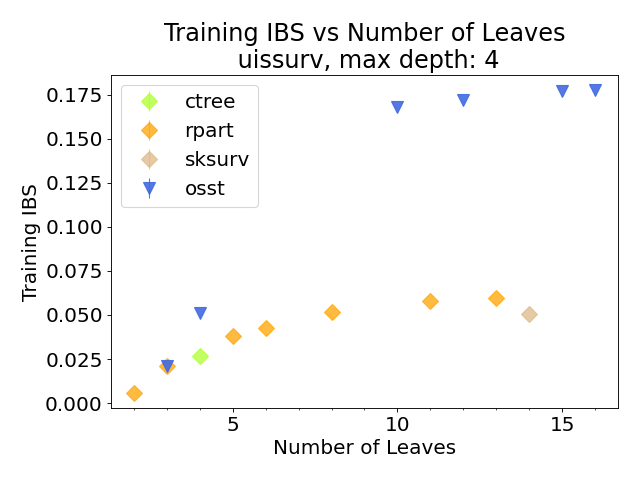}
    \includegraphics[width=0.42\textwidth]{figures/loss_vs_sparsity/uissurv/ibs/uissurv_depth_5.png}
    \includegraphics[width=0.42\textwidth]{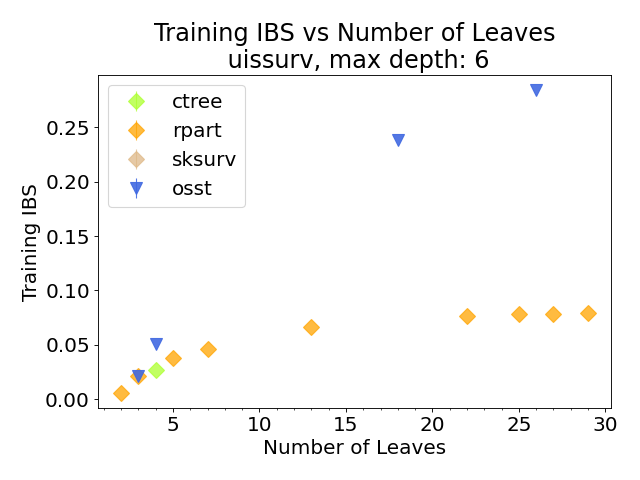}
    \includegraphics[width=0.42\textwidth]{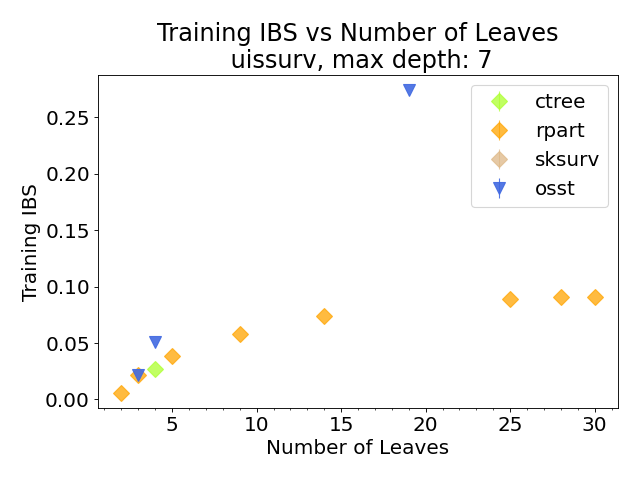}
    \includegraphics[width=0.42\textwidth]{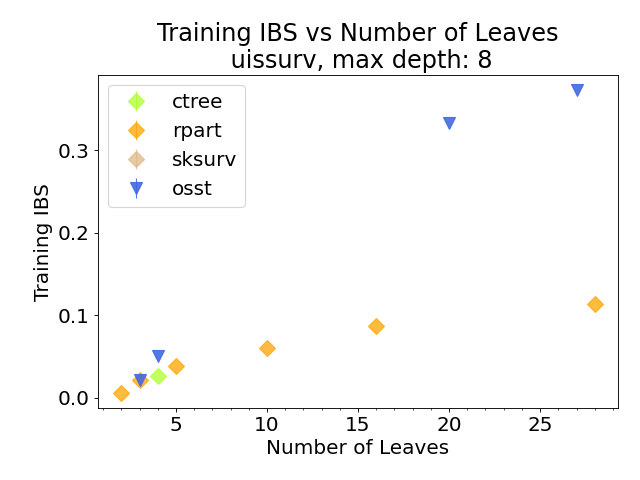}
    \includegraphics[width=0.42\textwidth]{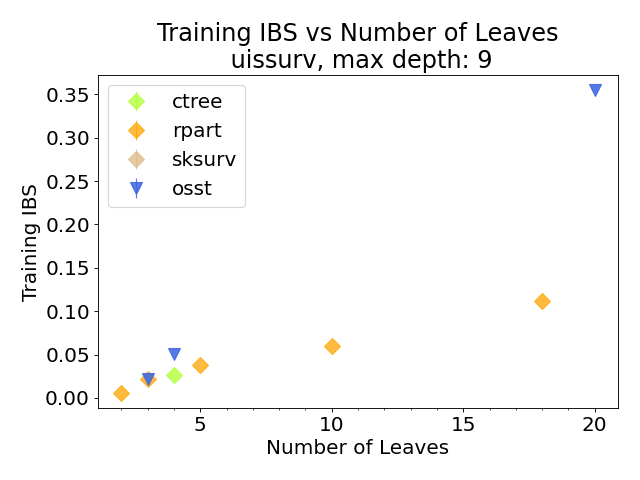}
    
    \caption{Training IBS achieved by CTree, RPART, SkSurv and OSST as a function of number of leaves on dataset: uissurv.}
    \label{fig:lvs:uissurv-ibs}
\end{figure*}

\begin{figure*}[htbp]
    \centering
    \includegraphics[width=0.42\textwidth]{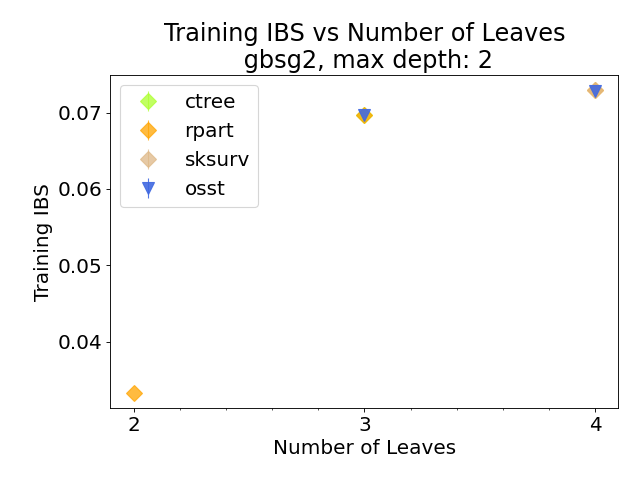}
    \includegraphics[width=0.42\textwidth]{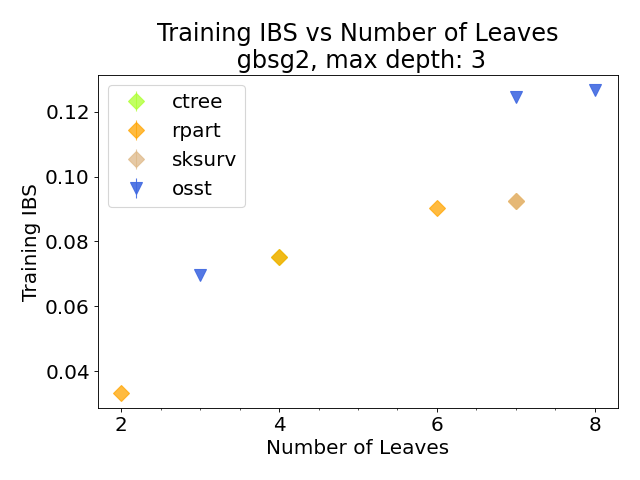}
    \includegraphics[width=0.42\textwidth]{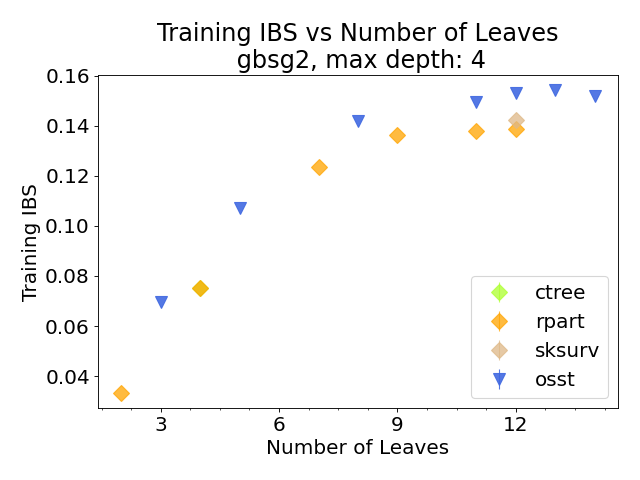}
    \includegraphics[width=0.42\textwidth]{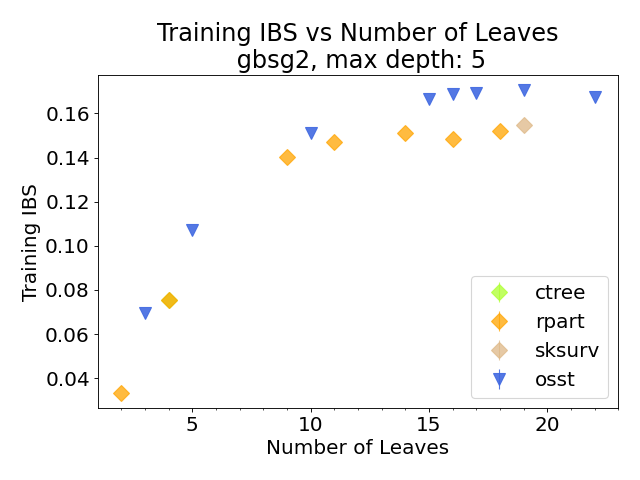}
    \includegraphics[width=0.42\textwidth]{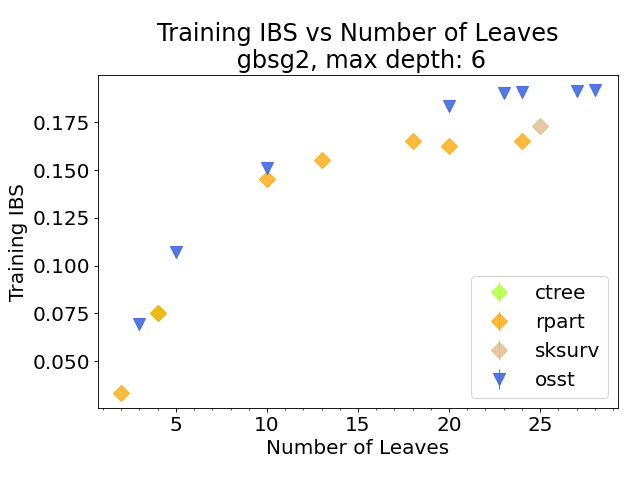}
    \includegraphics[width=0.42\textwidth]{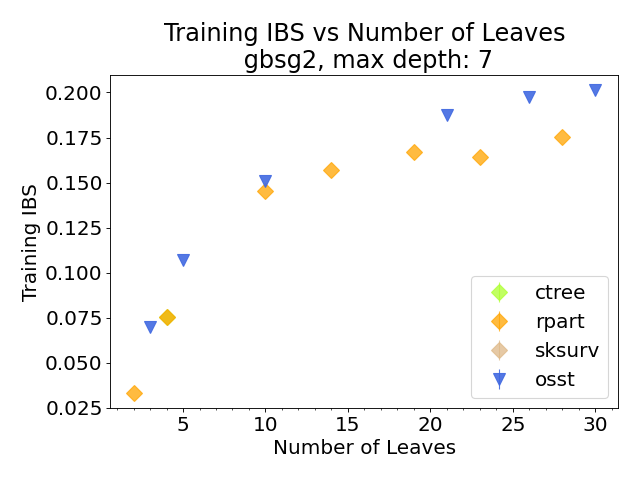}
    \includegraphics[width=0.42\textwidth]{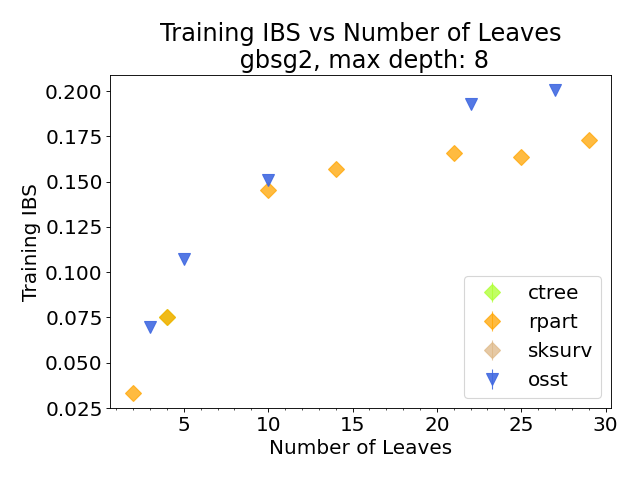}
    \includegraphics[width=0.42\textwidth]{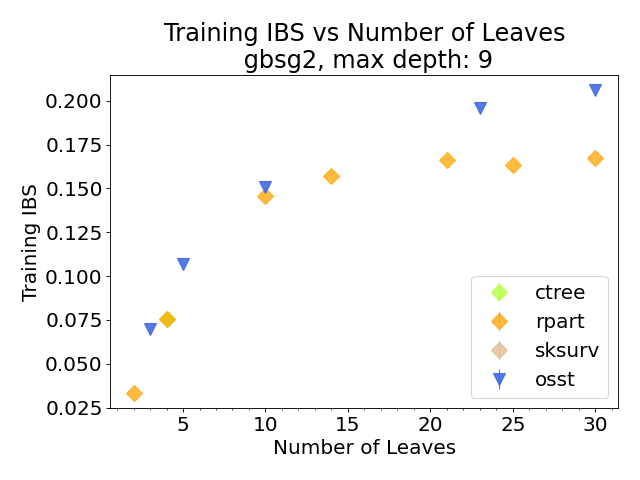}
    \caption{Training IBS achieved by CTree, RPART, SkSurv and OSST as a function of number of leaves on dataset: gbsg2.}
    \label{fig:lvs:gbsg2-ibs}
\end{figure*}

\begin{figure*}[htbp]
    \centering
    \includegraphics[width=0.42\textwidth]{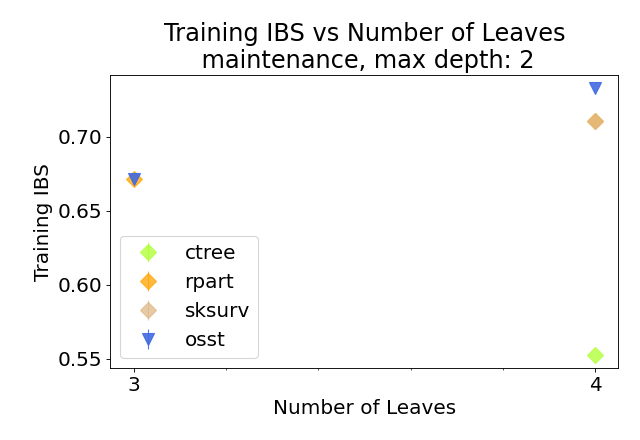}
    \includegraphics[width=0.42\textwidth]{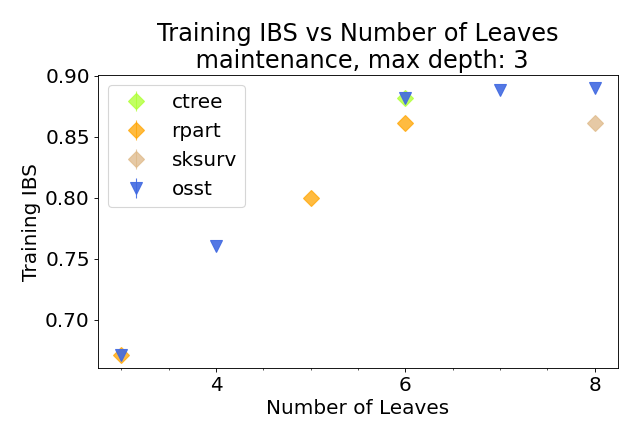}
    \includegraphics[width=0.42\textwidth]{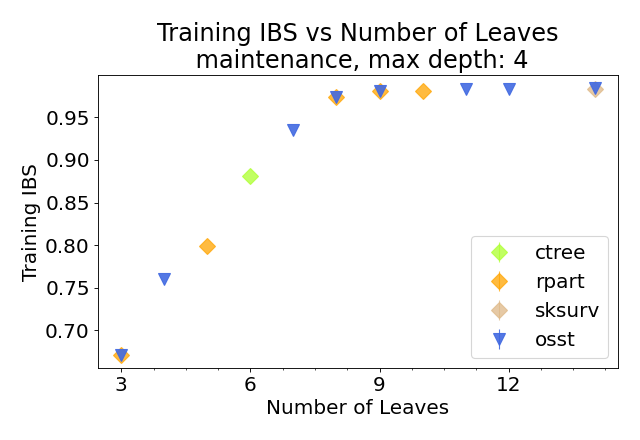}
    \includegraphics[width=0.42\textwidth]{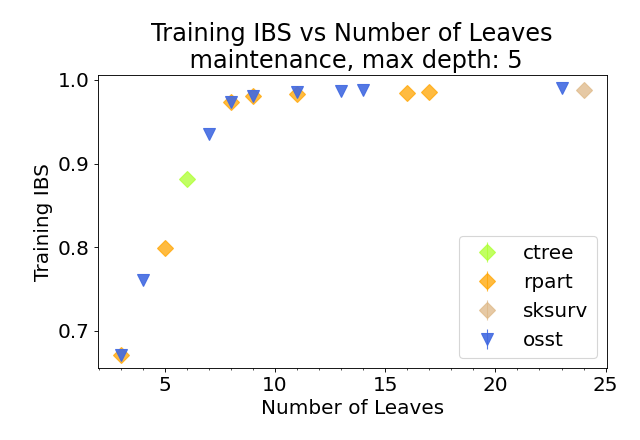}
    \includegraphics[width=0.42\textwidth]{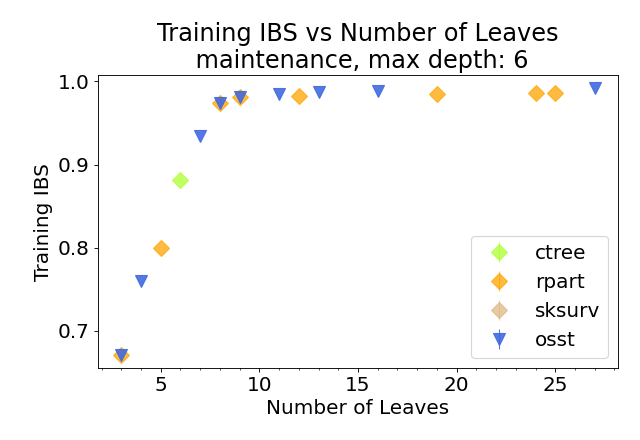}
    \includegraphics[width=0.42\textwidth]{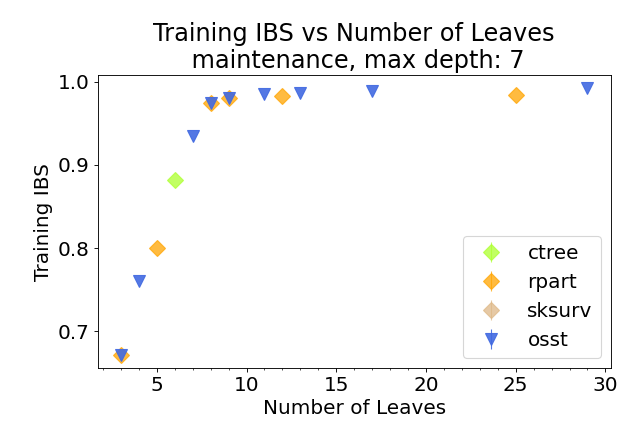}
    \includegraphics[width=0.42\textwidth]{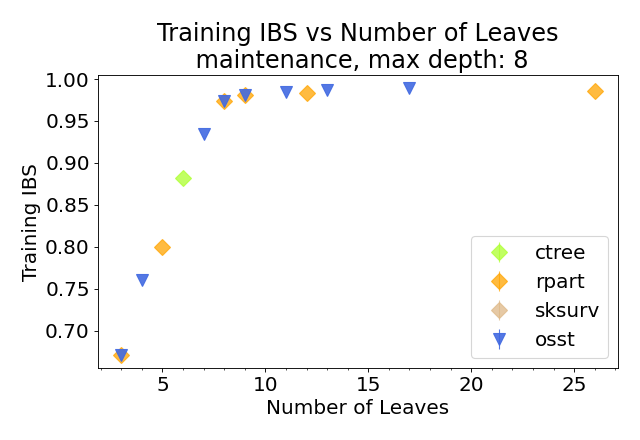}
    \includegraphics[width=0.42\textwidth]{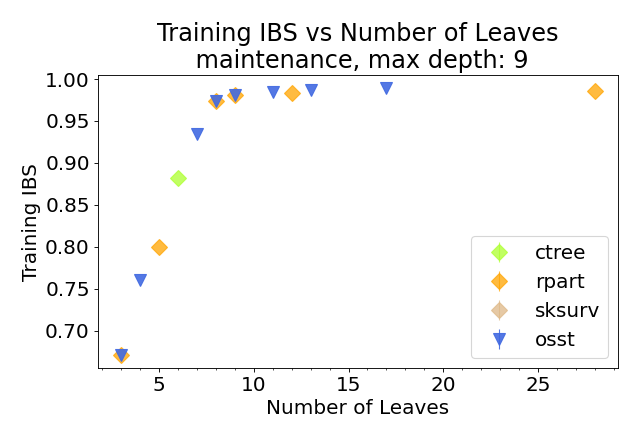}
    \caption{Training IBS achieved by CTree, RPART, SkSurv and OSST as a function of number of leaves on dataset: maintenance.}
    \label{fig:lvs:maintenance-ibs}
\end{figure*}

\begin{figure*}[htbp]
    \centering
    \includegraphics[width=0.42\textwidth]{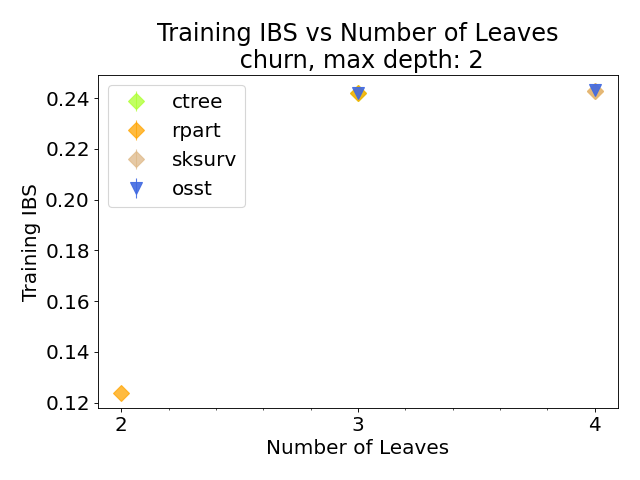}
    \includegraphics[width=0.42\textwidth]{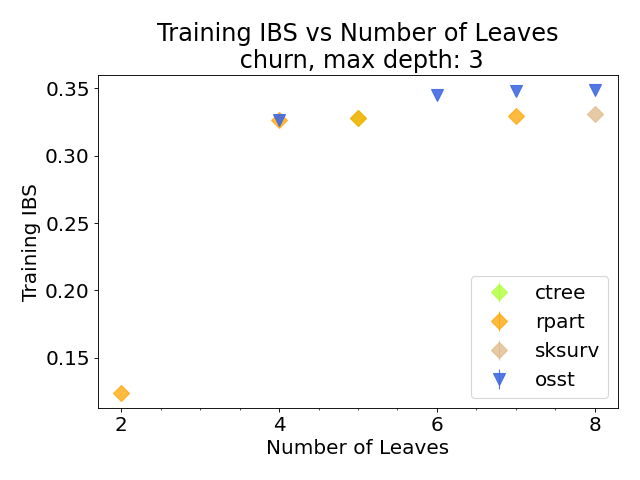}
    \includegraphics[width=0.42\textwidth]{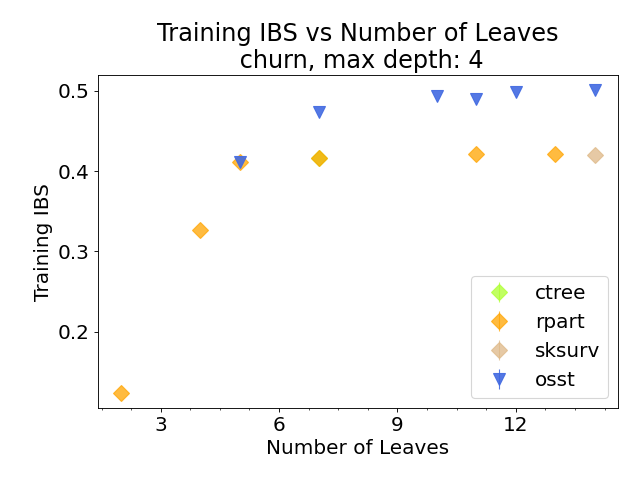}
    \includegraphics[width=0.42\textwidth]{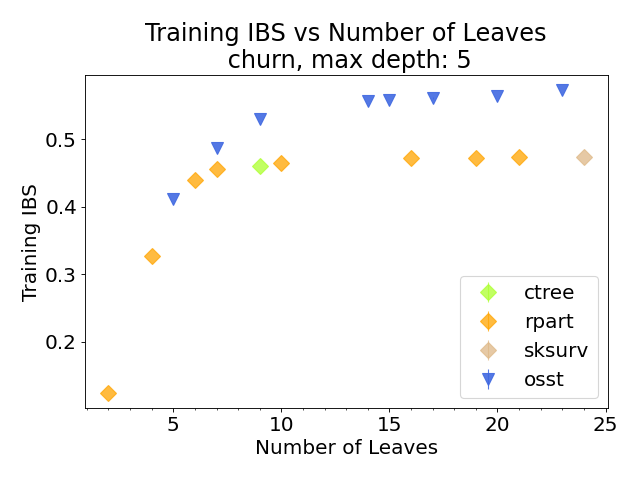}
    \includegraphics[width=0.42\textwidth]{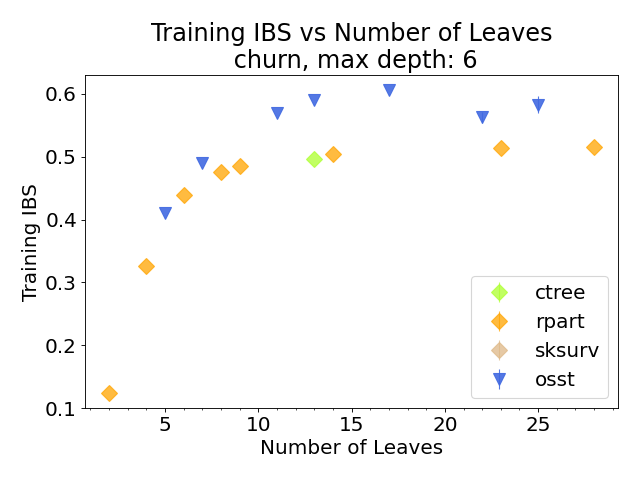}
    \includegraphics[width=0.42\textwidth]{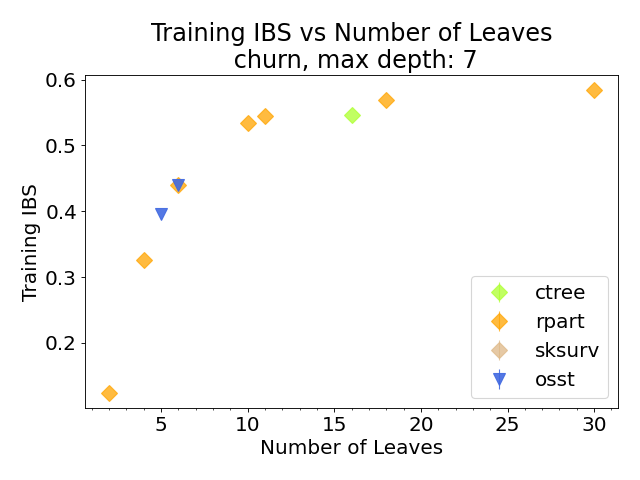}
    \includegraphics[width=0.42\textwidth]{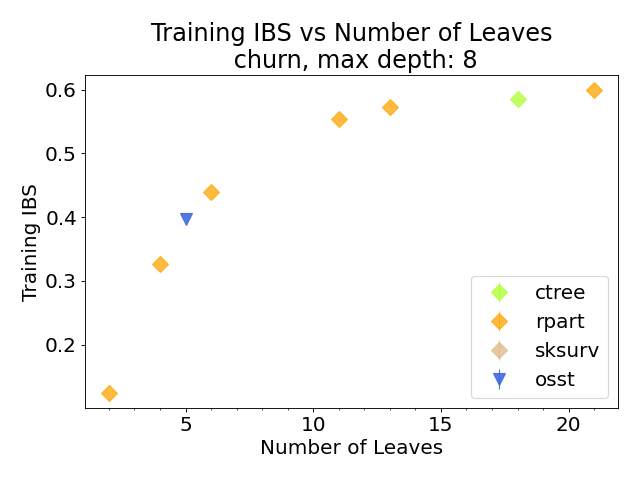}
    \includegraphics[width=0.42\textwidth]{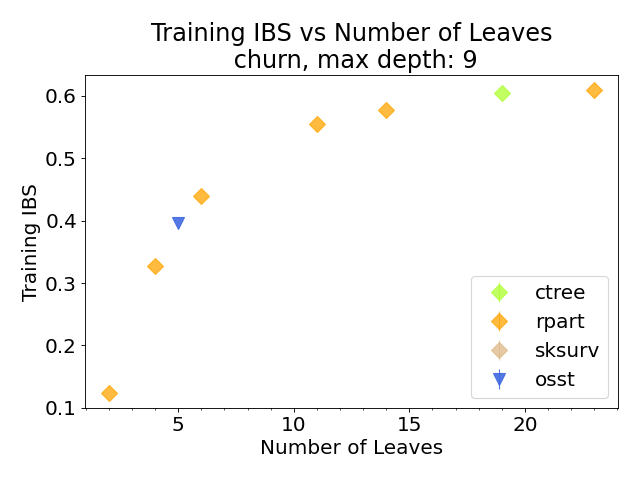}
    \caption{Training IBS achieved by CTree, RPART, SkSurv and OSST as a function of number of leaves on dataset: churn.}
    \label{fig:lvs:churn-ibs}
\end{figure*}

\begin{figure*}[htbp]
    \centering
    \includegraphics[width=0.42\textwidth]{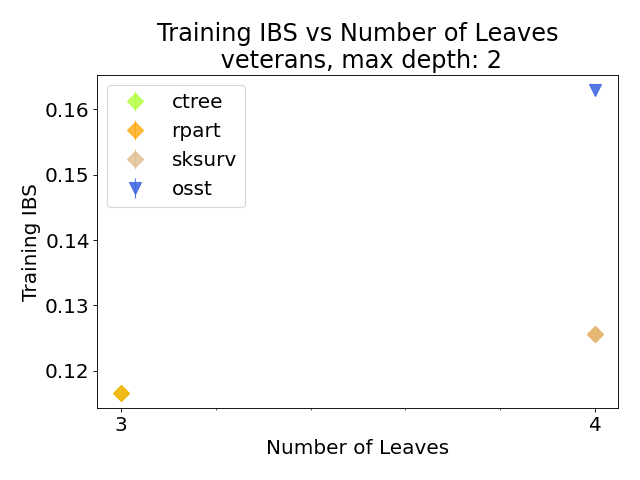}
    \includegraphics[width=0.42\textwidth]{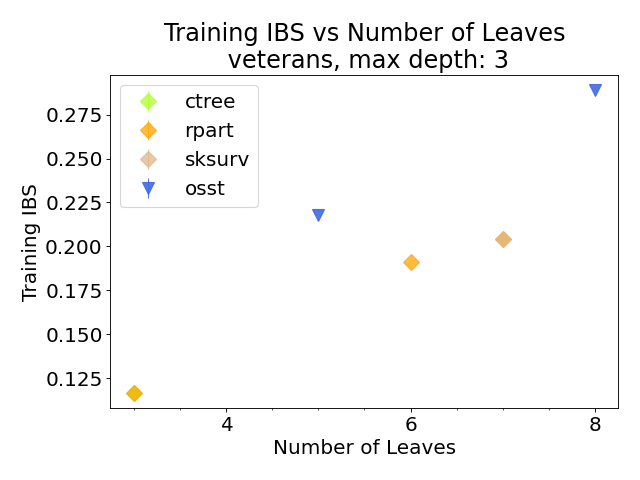}
    \includegraphics[width=0.42\textwidth]{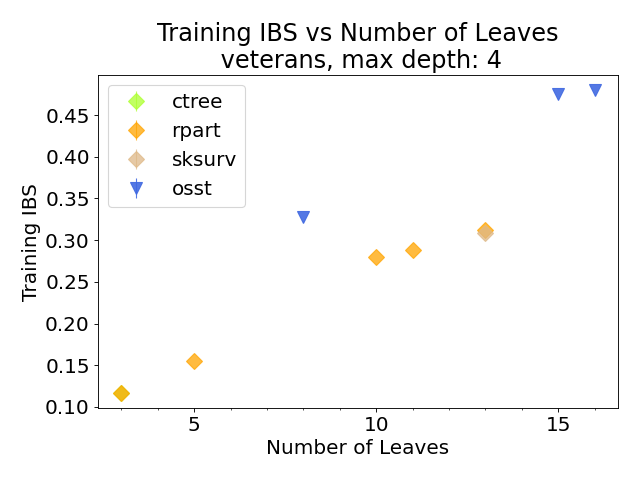}
    \includegraphics[width=0.42\textwidth]{figures/loss_vs_sparsity/veterans/ibs/veterans_depth_5.png}
    \includegraphics[width=0.42\textwidth]{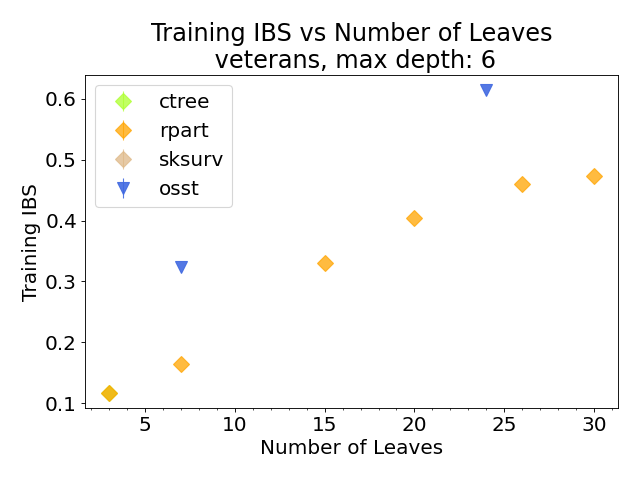}
    \includegraphics[width=0.42\textwidth]{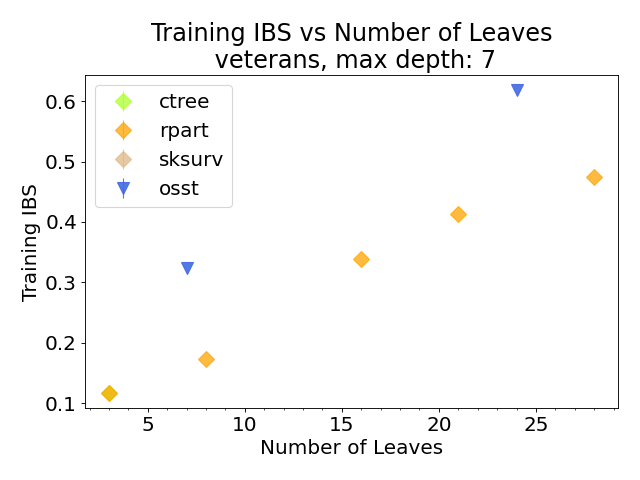}
    \includegraphics[width=0.42\textwidth]{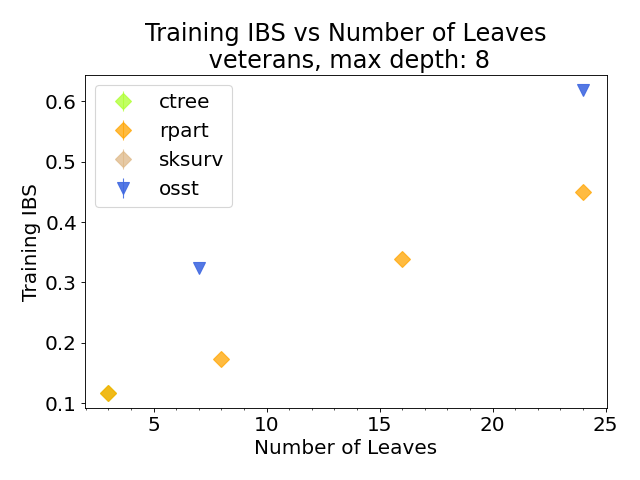}
    \includegraphics[width=0.42\textwidth]{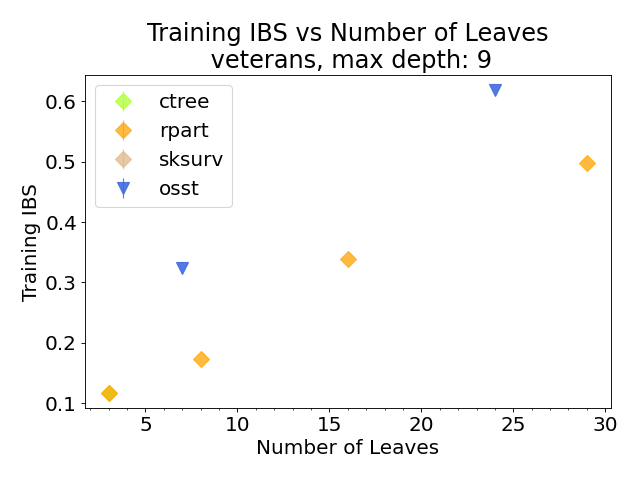}
    \caption{Training IBS achieved by CTree, RPART, SkSurv and OSST as a function of number of leaves on dataset: veterans.}
    \label{fig:lvs:veterans-ibs}
\end{figure*}

\begin{figure*}[htbp]
    \centering
    \includegraphics[width=0.42\textwidth]{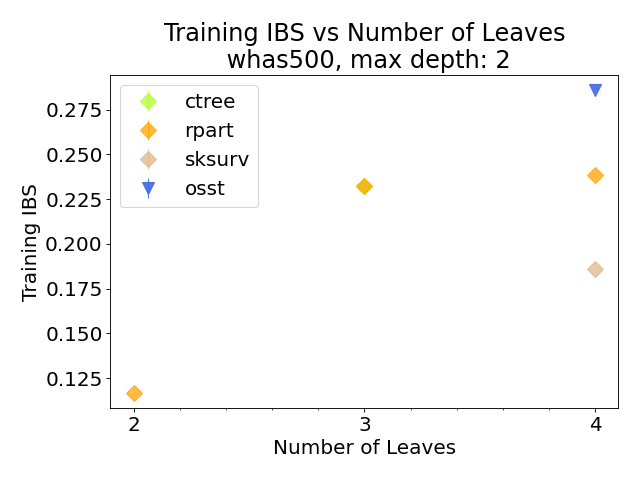}
    \includegraphics[width=0.42\textwidth]{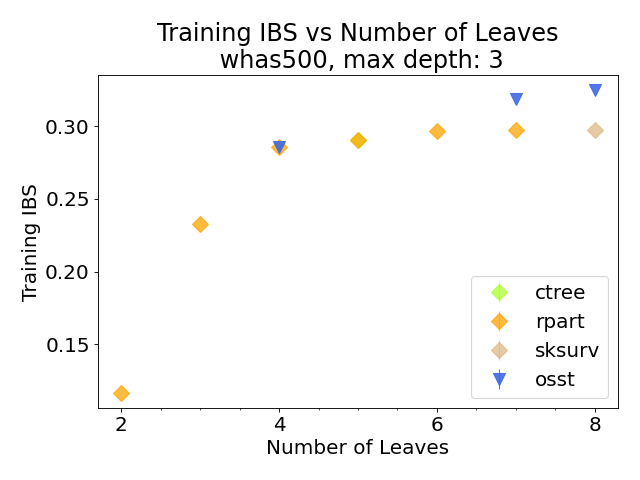}
    \includegraphics[width=0.42\textwidth]{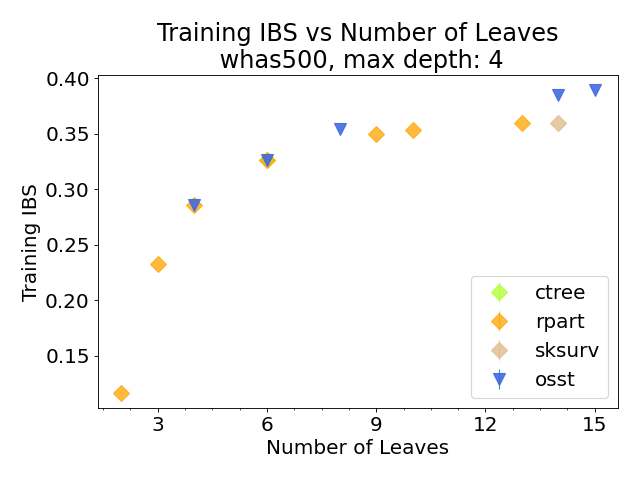}
    \includegraphics[width=0.42\textwidth]{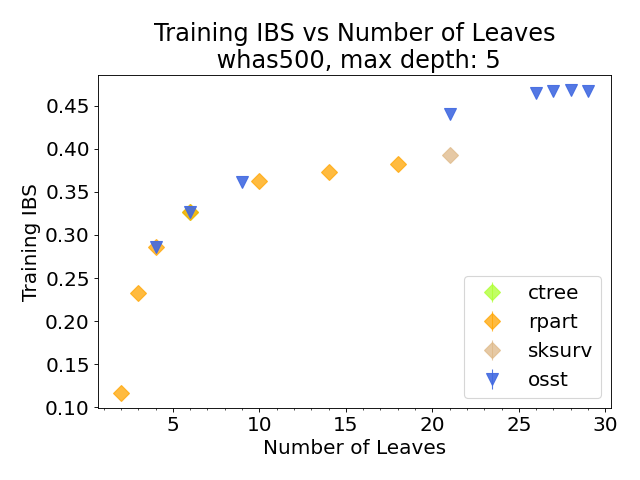}
    \includegraphics[width=0.42\textwidth]{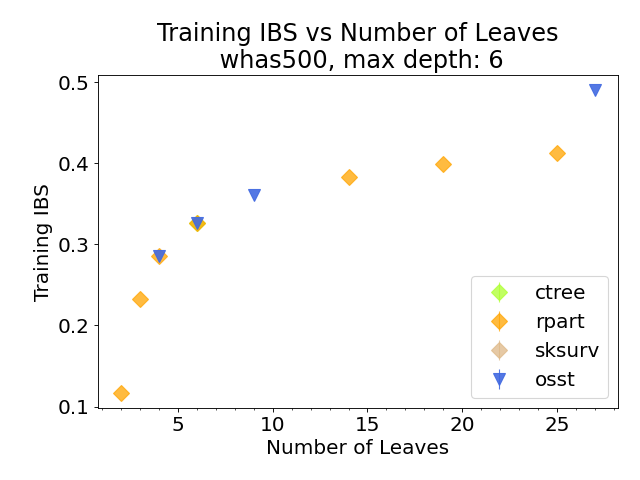}
    \includegraphics[width=0.42\textwidth]{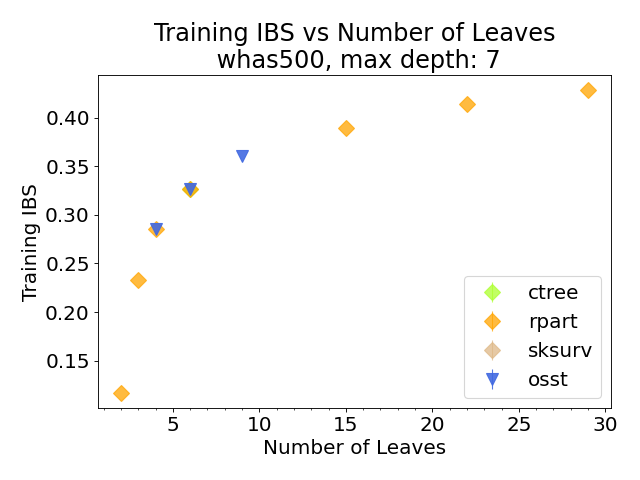}
    \includegraphics[width=0.42\textwidth]{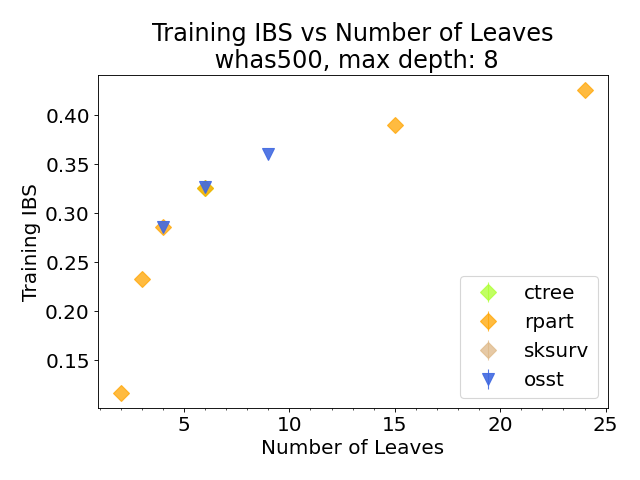}
    \includegraphics[width=0.42\textwidth]{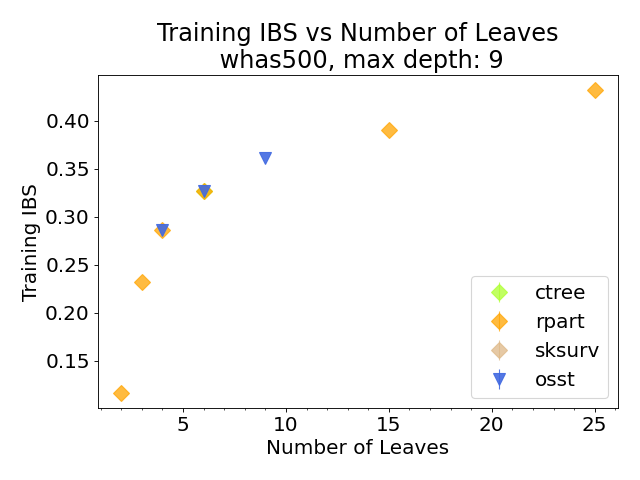}
    \caption{Training IBS achieved by CTree, RPART, SkSurv and OSST as a function of number of leaves on dataset: whas500.}
    \label{fig:lvs:whas500-ibs}
\end{figure*}

\noindent\textbf{Results:}
Figures \ref{fig:lvs:aids-ibs} to \ref{fig:lvs:whas500-ibs} show that \ourmethod{} consistently produces optimal trees that achieve the best IBS score, which means they minimize the objective defined in Equation \ref{eq:obj1}. \ourmethod{}, which produces provably optimal trees, defines a frontier between training loss and the number of leaves. These plots also show how far away other methods' objectives are from the optimal solution. \textit{OSST make it possible to quantify how far other methods are from the optimal solution.} We also noticed that the plots from depth 7 to 9 look similar or identical;  when the regularization coefficient is too small or the depth limit is too large,  OSST produces complex and overfitted trees with more than 30 leaves, which we excluded from the plots.

\newpage 
\section{Experiments: Generalization}\label{exp:cv}
\noindent\textbf{Collection and Setup:}
We ran 5-fold cross-validation on the 5 datasets: \textit{churn, credit, employee, maintenance, and servo}. The time limit was set to 60 minutes. For each dataset, we ran algorithms with different configurations:
\begin{itemize}[leftmargin=10pt]
    \item \textbf{CTree}: We ran this algorithm with 4 different configurations: depth limit, $d$, ranging from 2 to 5, and a corresponding maximum leaf limit $2^d$. All other parameters were set to the default.
    \item \textbf{SkSurv}: We ran this algorithm with 4 different configurations: depth limit, $d$, ranging from 2 to 5, and a corresponding maximum leaf limit $2^d$. The random state was set to 2023 and all other parameters were set to the default.
    \item \textbf{RPART}: We ran this algorithm with $4 \times 12$ different configurations: depth limits ranging from 2 to 5, and 12 different regularization coefficients (0.1, 0.05, 0.025, 0.01, 0.005, 0.0025, 0.001, 0.0005, 0.00025, 0.0001, 0.00005, 0.00001). All other parameters were set to the default.
    \item \textbf{OSST} (our method): We ran this algorithm with $4 \times 12$ different configurations: depth limits ranging from 2 to 5, and 12 different regularization coefficients (0.1, 0.05, 0.01, 0.005, 0.0025, 0.001, 0.0005, 0.00025, 0.0001, 0.00005, 0.000025, 0.00001). 
\end{itemize}
\noindent\textbf{Calculations:} We drew one plot per dataset and depth. For each combination of regularization coefficient and algorithm in the same plot, we produced a set of up to 5 trees, depending on if the runs exceeded the time limit. We summarized the measurements of training loss, testing loss and number of leaves across the set of up to 5 trees by plotting the median, showing the 25th and 75th quantiles for  number of leaves, training/testing error in the set as the lower and upper error values respectively. As in Section \ref{exp:lvs}, trees with more than 30 leaves and single root node were excluded. 

\begin{figure*}
\centering
\includegraphics[width=0.42\textwidth]{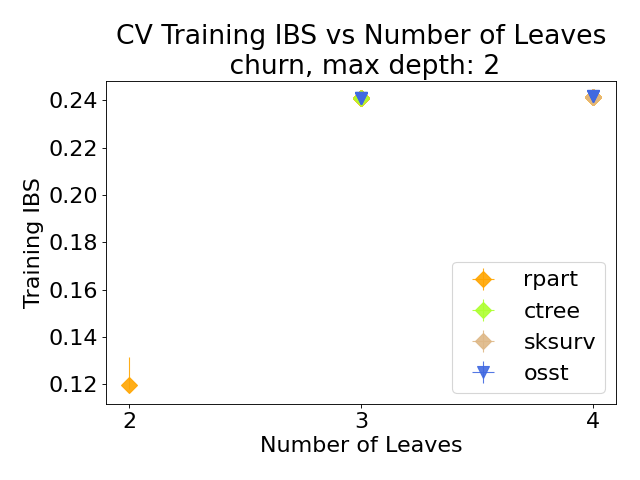}
\includegraphics[width=0.42\textwidth]{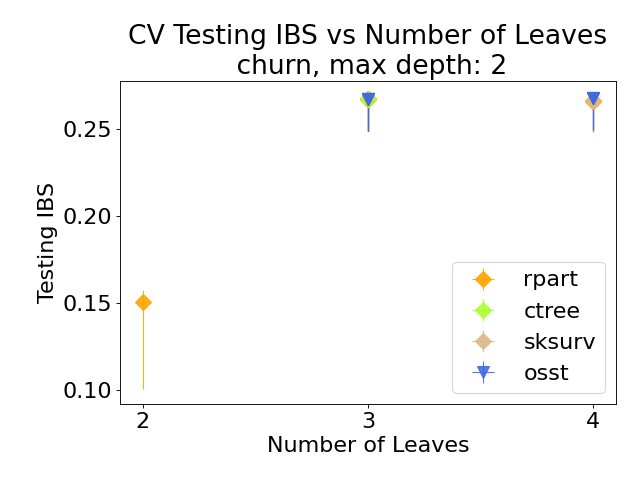}
\includegraphics[width=0.42\textwidth]{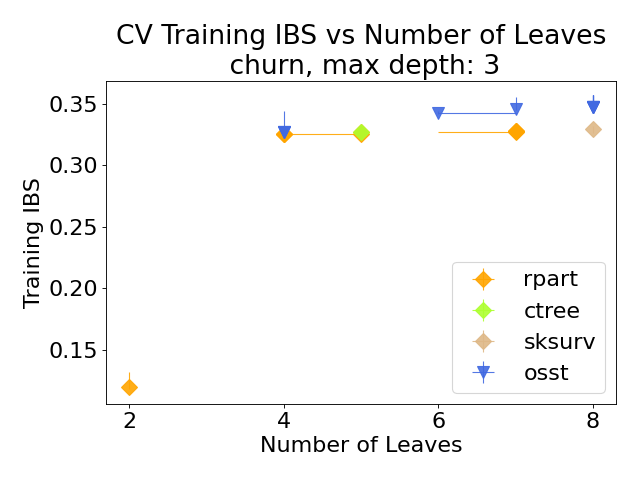}
\includegraphics[width=0.42\textwidth]{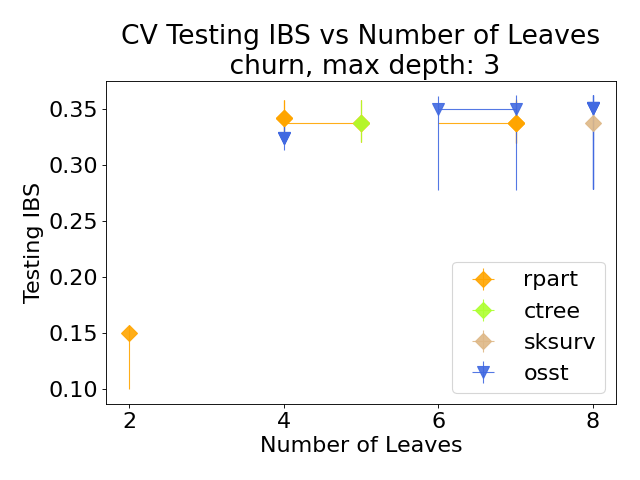}
\includegraphics[width=0.42\textwidth]{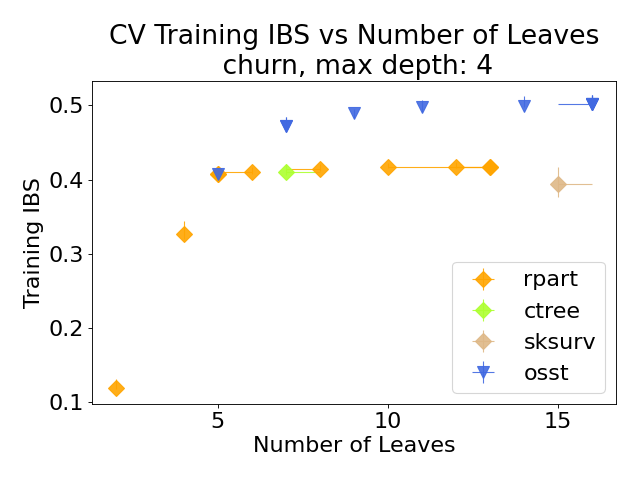}
\includegraphics[width=0.42\textwidth]{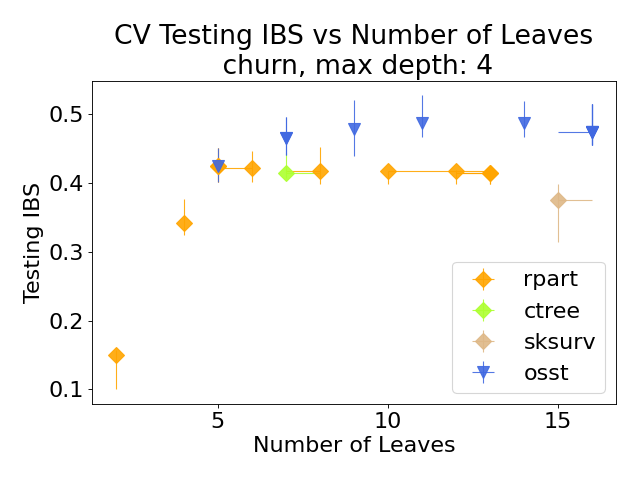}
\includegraphics[width=0.42\textwidth]{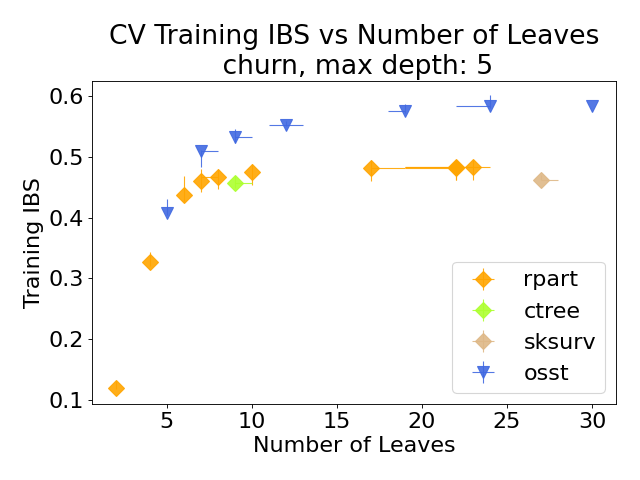}
\includegraphics[width=0.42\textwidth]{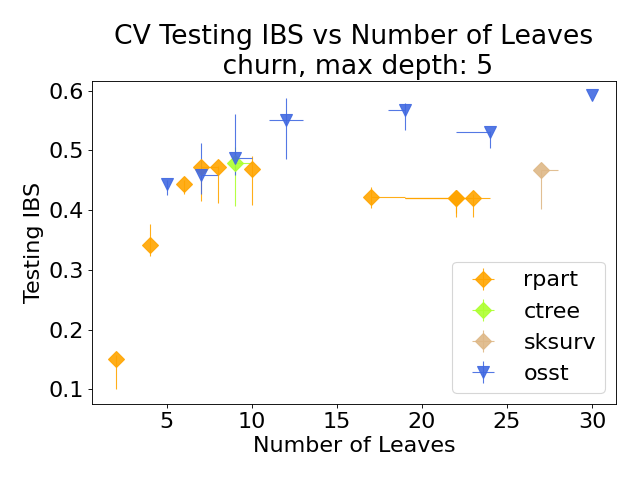}
\caption{5-fold CV of CTree, RPART, SkSurv and OSST as a function of number of leaves on dataset: churn.}
\label{fig:cv:churn-ibs}
\end{figure*}

\begin{figure*}
\centering
\includegraphics[width=0.42\textwidth]{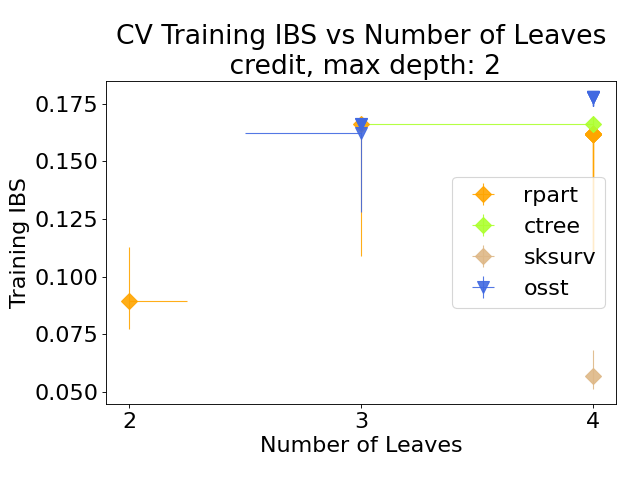}
\includegraphics[width=0.42\textwidth]{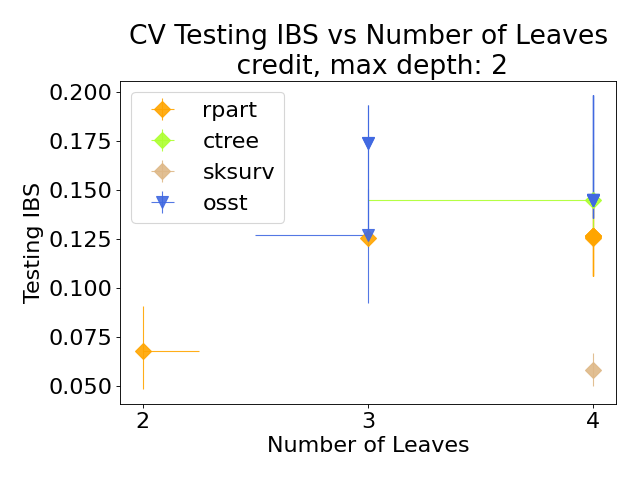}
\includegraphics[width=0.42\textwidth]{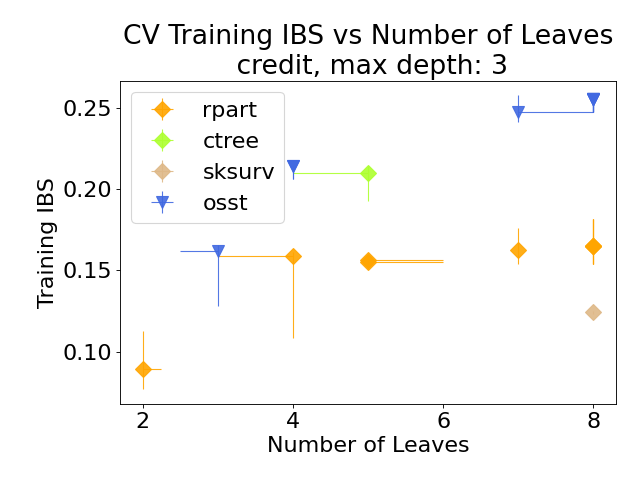}
\includegraphics[width=0.42\textwidth]{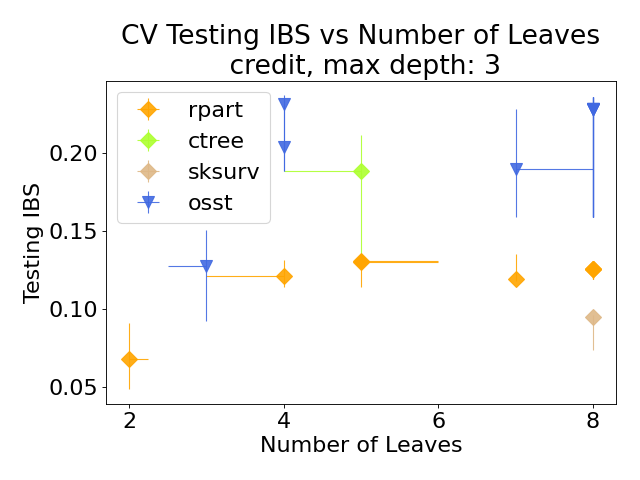}
\includegraphics[width=0.42\textwidth]{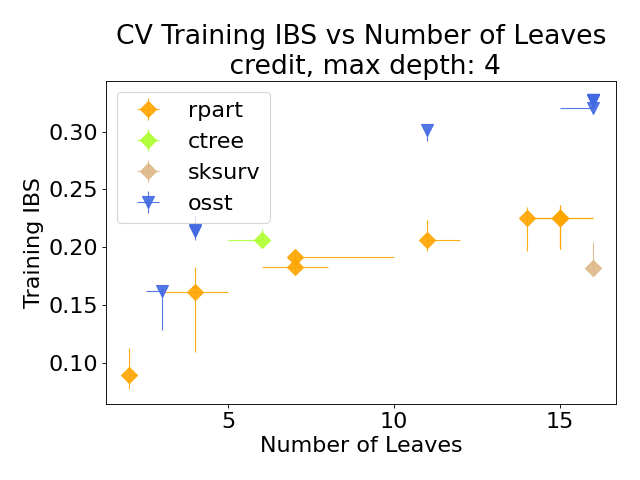}
\includegraphics[width=0.42\textwidth]{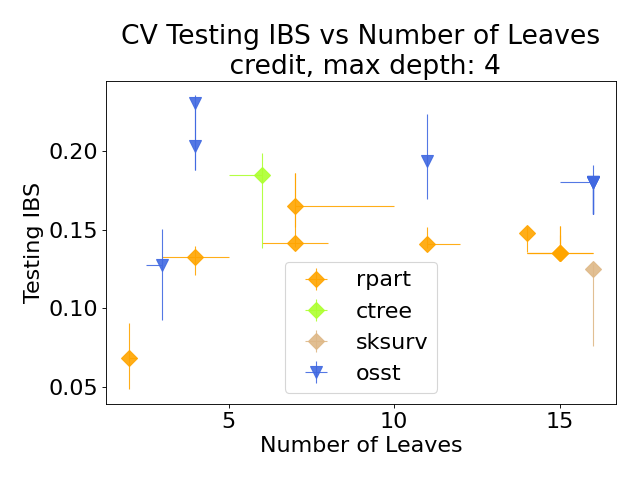}
\includegraphics[width=0.42\textwidth]{figures/cv/credit/ibs/credit_depth5_ibs_train.png}
\includegraphics[width=0.42\textwidth]{figures/cv/credit/ibs/credit_depth5_ibs_test.png}
\caption{5-fold CV of CTree, RPART, SkSurv and OSST as a function of number of leaves on dataset: credit.}
\label{fig:cv:credit-ibs}
\end{figure*}

\begin{figure*}
\centering
\includegraphics[width=0.42\textwidth]{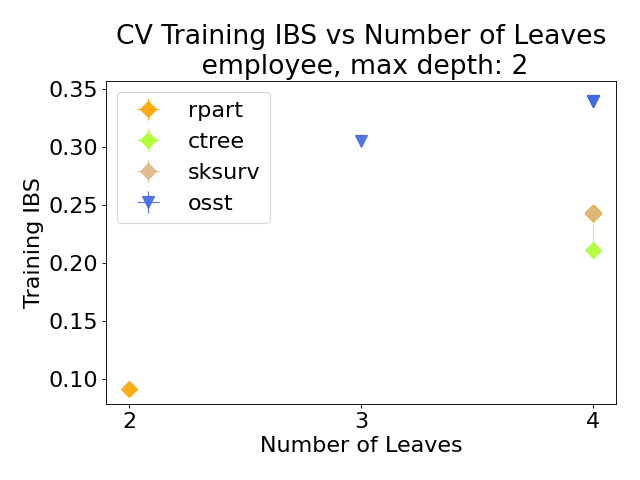}
\includegraphics[width=0.42\textwidth]{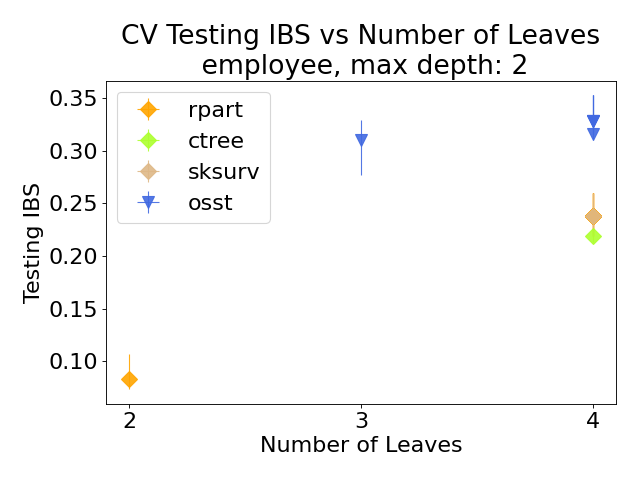}
\includegraphics[width=0.42\textwidth]{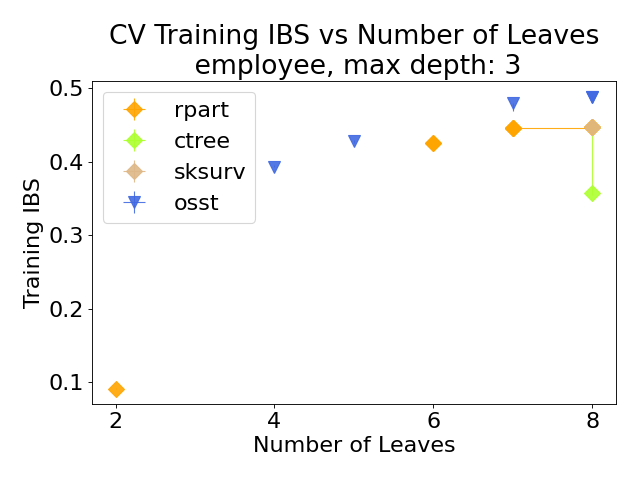}
\includegraphics[width=0.42\textwidth]{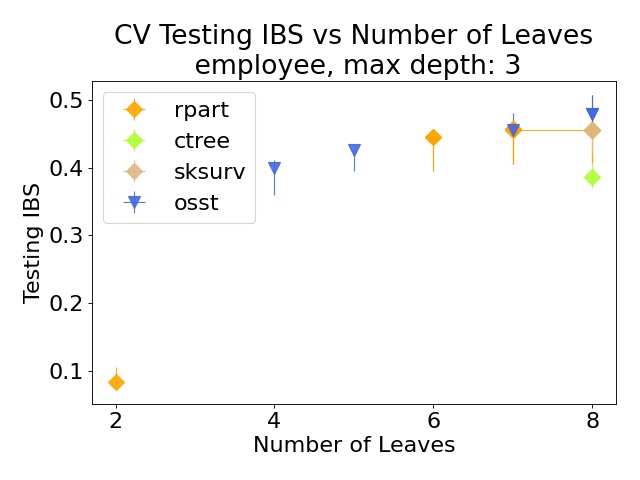}
\includegraphics[width=0.42\textwidth]{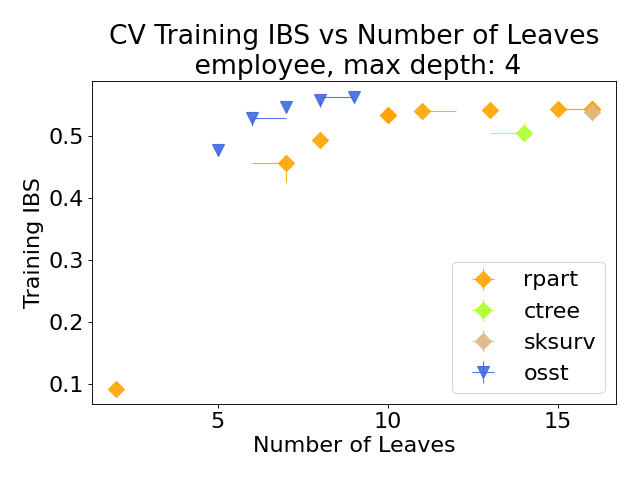}
\includegraphics[width=0.42\textwidth]{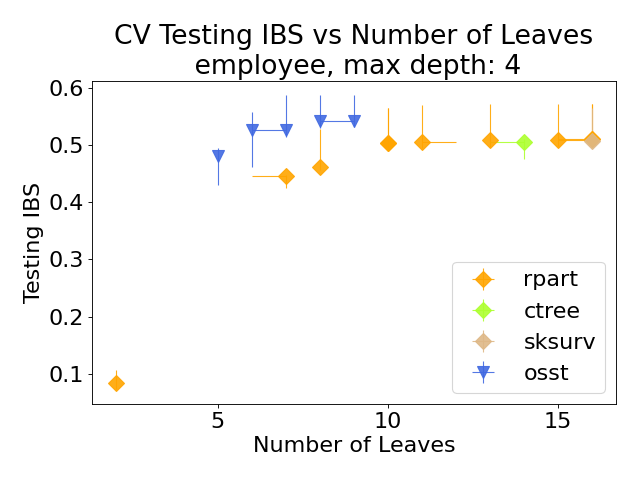}
\includegraphics[width=0.42\textwidth]{figures/cv/employee/ibs/employee_depth5_ibs_train.png}
\includegraphics[width=0.42\textwidth]{figures/cv/employee/ibs/employee_depth5_ibs_test.png}
\caption{5-fold CV of CTree, RPART, SkSurv and OSST as a function of number of leaves on dataset: employee.}
\label{fig:cv:employee-ibs}
\end{figure*}

\begin{figure*}
\centering
\includegraphics[width=0.42\textwidth]{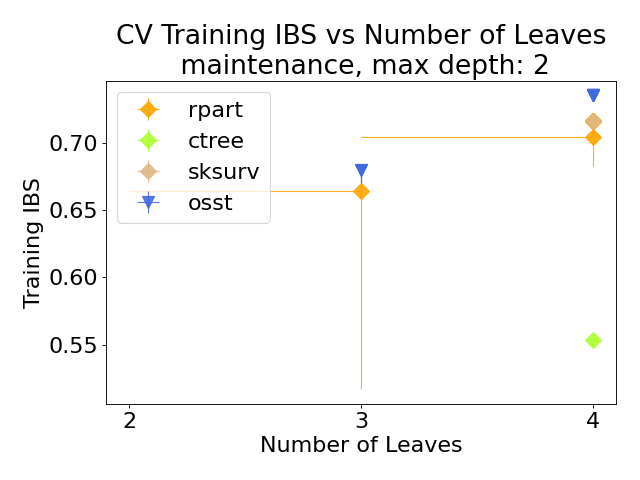}
\includegraphics[width=0.42\textwidth]{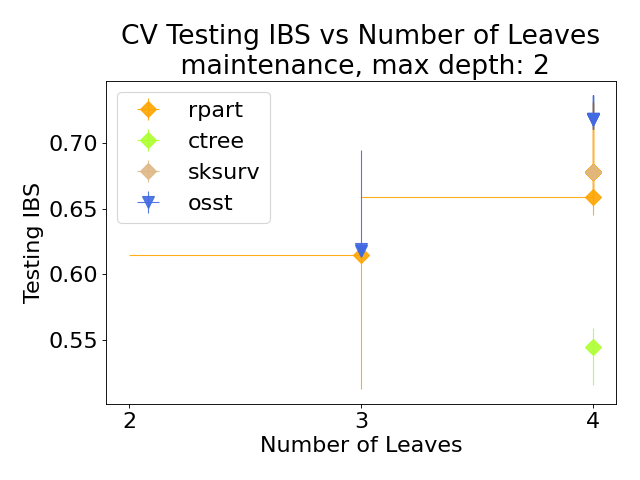}
\includegraphics[width=0.42\textwidth]{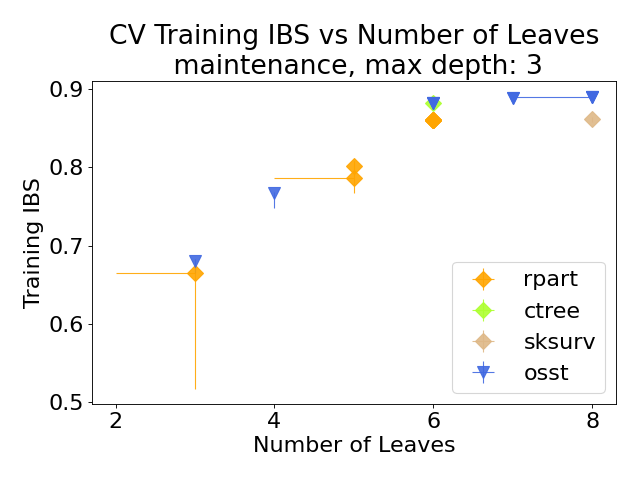}
\includegraphics[width=0.42\textwidth]{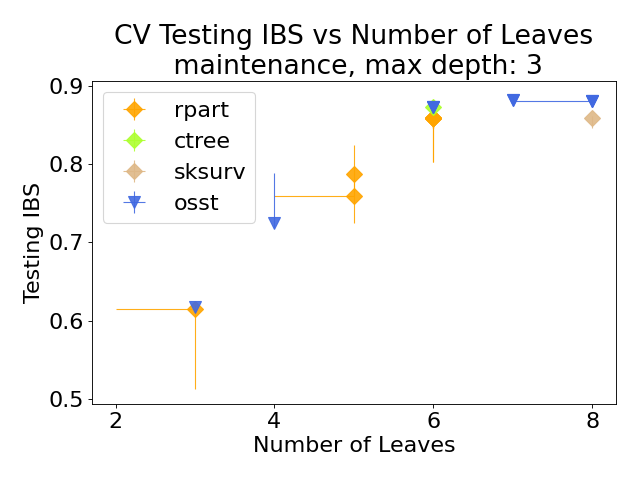}
\includegraphics[width=0.42\textwidth]{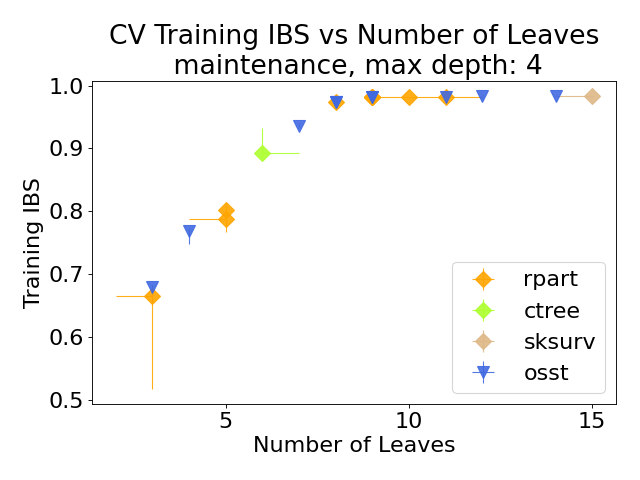}
\includegraphics[width=0.42\textwidth]{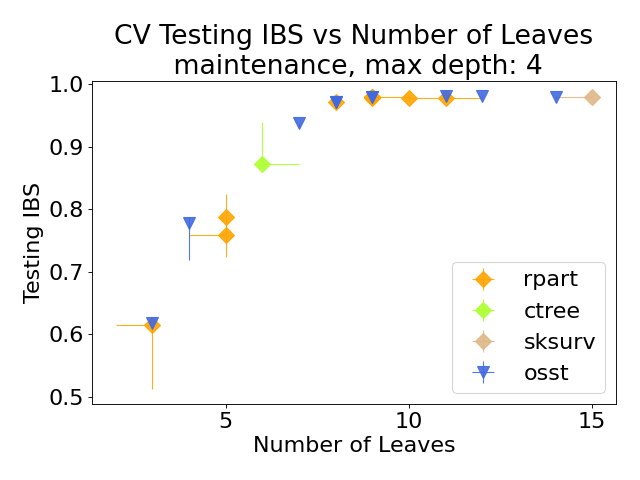}
\includegraphics[width=0.42\textwidth]{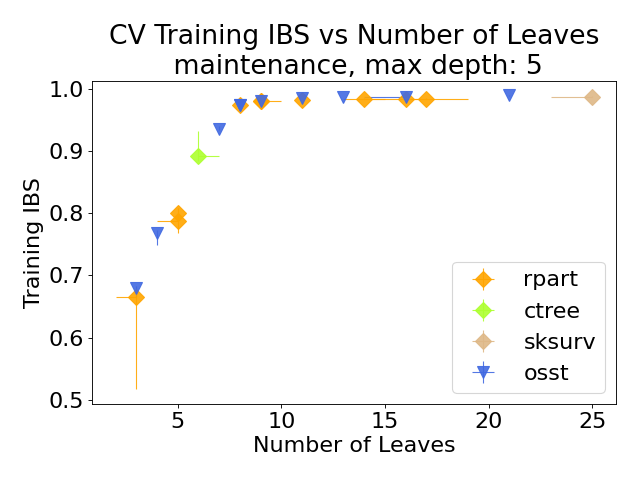}
\includegraphics[width=0.42\textwidth]{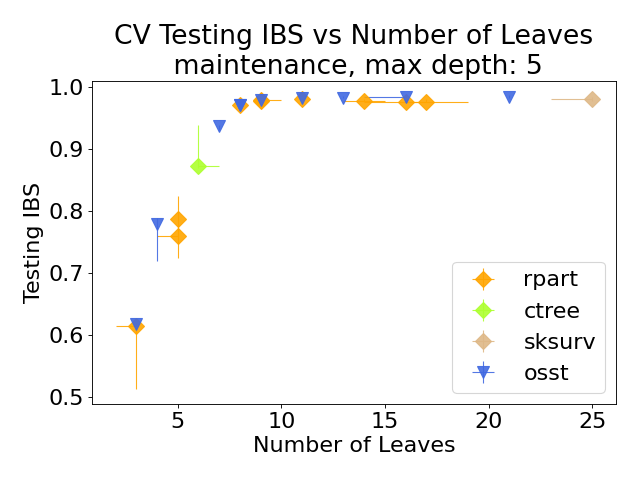}
\caption{5-fold CV of CTree, RPART, SkSurv and OSST as a function of number of leaves on dataset: maintenance.}
\label{fig:cv:maintenance-ibs}
\end{figure*}

\begin{figure*}
\centering
\includegraphics[width=0.42\textwidth]{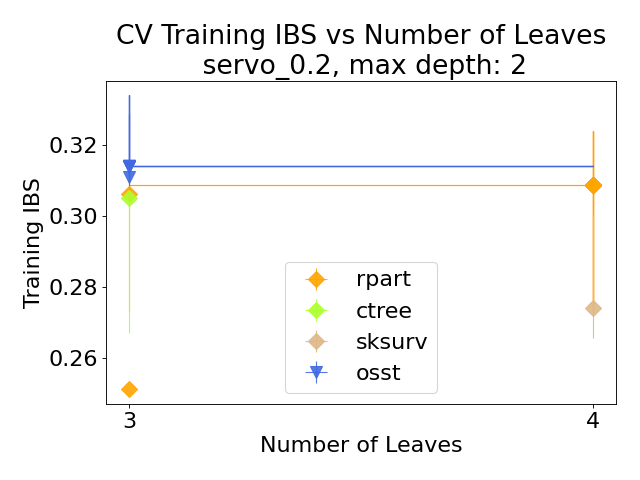}
\includegraphics[width=0.42\textwidth]{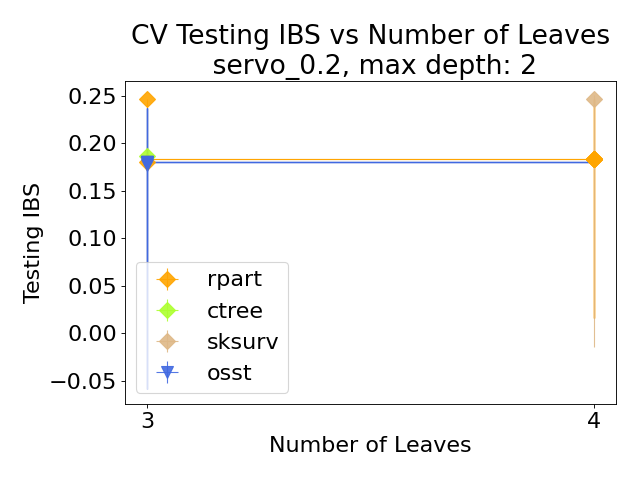}
\includegraphics[width=0.42\textwidth]{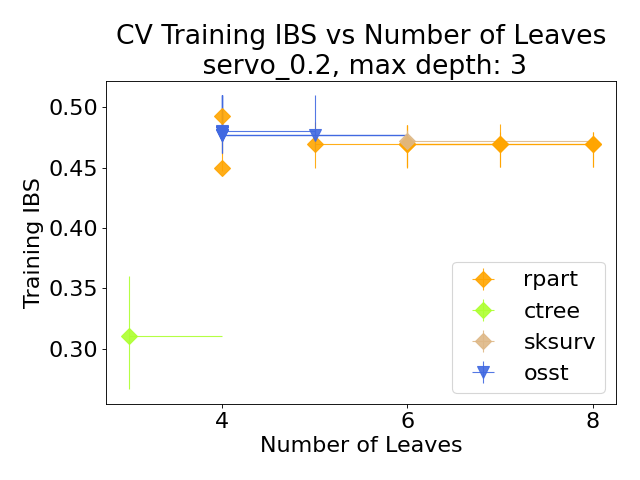}
\includegraphics[width=0.42\textwidth]{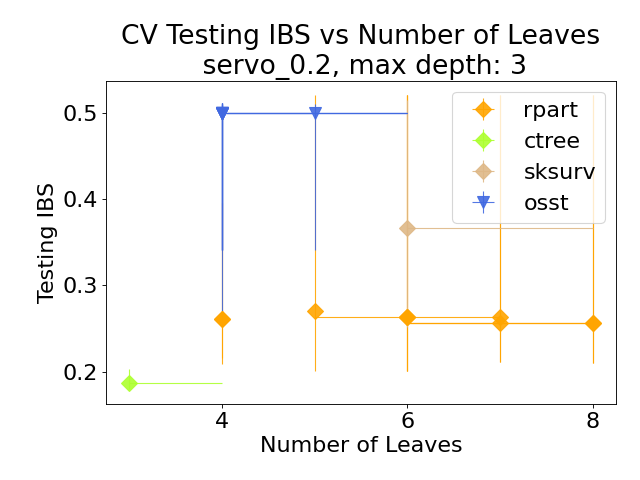}
\includegraphics[width=0.42\textwidth]{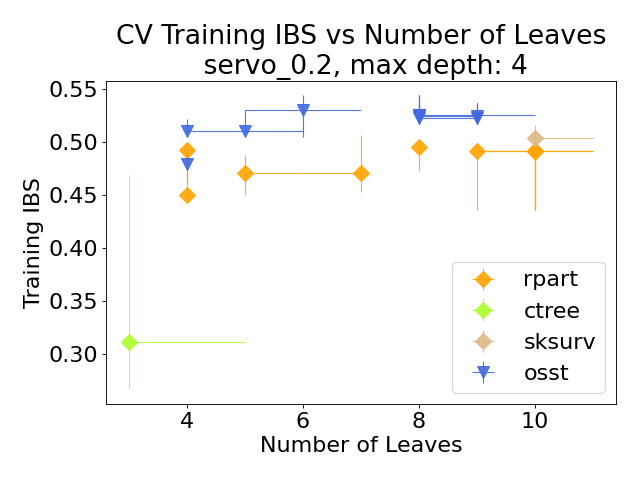}
\includegraphics[width=0.42\textwidth]{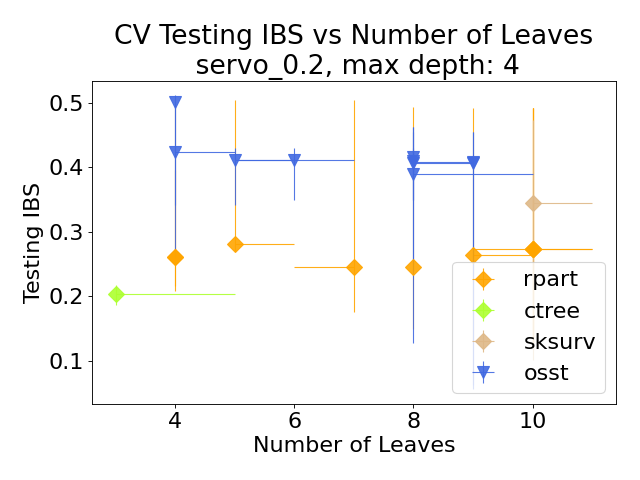}
\includegraphics[width=0.42\textwidth]{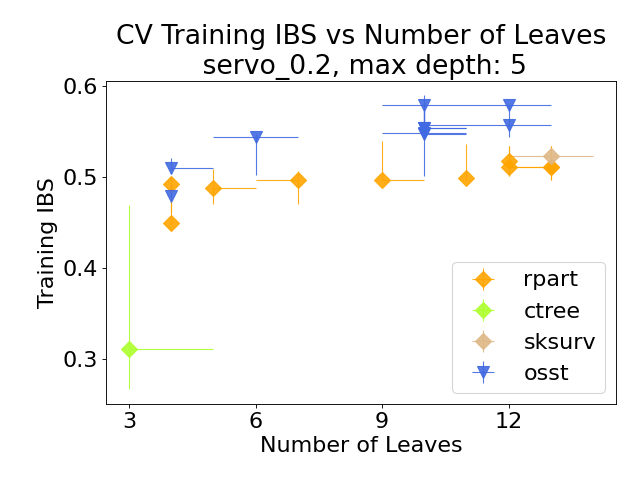}
\includegraphics[width=0.42\textwidth]{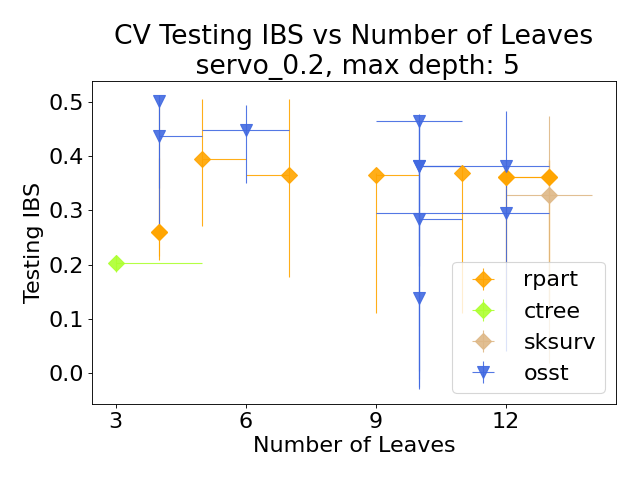}
\caption{5-fold CV of CTree, RPART, SkSurv and OSST as a function of number of leaves on dataset: servo. (Trees with more than 9 leaves overfitted at depth 5)}
\label{fig:cv:servo-ibs}
\end{figure*}

\noindent\textbf{Results:} Figures \ref{fig:cv:churn-ibs} to \ref{fig:cv:servo-ibs} show that \ourmethod{} trees produce the best testing score among all the survival trees. If an optimal tree significantly outperforms other sub-optimal trees in terms of training performance, it is also outperformed in testing (e.g., \textit{churn} depth 4 and 5; \textit{credit} depth 4), otherwise the difference in testing score becomes insignificant due to generalization error. Note that larger trees start overfitting when depth is greater than 4 or 5 (e.g., \textit{servo} depth 5), and sparse trees tend to have better generalization.

\newpage
\section{Experiment: Tree Quality}\label{exp:quality}
\noindent \textbf{Collection and Setup:} One limitation of the IBS metric is that it heavily penalizes models for being inaccurate in the early stages of follow-up, so we would like to see if the comprehensive quality of optimal sparse trees found by \ourmethod{} is actually better than other sub-optimal trees. We evaluated the experiments again in Section \ref{exp:lvs} and \ref{exp:cv} but using \textbf{Uno's C-index, Harrell's C-index} and \textbf{AUC} (note that OSST optimizes the IBS loss). We evaluate the optimization experiments on \textit{gbsg2, whas500} in Section \ref{exp:lvs} and the cross-validation experiments on \textit{churn, employee} in Section \ref{exp:cv}.

\noindent\textbf{Calculations:} Again, we drew one plot per dataset and depth. We replaced the performance metrics to Uno's C-index, Harrell's C-index and AUC.

\begin{figure*}[htbp]
    \centering
    \includegraphics[width=0.42\textwidth]{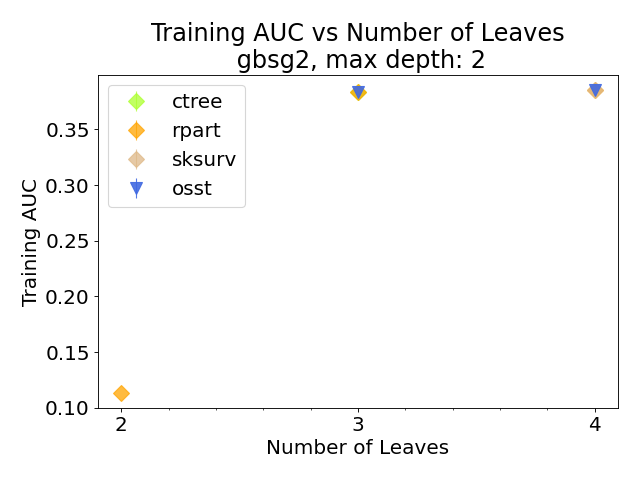}
    \includegraphics[width=0.42\textwidth]{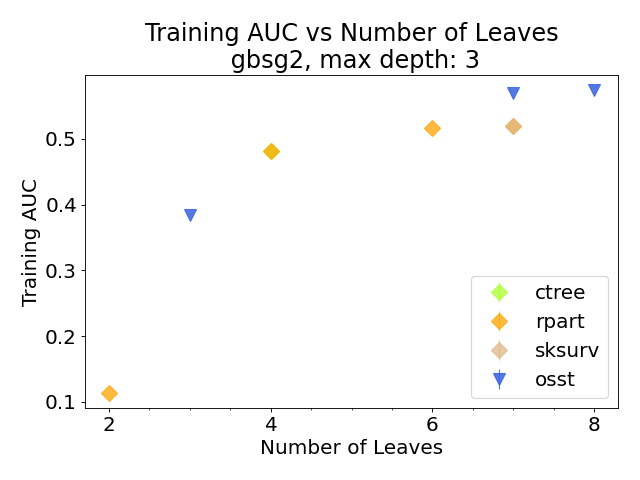}
    \includegraphics[width=0.42\textwidth]{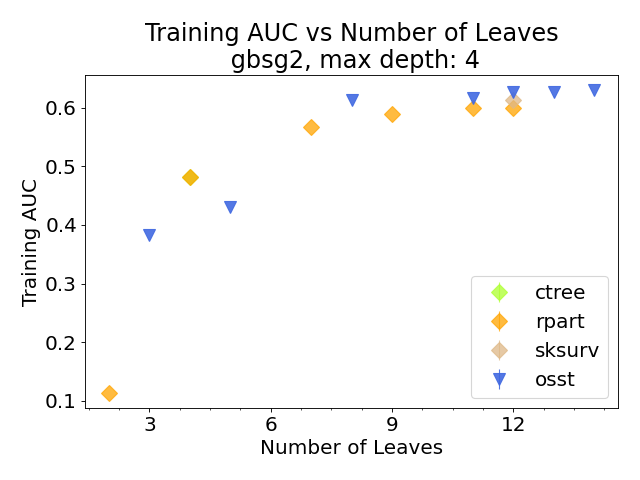}
    \includegraphics[width=0.42\textwidth]{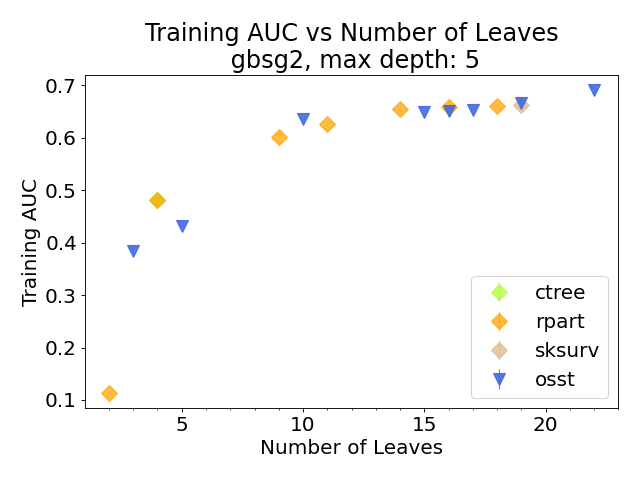}
    \includegraphics[width=0.42\textwidth]{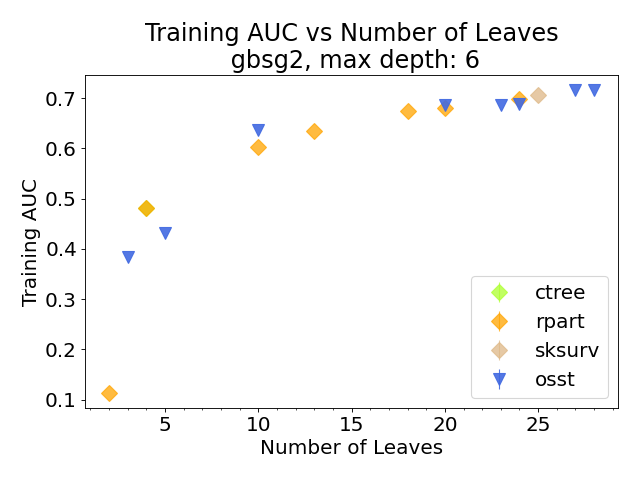}
    \includegraphics[width=0.42\textwidth]{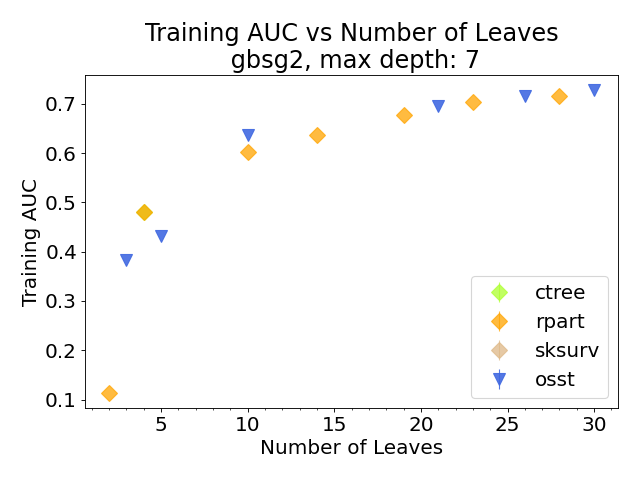}
    \includegraphics[width=0.42\textwidth]{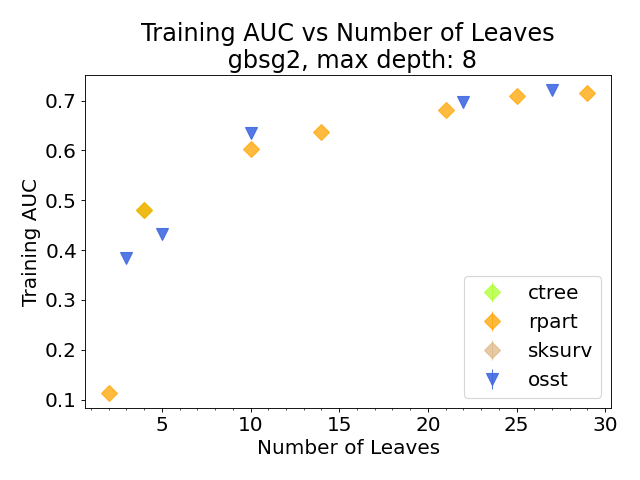}
    \includegraphics[width=0.42\textwidth]{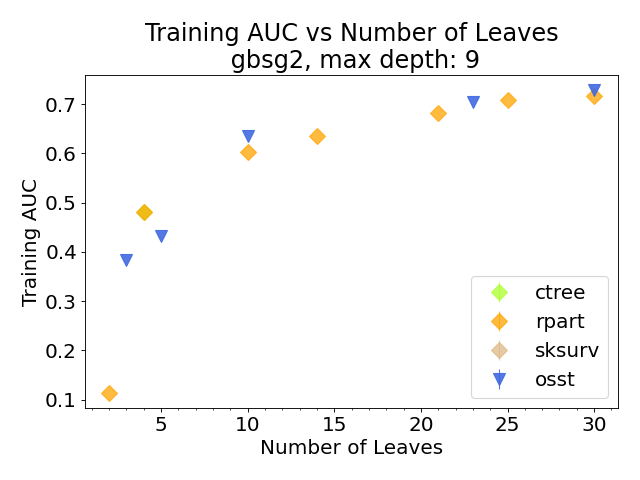}
    \caption{Training auc achieved by CTree, RPART, SkSurv and OSST as a function of number of leaves on dataset: gbsg2.}
    \label{fig:lvs:gbsg2-auc}
\end{figure*}

\begin{figure*}[htbp]
    \centering
    \includegraphics[width=0.42\textwidth]{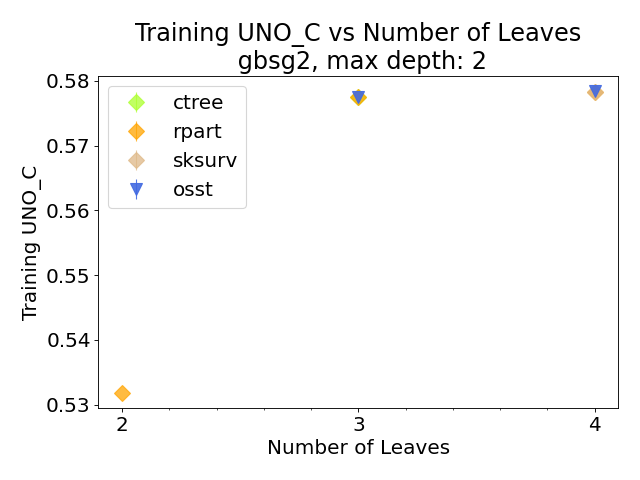}
    \includegraphics[width=0.42\textwidth]{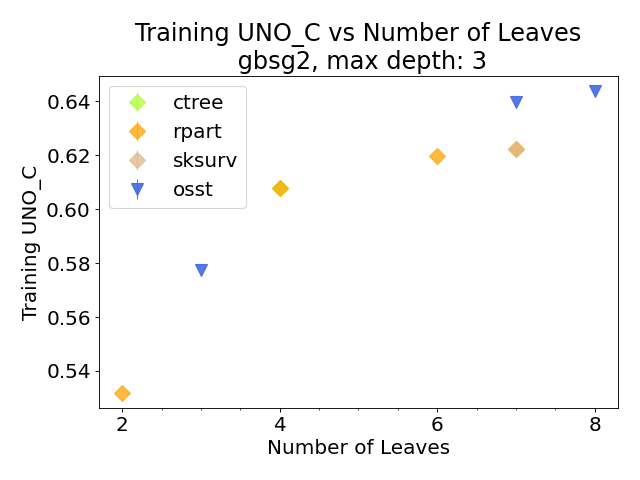}
    \includegraphics[width=0.42\textwidth]{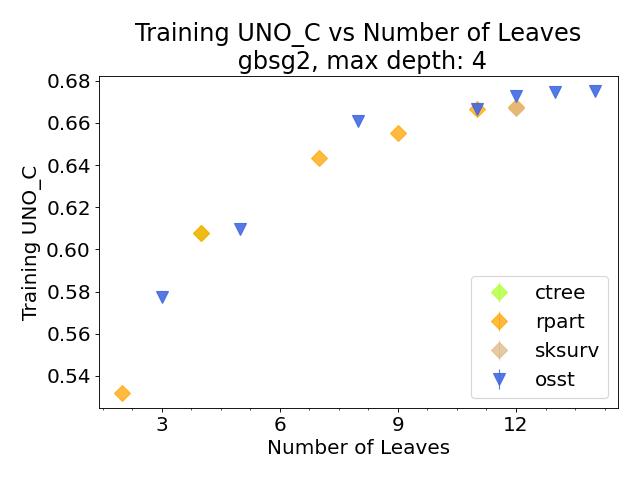}
    \includegraphics[width=0.42\textwidth]{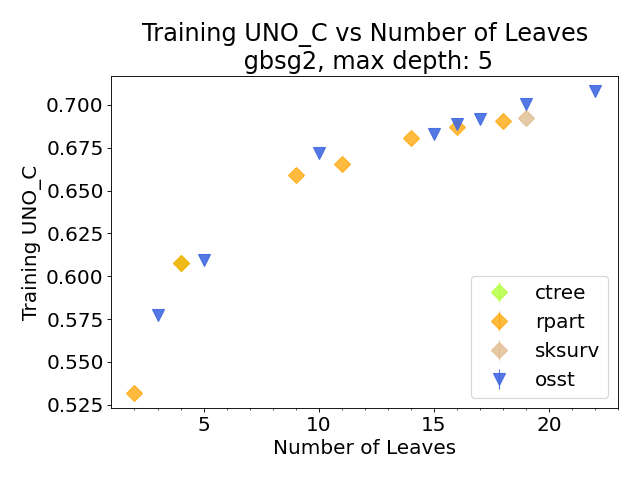}
    \includegraphics[width=0.42\textwidth]{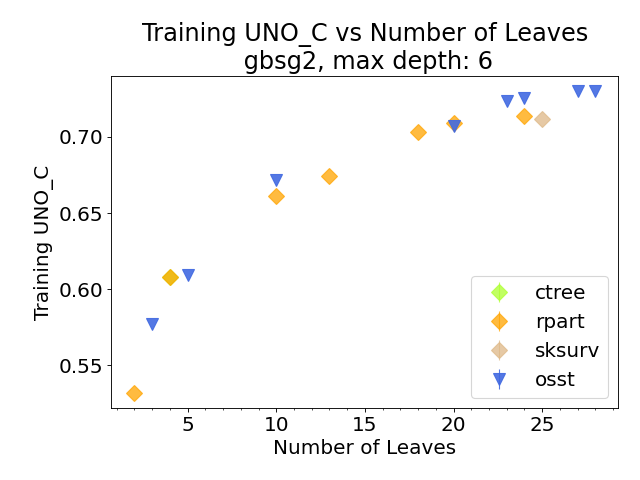}
    \includegraphics[width=0.42\textwidth]{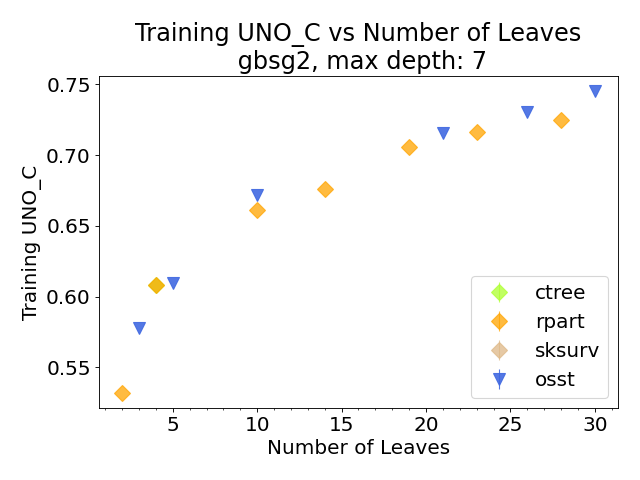}
    \includegraphics[width=0.42\textwidth]{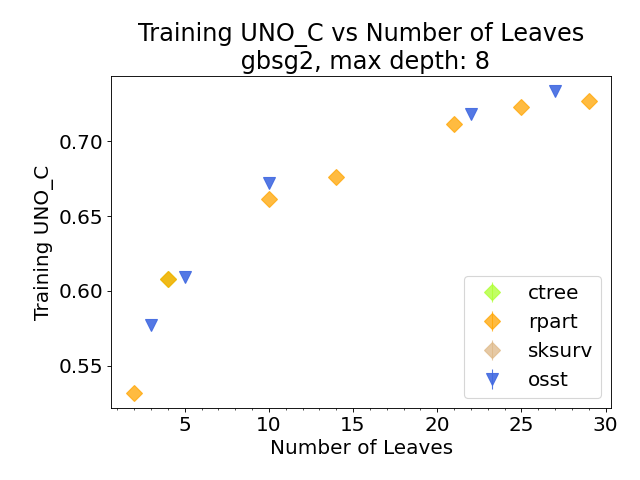}
    \includegraphics[width=0.42\textwidth]{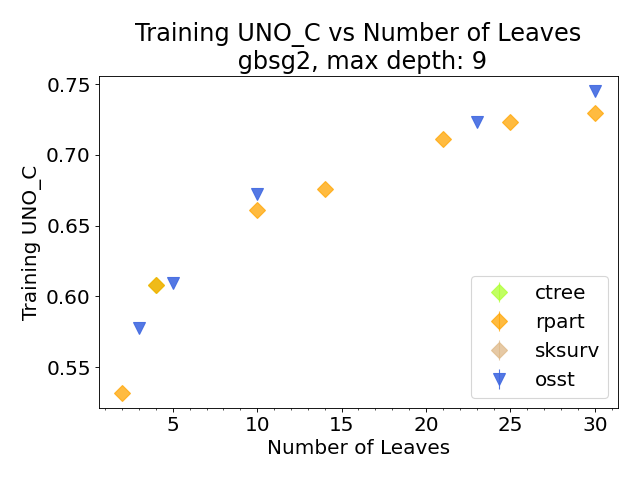}
    \caption{Training uno\_c achieved by CTree, RPART, SkSurv and OSST as a function of number of leaves on dataset: gbsg2.}
    \label{fig:lvs:gbsg2-uno_c}
\end{figure*}

\begin{figure*}[htbp]
    \centering
    \includegraphics[width=0.42\textwidth]{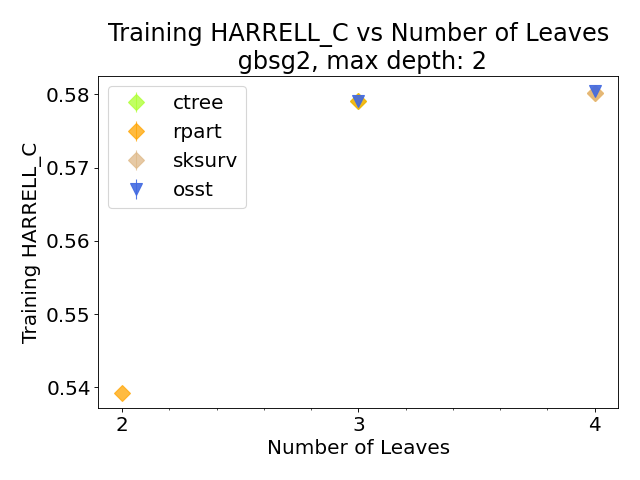}
    \includegraphics[width=0.42\textwidth]{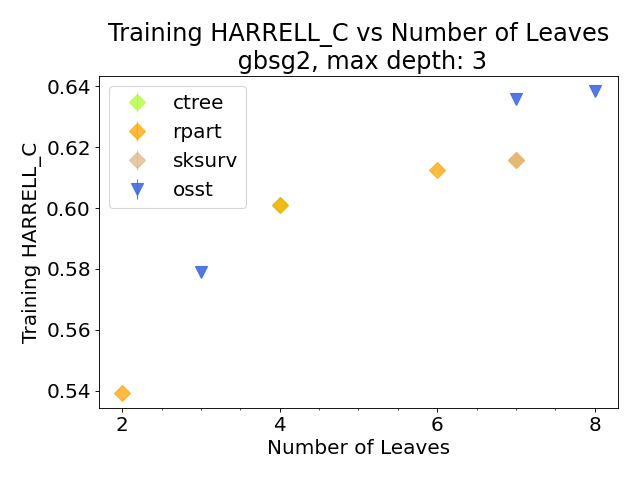}
    \includegraphics[width=0.42\textwidth]{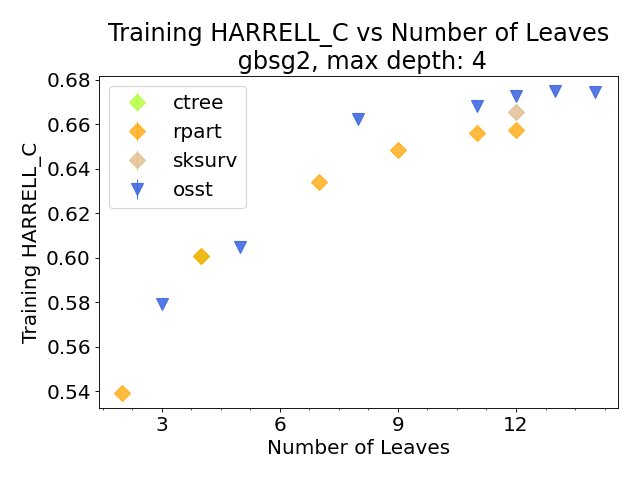}
    \includegraphics[width=0.42\textwidth]{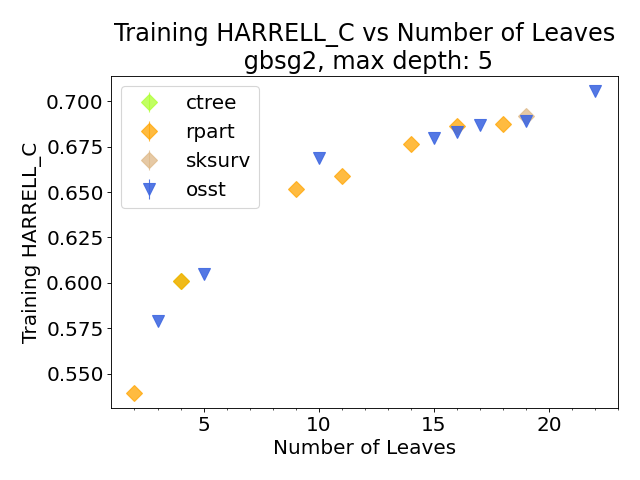}
    \includegraphics[width=0.42\textwidth]{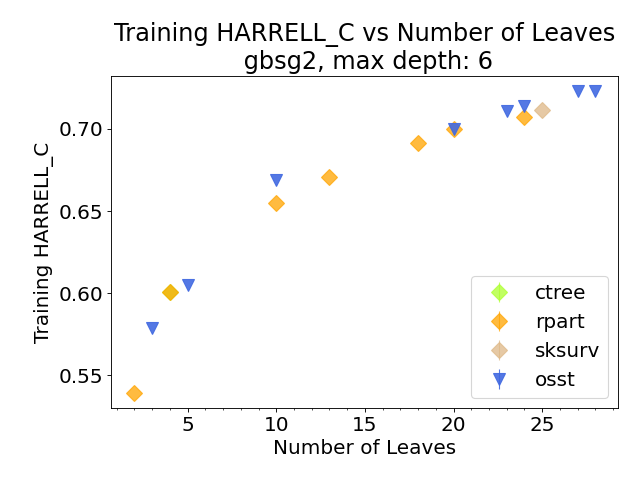}
    \includegraphics[width=0.42\textwidth]{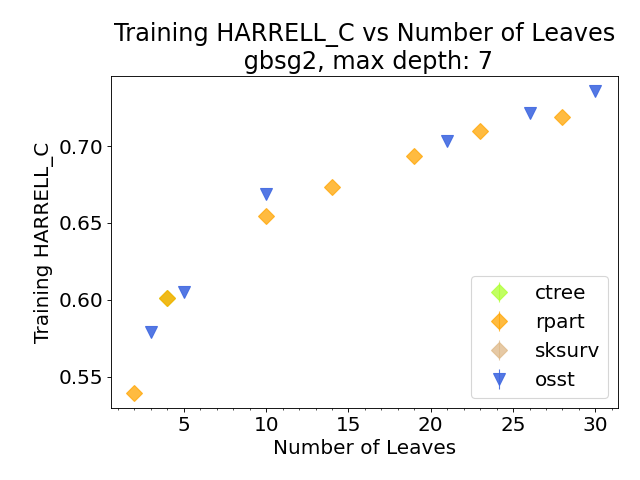}
    \includegraphics[width=0.42\textwidth]{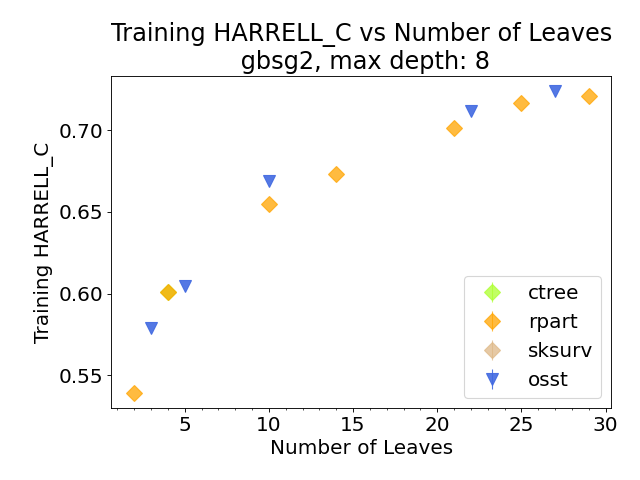}
    \includegraphics[width=0.42\textwidth]{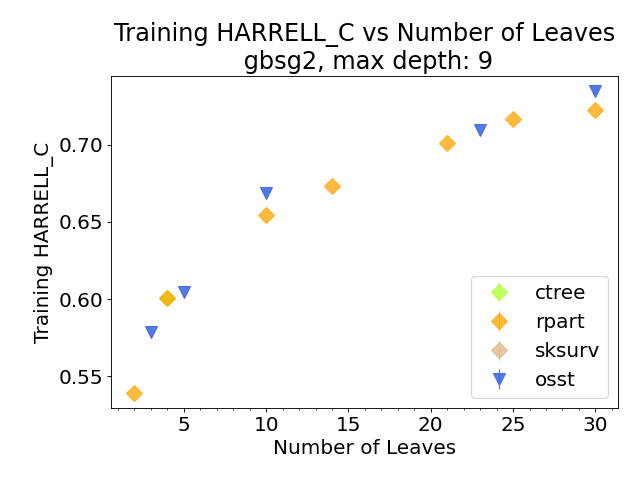}
    \caption{Training harrell\_c achieved by CTree, RPART, SkSurv and OSST as a function of number of leaves on dataset: gbsg2.}
    \label{fig:lvs:gbsg2-harrel_c}
\end{figure*}

\begin{figure*}[htbp]
    \centering
    \includegraphics[width=0.42\textwidth]{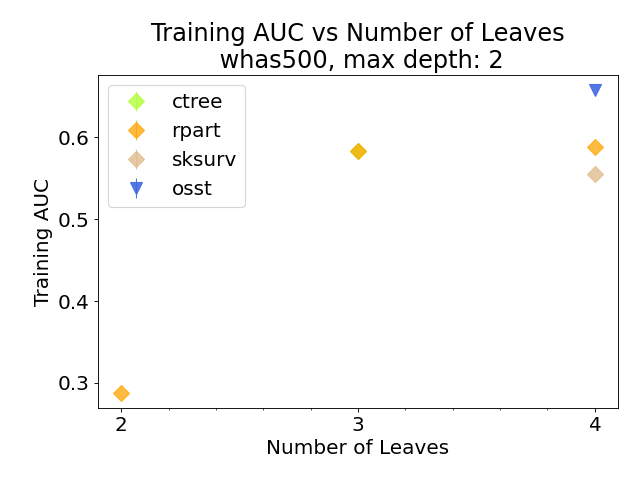}
    \includegraphics[width=0.42\textwidth]{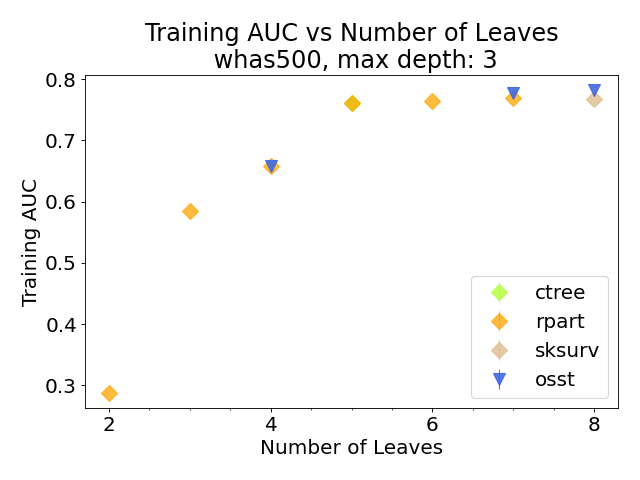}
    \includegraphics[width=0.42\textwidth]{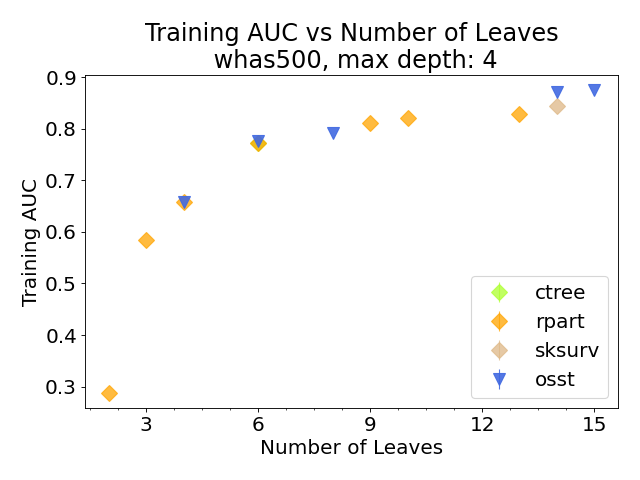}
    \includegraphics[width=0.42\textwidth]{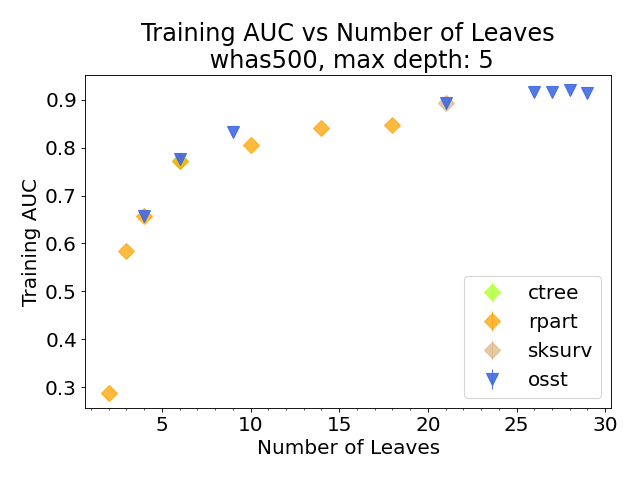}
    \includegraphics[width=0.42\textwidth]{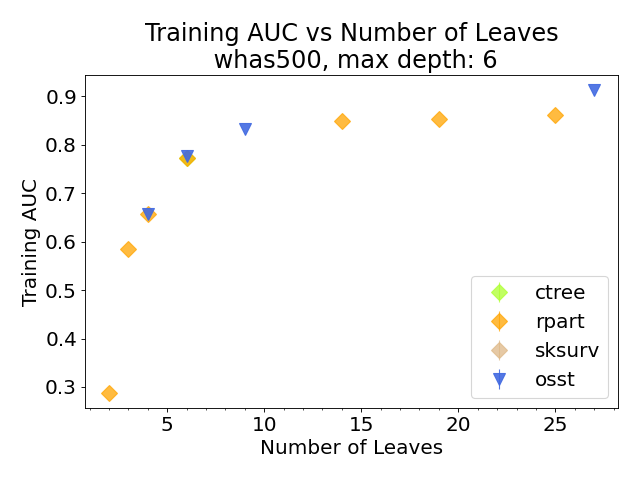}
    \includegraphics[width=0.42\textwidth]{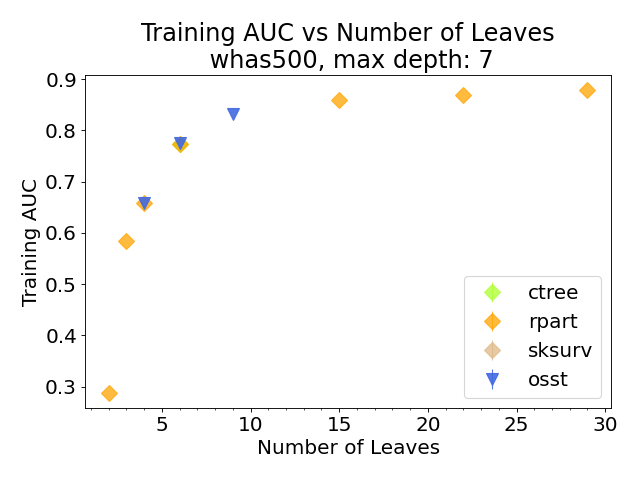}
    \includegraphics[width=0.42\textwidth]{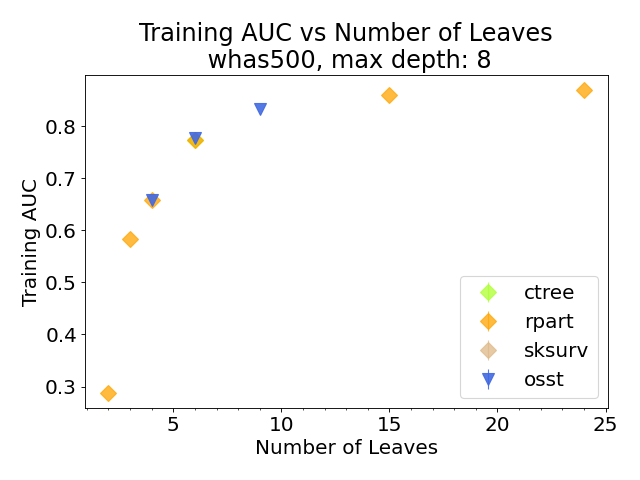}
    \includegraphics[width=0.42\textwidth]{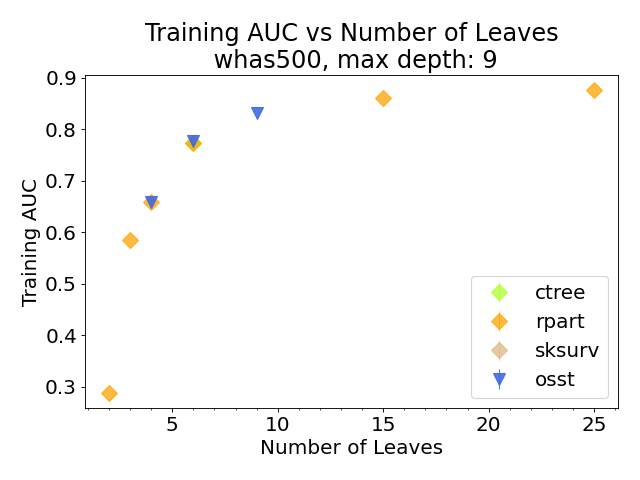}
    \caption{Training auc achieved by CTree, RPART, SkSurv and OSST as a function of number of leaves on dataset: whas500.}
    \label{fig:lvs:whas500-auc}
\end{figure*}

\begin{figure*}[htbp]
    \centering
    \includegraphics[width=0.42\textwidth]{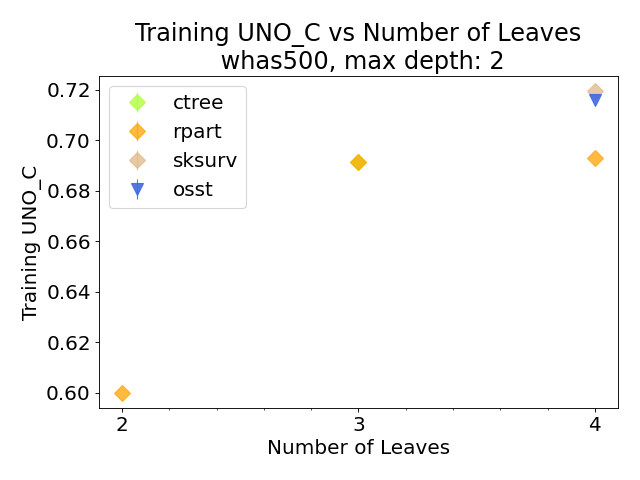}
    \includegraphics[width=0.42\textwidth]{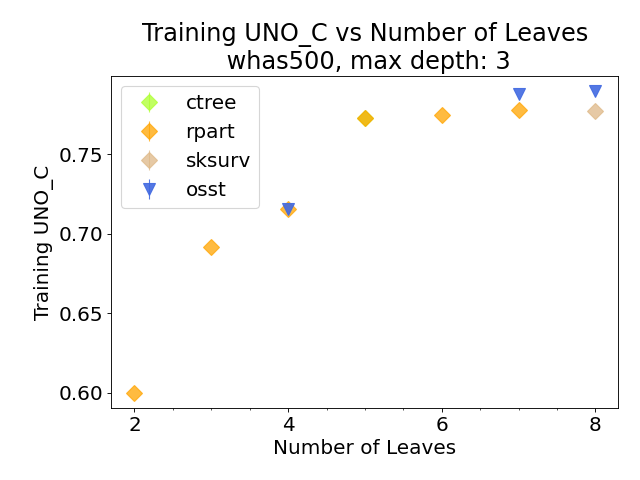}
    \includegraphics[width=0.42\textwidth]{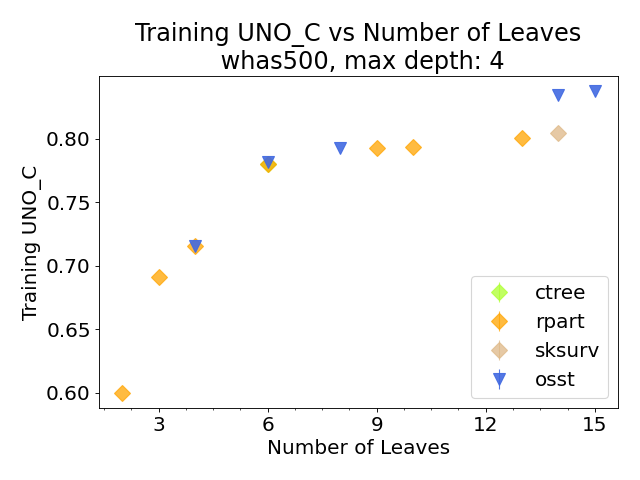}
    \includegraphics[width=0.42\textwidth]{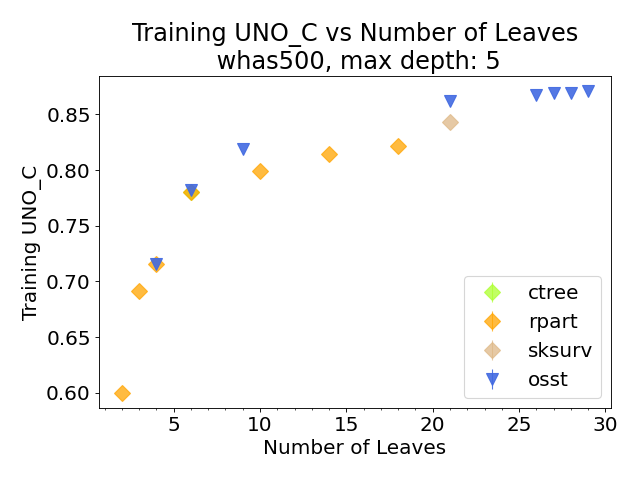}
    \includegraphics[width=0.42\textwidth]{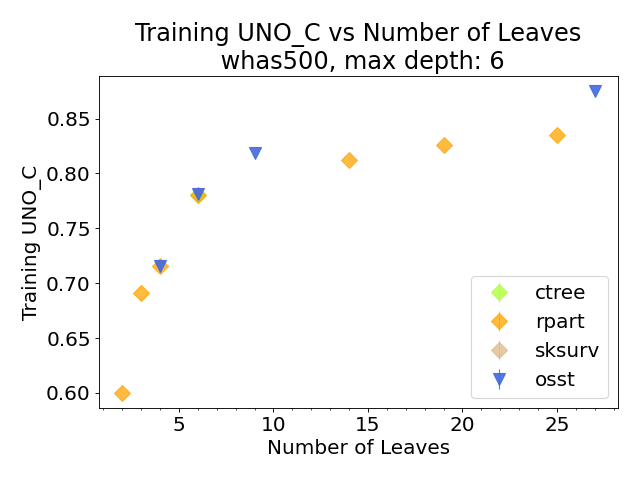}
    \includegraphics[width=0.42\textwidth]{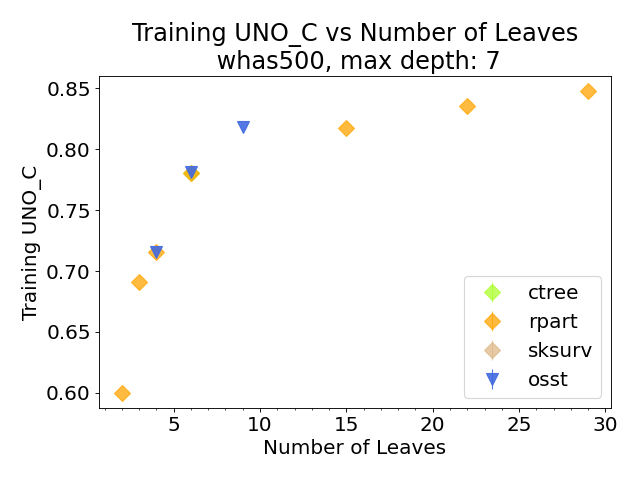}
    \includegraphics[width=0.42\textwidth]{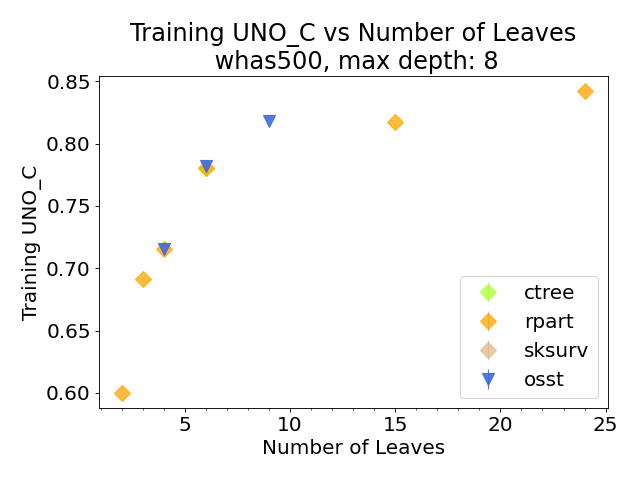}
    \includegraphics[width=0.42\textwidth]{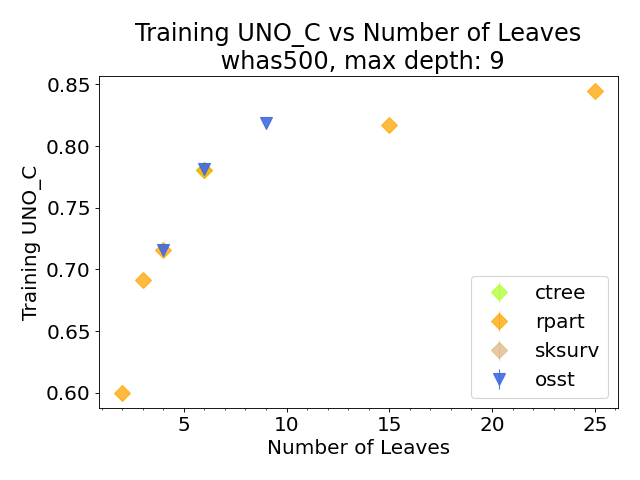}
    \caption{Training uno\_c achieved by CTree, RPART, SkSurv and OSST as a function of number of leaves on dataset: whas500.}
    \label{fig:lvs:whas500-uno_c}
\end{figure*}

\begin{figure*}[htbp]
    \centering
    \includegraphics[width=0.42\textwidth]{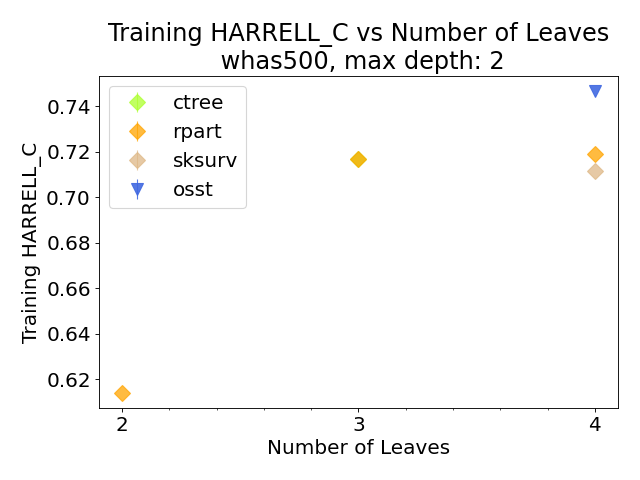}
    \includegraphics[width=0.42\textwidth]{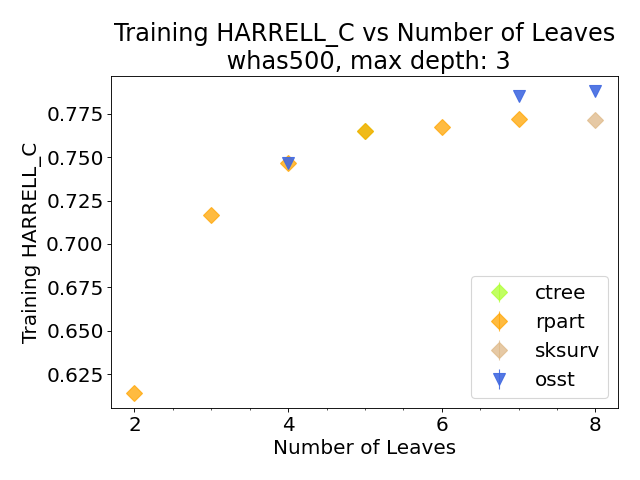}
    \includegraphics[width=0.42\textwidth]{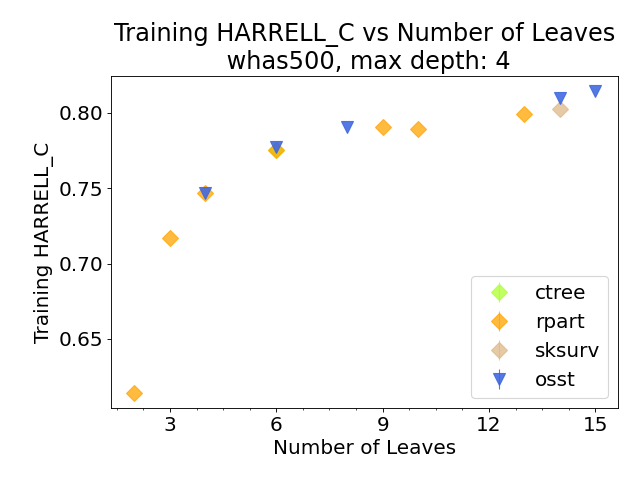}
    \includegraphics[width=0.42\textwidth]{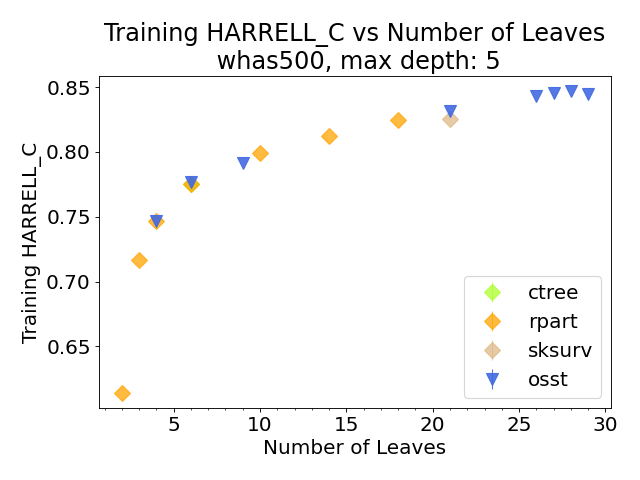}
    \includegraphics[width=0.42\textwidth]{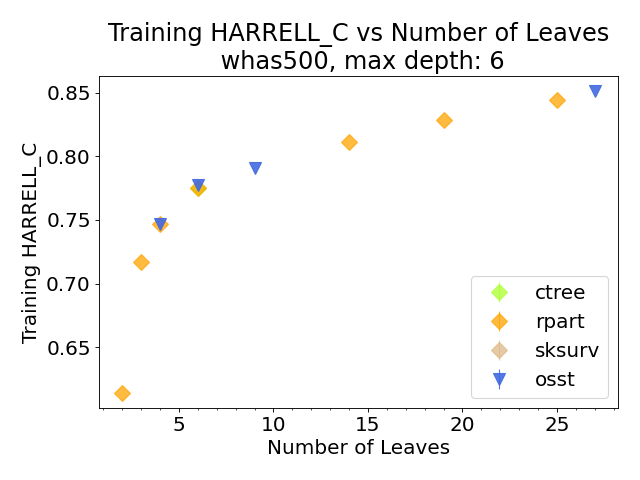}
    \includegraphics[width=0.42\textwidth]{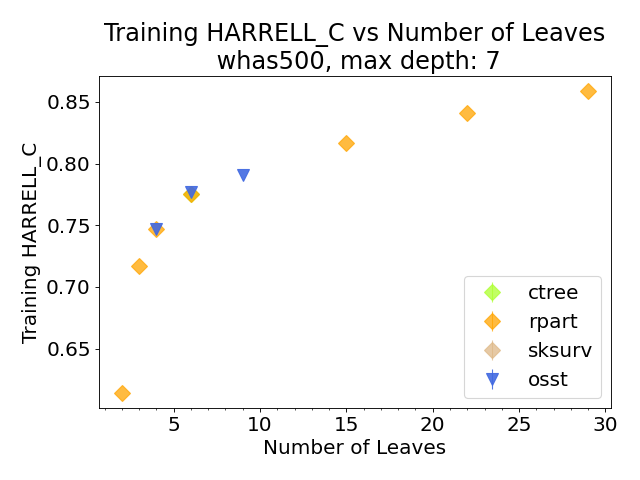}
    \includegraphics[width=0.42\textwidth]{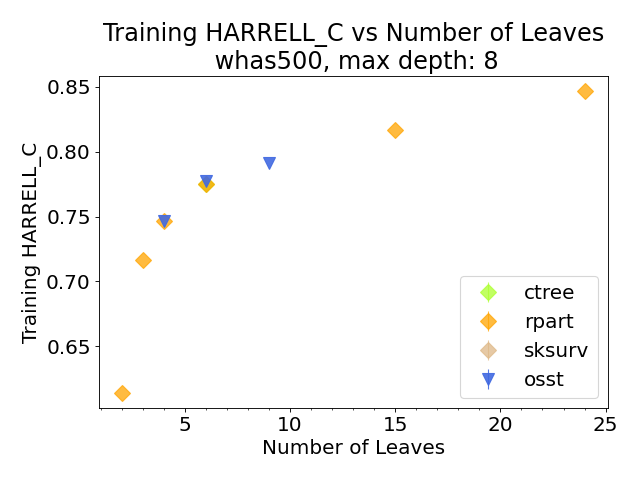}
    \includegraphics[width=0.42\textwidth]{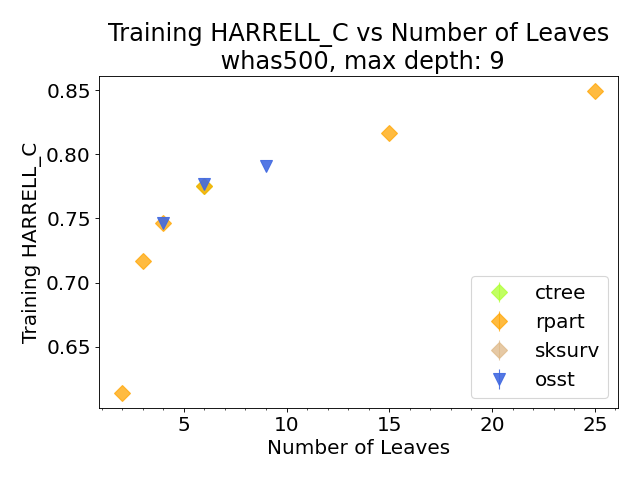}
    \caption{Training harrell\_c achieved by CTree, RPART, SkSurv and OSST as a function of number of leaves on dataset: whas500.}
    \label{fig:lvs:whas500-harrel_c}
\end{figure*}

\begin{figure*}
\centering
\includegraphics[width=0.42\textwidth]{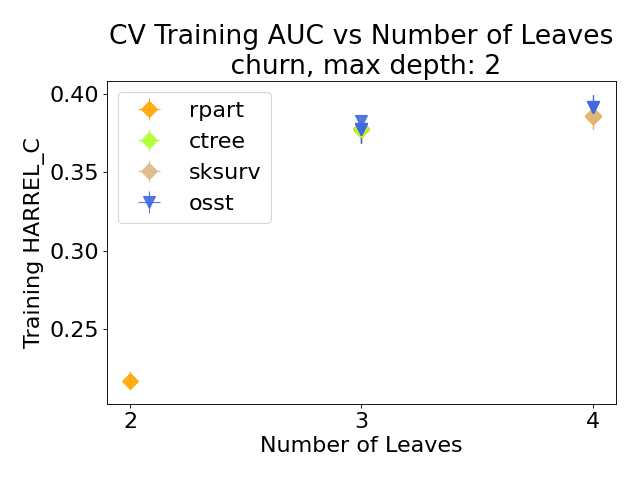}
\includegraphics[width=0.42\textwidth]{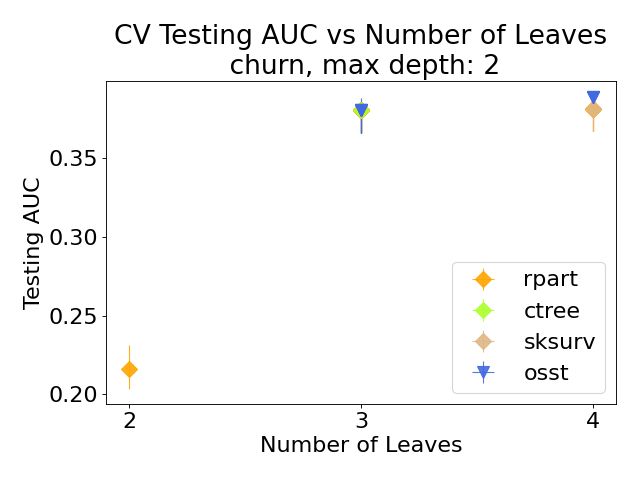}
\includegraphics[width=0.42\textwidth]{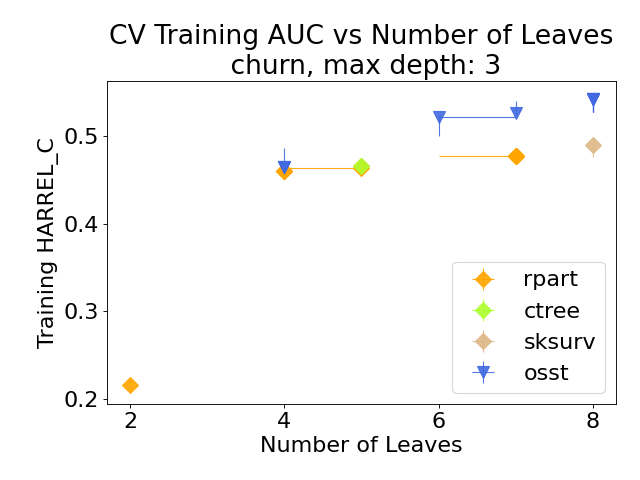}
\includegraphics[width=0.42\textwidth]{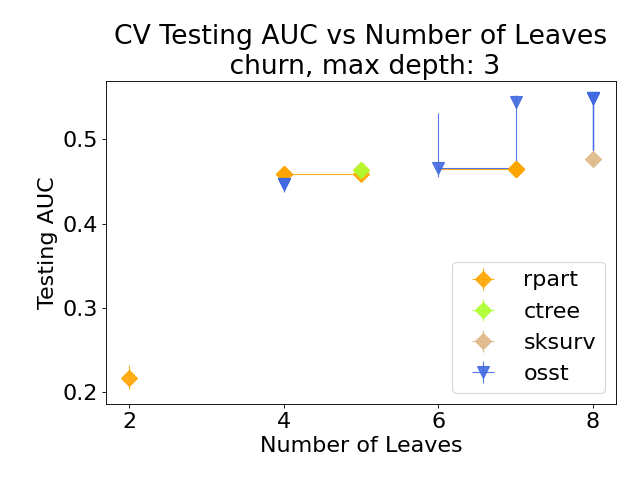}
\includegraphics[width=0.42\textwidth]{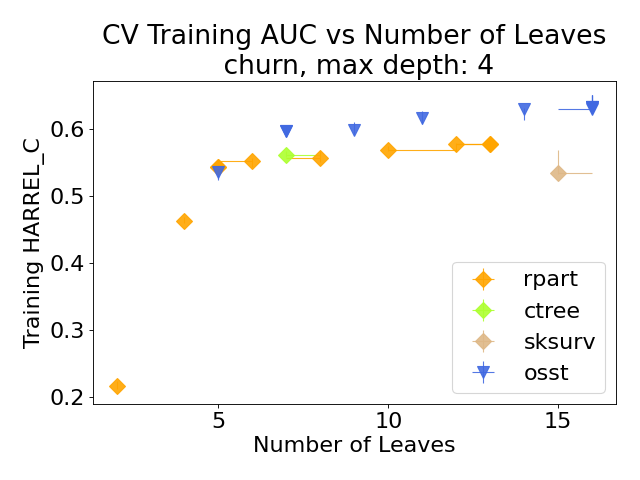}
\includegraphics[width=0.42\textwidth]{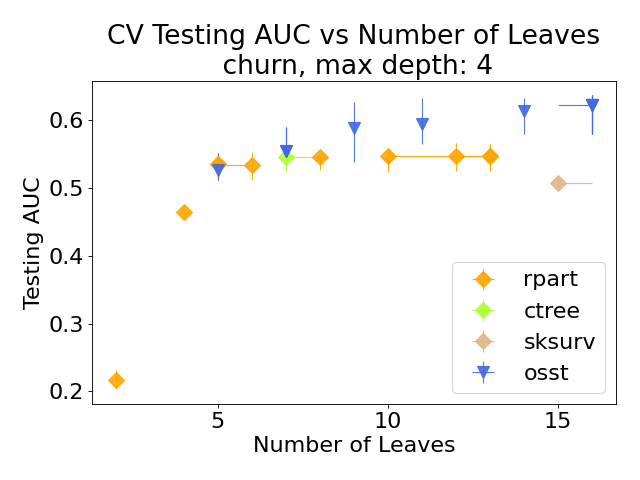}
\includegraphics[width=0.42\textwidth]{figures/cv/churn/auc/churn_depth5_auc_train.png}
\includegraphics[width=0.42\textwidth]{figures/cv/churn/auc/churn_depth5_auc_test.png}
\caption{5-fold CV of CTree, RPART, SkSurv and OSST as a function of number of leaves on dataset: churn, metric: auc.}
\label{fig:cv:churn-auc}
\end{figure*}

\begin{figure*}
\centering
\includegraphics[width=0.42\textwidth]{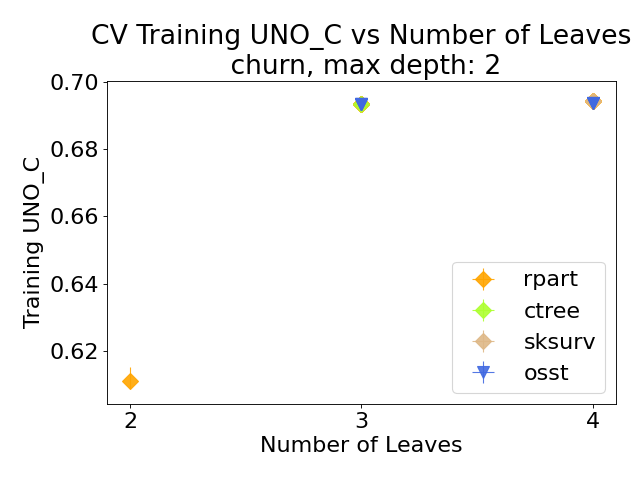}
\includegraphics[width=0.42\textwidth]{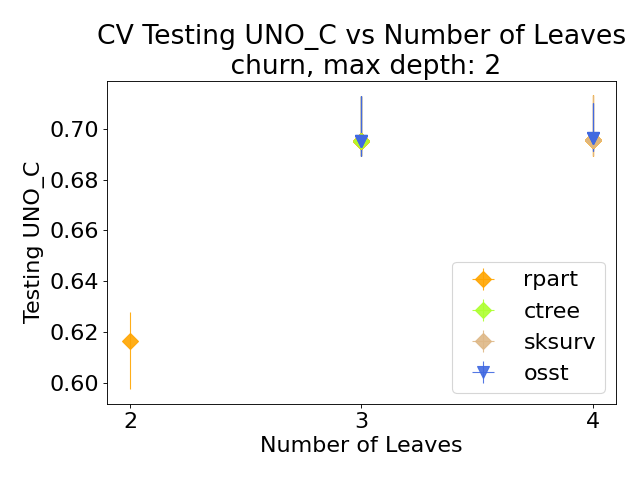}
\includegraphics[width=0.42\textwidth]{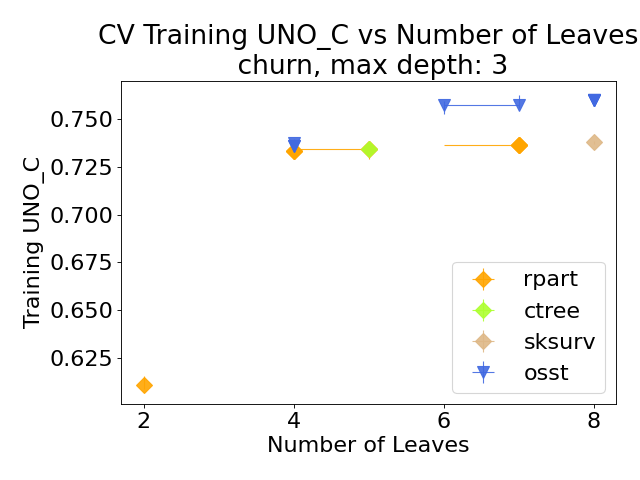}
\includegraphics[width=0.42\textwidth]{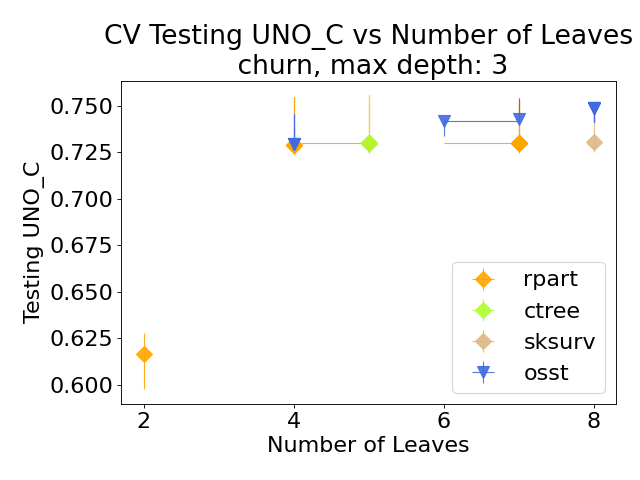}
\includegraphics[width=0.42\textwidth]{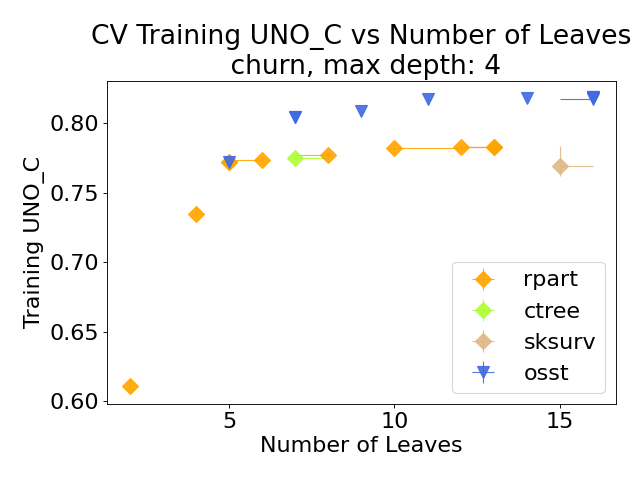}
\includegraphics[width=0.42\textwidth]{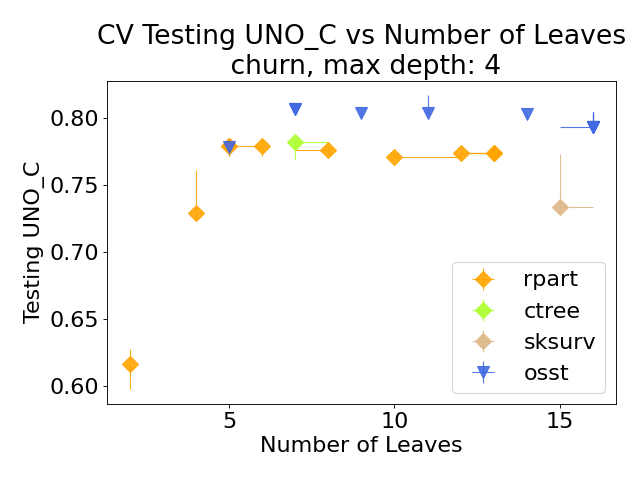}
\includegraphics[width=0.42\textwidth]{figures/cv/churn/uno_c/churn_depth5_uno_c_train.png}
\includegraphics[width=0.42\textwidth]{figures/cv/churn/uno_c/churn_depth5_uno_c_test.png}
\caption{5-fold CV of CTree, RPART, SkSurv and OSST as a function of number of leaves on dataset: churn, metric:uno\_c.}
\label{fig:cv:churn-uno_c}
\end{figure*}

\begin{figure*}
\centering
\includegraphics[width=0.42\textwidth]{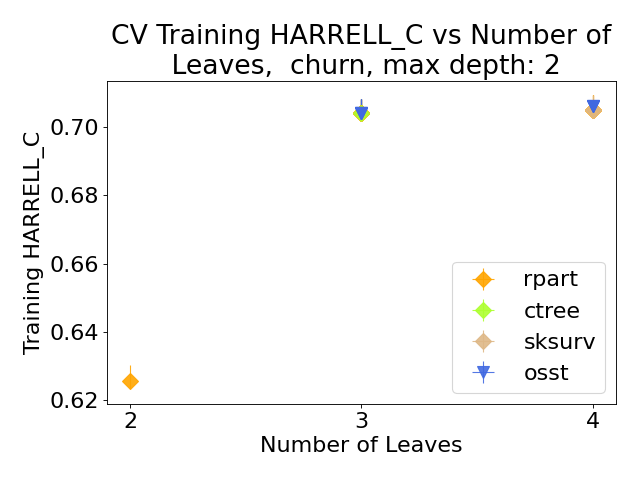}
\includegraphics[width=0.42\textwidth]{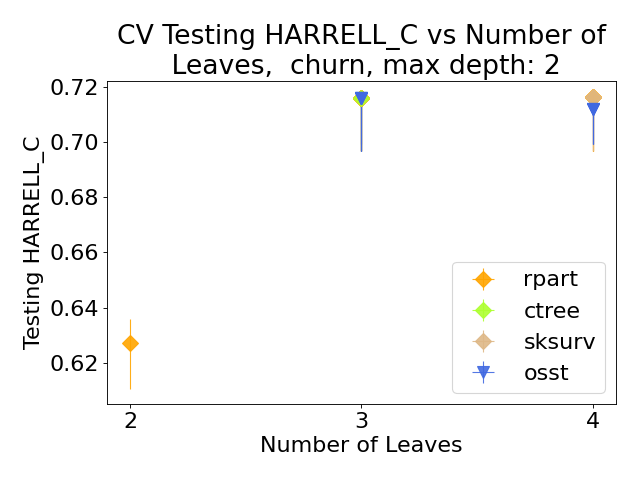}
\includegraphics[width=0.42\textwidth]{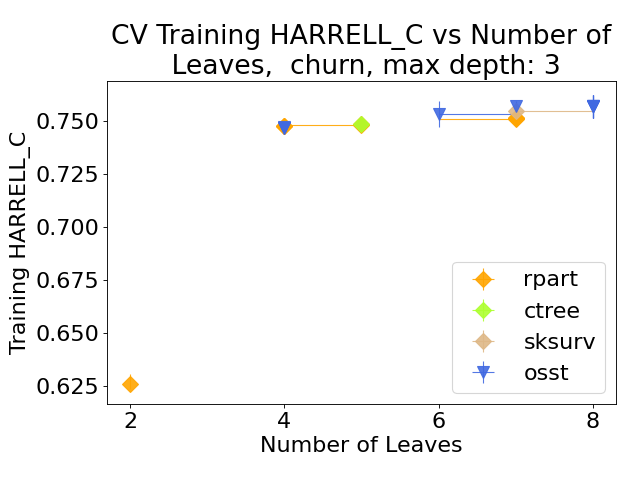}
\includegraphics[width=0.42\textwidth]{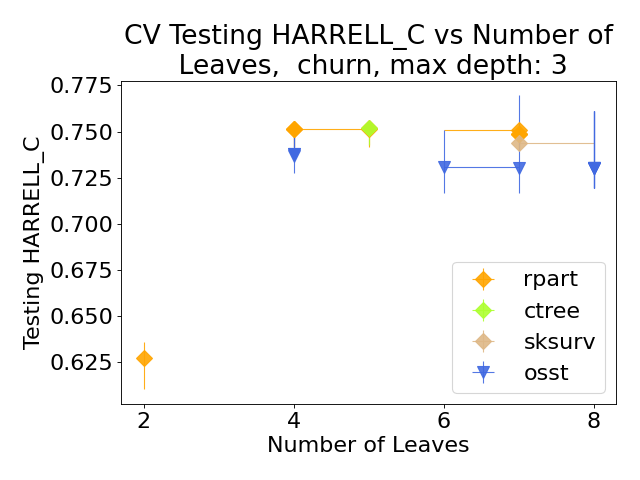}
\includegraphics[width=0.42\textwidth]{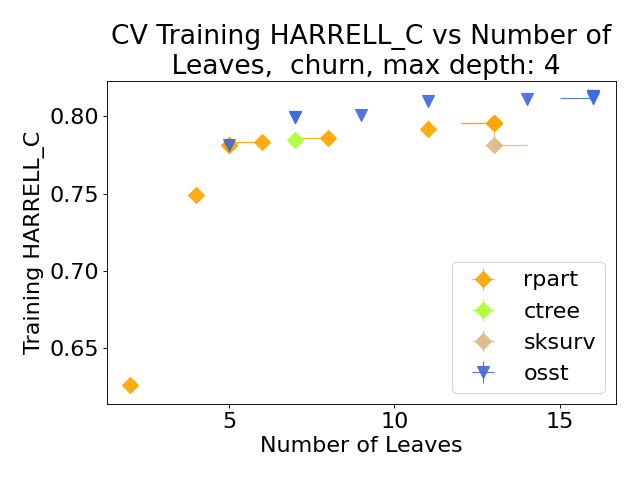}
\includegraphics[width=0.42\textwidth]{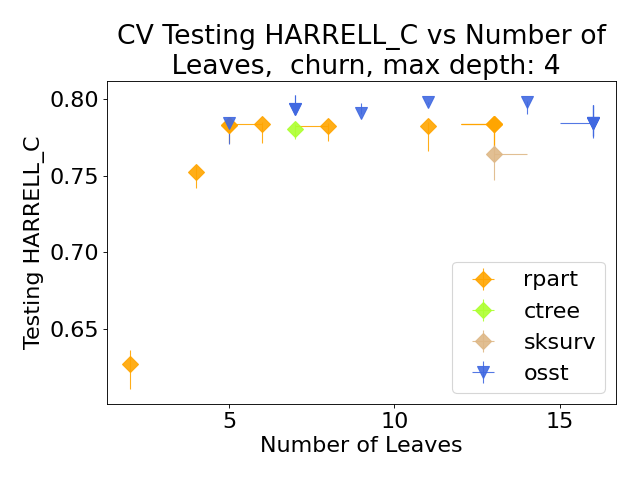}
\includegraphics[width=0.42\textwidth]{figures/cv/churn/harrel_c/churn_depth5_harrel_c_train.png}
\includegraphics[width=0.42\textwidth]{figures/cv/churn/harrel_c/churn_depth5_harrel_c_test.png}
\caption{5-fold CV of CTree, RPART, SkSurv and OSST as a function of number of leaves on dataset: churn, metric: harrell\_c.}
\label{fig:cv:churn-harrel_c}
\end{figure*}

\begin{figure*}
\centering
\includegraphics[width=0.42\textwidth]{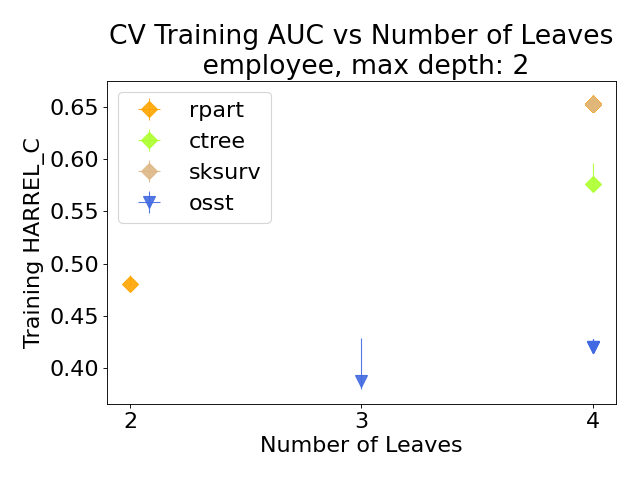}
\includegraphics[width=0.42\textwidth]{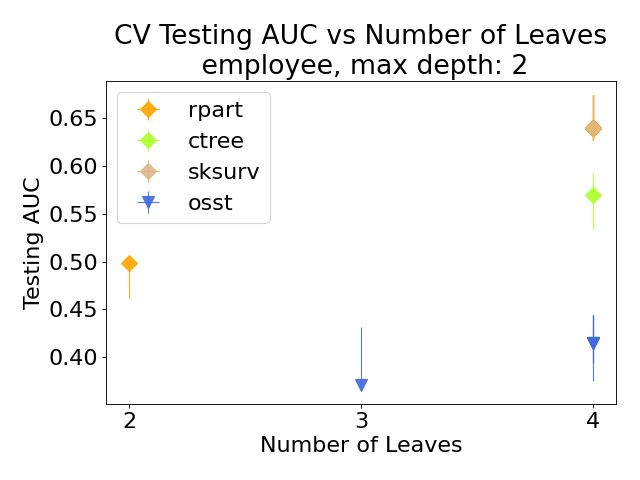}
\includegraphics[width=0.42\textwidth]{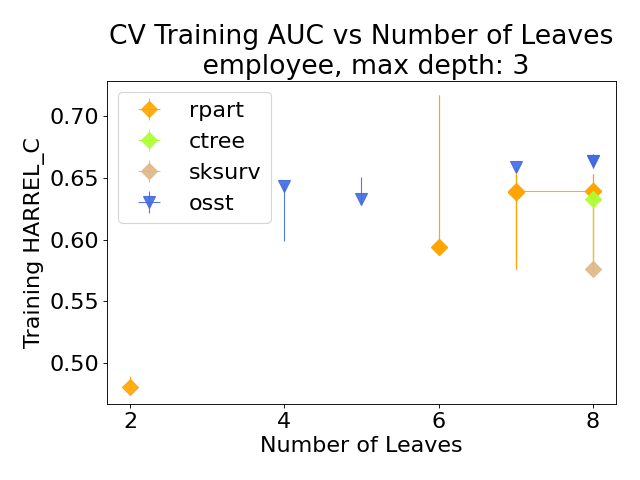}
\includegraphics[width=0.42\textwidth]{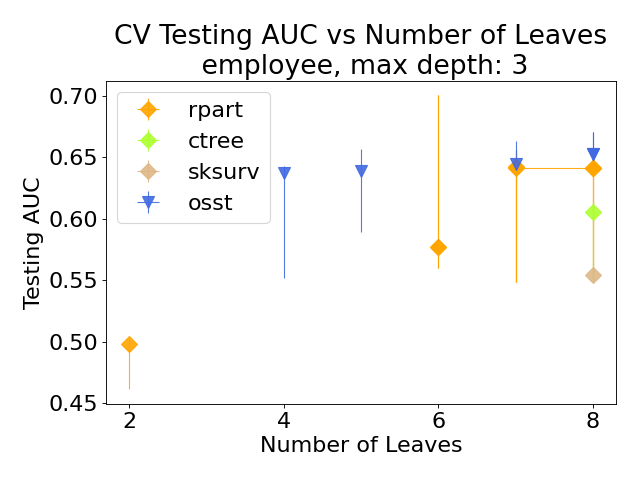}
\includegraphics[width=0.42\textwidth]{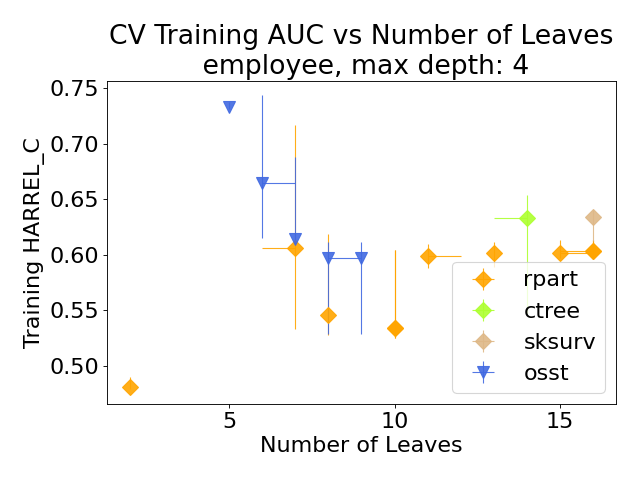}
\includegraphics[width=0.42\textwidth]{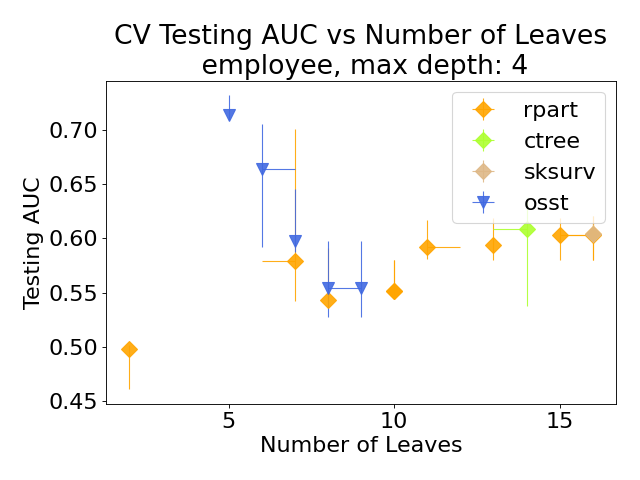}
\includegraphics[width=0.42\textwidth]{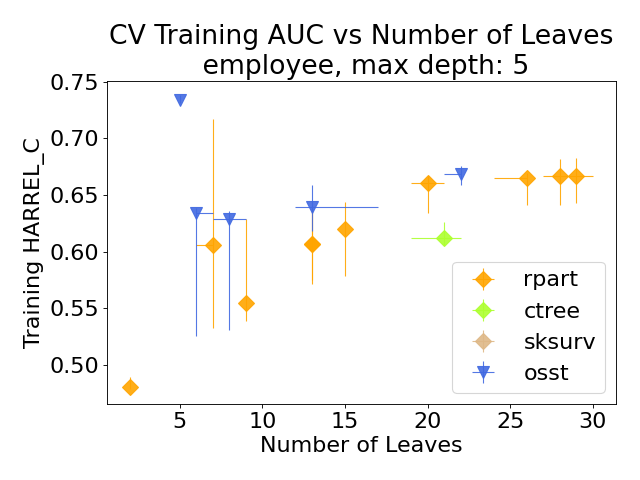}
\includegraphics[width=0.42\textwidth]{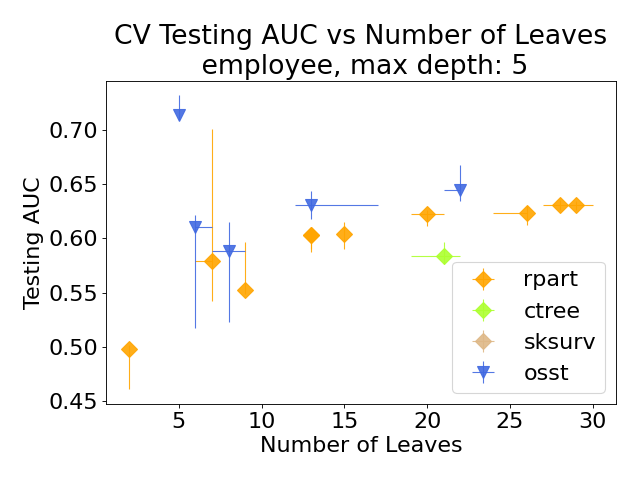}
\caption{5-fold CV of CTree, RPART, SkSurv and OSST as a function of number of leaves on dataset: employee, metric: auc.}
\label{fig:cv:employee-auc}
\end{figure*}

\begin{figure*}
\centering
\includegraphics[width=0.42\textwidth]{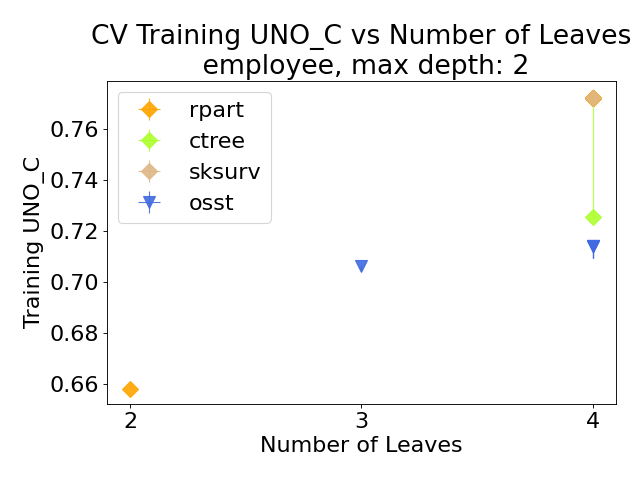}
\includegraphics[width=0.42\textwidth]{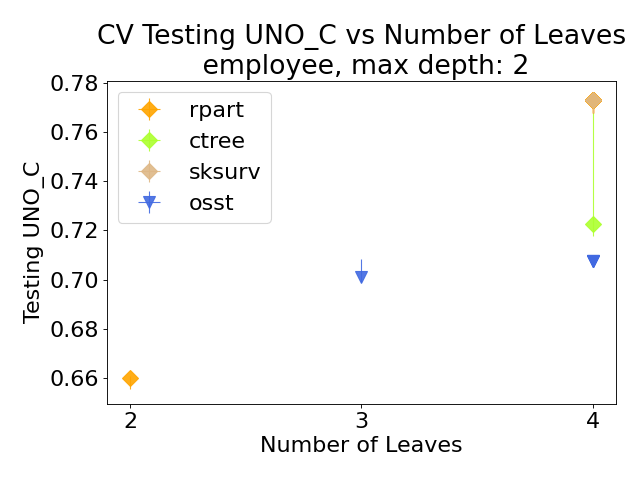}
\includegraphics[width=0.42\textwidth]{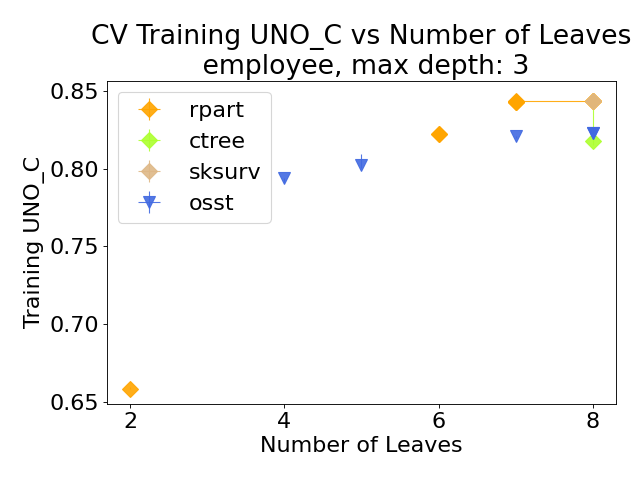}
\includegraphics[width=0.42\textwidth]{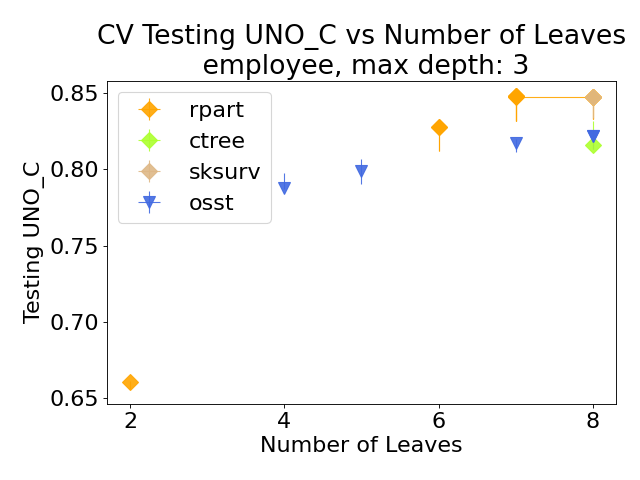}
\includegraphics[width=0.42\textwidth]{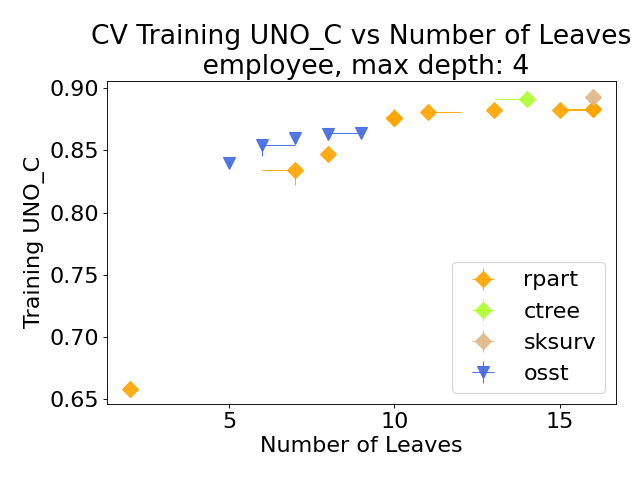}
\includegraphics[width=0.42\textwidth]{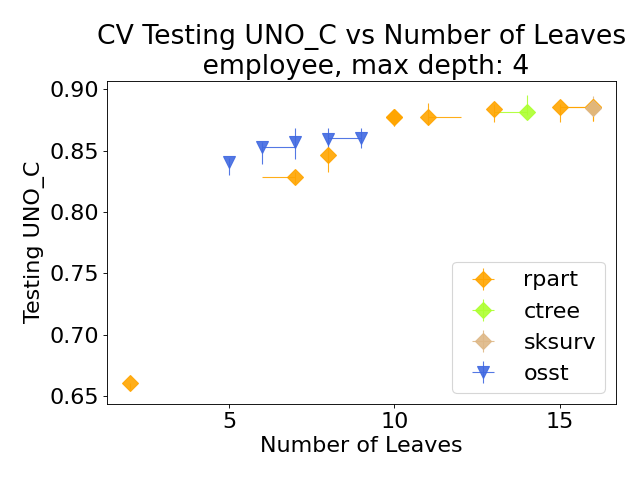}
\includegraphics[width=0.42\textwidth]{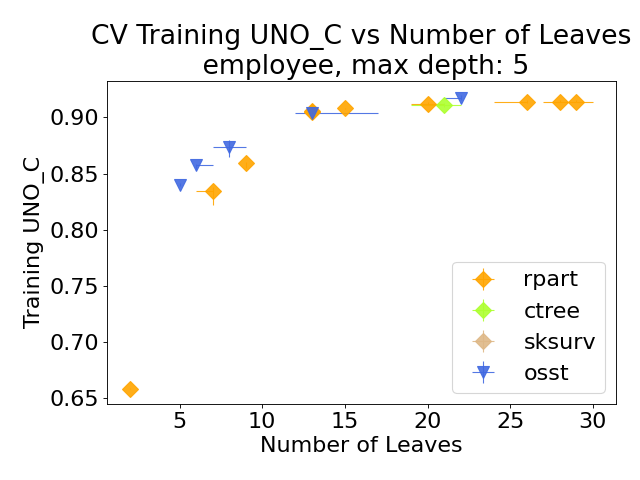}
\includegraphics[width=0.42\textwidth]{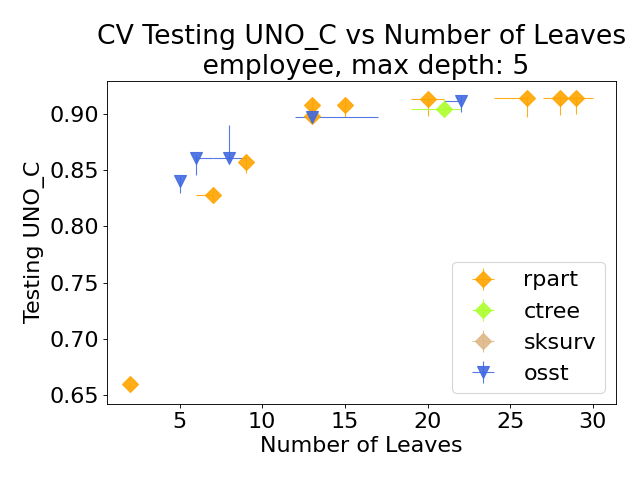}
\caption{5-fold CV of CTree, RPART, SkSurv and OSST as a function of number of leaves on dataset: employee, metric: uno\_c.}
\label{fig:cv:employee-uno_c}
\end{figure*}

\begin{figure*}
\centering
\includegraphics[width=0.42\textwidth]{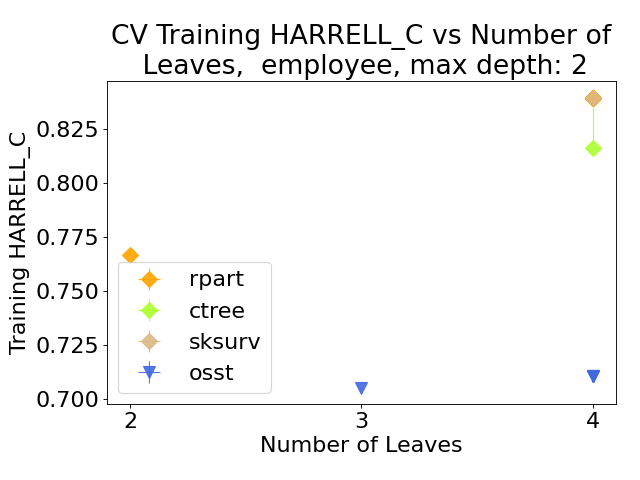}
\includegraphics[width=0.42\textwidth]{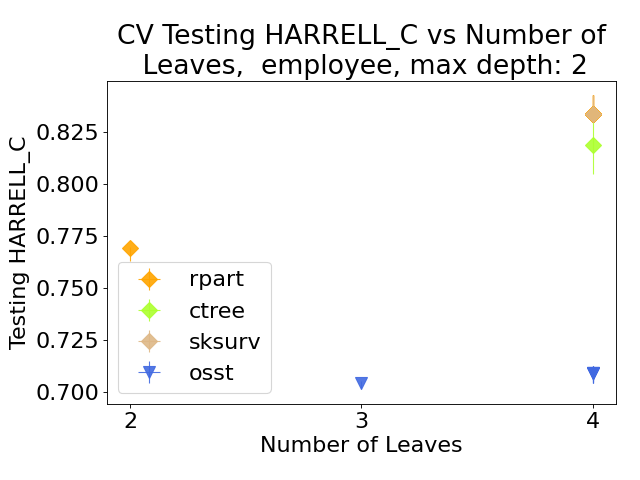}
\includegraphics[width=0.42\textwidth]{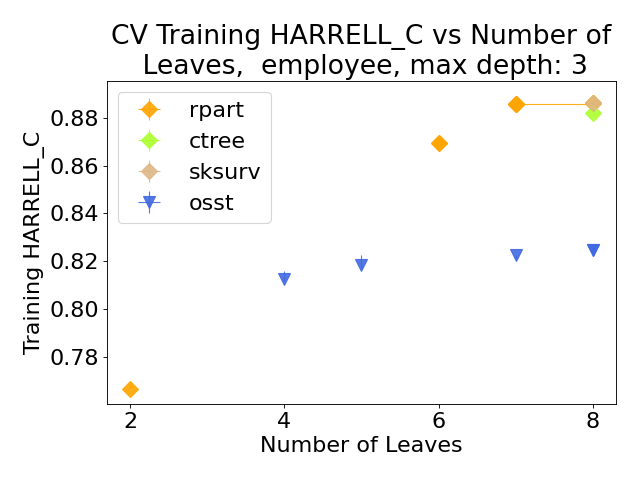}
\includegraphics[width=0.42\textwidth]{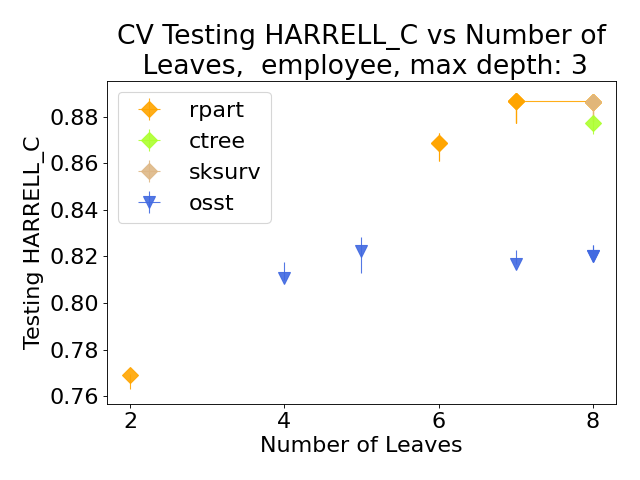}
\includegraphics[width=0.42\textwidth]{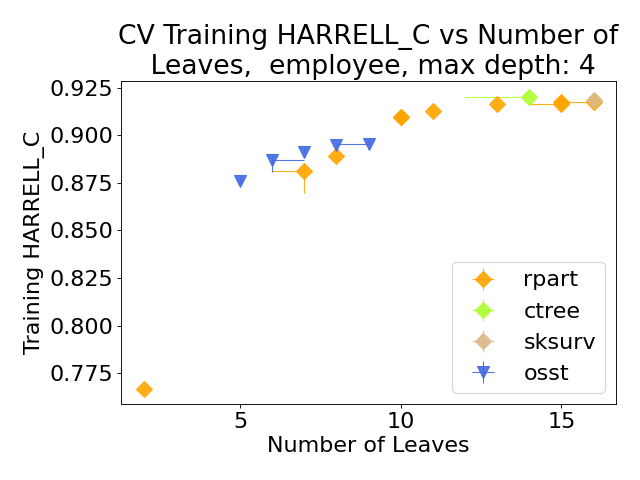}
\includegraphics[width=0.42\textwidth]{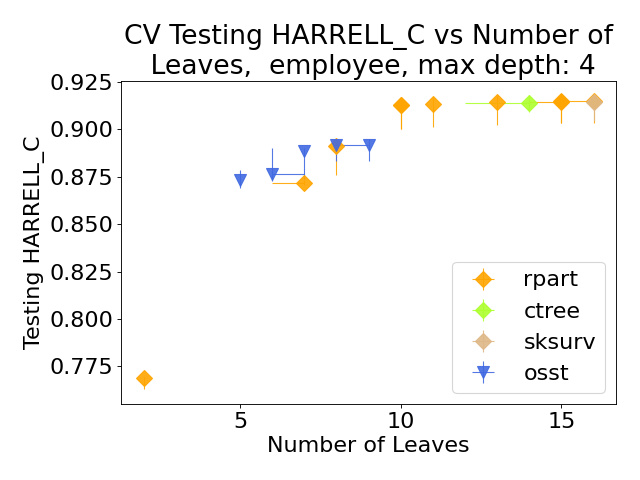}
\includegraphics[width=0.42\textwidth]{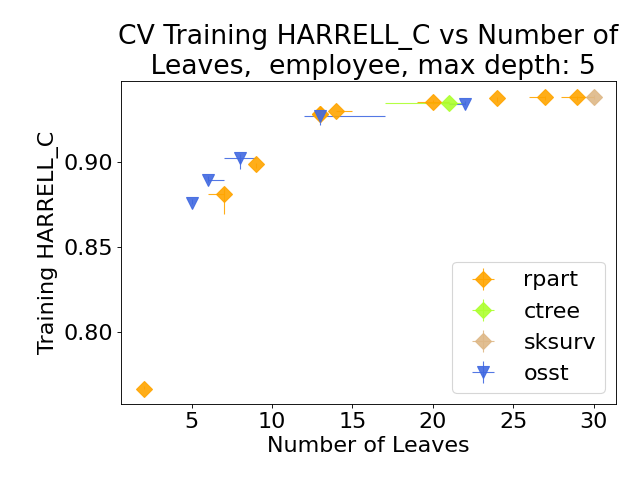}
\includegraphics[width=0.42\textwidth]{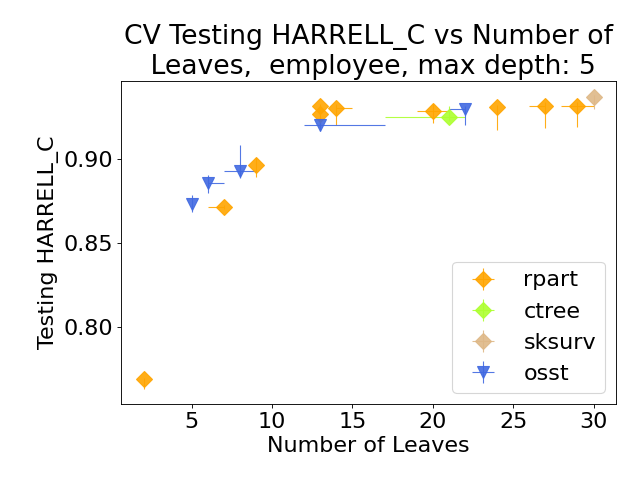}
\caption{5-fold CV of CTree, RPART, SkSurv and OSST as a function of number of leaves on dataset: employee, metric: harrell\_c.}
\label{fig:cv:employee-harrel_c}
\end{figure*}

\noindent \textbf{Results:}
Figure \ref{fig:lvs:gbsg2-auc} to \ref{fig:cv:employee-harrel_c} show that OSST generally obtains better training and testing performance of other metrics as well, even though it is optimized on the IBS Loss. This indicates that the sparse optimal trees found by OSST \textit{have better overall quality} over trees returned by other methods. In other word, OSST is better in capturing the patterns in survival data than other methods.




\newpage 
\newpage 
\section{Experiment: Running Time}\label{exp:time}
\noindent \textbf{Collection and Setup:} The experiments in Section \ref{exp:lvs} and \ref{exp:cv} suggest that configurations with large depth limit (e.g., $d = 7, 8, 9$) and small regularization coefficient often result in very large trees (with more than 40 leaves), which lose generalization and interpretability. We compared the running time of different methods on each dataset from depth 2 to 6, with the time limit of 60 minutes, using following configurations:
\begin{itemize}[leftmargin=10pt]
    \item \textbf{CTree}: We ran this algorithm with 5 different configurations: depth limit, $d$, ranging from 2 to 6, and a corresponding maximum leaf limit $2^d$. All other parameters were set to the default.
    \item \textbf{SkSurv}: We ran this algorithm with 5 different configurations: depth limit, $d$, ranging from 2 to 6, and a corresponding maximum leaf limit $2^d$. The random state was set to 2023 and all other parameters were set to the default.
    \item \textbf{RPART}: We ran this algorithm with $5 \times 12$ different configurations: depth limits ranging from 2 to 6, and 12 different regularization coefficients (0.1, 0.05, 0.025, 0.01, 0.005, 0.0025, 0.001, 0.0005, 0.00025, 0.0001, 0.00005, 0.00001). All other parameters were set to the default.
    \item \textbf{OSST} (our method): We ran this algorithm with $5 \times 12$ different configurations: depth limits ranging from 2 to 6, and 12 different regularization coefficients (0.1, 0.05, 0.01, 0.005, 0.0025, 0.001, 0.0005, 0.00025, 0.0001, 0.00005, 0.000025, 0.00001). We used a random survival forest that consists of 100 trees with maximum depth 9 (in order to preserve high quality) as a reference model on datasets: \textit{airfoil, insurance, real-estate, sync, servo, flchain} for all depths, \textit{churn, credit, employee} for depths 5-6. No reference model was used on  other datasets.
\end{itemize}

\noindent \textbf{Results:} Tables \ref{tb:time1} to \ref{tb:time3} summarize the average running time and its 95\% confidence interval for different methods. \textit{OSST often finds the optimal survival tree in seconds}. Greedy methods are generally faster, but fail to produce an optimal tree. Interestingly, we found that the running speed of OSST on real-world survival data is generally faster than on synthetic data if no guessing technique was applied.
\begin{table}[htbp]
\centering
\begin{tabularx}{\textwidth}{|c|c|X|X|X|X|}
\hline
\textbf{Dataset} & \textbf{Depth} & \textbf{CTree} & \textbf{RPART} & \textbf{SkSurv} & \textbf{OSST} \\ \hline
\multirow{5}{*}{aids} & 2 & $0.192 (\pm 0)$ & $0.033 (\pm 0.003)$ & $0.006 (\pm 0)$ & $0.011 (\pm 0.000)$ \\ \cline{2-6} 
& 3 & $0.201 (\pm 0)$ & $0.039 (\pm 0.005)$ & $0.007 (\pm 0)$ & $0.083 (\pm 0.003)$ \\ \cline{2-6} 
& 4 & $0.217 (\pm 0)$ & $0.045 (\pm 0.006)$ & $0.008 (\pm 0)$ & $0.457 (\pm 0.056)$ \\ \cline{2-6} 
& 5 & $0.214 (\pm 0)$ & $0.048 (\pm 0.007)$ & $0.009 (\pm 0)$ & $2.215 (\pm 0.533)$ \\ \cline{2-6} 
& 6 & $0.206 (\pm 0)$ & $0.052 (\pm 0.008)$ & $0.010 (\pm 0)$ & $8.650 (\pm 2.530)$ \\ \hline
\multirow{5}{*}{aids\_death} & 2 & $0.192 (\pm 0)$ & $0.033 (\pm 0.003)$ & $0.006 (\pm 0)$ & $0.011 (\pm 0.000)$ \\ \cline{2-6} 
& 3 & $0.201 (\pm 0)$ & $0.039 (\pm 0.005)$ & $0.007 (\pm 0)$ & $0.083 (\pm 0.003)$ \\ \cline{2-6} 
& 4 & $0.217 (\pm 0)$ & $0.045 (\pm 0.006)$ & $0.008 (\pm 0)$ & $0.457 (\pm 0.056)$ \\ \cline{2-6} 
& 5 & $0.214 (\pm 0)$ & $0.048 (\pm 0.007)$ & $0.009 (\pm 0)$ & $2.215 (\pm 0.533)$ \\ \cline{2-6} 
& 6 & $0.206 (\pm 0)$ & $0.052 (\pm 0.008)$ & $0.010 (\pm 0)$ & $8.650 (\pm 2.530)$ \\ \hline
\multirow{5}{*}{gbsg2} & 2 & $0.201 (\pm 0)$ & $0.023 (\pm 0.002)$ & $0.003 (\pm 0)$ & $0.003 (\pm 0.000)$ \\ \cline{2-6} 
& 3 & $0.209 (\pm 0)$ & $0.025 (\pm 0.003)$ & $0.004 (\pm 0)$ & $0.019 (\pm 0.001)$ \\ \cline{2-6} 
& 4 & $0.213 (\pm 0)$ & $0.027 (\pm 0.003)$ & $0.004 (\pm 0)$ & $0.078 (\pm 0.006)$ \\ \cline{2-6} 
& 5 & $0.225 (\pm 0)$ & $0.030 (\pm 0.004)$ & $0.005 (\pm 0)$ & $0.264 (\pm 0.026)$ \\ \cline{2-6} 
& 6 & $0.214 (\pm 0)$ & $0.031 (\pm 0.500)$ & $0.005 (\pm 0)$ & $0.667 (\pm 0.077)$ \\ \hline
\multirow{5}{*}{maintenance} & 2 & $0.188 (\pm 0)$ & $0.027 (\pm 0.004)$ & $0.003 (\pm 0)$ & $0.033 (\pm 0.000)$ \\ \cline{2-6} 
& 3 & $0.205 (\pm 0)$ & $0.031 (\pm 0.005)$ & $0.004 (\pm 0)$ & $0.141 (\pm 0.022)$ \\ \cline{2-6} 
& 4 & $0.218 (\pm 0)$ & $0.034 (\pm 0.001)$ & $0.005 (\pm 0)$ & $0.390 (\pm 0.090)$ \\ \cline{2-6} 
& 5 & $0.220 (\pm 0)$ & $0.037 (\pm 0.003)$ & $0.005 (\pm 0)$ & $0.950 (\pm 0.320)$ \\ \cline{2-6} 
& 6 & $0.229 (\pm 0)$ & $0.039 (\pm 0.004)$ & $0.006 (\pm 0)$ & $2.400 (\pm 1.110)$ \\ \hline
\end{tabularx}
\caption{Running time in seconds of CTree, RPART, SkSurv and OSST.}
\label{tb:time1}
\end{table}
\begin{table}[htbp]
\begin{tabularx}{\textwidth}{|c|c|X|X|X|X|}
\hline
\textbf{Dataset} & \textbf{Depth} & \textbf{CTree} & \textbf{RPART} & \textbf{SkSurv} & \textbf{OSST} \\ \hline
\multirow{5}{*}{veterans} & 2 & $0.203 (\pm 0)$ & $0.013 (\pm 0.000)$ & $0.001 (\pm 0)$ & $0.007 (\pm 0.001)$ \\ \cline{2-6} 
& 3 & $0.200 (\pm 0)$ & $0.014 (\pm 0.001)$ & $0.002 (\pm 0)$ & $0.024 (\pm 0.001)$ \\ \cline{2-6} 
& 4 & $0.202 (\pm 0)$ & $0.014 (\pm 0.001)$ & $0.001 (\pm 0)$ & $0.095 (\pm 0.009)$ \\ \cline{2-6} 
& 5 & $0.208 (\pm 0)$ & $0.015 (\pm 0.001)$ & $0.001 (\pm 0)$ & $0.380 (\pm 0.080)$ \\ \cline{2-6} 
& 6 & $0.204 (\pm 0)$ & $0.015 (\pm 0.001)$ & $0.001 (\pm 0)$ & $1.270 (\pm 0.380)$ \\ \hline
\multirow{5}{*}{whas500} & 2 & $0.217 (\pm 0)$ & $0.026 (\pm 0.001)$ & $0.003 (\pm 0)$ & $0.006 (\pm 0.000)$ \\ \cline{2-6} 
& 3 & $0.203 (\pm 0)$ & $0.029 (\pm 0.005)$ & $0.003 (\pm 0)$ & $0.007 (\pm 0.001)$ \\ \cline{2-6} 
& 4 & $0.229 (\pm 0)$ & $0.033 (\pm 0.006)$ & $0.004 (\pm 0)$ & $0.690 (\pm 0.020)$ \\ \cline{2-6} 
& 5 & $0.224 (\pm 0)$ & $0.036 (\pm 0.001)$ & $0.005 (\pm 0)$ & $6.040 (\pm 0.470)$ \\ \cline{2-6} 
& 6 & $0.234 (\pm 0)$ & $0.038 (\pm 0.002)$ & $0.005 (\pm 0)$ & $37.01 (\pm 4.530)$ \\ \hline
\multirow{5}{*}{uissurv} & 2 & $0.198 (\pm 0)$ & $0.046 (\pm 0.001)$ & $0.004 (\pm 0)$ & $0.006 (\pm 0.000)$ \\ \cline{2-6} 
& 3 & $0.206 (\pm 0)$ & $0.051 (\pm 0.001)$ & $0.005 (\pm 0)$ & $0.058 (\pm 0.001)$ \\ \cline{2-6} 
& 4 & $0.212 (\pm 0)$ & $0.056 (\pm 0.006)$ & $0.006 (\pm 0)$ & $0.410 (\pm 0.010)$ \\ \cline{2-6} 
& 5 & $0.225 (\pm 0)$ & $0.060 (\pm 0.006)$ & $0.007 (\pm 0)$ & $2.680 (\pm 0.140)$ \\ \cline{2-6} 
& 6 & $0.214 (\pm 0)$ & $0.065 (\pm 0.008)$ & $0.008 (\pm 0)$ & $13.88 (\pm 1.200)$ \\ \hline
\multirow{5}{*}{airfoil} & 2 & $0.230 (\pm 0)$ & $0.032 (\pm 0.013)$ & $0.190 (\pm 0)$ & $38.12 (\pm 1.540)$ \\ \cline{2-6} 
& 3 & $0.286 (\pm 0)$ & $0.139 (\pm 0.025)$ & $0.240 (\pm 0)$ & $32.00 (\pm 1.250)$ \\ \cline{2-6} 
& 4 & $0.448 (\pm 0)$ & $0.289 (\pm 0.066)$ & $0.288 (\pm 0)$ & $31.90 (\pm 2.560)$ \\ \cline{2-6} 
& 5 & $0.272 (\pm 0)$ & $0.348 (\pm 0.047)$ & $0.320 (\pm 0)$ & $31.97 (\pm 1.533)$ \\ \cline{2-6} 
& 6 & $0.335 (\pm 0)$ & $0.453 (\pm 0.088)$ & $0.360 (\pm 0)$ & $31.98 (\pm 2.510)$ \\ \hline
\multirow{5}{*}{real-estate} & 2 & $0.273 (\pm 0)$ & $0.017 (\pm 0.003)$ & $0.018 (\pm 0)$ & $0.824 (\pm 0.005)$ \\ \cline{2-6} 
& 3 & $0.211 (\pm 0)$ & $0.018 (\pm 0.004)$ & $0.020 (\pm 0)$ & $100.3 (\pm 13.92)$ \\ \cline{2-6} 
& 4 & $0.215 (\pm 0)$ & $0.02 (\pm 0.006)$ & $0.030 (\pm 0)$ & $285.3 (\pm 18.61)$ \\ \cline{2-6} 
& 5 & $0.222 (\pm 0)$ & $0.022 (\pm 0.001)$ & $0.030 (\pm 0)$ & $290.4 (\pm 16.42)$ \\ \cline{2-6} 
& 6 & $0.221 (\pm 0)$ & $0.022 (\pm 0.001)$ & $0.030 (\pm 0)$ & $289.4 (\pm 36.42)$ \\ \hline
\multirow{5}{*}{sync} & 2 & $0.388 (\pm 0)$ & $0.017 (\pm 0.003)$ & $0.026 (\pm 0)$ & $1.920 (\pm 0.020)$ \\ \cline{2-6} 
& 3 & $0.201 (\pm 0)$ & $0.019 (\pm 0.000)$ & $0.030 (\pm 0)$ & $2.120 (\pm 0.030)$ \\ \cline{2-6} 
& 4 & $0.269 (\pm 0)$ & $0.019 (\pm 0.006)$ & $0.053 (\pm 0)$ & $3.788 (\pm 0.056)$ \\ \cline{2-6} 
& 5 & $0.204 (\pm 0)$ & $0.021 (\pm 0.007)$ & $0.036 (\pm 0)$ & $3.804 (\pm 0.033)$ \\ \cline{2-6} 
& 6 & $0.261 (\pm 0)$ & $0.021 (\pm 0.008)$ & $0.040 (\pm 0)$ & $3.785 (\pm 0.053)$ \\ \hline
\multirow{5}{*}{credit} & 2 & $0.221 (\pm 0)$ & $0.043 (\pm 0.007)$ & $0.011 (\pm 0)$ & $0.056 (\pm 0.010)$ \\ \cline{2-6} 
& 3 & $0.246 (\pm 0)$ & $0.055 (\pm 0.010)$ & $0.014 (\pm 0)$ & $1.473 (\pm 0.482)$ \\ \cline{2-6} 
& 4 & $0.273 (\pm 0)$ & $0.065(\pm 0.010)$ & $0.020 (\pm 0)$ & $33.06 (\pm 12.90)$ \\ \cline{2-6} 
& 5 & $0.283 (\pm 0)$ & $0.074 (\pm 0.017)$ & $0.020 (\pm 0)$ & $610.9 (\pm 272.3)$ \\ \cline{2-6} 
& 6 & $0.317 (\pm 0)$ & $0.082 (\pm 0.020)$ & $0.022 (\pm 0)$ & $930.8 (\pm 398.4)$ \\ \hline
\multirow{5}{*}{employee} & 2 & $0.260 (\pm 0)$ & $0.412 (\pm 0.001)$ & $0.087 (\pm 0)$ & $0.226 (\pm 0.018)$ \\ \cline{2-6} 
& 3 & $0.272 (\pm 0)$ & $0.492 (\pm 0.005)$ & $0.122 (\pm 0)$ & $2.090 (\pm 0.110)$ \\ \cline{2-6} 
& 4 & $0.335 (\pm 0)$ & $0.552 (\pm 0.012)$ & $0.154 (\pm 0)$ & $16.20 (\pm 1.370)$ \\ \cline{2-6} 
& 5 & $0.370 (\pm 0)$ & $0.610 (\pm 0.030)$ & $0.183 (\pm 0)$ & $101.8 (\pm 15.36)$ \\ \cline{2-6} 
& 6 & $0.443 (\pm 0)$ & $0.6594 (\pm 0.04)$ & $0.210 (\pm 0)$ & $538.3 (\pm 125.7)$ \\ \hline
\multirow{5}{*}{churn} & 2 & $0.211 (\pm 0)$ & $0.060 (\pm 0.006)$ & $0.016 (\pm 0)$ & $0.138 (\pm 0.000)$ \\ \cline{2-6} 
& 3 & $0.225 (\pm 0)$ & $0.074 (\pm 0.001)$ & $0.022 (\pm 0)$ & $0.880 (\pm 0.070)$ \\ \cline{2-6} 
& 4 & $0.250 (\pm 0)$ & $0.090 (\pm 0.006)$ & $0.030 (\pm 0)$ & $11.67 (\pm 2.360)$ \\ \cline{2-6} 
& 5 & $0.281 (\pm 0)$ & $0.048 (\pm 0.007)$ & $0.009 (\pm 0)$ & $125.2 (\pm 35.53)$ \\ \cline{2-6} 
& 6 & $0.306 (\pm 0)$ & $0.117 (\pm 0.008)$ & $0.038 (\pm 0)$ & $1197 (\pm 410.5)$ \\ \hline
\multirow{5}{*}{insurance} & 2 & $0.220 (\pm 0)$ & $0.043 (\pm 0.003)$ & $0.180 (\pm 0)$ & $0.083 (\pm 0.010)$ \\ \cline{2-6} 
& 3 & $0.234 (\pm 0)$ & $0.049 (\pm 0.001)$ & $0.230 (\pm 0)$ & $0.083 (\pm 0.010)$ \\ \cline{2-6} 
& 4 & $0.258 (\pm 0)$ & $0.058 (\pm 0.002)$ & $0.261 (\pm 0)$ & $0.084 (\pm 0.010)$ \\ \cline{2-6} 
& 5 & $0.259 (\pm 0)$ & $0.064 (\pm 0.003)$ & $0.277 (\pm 0)$ & $0.084 (\pm 0.010)$ \\ \cline{2-6} 
& 6 & $0.250 (\pm 0)$ & $0.064 (\pm 0.005)$ & $0.290 (\pm 0)$ & $0.084 (\pm 0.010)$ \\ \hline
\end{tabularx}
\caption{Running time in seconds of CTree, RPART, SkSurv and OSST (continued).}
\label{tb:time2}
\end{table}
\begin{table}
    \centering
    \begin{tabularx}{\textwidth}{|c|c|X|X|X|X|}
\hline
\textbf{Dataset} & \textbf{Depth} & \textbf{CTree} & \textbf{RPART} & \textbf{SkSurv} & \textbf{OSST} \\ \hline
\multirow{5}{*}{flchain} & 2 & $0.418 (\pm 0)$ & $0.523 (\pm 0.013)$ & $0.109 (\pm 0)$ & $0.481 (\pm 0.022)$ \\ \cline{2-6} 
& 3 & $0.460 (\pm 0)$ & $0.521 (\pm 0.001)$ & $0.110 (\pm 0)$ & $0.336 (\pm 0.240)$ \\ \cline{2-6} 
& 4 & $0.523 (\pm 0)$ & $0.652 (\pm 0.042)$ & $0.209 (\pm 0)$ & $0.419 (\pm 0.110)$ \\ \cline{2-6} 
& 5 & $0.559 (\pm 0)$ & $0.824 (\pm 0.103)$ & $0.244 (\pm 0)$ & $1.264 (\pm 0.510)$ \\ \cline{2-6} 
& 6 & $0.601 (\pm 0)$ & $0.064 (\pm 0.005)$ & $0.285 (\pm 0)$ & $7.024 (\pm 3.230)$ \\ \hline
\multirow{5}{*}{servo} & 2 & $0.198 (\pm 0)$ & $0.014 (\pm 0.000)$ & $0.004 (\pm 0)$ & $0.075 (\pm 0.001)$ \\ \cline{2-6} 
& 3 & $0.260 (\pm 0)$ & $0.015 (\pm 0.002)$ & $0.004 (\pm 0)$ & $6.100(\pm 0.360)$ \\ \cline{2-6} 
& 4 & $0.199 (\pm 0)$ & $0.015 (\pm 0.006)$ & $0.004 (\pm 0)$ & $227.4 (\pm 45.10)$ \\ \cline{2-6} 
& 5 & $0.206 (\pm 0)$ & $0.016 (\pm 0.007)$ & $0.004 (\pm 0)$ & $428.4 (\pm 85.30)$ \\ \cline{2-6} 
& 6 & $0.206 (\pm 0)$ & $0.016 (\pm 0.006)$ & $0.004 (\pm 0)$ & $2192 (\pm 586.0)$ \\ \hline
\end{tabularx}
\caption{Running time in seconds of CTree, RPART, SkSurv and OSST (continued).\vspace*{5pt}}
\label{tb:time3}
\end{table}

\section{Experiment: Scalability}\label{exp:scalability}
\noindent\textbf{Collection and Setup:} We ran this experiment on the dataset \textit{household} for CTree, RPART, SkSurv and OSST. We subsampled 100, 500, 1000, 5000, 10,000, 50,000, 100,000, 500,000, 1000,000 samples from it to form 10 datasets (including the original dataset). We randomly censored samples in each dataset with censoring rates of 20\%, 50\% and 80\%. The depth limit was set to 5 (for all methods) and regularization coefficient to 0.01 for OSST and 0.05 for RPART, in order to generate trees with similar size. OSST used an extreme random survival forest that consists of 20 trees with maximum depth 3 as the reference model. We set the time limit of each run to 60 minutes in this experiment.\\

\begin{figure}[htbp]
    \centering
    \includegraphics[width=0.32\textwidth]{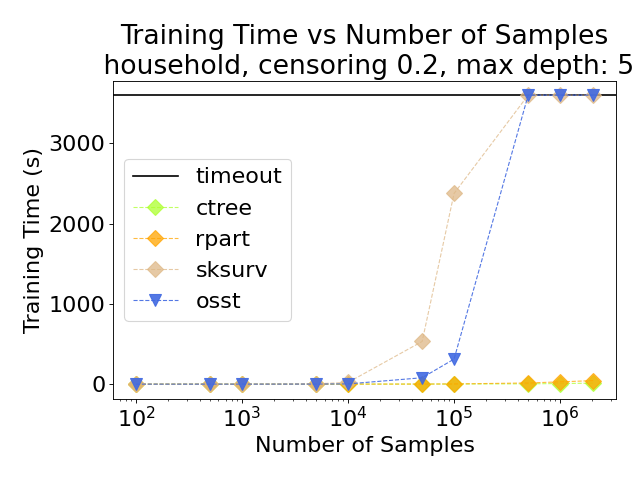}
    \includegraphics[width=0.32\textwidth]{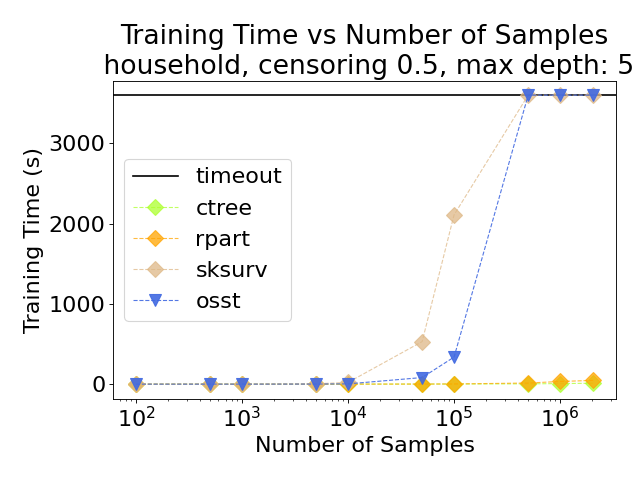}
    \includegraphics[width=0.32\textwidth]{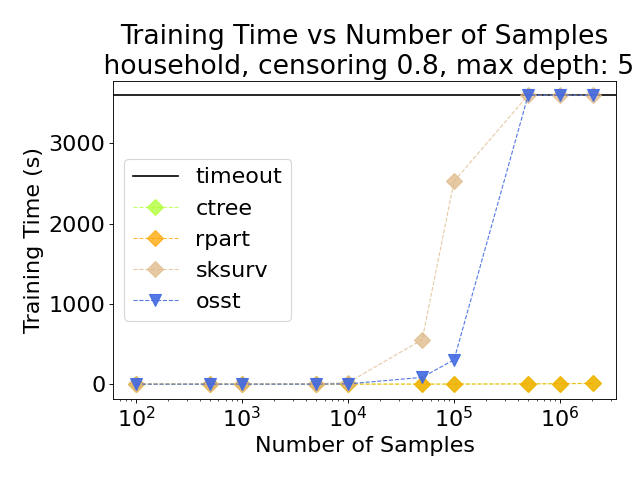}
    \caption{Training time of CTree, RPART, SkSurv and OSST as a function of sample size on dataset \textit{household}, $d=5,\lambda=0.01$. (60-minutes time limit)}
    \label{fig:scalability}
\end{figure}

\noindent\textbf{Results:} Figure \ref{fig:scalability} shows that OSST achieves similar performance with other greedy methods when the number of samples is no more than ten thousand. It scales better than SkSurv on datasets with more than ten thousand samples (8-10 times faster than SkSurv). OSST and SkSurv timed out when there are more than 0.1 million samples in the dataset. There is no obvious difference in training time when the rate of censoring varies. 

\section{Ablation Study: Guessing}
We explored if OSST is still able to find optimal trees when our guessing technique is applied and how much time it saves.

\noindent\textbf{Collection and Setup:} We ran this experiment on 4 datasets \textit{(churn, employee, servo, sync)} for variations of \ourmethod.
\begin{itemize}[leftmargin=*, topsep=0pt, noitemsep]
    \item \textit{churn}: We set depth limit to 6 and regularization coefficient to 0.0025.
    \item \textit{employee}: We set depth limit to 6 and regularization coefficient to 0.01.
    \item \textit{servo}: We set depth limit to 5 and regularization coefficient to 0.01.
    \item \textit{sync}: We set depth limit to 5 and regularization coefficient to 0.01.
\end{itemize} 

For each dataset, we ran \ourmethod{} twice, once using the \textit{reference model lower bound} and once using the \textit{equivalent points lower bound}. We used random survival forests with 100 trees of depth 9 as reference models.

\noindent\textbf{Calculations:} We recorded the elapsed time (includes the training time of reference models), IBS scores, iterations, and size of dependency graph for each run of \ourmethod{}. We plot the training IBS against training time for each dataset.

\begin{figure}[htbp]
    \centering
    \includegraphics[width=0.45\textwidth]{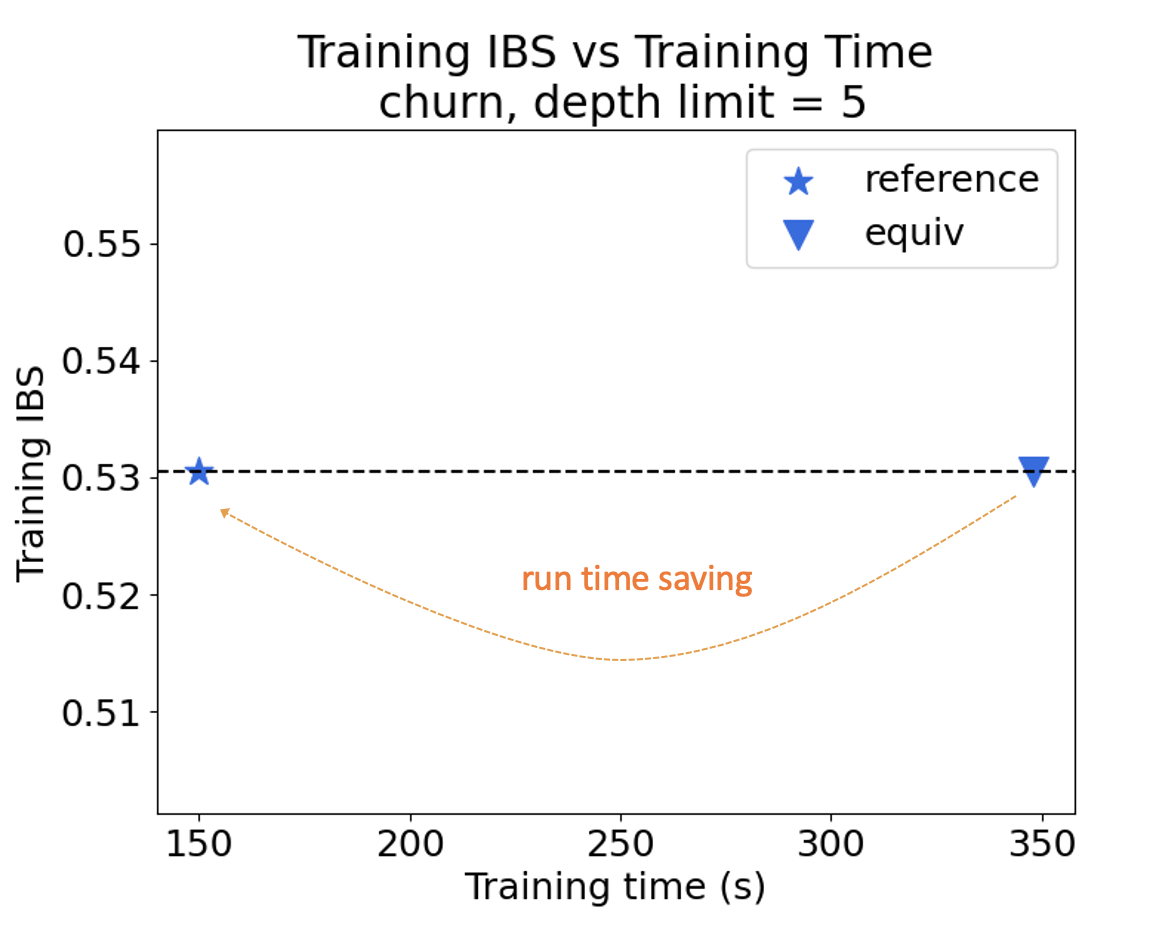}
    \includegraphics[width=0.45\textwidth]{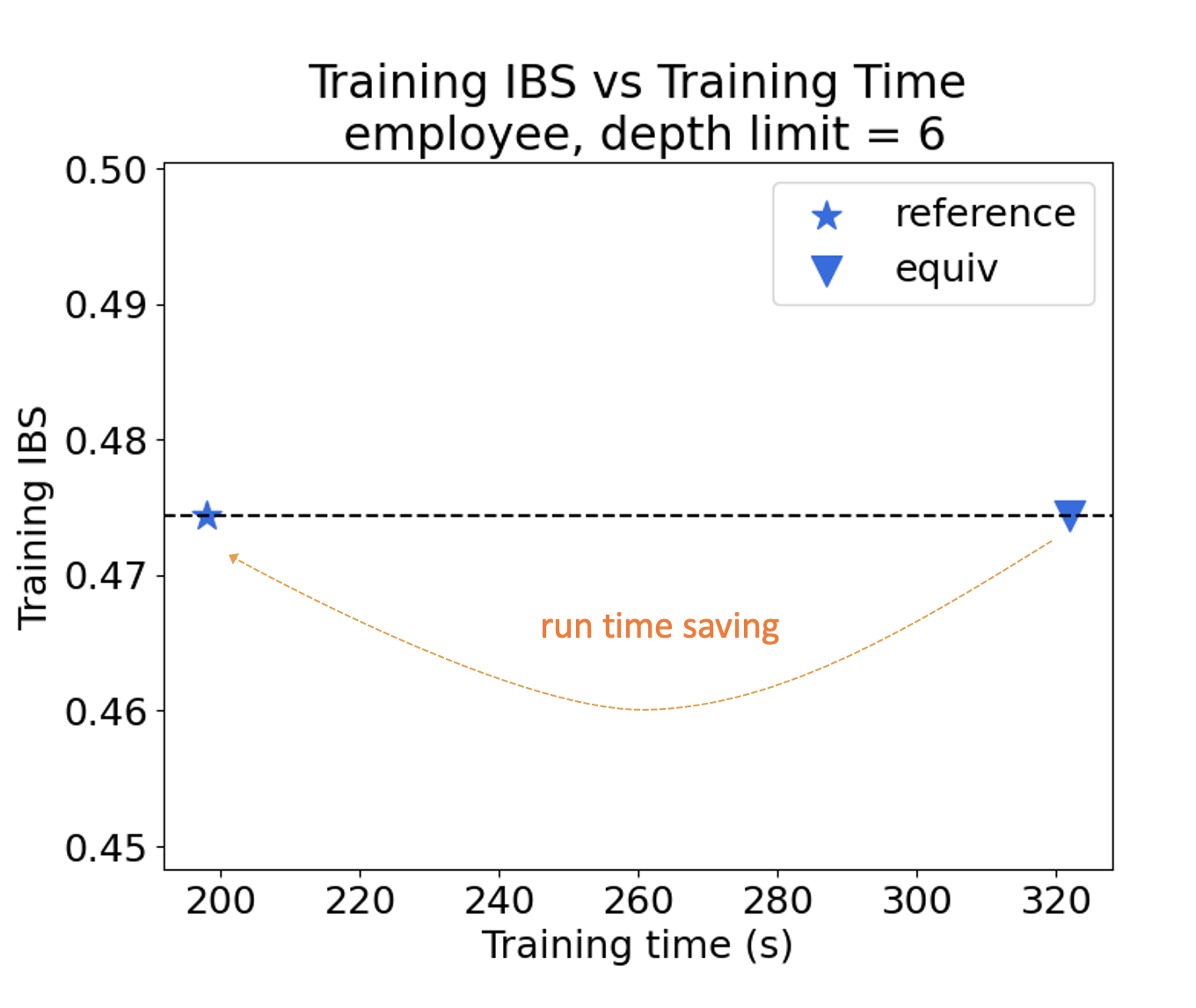}
    \includegraphics[width=0.45\textwidth]{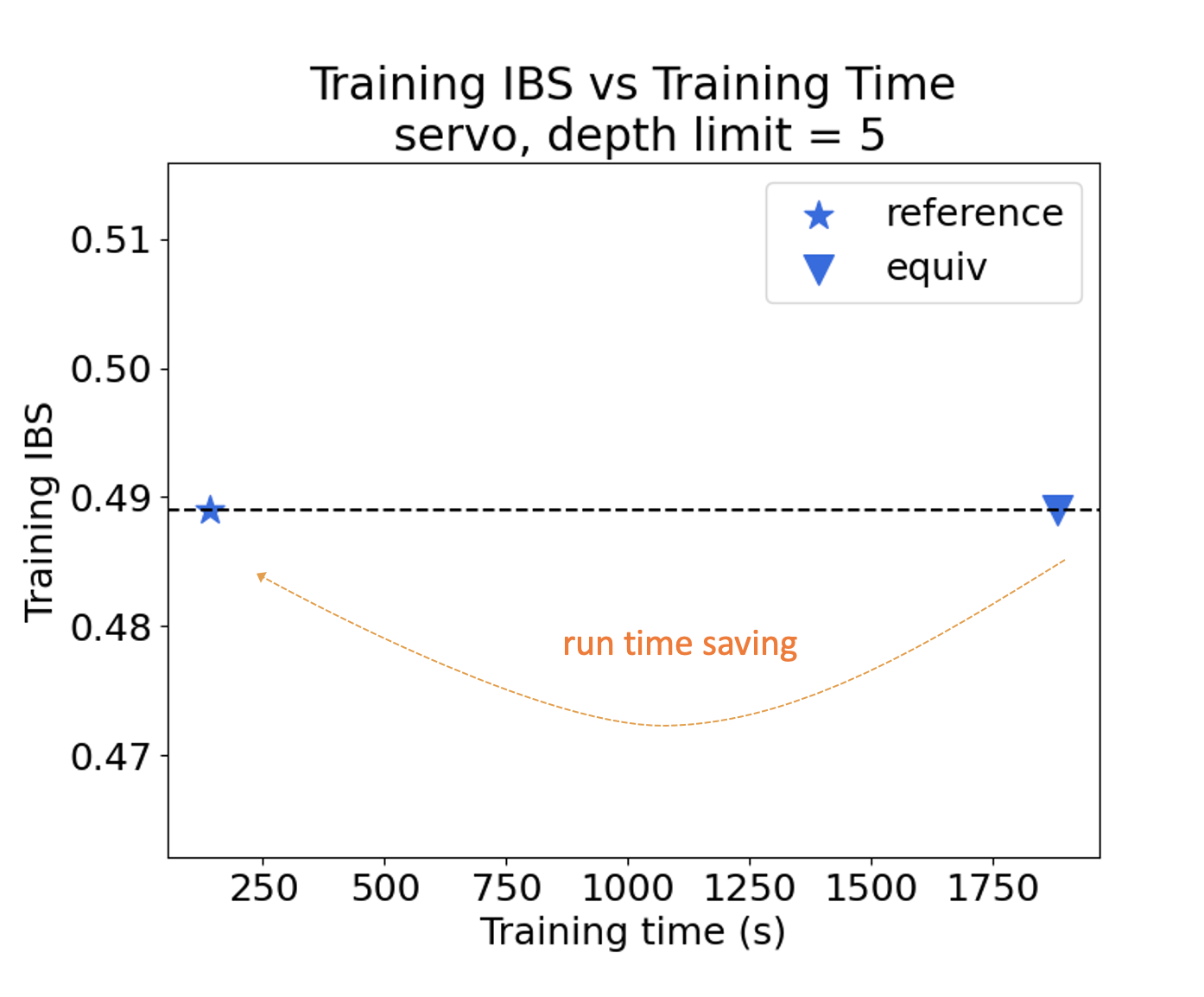}
    \includegraphics[width=0.45\textwidth]{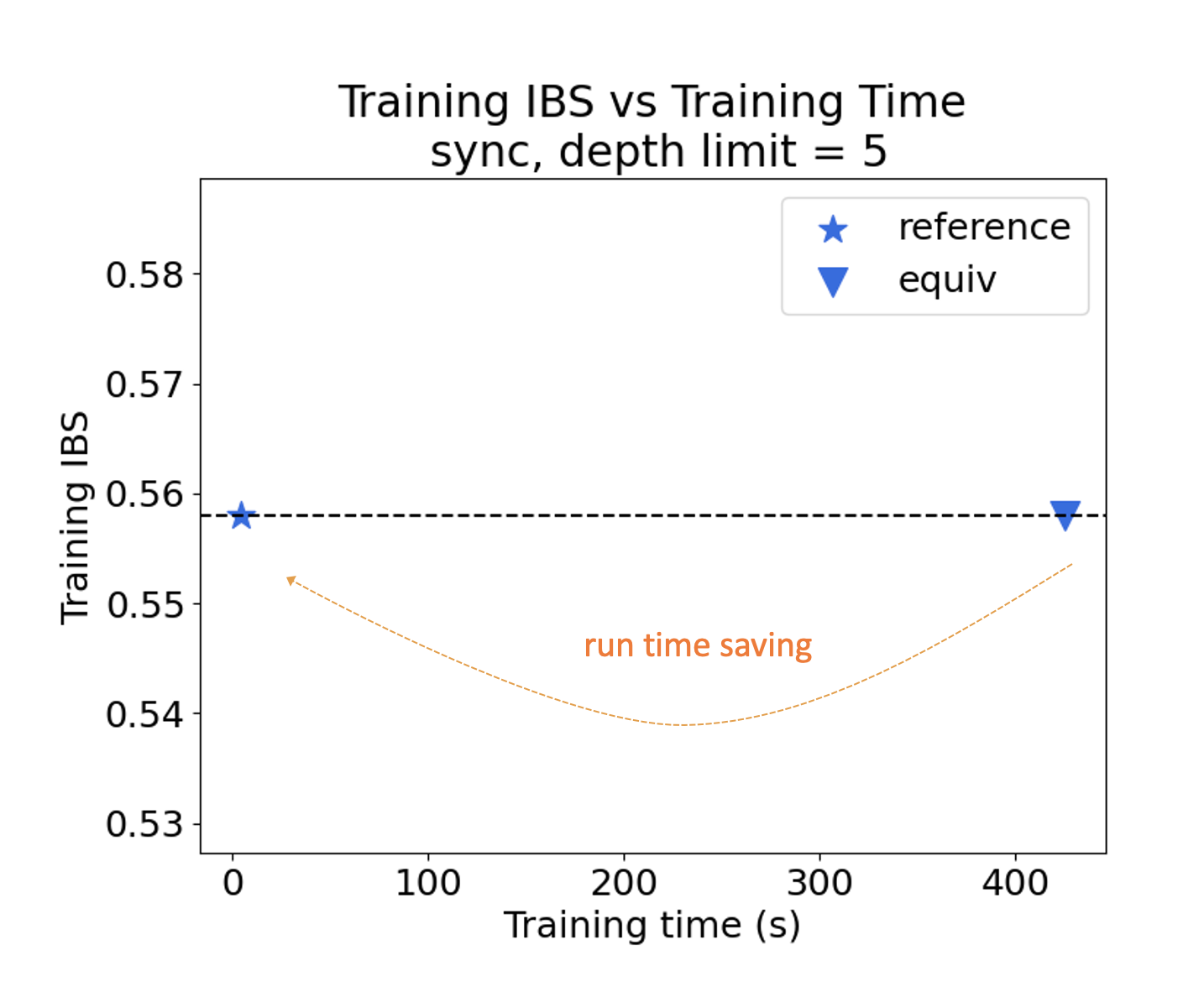}
    \caption{Training time savings of OSST using reference model lower bound on dataset \textit{churn, employee, servo, sync}.}
    \label{fig:ablation}
\end{figure}

\noindent\textbf{Results:} Figure \ref{fig:ablation} shows that our reference model lower bound substantially reduces the training time (reduction ranges from 40\% to 95\%), without losing the optimality of generated trees. 
\newpage
\section{Optimal Survival Trees}\label{exp:opt_trees}
\begin{figure}[htbp]
    \centering
    \includegraphics[height = 3in]{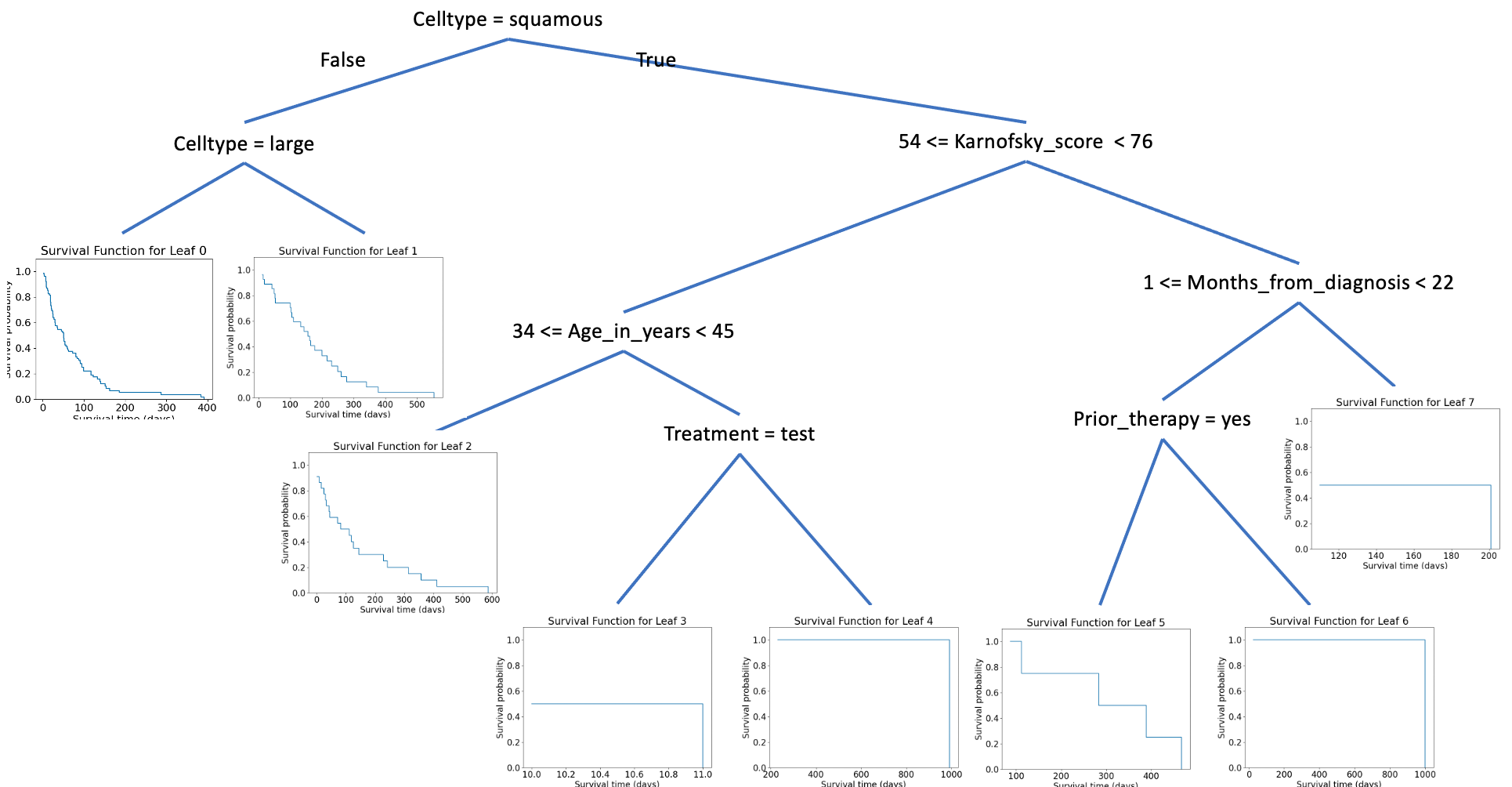}
    \caption{Optimal survival tree produced by OSST for \textbf{veterans} dataset with 8 leaves, IBS: $32.83\%$.}
    \label{fig:veterans_osst}
\end{figure}

\begin{figure}[htbp]
    \centering
    \includegraphics[height = 3in]{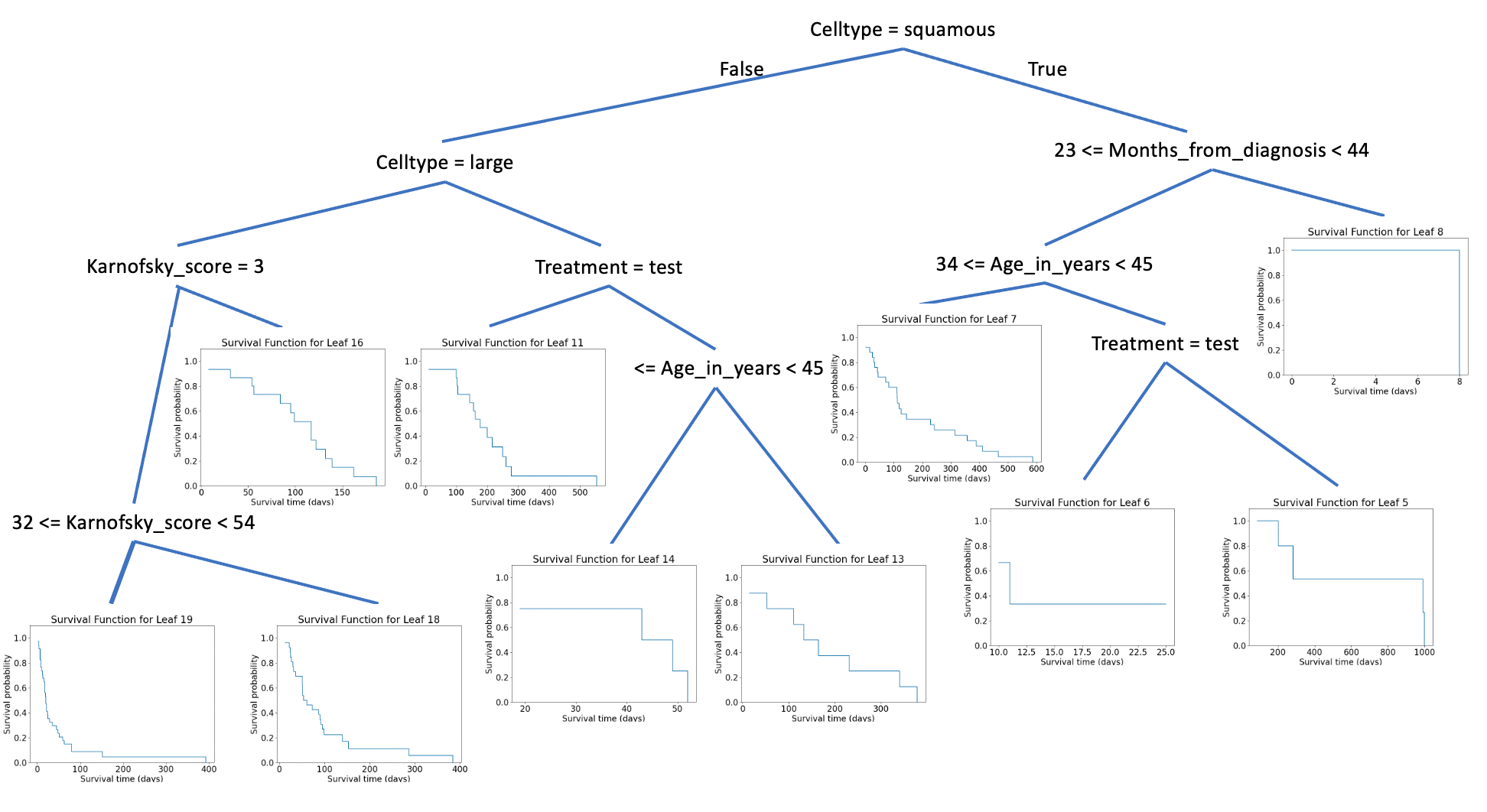}
    \caption{Sub-optimal survival tree produced by RPART for \textbf{veterans} dataset with 10 leaves, IBS: $27.95\%$.}
    \label{fig:veterans_rpart}
\end{figure}

\begin{figure}[htbp]
    \centering
    \includegraphics[height = 2in]{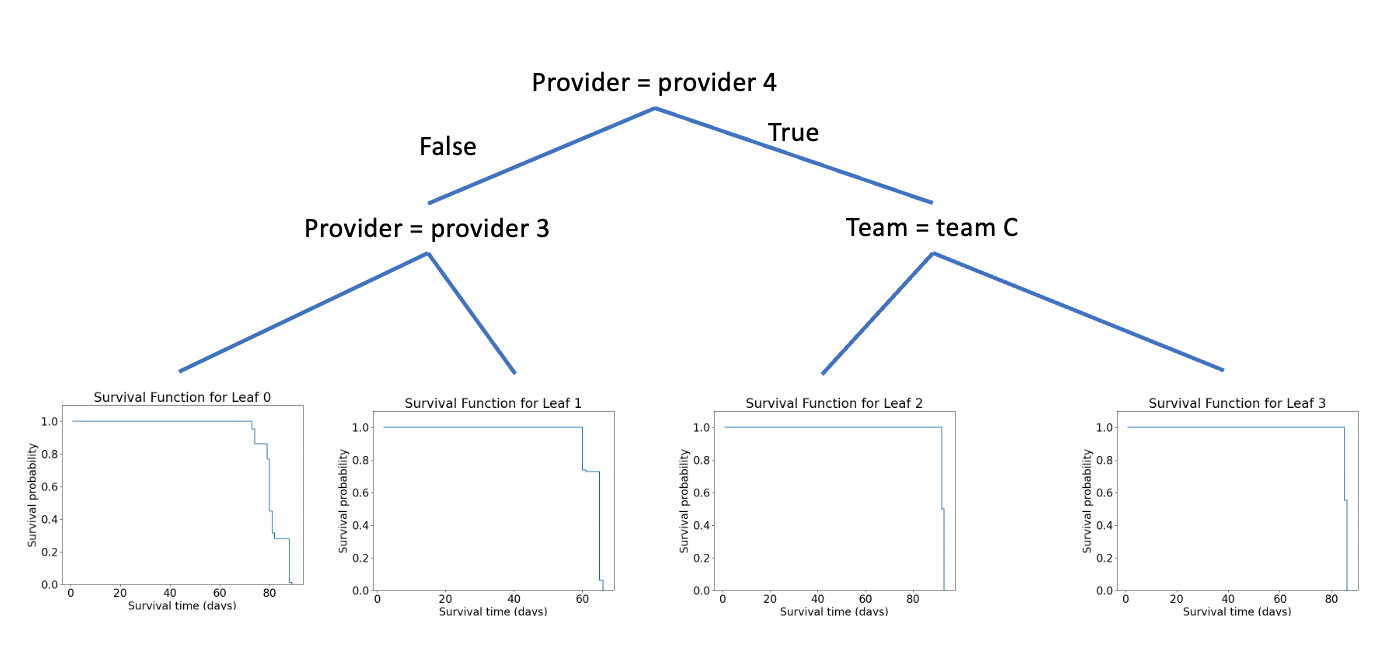}
    \caption{Optimal survival tree produced by OSST for \textbf{maintenance} dataset with 4 leaves, IBS: $73.25\%$.}
    \label{fig:maintenance_osst}
\end{figure}

\begin{figure}[htbp]
    \centering
    \includegraphics[height = 2in]{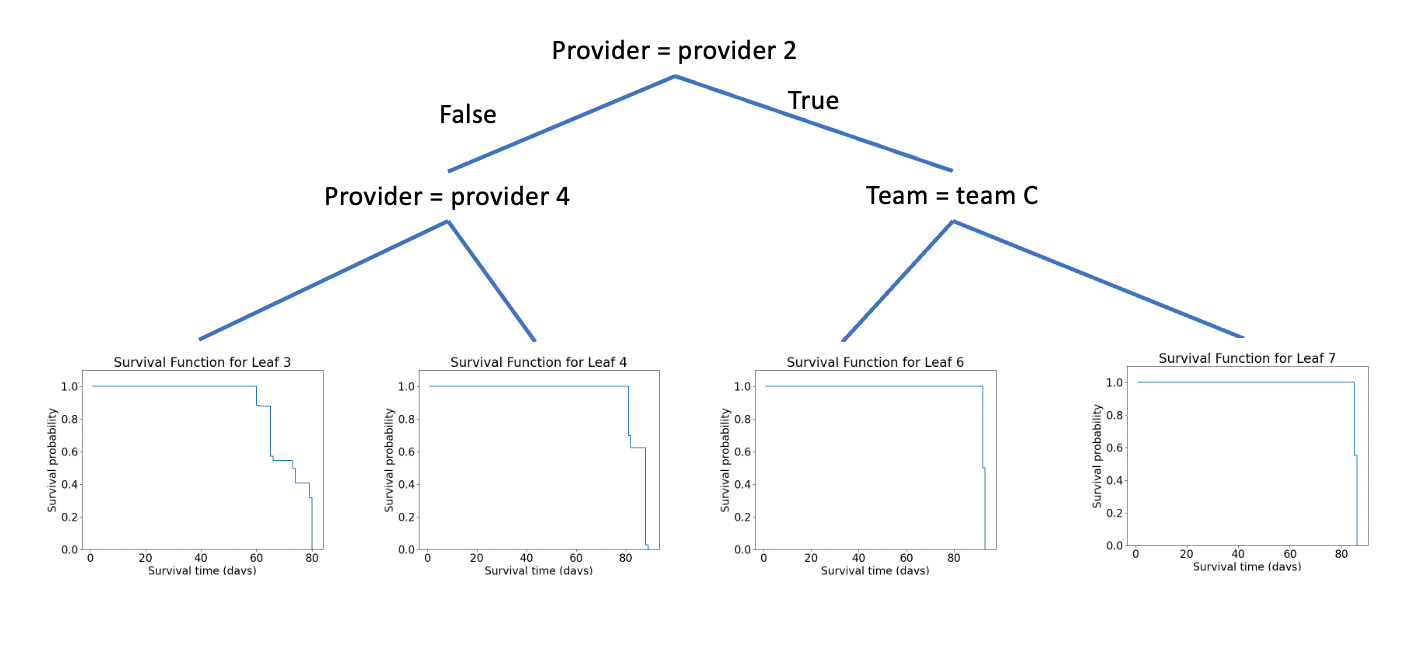}
    \caption{Sub-optimal survival tree produced by CTree for \textbf{maintenance} dataset with 4 leaves, IBS: $52.26\%$.}
    \label{fig:maintenance_ctree}
\end{figure}

\begin{figure}[htbp]
    \centering
    \includegraphics[height = 2in]{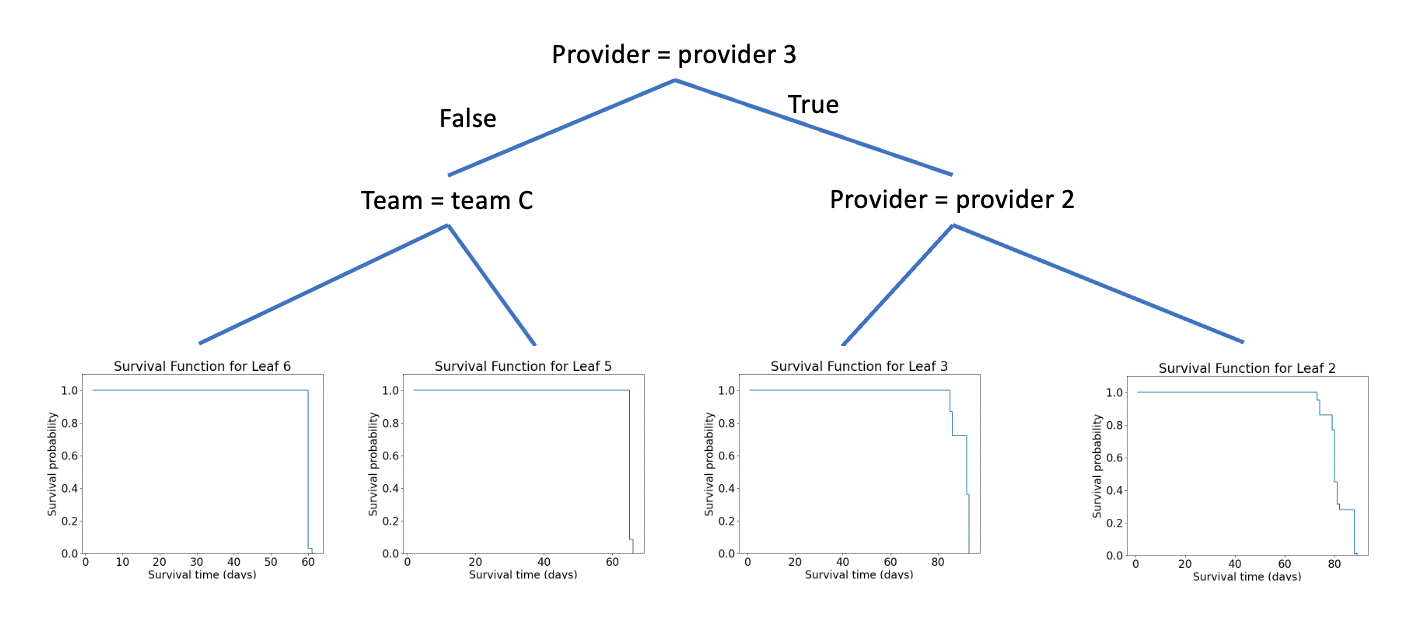}
    \caption{Sub-optimal survival tree produced by SkSurv for \textbf{maintenance} dataset with 4 leaves, IBS: $71.05\%$.}
    \label{fig:maintenance_sksurv}
\end{figure}

We visually compare the optimal survival trees returned by OSST and sub-optimal trees returned by other methods. Figure \ref{fig:veterans_osst} shows an optimal tree found by OSST with 8 leaves (IBS ratio of $32.83\%$) and Figure \ref{fig:veterans_rpart} shows a sub-optimal tree found by RPART with 10 leaves (IBS = $27.95\%$). Here, OSST found a tree with fewer leaves yet better performance. Figures \ref{fig:maintenance_osst} to \ref{fig:maintenance_sksurv} show three 4-leaf survival trees returned by OSST, CTree and SkSurv respectively, among which OSST achieves the best performance. 


\clearpage 

\section{Comparison with Interpretable AI}\label{exp:iai}
Interpretable AI implements the Optimal Survival Tree (OST) algorithm of \citet{bertsimas2022optimal} which optimizes an objective that is totally different from \ourmethod{}. It does not use any of the survival metrics defined in Section \ref{sec:metrics}. OST assumes the cumulative hazard function of each sample is proportional to the baseline cumulative hazard function, like Cox models do, and samples in the same leaf have the same adjusted coefficient to the baseline. It considers the `ground truth' of each sample's adjusted coefficient as the coefficient when each sample stays in its own leaf, which is not always possible due to the existence of equivalent points as we discussed before in Theorem \ref{thm:equiv_lb}. OST uses the log-likelihood difference between the fitted leaf node coefficients and the `true' coefficients as the prediction error and penalizes it with the number of leaves. 

\begin{figure}[htbp]
    \centering
    \includegraphics[width=0.45\textwidth]{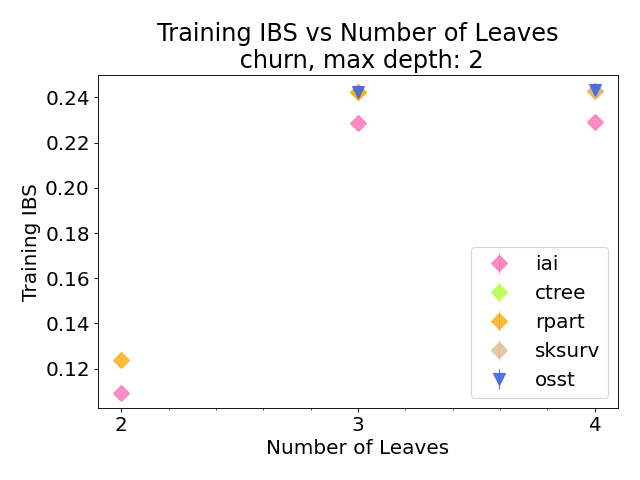}
    \includegraphics[width=0.45\textwidth]{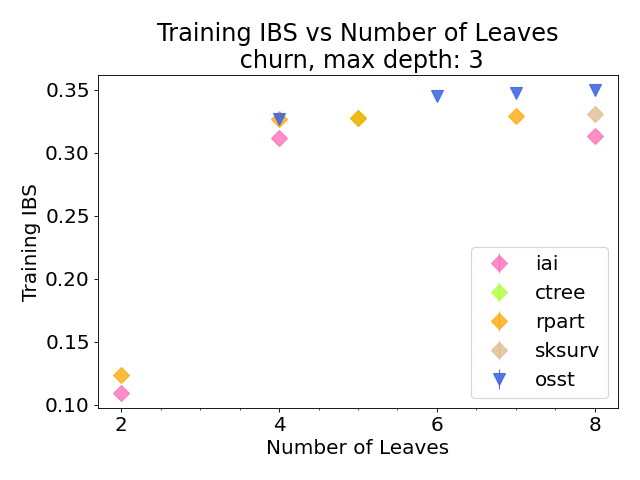}
    \includegraphics[width=0.45\textwidth]{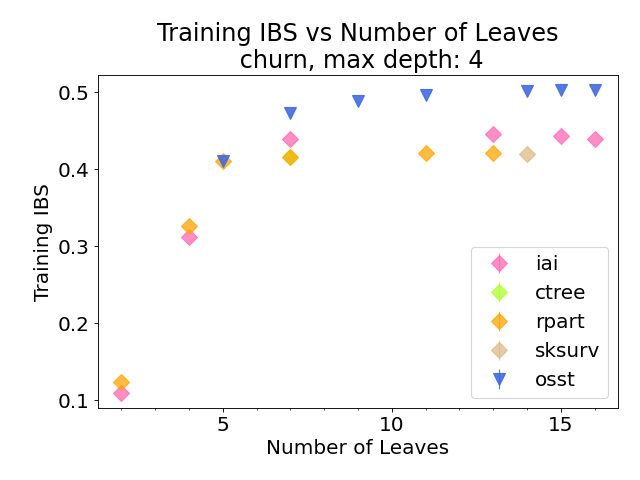}
    \includegraphics[width=0.45\textwidth]{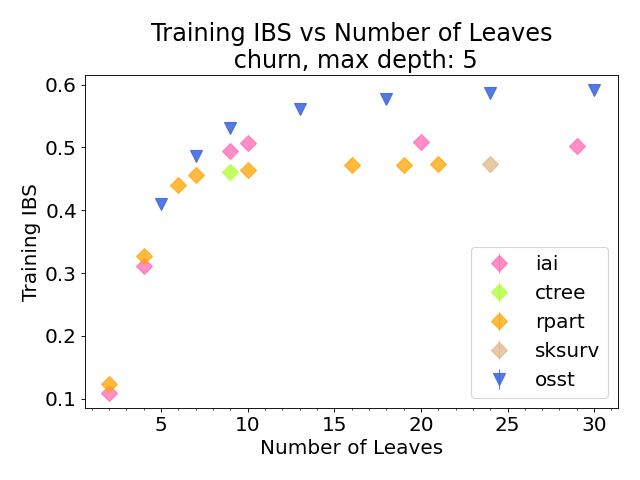}
    \caption{Training IBS achieved by CTree, RPART, SkSurv, IAI and OSST as a function of number of leaves on dataset: churn.}
    \label{fig:iai}
\end{figure}

OST frequently crashed in the experiments in Section \ref{exp:lvs} to \ref{exp:scalability} (reporting segment faults, see Listing \ref{iai_error}). With limited results returned by IAI, we found that trees generated by IAI are \textit{not optimal} in terms of IBS (ratio). Figure \ref{fig:iai} shows that the trees of IAI sometimes are worse than those of the greedy methods. 

\clearpage 
\begin{lstlisting}[caption={IAI Error Code},label={iai_error},language=Python]
[ Warning: This copy of Interpretable AI software is for academic purposes
only and not for commercial use.

signal (11): Segmentation fault
in expression starting at none:0
_Py_DECREF at /usr/local/src/conda/python-3.9.12/Include/object.h:422 [inlined]
GetResult at /usr/local/src/conda/python-3.9.12/Modules/_ctypes/callproc.c:989 [inlined]
_ctypes_callproc at /usr/local/src/conda/python-3.9.12/Modules/_ctypes/callproc.c:1301
PyCFuncPtr_call at /usr/local/src/conda/python-3.9.12/Modules/_ctypes/_ctypes.c:4201
_PyObject_MakeTpCall at python3 (unknown line)
_PyEval_EvalFrameDefault at python3 (unknown line)
_PyFunction_Vectorcall at python3 (unknown line)
unknown function (ip: 0x55cf104ac72e)
_PyFunction_Vectorcall at python3 (unknown line)
unknown function (ip: 0x55cf104ac72e)
_PyFunction_Vectorcall at python3 (unknown line)
unknown function (ip: 0x55cf104ac72e)
unknown function (ip: 0x55cf10544bf8)
unknown function (ip: 0x55cf104ac754)
unknown function (ip: 0x55cf10544bf8)
unknown function (ip: 0x55cf104f5286)
unknown function (ip: 0x55cf104f54dc)
unknown function (ip: 0x55cf1053a102)
PyObject_GetAttr at python3 (unknown line)
_PyObject_GetMethod at python3 (unknown line)
_PyEval_EvalFrameDefault at python3 (unknown line)
_PyFunction_Vectorcall at python3 (unknown line)
unknown function (ip: 0x55cf1056ef4e)
PyObject_GetItem at python3 (unknown line)
_PyEval_EvalFrameDefault at python3 (unknown line)
_PyFunction_Vectorcall at python3 (unknown line)
unknown function (ip: 0x55cf104aaae5)
unknown function (ip: 0x55cf10543662)
_PyFunction_Vectorcall at python3 (unknown line)
unknown function (ip: 0x55cf104aaae5)
unknown function (ip: 0x55cf10543662)
PyEval_EvalCodeEx at python3 (unknown line)
PyEval_EvalCode at python3 (unknown line)
unknown function (ip: 0x55cf105f050a)
unknown function (ip: 0x55cf10620f74)
unknown function (ip: 0x55cf104c1986)
PyRun_SimpleFileExFlags at python3 (unknown line)
Py_RunMain at python3 (unknown line)
Py_BytesMain at python3 (unknown line)
__libc_start_main at /lib/x86_64-linux-gnu/libc.so.6 (unknown line)
unknown function (ip: 0x55cf105ae09f)
Allocations: 57922925 (Pool: 57903382; Big: 19543); GC: 62
Segmentation fault (core dumped)
\end{lstlisting}
\clearpage 
\section{Comparison with Other Interpretable Survival Models}
We also compared OSST with other interpretable models. The Cox proportional hazards model provides more flexible predictions of survival functions but fails to capture non-linear relationships among the features. More importantly, standard Cox models generally do not have sparsity regularization and they are not sparse even with ridge ($\ell_2$) regularization. We consider a single Cox model with many non-zero coefficients less interpretable than a single sparse tree model. Generalized Additive Models (GAMs) is another interpretable survival model form that is not as popular as Cox models. \citet{bender2018generalized} provide one possible way to apply GAMs to survival analysis: the survival data should first be transformed into a format that is suitable to fit Piecewise Exponential Additive Mixed Models (PAMMs); fitted PAMMs can be represented as GAMs and be estimated using generic GAM software.

\noindent\textbf{Collection and Setup:} We ran this experiment on 5 datasets \textit{(churn, credit, employee, maintenance, servo)} for Cox, PAMM and OSST.
We split each dataset into halves (training and testing sets). For each method, we performed 5-fold cross-validation on the training set to choose hyper-parameters, trained models with the chosen hyper-parameters on the training set and tested the performance (IBS ratio) on the testing set. There are no available hyper-parameters to tune in \textit{pammtools}. 

\textbf{Cox:} We used \textit{CoxPHSurvivalAnalysis} from scikit-survival.\\
\textbf{PAMM:} We used \textit{pam} from \textit{pammtools} package, \url{https://cran.r-project.org/web/packages/pammtools/index.html}.

\noindent\textbf{Results:} We found that OSST approaches the performance of Cox while PAMM is dominated by OSST and a single Kaplan-Meier curve. Table \ref{tb:interpretable-models} shows the best hyper-parameters, training accuracies and testing accuracies of each method on various datasets. 

\begin{table}[htbp]
\begin{tabular}{|c|c|cc|c|c|}
\hline
Dataset                & Method                & \multicolumn{2}{c|}{Tuned Hyper-parameters}      & Training IBS Ratio      & Testing IBS Ratio        \\ \hline\hline
\multirow{6}{*}{churn} & \multirow{2}{*}{OSST} & \multicolumn{1}{c|}{regularization} & max\_depth & \multirow{2}{*}{0.602}  & \multirow{2}{*}{0.470}   \\ \cline{3-4}
                       &                       & \multicolumn{1}{c|}{0.001}          & 5          &                         &                          \\ \cline{2-6} 
                       & \multirow{2}{*}{Cox}  & \multicolumn{2}{c|}{alpha}                       & \multirow{2}{*}{0.571}  & \multirow{2}{*}{0.495}   \\ \cline{3-4}
                       &                       & \multicolumn{2}{c|}{1}                           &                         &                          \\ \cline{2-6} 
                       & \multirow{2}{*}{PAMM} & \multicolumn{2}{c|}{N/A}                         & \multirow{2}{*}{-0.500} & \multirow{2}{*}{0.490}   \\ \cline{3-4}
                       &                       & \multicolumn{2}{c|}{N/A}                         &                         &                          \\ \hline\hline
\multirow{6}{*}{credit}      & \multirow{2}{*}{OSST} & \multicolumn{1}{c|}{regularization} & max\_depth & \multirow{2}{*}{0.199} & \multirow{2}{*}{0.170} \\ \cline{3-4}
                       &                       & \multicolumn{1}{c|}{0.0025}         & 5          &                         &                          \\ \cline{2-6} 
                       & \multirow{2}{*}{Cox}  & \multicolumn{2}{c|}{alpha}                       & \multirow{2}{*}{0.323}  & \multirow{2}{*}{0.135}   \\ \cline{3-4}
                       &                       & \multicolumn{2}{c|}{1}                           &                         &                          \\ \cline{2-6} 
                       & \multirow{2}{*}{PAMM} & \multicolumn{2}{c|}{N/A}                         & \multirow{2}{*}{-0.443} & \multirow{2}{*}{error}   \\ \cline{3-4}
                       &                       & \multicolumn{2}{c|}{N/A}                         &                         &                          \\ \hline\hline
\multirow{6}{*}{employee}    & \multirow{2}{*}{OSST} & \multicolumn{1}{c|}{regularization} & max\_depth & \multirow{2}{*}{0.691} & \multirow{2}{*}{0.634} \\ \cline{3-4}
                       &                       & \multicolumn{1}{c|}{0.0005}         & 5          &                         &                          \\ \cline{2-6} 
                       & \multirow{2}{*}{Cox}  & \multicolumn{2}{c|}{alpha}                       & \multirow{2}{*}{0.441}  & \multirow{2}{*}{0.376}   \\ \cline{3-4}
                       &                       & \multicolumn{2}{c|}{5}                           &                         &                          \\ \cline{2-6} 
                       & \multirow{2}{*}{PAMM} & \multicolumn{2}{c|}{N/A}                         & \multirow{2}{*}{error}  & \multirow{2}{*}{error}   \\ \cline{3-4}
                       &                       & \multicolumn{2}{c|}{N/A}                         &                         &                          \\ \hline\hline
\multirow{6}{*}{maintenance} & \multirow{2}{*}{OSST} & \multicolumn{1}{c|}{regularization} & max\_depth & \multirow{2}{*}{0.985} & \multirow{2}{*}{0.983} \\ \cline{3-4}
                       &                       & \multicolumn{1}{c|}{0.0001}         & 5          &                         &                          \\ \cline{2-6} 
                       & \multirow{2}{*}{Cox}  & \multicolumn{2}{c|}{alpha}                       & \multirow{2}{*}{0.901}  & \multirow{2}{*}{0.898}   \\ \cline{3-4}
                       &                       & \multicolumn{2}{c|}{1}                           &                         &                          \\ \cline{2-6} 
                       & \multirow{2}{*}{PAMM} & \multicolumn{2}{c|}{N/A}                         & \multirow{2}{*}{-6.54}  & \multirow{2}{*}{-5.89}   \\ \cline{3-4}
                       &                       & \multicolumn{2}{c|}{N/A}                         &                         &                          \\ \hline\hline
\multirow{6}{*}{servo} & \multirow{2}{*}{OSST} & \multicolumn{1}{c|}{regularization} & max\_depth & \multirow{2}{*}{0.544}  & \multirow{2}{*}{0.360}   \\ \cline{3-4}
                       &                       & \multicolumn{1}{c|}{0.01}           & 4          &                         &                          \\ \cline{2-6} 
                       & \multirow{2}{*}{Cox}  & \multicolumn{2}{c|}{alpha}                       & \multirow{2}{*}{0.624}  & \multirow{2}{*}{0.487}   \\ \cline{3-4}
                       &                       & \multicolumn{2}{c|}{1}                           &                         &                          \\ \cline{2-6} 
                       & \multirow{2}{*}{PAMM} & \multicolumn{2}{c|}{N/A}                         & \multirow{2}{*}{error}  & \multirow{2}{*}{-10.963} \\ \cline{3-4}
                       &                       & \multicolumn{2}{c|}{N/A}                         &                         &                          \\ \hline
\end{tabular}
\setlength{\abovecaptionskip}{10pt}
\caption{The best hyper-parameters, training accuracies and testing accuracies of OSST, Cox and PAMM on datasets churn, credit, employee, maintenance, servo.}
\label{tb:interpretable-models}
\end{table}

\clearpage
\section{Black Box Survival Models}\label{sec:black_box}
We further explored whether optimal sparse survival trees achieve the performance of black box models. Random Survival Forest (RSF) improves predictive performance and reduces overfitting by aggregating the predicted survival functions from individual survival trees. Gradient-boosted Cox proportional hazard loss with regression trees as the base learner can capture complex non-linear relationships and interactions between predictors, producing more accurate predictions of survival outcomes. However, it significantly increases model complexity and makes the model extremely hard to interpret. We tried to compare to several deep learning survival models implemented in package \textit{survivalmodels} (\url{https://cran.r-project.org/web/packages/survivalmodels/index.html}), including Deepsurv \citep{katzman2018deepsurv}, Deephit \citep{lee2018deephit}, Coxtime \citep{kvamme2019time}, DNNSurv \citep{zhao2019dnnsurv}, but found that they were no longer able to install or run even their example codes. 

\noindent\textbf{Collection and Setup:} We ran this experiment on 5 datasets \textit{(churn, credit, employee, maintenance, servo)} for RSF, GB-Cox.
We split each dataset into halves (training and testing sets). For each method, we performed 5-fold cross-validation on the training set to choose hyper-parameters, trained models with the chosen hyper-parameters on the training set and tested the performance (IBS ratio) on the testing set. \\

\textbf{RSF:} We used \textit{RandomSurvivalForest} from scikit-survival.\\
\textbf{GB-Cox:} We used \textit{GradientBoostingSurvivalAnalysis} from scikit-survival.\\

\noindent\textbf{Results:} We found a single optimal sparse survival tree generally outperforms a random survival forest and approaches the performance of  GB-Cox. Remember that these methods sacrifice intepretability.  Table \ref{tb:black-box} shows the best hyper-parameters, training accuracies and testing accuracies of each method on various datasets. 

\begin{table}[htbp]
\begin{tabular}{|c|c|ccc|c|c|}
\hline
Dataset &
  Method &
  \multicolumn{3}{c|}{Tuned Hyper-parameters} &
  Training IBS Ratio &
  Testing IBS Ratio \\ \hline\hline
\multirow{6}{*}{churn} &
  \multirow{2}{*}{OSST} &
  \multicolumn{2}{c|}{regularization} &
  max\_depth &
  \multirow{2}{*}{0.602} &
  \multirow{2}{*}{0.470} \\ \cline{3-5}
 &
   &
  \multicolumn{2}{c|}{0.001} &
  5 &
   &
   \\ \cline{2-7} 
 &
  \multirow{2}{*}{RSF} &
  \multicolumn{1}{c|}{max\_features} &
  \multicolumn{1}{c|}{n\_estimators} &
  max\_depth &
  \multirow{2}{*}{0.533} &
  \multirow{2}{*}{0.398} \\ \cline{3-5}
 &
   &
  \multicolumn{1}{c|}{sqrt} &
  \multicolumn{1}{c|}{200} &
  9 &
   &
   \\ \cline{2-7} 
 &
  \multirow{2}{*}{GB-Cox} &
  \multicolumn{1}{c|}{max\_features} &
  \multicolumn{1}{c|}{n\_estimators} &
  max\_depth &
  \multirow{2}{*}{0.677} &
  \multirow{2}{*}{0.561} \\ \cline{3-5}
 &
   &
  \multicolumn{1}{c|}{log2} &
  \multicolumn{1}{c|}{500} &
  3 &
   &
   \\ \hline\hline
\multirow{6}{*}{credit} &
  \multirow{2}{*}{OSST} &
  \multicolumn{2}{c|}{regularization} &
  max\_depth &
  \multirow{2}{*}{0.199} &
  \multirow{2}{*}{0.170} \\ \cline{3-5}
 &
   &
  \multicolumn{2}{c|}{0.0025} &
  5 &
   &
   \\ \cline{2-7} 
 &
  \multirow{2}{*}{RSF} &
  \multicolumn{1}{c|}{max\_features} &
  \multicolumn{1}{c|}{n\_estimators} &
  max\_depth &
  \multirow{2}{*}{0.221} &
  \multirow{2}{*}{0.154} \\ \cline{3-5}
 &
   &
  \multicolumn{1}{c|}{sqrt} &
  \multicolumn{1}{c|}{100} &
  5 &
   &
   \\ \cline{2-7} 
 &
  \multirow{2}{*}{GB-Cox} &
  \multicolumn{1}{c|}{max\_features} &
  \multicolumn{1}{c|}{n\_estimators} &
  max\_depth &
  \multirow{2}{*}{0.309} &
  \multirow{2}{*}{0.209} \\ \cline{3-5}
 &
   &
  \multicolumn{1}{c|}{sqrt} &
  \multicolumn{1}{c|}{100} &
  3 &
   &
   \\ \hline\hline
\multirow{6}{*}{employee} &
  \multirow{2}{*}{OSST} &
  \multicolumn{2}{c|}{regularization} &
  max\_depth &
  \multirow{2}{*}{0.691} &
  \multirow{2}{*}{0.634} \\ \cline{3-5}
 &
   &
  \multicolumn{2}{c|}{0.0005} &
  5 &
   &
   \\ \cline{2-7} 
 &
  \multirow{2}{*}{RSF} &
  \multicolumn{1}{c|}{max\_features} &
  \multicolumn{1}{c|}{n\_estimators} &
  max\_depth &
  \multirow{2}{*}{0.676} &
  \multirow{2}{*}{0.616} \\ \cline{3-5}
 &
   &
  \multicolumn{1}{c|}{log2} &
  \multicolumn{1}{c|}{50} &
  8 &
   &
   \\ \cline{2-7} 
 &
  \multirow{2}{*}{GB-Cox} &
  \multicolumn{1}{c|}{max\_features} &
  \multicolumn{1}{c|}{n\_estimators} &
  max\_depth &
  \multirow{2}{*}{0.784} &
  \multirow{2}{*}{0.625} \\ \cline{3-5}
 &
   &
  \multicolumn{1}{c|}{log2} &
  \multicolumn{1}{c|}{500} &
  12 &
   &
   \\ \hline\hline
\multirow{6}{*}{maintenance} &
  \multirow{2}{*}{OSST} &
  \multicolumn{2}{c|}{regularization} &
  max\_depth &
  \multirow{2}{*}{0.985} &
  \multirow{2}{*}{0.983} \\ \cline{3-5}
 &
   &
  \multicolumn{2}{c|}{0.0001} &
  5 &
   &
   \\ \cline{2-7} 
 &
  \multirow{2}{*}{RSF} &
  \multicolumn{1}{c|}{max\_features} &
  \multicolumn{1}{c|}{n\_estimators} &
  max\_depth &
  \multirow{2}{*}{0.565} &
  \multirow{2}{*}{0.545} \\ \cline{3-5}
 &
   &
  \multicolumn{1}{c|}{log2} &
  \multicolumn{1}{c|}{50} &
  3 &
   &
   \\ \cline{2-7} 
 &
  \multirow{2}{*}{GB-Cox} &
  \multicolumn{1}{c|}{max\_features} &
  \multicolumn{1}{c|}{n\_estimators} &
  max\_depth &
  \multirow{2}{*}{0.880} &
  \multirow{2}{*}{0.866} \\ \cline{3-5}
 &
   &
  \multicolumn{1}{c|}{log2} &
  \multicolumn{1}{c|}{500} &
  3 &
   &
   \\ \hline\hline
\multirow{6}{*}{servo} &
  \multirow{2}{*}{OSST} &
  \multicolumn{2}{c|}{regularization} &
  max\_depth &
  \multirow{2}{*}{0.544} &
  \multirow{2}{*}{0.360} \\ \cline{3-5}
 &
   &
  \multicolumn{2}{c|}{0.01} &
  4 &
   &
   \\ \cline{2-7} 
 &
  \multirow{2}{*}{RSF} &
  \multicolumn{1}{c|}{max\_features} &
  \multicolumn{1}{c|}{n\_estimators} &
  max\_depth &
  \multirow{2}{*}{0.360} &
  \multirow{2}{*}{0.260} \\ \cline{3-5}
 &
   &
  \multicolumn{1}{c|}{log2} &
  \multicolumn{1}{c|}{500} &
  3 &
   &
   \\ \cline{2-7} 
 &
  \multirow{2}{*}{GB-Cox} &
  \multicolumn{1}{c|}{max\_features} &
  \multicolumn{1}{c|}{n\_estimators} &
  max\_depth &
  \multirow{2}{*}{0.803} &
  \multirow{2}{*}{0.499} \\ \cline{3-5}
 &
   &
  \multicolumn{1}{c|}{log2} &
  \multicolumn{1}{c|}{200} &
  6 &
   &
   \\ \hline
\end{tabular}
\setlength{\abovecaptionskip}{10pt}
\caption{The best hyper-parameters, training accuracies and testing accuracies of OSST, RSF and GB-Cox on datasets churn, credit, employee, maintenance, servo.}
\label{tb:black-box}
\end{table}

\end{document}